\documentclass[10pt]{report} 

\usepackage[authoryear,round]{natbib}
\usepackage{url}
\usepackage{graphicx}
\usepackage[labelfont=bf]{caption}
\usepackage[labelformat=simple]{subcaption}

\usepackage{times}
\usepackage{epsfig}
\usepackage{graphicx}
\usepackage{amsmath}
\usepackage{amssymb}
\usepackage{float}
\usepackage{tabularx}
\usepackage{rotating}

\usepackage{algorithm}
\usepackage[algo2e]{algorithm2e}
\usepackage{booktabs}
\usepackage{bbm}
\usepackage{tikz,tikz-3dplot}
\usetikzlibrary{positioning,shapes,arrows,calc,backgrounds}
\usetikzlibrary{arrows.meta,shapes.arrows}

\usepackage{enumitem}

\definecolor{redplot}{rgb}{0.81, 0.00, 0.00}
\definecolor{blueplot}{rgb}{0.157, 0.43, 0.678}
\definecolor{greenplot}{rgb}{0.32, 0.647, 0.223}
\usepackage[T1]{fontenc}
\usepackage[utf8]{inputenc}
\usepackage{xcolor}
\color{black} 

\usepackage{newpxtext}
\usepackage{newpxmath}

\usepackage{setspace}
\usepackage[
    margin=1.5in,                
    footskip=0.75in,           
    includeheadfoot=false      
]{geometry}

\usepackage{hyperref}
\definecolor{tticblue}{HTML}{006DB6}
\definecolor{tticgrey}{HTML}{666666}
\hypersetup{
    pdfborder={0 0 0},      
    colorlinks=true,        
    linkcolor=black,        
    citecolor=tticblue,     
    urlcolor=magenta        
}

\usepackage{fancyhdr}
\fancypagestyle{plain}{
  \fancyhf{}
  \fancyfoot[C]{\thepage} 
  \renewcommand{\headrulewidth}{0pt}
}
\usepackage{chngcntr}
\counterwithout{page}{chapter} 

\usepackage{amsmath,amsfonts,bm}

\def\eqref#1{equation~\ref{#1}}

\def\1{\bm{1}}

\DeclareMathAlphabet{\mathsfit}{\encodingdefault}{\sfdefault}{m}{sl}
\SetMathAlphabet{\mathsfit}{bold}{\encodingdefault}{\sfdefault}{bx}{n}

\DeclareMathOperator*{\argmin}{arg\,min}

\begin{document}

\begin{titlepage}
    \begin{center}
        \singlespacing

        {\Large \textbf{FROM THE LOSS LANDSCAPE TO DIVERSE FEATURE}} \\
        \vspace{0.2cm}
        {\Large \textbf{LEARNING IN NEURAL NETWORKS}} \\

        \vspace{1cm}

        BY \\
        DAVID ARAM YUNIS \\

        \vfill

        A thesis submitted \\
        in partial fulfillment of the requirements \\
        for the degree of \\

        \vspace{1cm}

        Doctor of Philosophy in Computer Science \\

        \vspace{1cm}

        at the \\

        \vspace{1cm}

        TOYOTA TECHNOLOGICAL INSTITUTE AT CHICAGO \\
        Chicago, Illinois \\

        \vspace{1cm}

        September, 2026 \\

        \vfill

        Thesis Committee: \\
        Matthew R. Walter (Thesis Advisor) \\
        Michael Maire \\
        David McAllester \\
        Mikhail Belkin \\

    \end{center}
\end{titlepage}

\pagenumbering{roman}
\setcounter{page}{2} 

\chapter*{Abstract}
\addcontentsline{toc}{chapter}{Abstract}

\begin{figure}[h]
\centering
\includegraphics[width=0.5\textwidth]{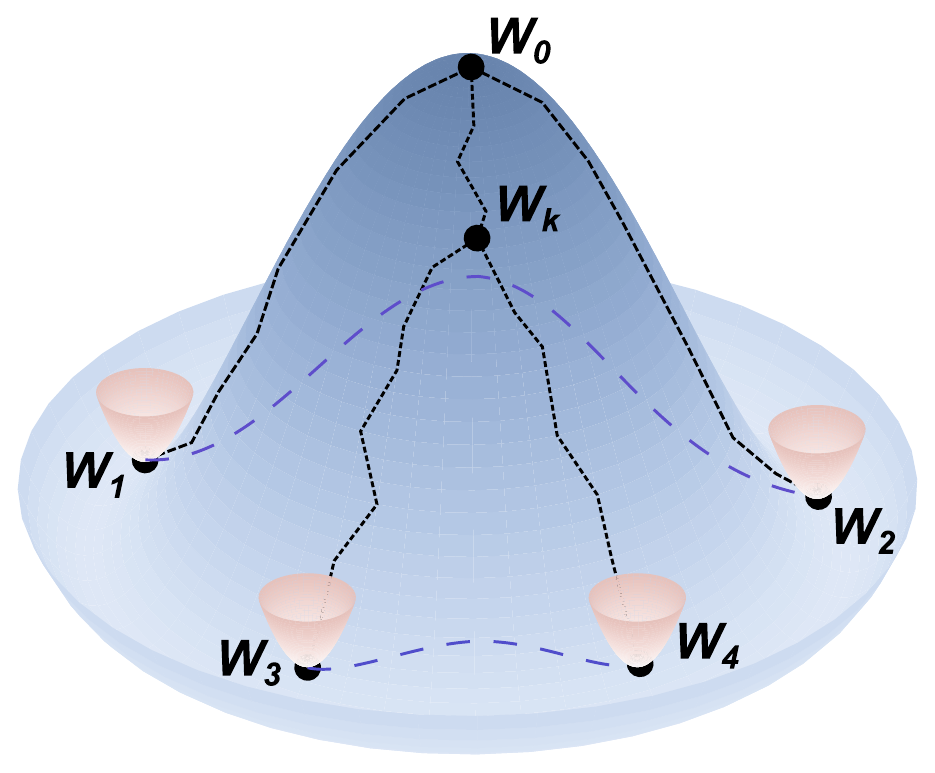}
\caption{The hat structure of the loss landscape of neural networks. We demonstrate this structure, provide a preliminary explanation for it, and leverage the explanation to improve practical neural networks.}
\label{fig:abstract_fig}
\end{figure}

{
Over the course of the last decade, neural networks have grown from an academic curiosity to moving the markets of nations. Despite this explosion in both research and deployment, relatively little is understood about how they achieve the solutions they do. This is both scientifically relevant, and pressing for society. When neural networks make decisions across self-driving, construction, law, hiring and health, there have been and will continue to be unintended consequences.

However, attempting to generalize the failures of the largest and most important production systems makes for a very difficult task. Yet signs of these failures exist at all scales of neural networks, so we should be able to study a much more tractable setting. All neural networks must undergo an optimization process, called training, to be useful. To a great degree, understanding neural networks is understanding their optimization: through what process and exposure to which data did they arrive at their results. Yet our knowledge on this topic as a field is quite imprecise. In particular, a curious phenomenon called mode connectivity, the ability to connect neural networks in the loss surface (Figure~\ref{fig:abstract_fig}), defies explanation entirely.

This dissertation elucidates, explains and exploits this special structure in the loss landscape.

We begin by deepening the phenomenon of mode connectivity. We show that it is both more complex and general than previously expected, and also that it defies common theoretical explanations previously put forth. Mode connectivity separates neural networks from the simple models of the past.

To explain the phenomenon, we propose an empirical approach centered on the spectral dynamics of weights---the behavior of singular values and vectors during optimization. Beyond offering an explanation for the strange structure of the loss landscape, these spectral dynamics connect a broad variety of phenomena, including grokking, lottery tickets, memorization and regularization. Such a large degree of common connections has been lacking in neural network phenomenology.

Armed with the powerful lens of spectral dynamics, we turn our attention to diverse feature learning, aiming to enlarge the information neural networks extract from data. We find that this spectral perspective allows us to overcome spurious correlations in simple settings, without any specialized assumptions common to prior work about task, loss or architecture. We then build on this understanding to improve the training of more practical neural networks, scaling to language modeling.

We hope these results can serve as a blueprint for the study of neural networks in the years to come.
}

\clearpage
\vspace*{\fill}
\begin{center}
    {\small\itshape A mi papá.}
\end{center}
\vspace*{\fill}

\chapter*{Acknowledgments}
It is hard for me to say that my PhD was very enjoyable. Between the many twists and turns, failed submissions, and even the sheer amount of doubt about the work in the coming pages, I had a tough time. But it was deeply fulfilling. For the vast majority of my degree, I suffered in various ways for various reasons, and in many of those moments of darkness the brightness came from the people around me. There is meaning in the struggle with others around. I owe these people many thanks.

First, to my advisor Matt Walter... what can I say? It's no exaggeration that I owe my career to Matt. As an overconfident bright-eyed undergrad, he took me on and gave me a space to grow. In the early days of graduate school, I went astray and spent years working on other topics, but he was happy to take me back into the lab even when I was a washed-up third-year with no papers to my name. He indulged my stubbornness and my curiosity for the wide range of topics I explored, and was willing to let me run with this thesis among many who doubted my work. I'm still not sure how it happened, but he has my deep, deep gratitude. He has been a role model for research, for collaboration, for interdisciplinary work, for outreach, and for balance between work and life. I hope you have the pleasure of working with him. Bonus points if it's on actual robots.

To each of my committee members, I also owe thanks. Like Matt, Michael Maire was willing to take me on in that terrible third year when I was lacking papers and low on confidence. I appreciated his precise insight, his support in my transition and his ability to always dig deeper for a new experiment. I hope a little of that has rubbed off on me. On my first interaction with David McAllester on visit day, he turned his monitor around, pointed at GPT-2 outputs, and told me ``don't do theory.'' Perhaps without that push I wouldn't have found my current love of experiments. Certainly, this dissertation wouldn't exist in its current form. My conversations with David in his corner office were a particular joy, whether on AI and society, information theory, or automated theorem proving. They broadened my horizons. Lastly, thank you to Misha Belkin. Specifically for optimism and interest when this work was a few plots on a laptop. His presentation of double descent at TTIC was one of the best talks I saw during graduate school, and a positive conversation for 30 minutes in a hotel lobby before he had to head to the airport carried me through years of doubts from others. It has been my deep satisfaction to involve him more formally in the tail end of this work, to be able to discuss like a peer, and to visit UCSD for a fantastic week in the California sun dreaming of feature learning. Thank you.

The TTIC and UChicago intellectual communities were enabling places to do research. I was able to touch so many aspects of machine learning in such a small building, and I'm thankful for the people, especially the students, who helped me to do this. Thanks to Falcon Dai for introducing me to machine learning and bringing me into the lab. For many discussions on research and learning to dream big. To Chip Schaff for showing me the ropes, and a target for what research engineering should look like. To Andrea Daniele for his company, for filling the lab with robots and making the rest of us look better for it. Thank you to Shane Settle and Shubham Toshniwal for their advice during COVID, for their support in sticking with the program when everything lost all structure, and for reminding me that whatever I had was plenty to get through it. I am thankful they stood in my corner. Thanks to Pedro Savarese for being a role model a few years ahead, and inspiring me to take a leap for my own interests to see where I would land. For his thoughtfulness and the seriousness with which he took my early and dumb ideas. To Takuma Yoneda who first supported my strange ideas when I moved back to Matt's group, at a time when I had very low confidence. And for a colleague to discuss with. To Sudarshan Babu for late night conversations, for humor, for sharing the suffering, for company. To Kshitij Patel for his early collaboration in all of this work, for his groundedness and willingness to try to see the value. Thank you to Gal Vardi for his dogged support of the middle chapter of this dissertation. To Jungo Kasai for his help in navigating the internship market, and for demonstrating the value of a network. To Kevin Stangl and Nick Kolkin for breakfasts, and help with orienting myself on the tail end of the program. To Sam Buchanan for curiosity and lots of enriching conversations. Thanks to Pushkar Shukla for his friendship, for his parallel experience that helped me to see I was not alone.

I also want to thank the broader TTIC community. Rose Bradford made a huge effort over Christmas break one year sorting out my fellowship when it was standing in the way of an internship. She continued to sacrifice at other times for me and I am thankful. Thank you to Asha for being a pleasure on the late nights and for making TTIC comfortable to work in. Thanks to Mary Marre for her steadfast support of me throughout the program, for alerting me to any leftovers when I was in need. Thank you to Chrissy Novak, Amy Minick, Erica Cocom and Alicia McClarin for their super-responsiveness and for making lots of administrative work breezy that could easily become a complete drag. Thanks to Adam Bohlander for his tireless teachings from a land before LLMs. It was a unique pleasure to get a message from him about trying new tech in his office, and I always walked out knowing something more than when I walked in. Thank you to Jerry for his bright energy, in the dark of winter, year after year, even when I was deep in the hole. I am grateful.

A special thanks to Xiao Zhang for our collaboration midway through the PhD. When we started working together, I was scrambling for a way to survive, and I'm very glad I caught your hand when I did. That project put wind in my sails, and though it was unrelated, gave me the confidence to attack this thesis again. Thank you for the reminder that sometimes all one needs is a great collaborator, and the rest will come.

From my professional colleagues, I want to thank Rui Shen for hiring me for an internship when I was having trouble in the program, and for encouraging me to go back to graduate school and fight for my degree. Thanks to Andrew Anderson for his faith in me as a colleague, even early on, and for some friendship in the suburban summer of New York. Thank you to Fantine Huot for her support, groundedness, and the way she showed me how to achieve professional balance. Thanks to Joshua Maynez, Sian Gooding, and Q Green for a sense of belonging in a workplace, a strange thing I did not imagine. Thank you to Mirella Lapata for cutting humor, and straight and practical advice for job searching. Thanks to Sohee Wang for camaraderie during the late nights in the office chasing a dream.

It took a long time, but as the PhD wrapped up, things started to improve. In particular, I spent a delightful final year working in Matt's lab, and besides the fun of doing my work and knowing it was my work, I enjoyed interacting with a new group of students. Thank you to Xiaodan Du, Tianchong Jiang, Luzhe Sun, Vincent Tan, Teddy Ayalew, Haoran Chen, Hung Le, Sam Wheeler, Jingtian Ji, and Richard Xu. It was a pleasure to work with you all, give a little bit of advice, and learn from another generation while I reflected on all of my early mistakes. I miss the long walks to lunch in the summertime. I miss the heated conversations even after two hours of sitting at group meetings. I hope that you all drew something meaningful from our interactions, and that my experience was useful beyond myself. I am thankful for your faith in me.

Besides all of the academic and professional ties above, I have personal ones to thank. My dear friends: Miles Grogger, Lucas Penido, Kevin Li, Lucas Fagen, Jainaha Srikumar, Phil Blok. Thank you for being there through my worst times, and for being there in my finest hour. I will treasure the memory of the full room at my defense very fondly. I am grateful you put up with me. I am grateful for your undying faith in me. Thank you to Sensei Winston, and the rest of the Karate Club for a constancy, a practice, and an iron will that applied directly to this degree. Osu! Gracias a mi Abuela por su confianza en mí. Y al resto de la familia en Colombia, en particular mi tío Iván, mi prima Alicia y mi Abuelo. Estar y hablar con ustedes siempre fue un descanso. Thanks to my sister Catherine for her lightness, and for a series of vignettes in the sunsets of LA. Meals over a shared table, a reminder of who we are, grounding me even when I was floating aimlessly in the work. To my mom for being an example in letting the world glance off of her. For the plants, the garden, the kitchen, and an attitude toward work of getting it done. To my dad, for reminding me over the years, and always in the way I needed but not always how I wanted, that the thing that matters most is that you never give up.

Like other journeys before mine, my PhD has been a series of decisions to not drop out. One in the first year with COVID. Another in the second year with nothing to show for a research agenda while many peers moved forward. Another after an internship when it seemed I had little to come back to. Another when very senior researchers questioned whether what I was doing was at all worthwhile, and I had to find a new agenda. Another when a series of papers all went through three to four rounds of rejection. Another when the reception after acceptance was nonexistent. I think now that the most important thing was to hang on, and it is only because of the people above and those I'm inevitably forgetting that I was able to do so. I am thankful that I was surrounded by so many great people during these times. It would have been much harder to keep going without their hands on my back. I hope you enjoy the result of their support in the following pages.

\tableofcontents
\listoftables
\listoffigures

\clearpage
\pagenumbering{arabic} 

\chapter{Introduction}

In the past decade, neural networks have exploded onto the scene. What started with early experiments in speech recognition~\citep{seide2011feature, hinton2012deep} and image classification~\citep{krizhevsky2012imagenet, russakovsky2015imagenet} has led to massive language models~\citep{achiam2023gpt}. The transition from research to product is firmly underway: ChatGPT, at its core a language model, is one of the fastest growing applications ever~\citep{ouyang2022training, openai2022chatgpt}. In the past few years the model deployment has continued relentlessly, with the rise of coding assistants~\citep{chen2021evaluating, anthropic2025claudecode}. Self-driving cars are deployed in many cities~\citep{kusano2025comparison}. Weather models based on large-scale neural networks have supplanted traditional methods~\citep{alet2025skillful}. Large campaigns to revolutionize fields are underway in drug discovery~\citep{jumper2021highly} and materials science~\citep{merchant2023scaling}. Genuinely novel mathematical results have now been proven entirely by machine~\citep{openai2026tenadvances}. We are in the midst of a dizzying and breathtaking transformation in technology.

Along with the tremendous amount of research in the field in the past decade, there has been an enormous capital investment~\citep{kkr2025beyondbubble}. This is due to a fundamental observation since the ImageNet competition~\citep{russakovsky2015imagenet} that larger models tend to perform better~\citep{krizhevsky2012imagenet, simonyan2014very, he2016deep}. Such a trend continued~\citep{sun2017revisiting}, leading to its formalization as scaling laws~\citep{hestness2017deep, kaplan2020scaling} for predicting increase in performance with both model size and data. The natural conclusion, then, is that more data and more computation means a better problem solver. Thus there is a straightforward recipe for converting dollars into better models by buying more computers and paying for more data. Taken to its limit, this has led neural networks to occupy a gigantic place in our economy~\citep{tunguz2026twelvexbet}.

Yet, as these algorithms penetrate all aspects of our societies, with the benefits there are commensurately large risks and social repercussions~\citep{bender2021dangers, bommasani2021opportunities}. When employment, health, and legal questions are decided by these algorithms~\citep{d2022underspecification}, it is critical to have some sense of how they work. Moreover, without a fundamental understanding of deep learning, it will be very difficult to solve such questions.

As an example, take simplicity bias~\citep{rahaman2019spectral, shah2020pitfalls, pezeshki2021gradient}. This very basic issue, where neural networks latch on to simple aspects of the data even when more complex features may be useful, was first investigated at a small scale. Yet at a large scale, it transforms into discussions of biases against social groups~\citep{bender2021dangers, d2022underspecification} and adversarial examples~\citep{hendrycks2021natural}. The fundamental hammer that we have to combat something like simplicity bias is a relatively simple one: collect more data and hope scaling laws drive performance~\citep{hestness2017deep, kaplan2020scaling, mazeika2024harmbench}. Through this process, we close the gap between train and test scenarios. Yet such an approach hits practical walls in changing distributions \citep{zhang2023learning}, requires extreme amounts of data, and is scientifically unsatisfying. That is to say, despite the ever-growing investment, the hole in our understanding bites us.

And this is only one such issue. We know that very simple neural networks can represent any function of interest~\citep{cybenko1989approximation}, but there remain many mysteries in actually finding these solutions. Even though it can be framed very straightforwardly, the actual optimization problem can have complex hidden qualities, often called implicit bias~\citep{neyshabur2014search}. Though ridge regression is well understood~\citep{marquardt1975ridge}, basic questions on how to generalize this to neural networks with weight decay~\citep{hanson1988comparing, krogh1991simple, zhang2018three} have only partial answers~\citep{van2017l2, andriushchenko2023we, yaras2023law}. Perhaps most vexing, we lack a complete explanation for how neural networks generalize, despite having the capacity to perfectly memorize the training data~\citep{zhang2021understanding}. Such an explanation may allow us to design better algorithms. However, a lack of understanding makes the deployment of neural networks vulnerable to uninterpretable errors~\citep{szegedy2013intriguing, ilyas2019adversarial, hendrycks2021natural, d2022underspecification, zou2023universal}.

Much work has been done toward this understanding. There is a large body of theoretical work on many of the preceding topics, but these studies are often limited to special settings like deep linear networks~\citep{arora2018optimization, arora2019implicit} or infinite-width systems~\citep{jacot2018neural}, and arguments rely on unsubstantiated or impractical assumptions like near-zero initialization. On the empirical side, a growing body of work in interpretability has attempted to reverse-engineer neural networks~\citep{rahaman2019spectral, barak2022hidden, nanda2023progress}, but given the difficulty of the task, researchers focus on small scale with bespoke methodology that is challenging to scale. A body of empirical work aims to understand the behavior of larger networks~\citep{zhang2021understanding, huh2022low, yu2023compressing}, and compromises by focusing on more abstract objects like the gram matrix~\citep{huh2022low} or the Neural Tangent Kernel (NTK)~\citep{fort2020deep}. But due to the complexity of contemporary networks, it is very difficult to connect this broad body of literature into a coherent view, and many phenomena remain poorly understood.

We are interested in a broad understanding of neural networks that can be applied to practical systems. Through that understanding we aim at a more principled way to deal with distribution shifts and low data scenarios. In the following chapters, we describe a line of work that starts with a confusion regarding the loss landscape and ends with a simple definition of feature learning. We close with a method that uses this definition to overcome simplicity bias and break the data wall of deep learning.

\section{Contributions}

This dissertation is structured as follows:
\begin{itemize}
  \item \textbf{Chapter~\ref{chp:background}}: We give introductory background around the core phenomenon of Linear Mode Connectivity (LMC) that drives the trajectory of the following chapters.
  \item \textbf{Chapter~\ref{chp:convex_mode_connectivity}}: We show that there are high dimensional, diverse convex modes in the loss landscape of many neural networks that cannot be trivially explained, extending the phenomenon of linear mode connectivity.
  \item \textbf{Chapter~\ref{chp:spectral_dynamics}}: We find that these high dimensional modes can be explained by stable low rank dynamics of top singular vectors (spectral dynamics) and that these dynamics unify a host of confusing phenomena related to generalization and regularization.
  \item \textbf{Chapter~\ref{chp:diverse_features}}: We use spectral dynamics to improve the training of neural networks, in particular by overcoming simplicity bias through the use of ensembles, with direct implications for low data regimes.
  \item \textbf{Chapter~\ref{chp:conclusion}}: We conclude by discussing the implications of these results at the time of their development, and in the years to come.
\end{itemize}

\chapter{Background}\label{chp:background}

Neural networks are omnipresent today, and we have known for a very long time that they can approximate any function of interest~\citep{cybenko1989approximation}. Yet in theory their optimization is difficult~\citep{blum1988training}, so the critical question is \textit{how} we actually find the optimal solution. It is well known that many, often infinite minima exist in the neural network loss landscape~\citep{choromanska2015loss, kawaguchi2016deep}. Yet, only some solutions generalize well~\citep{zhang2021understanding,keskar2016large}, raising questions about how and why common optimization algorithms find the solutions they do and how one might design algorithms that find better solutions. 

One way to understand optimization is through the loss landscape, the total set of trajectories a neural network can traverse. Many researchers have studied this landscape in the past~\citep{keskar2016large, dinh2017sharp, li2018visualizing, jiang2019fantastic, fort2019large}. The vast literature is due to a variety of strange properties neural network loss landscapes exhibit which are not present in prior models, due to nonconvex structure of the landscape. To explain this strangeness we have to define convex structure.

\begin{figure}[t]
  \centering
  \begin{subfigure}[b]{0.44\columnwidth}
    \centering
    \includegraphics[width=\columnwidth]{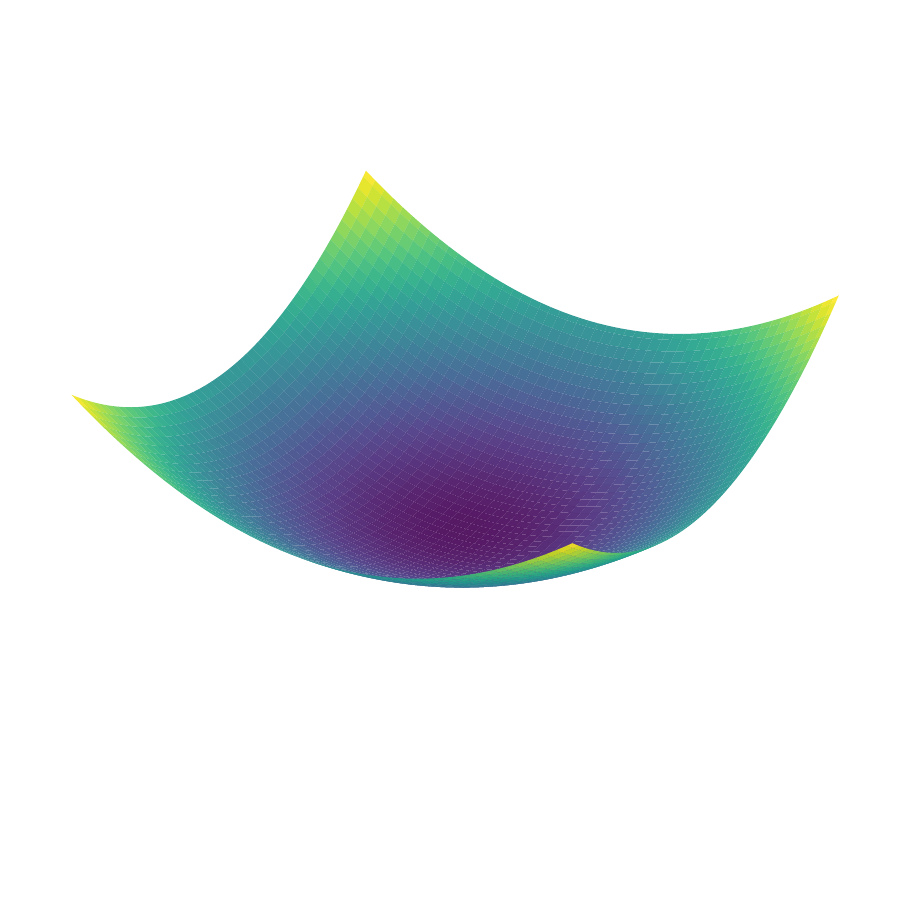}
    \caption{Convex Loss}
  \end{subfigure}
  \begin{subfigure}[b]{0.54\columnwidth}
    \centering
    \includegraphics[width=\columnwidth]{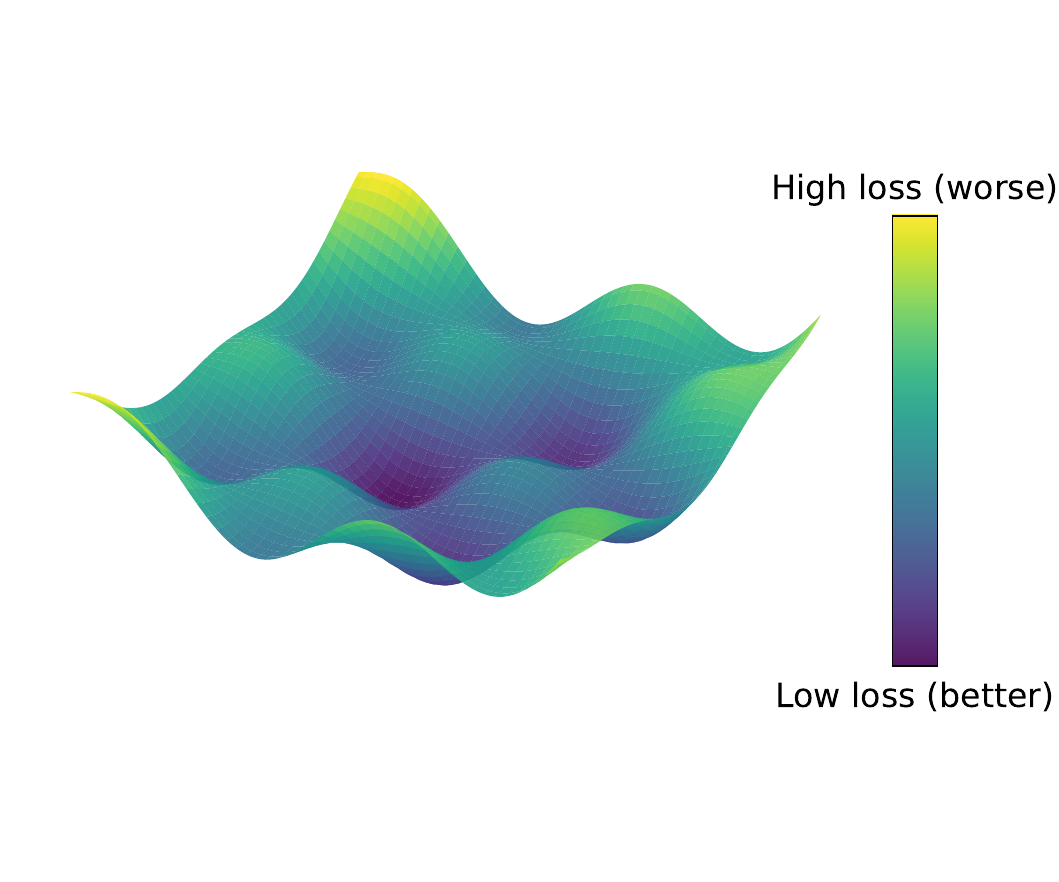}
    \caption{Nonconvex Loss}
  \end{subfigure}
  \caption{Schematic of different varieties of loss landscapes. Color denotes height, where lower is better. \textbf{(a)}: Convex loss. It is possible to find a path to the global minimum through local steps, when a ball is dropped in the landscape it will roll to the bottom. \textbf{(b)}: Nonconvex loss. Unlike the convex case, a ball may get stuck in a particular local minimum without finding the global minimum. Neural networks are generally nonconvex, making optimization more difficult than the convex case.}
  \label{fig:convex_nonconvex}
\end{figure}

Prior to neural networks, a large body of research focused on \textit{convex} optimization problems. Such problems are, intuitively, defined by the ability to connect any two minima by a linear path, and find equivalent or better solutions along that path. To be precise, if $\mathcal{L}$ is a \textbf{loss function} where lower indicates better quality, $W_1$ and $W_2$ are two sets of parameters for a neural network, and $\alpha \in [0, 1]$:
\begin{equation}
\mathcal{L}((1 - \alpha) W_1 + \alpha W_2) \leq (1 - \alpha) \mathcal{L}(W_1) + \alpha \mathcal{L}(W_2).
\end{equation}
In particular, this implies that no matter where one begins in the landscape, it is always possible to find a path to an optimum via small steps without getting stuck. If you drop a ball in this landscape, it will eventually roll to the bottom. Such a property allows for a complex and beautiful theory of optimization, and extremely efficient algorithms~\citep{vandenberghe2004convex}. See Figure~\ref{fig:convex_nonconvex} (a) for an intuitive picture.

Unfortunately, useful neural networks are nonconvex. This means that there are many possible optima in the loss landscape not connected by paths, like separate craters in the surface (see Figure~\ref{fig:convex_nonconvex} (b)). In particular, it is possible to get stuck at suboptimal points depending on where one starts, and how one descends. Besides being provable in specific cases~\citep{sontag1989backpropagation}, this is also empirically confirmed for neural networks of interest~\citep{li2018visualizing, frankle2020linear}. The rough intuition is that, due to the multiple layers of linear transforms, it is possible to permute one layer, then permute the following in a corresponding way such as to not alter the solution. The same is true for rescaling, but linearly interpolating between these solutions is worse than either endpoint~\citep{dinh2017sharp}. Moreover, different hyperparameters can lead to different minima \citep{keskar2016large}.

Despite the nonconvexity in the loss landscape, one interesting empirical property that has been observed is mode connectivity which suggests more structure than arbitrary complexity. Mode connectivity refers to the ability to find \textit{nonlinear} paths of low loss that connect two minima in the loss landscape that were derived from the same initialization. \citet{draxler2018essentially} and \citet{garipov2018loss} found that such paths could be quite simple in structure: piecewise continuous with only a few segments. The ability to connect different parts of the loss landscape may appear surprising due to the nonconvexity, but theoretical work showed that, with sufficient capacity, it is possible to find paths of even piecewise linear structure~\citep{kuditipudi2019explaining} through smoothly pushing one optimum into the excess capacity, then slowly interpolating to the other.

However, such a theoretical argument still requires nonlinear paths. More surprising, then, was the discovery of \textit{linear} mode connectivity (LMC), where two optima started from the same initialization could be connected by a \textit{linear} path of low loss \citep{nagarajan2019uniform}. The existence of this linear path implies that optimization appears to stay in a convex subspace, contrary to the seeming nonconvexity we have been discussing. Instead of exploring the entire landscape, the behavior of optimization appears quite structured.

Yet Nagarajan and Kolter's results only applied to a very small-scale experiment. It was unclear at the time whether they were specific to some experimental details, or a more general property of neural networks. \citet{frankle2020linear} showed that linear mode connectivity indeed held true for a larger class of neural networks under a modified protocol, indicating the observation was nontrivial. Their modified experiment added a small amount of shared optimization before splitting the two trajectories. After splitting, they train both as usual to their stopping points. This process differs slightly from that of Nagarajan and Kolter due to the shared optimization, but still indicates that optimization is quite well-behaved after a short amount of time.

It is worth asking whether the observation made by \citet{frankle2020linear} is trivial: is the apparent convexity due to very small distances after the shared optimization? Even the nonconvex loss functions should be locally convex when one zooms in far enough, so this explanation could make the phenomenon completely uninteresting. The evidence, however, suggests this is not the case as the distance between initialization and the end of shared training (the split point) is of the same magnitude as that between the split point and optima, and between optima. So LMC is a nontrivial kind of convexity, a point which we will reinforce in our experiments.

Additionally, LMC seems to appear broadly across more optimization settings and architectures. \citet{neyshabur2020being} showed that LMC is also present when the initialization is taken from a pretrained model on a separate task. Thus the optimization landscape is limited by prior training on very loosely-related tasks. \citet{juneja2022linear} demonstrated similar results for Transformers in language tasks, beyond the simple image classification setting used for all prior work.

Though the phenomenon is intriguing in its own right, there are also practical applications to LMC beyond understanding the loss landscape. \citet{wortsman2022model} and \citet{li2022branch} showed that the convexity suggested by LMC can lead to better model performance via finding the optimal point in between solutions (model averaging). \citet{rame2022diverse} found that such model averaging could be useful beyond a single task for out-of-distribution generalization. Moreover \citet{ilharco2022editing} demonstrated that model-averaging could be used to combine tasks into a single model.

So LMC is a surprising phenomenon given the seeming nonconvexity of optimization. It is also a general phenomenon and does not have trivial theoretical explanations. Additionally, LMC can be leveraged for practical benefits. Yet the question remains: why does LMC occur? In the following chapters, we investigate this question in detail, finding that the explanation is connected to a host of other phenomena in the literature through low-rank dynamics. Which in turn allows us to propose a new method for improved feature learning. Let us begin with LMC.

\chapter{On Convexity and Linear Mode Connectivity in Neural Networks \\ \citet{yunis2022convexity}}\label{chp:convex_mode_connectivity}
{In many cases, neural networks trained with stochastic gradient descent (SGD) that share an early and often small portion of the training trajectory have solutions connected by a linear path of low loss. This phenomenon, called linear mode connectivity (LMC), has been leveraged for pruning and model averaging in large neural network models, but it is not well understood how broadly or why it occurs. LMC suggests that SGD trajectories somehow end up in a \textit{``convex''} region of the loss landscape and stay there. In this work, we confirm that this eventually does happen by finding a high-dimensional convex hull of low loss between the endpoints of several SGD trajectories. But to our surprise, simple measures of convexity do not show any obvious transition at the point when SGD will converge into this region. To understand this convex hull better, we investigate the functional behaviors of its endpoints. We find that only a small number of correct predictions are shared between all endpoints of a hull, and an even smaller number of correct predictions are shared between the hulls, even when the final accuracy is high for every endpoint. Thus, we tie LMC more tightly to convexity, and raise several new questions about the source of this convexity in neural network optimization.}
\section{Introduction}
Many researchers have studied the loss landscape of neural networks~\citep{keskar2016large, li2018visualizing, dinh2017sharp, jiang2019fantastic, neyshabur2020being, fort2019large} in the past decade. Linear mode connectivity (LMC), i.e., the existence of linear paths of low loss between solutions of optimization, was likely first observed by \citet{nagarajan2019uniform} between two small networks with the same initialization but different data orders for stochastic gradient descent (SGD). \citet{frankle2020linear} subsequently demonstrated that LMC occurs in larger vision models when branch trajectories with different data orders are split from a shared early portion of training with a common data order.

Such LMC behavior would be expected if the loss landscape were convex, as, by definition, any two points in a convex region could be connected by a linear path of lower loss. Even if the entire loss landscape were non-convex, but SGD converged to a convex region at some point in training, LMC would still hold due to local convexity. Given this simple explanation, and possible specificity of prior work, we explore across a broader range of architectures and optimizers, asking several questions regarding LMC:

\begin{itemize}
    \item \textbf{Is LMC really indicating that the loss surface is convex between the endpoints?} We find a convex hull of low loss defined by the endpoints of many SGD trajectories and show that its dimension is large. This shows that LMC is not a property of any two endpoints, but the entire loss-landscape around a \textit{``mode''} indeed looks convex. This gives more credence to the explanation for LMC through convexity.
    \item \textbf{Is the optimization problem approximately convex after the point at which we expect LMC?} We see that the Hessian nearly always has negative eigenvalues but their relative magnitude doesn't change much, and in between stochastic updates the training loss looks convex. Neither of these metrics changes significantly throughout training, so even if they indicate optimization is nearly convex, they fail to identify the start of LMC.
    \item \textbf{How are the functions corresponding to linearly connected parameters related?} We probe deeper into the source of LMC. We see that the number of shared correct predictions between convex hulls is not close to the shared correct predictions among endpoints of a hull, even when the training accuracy is high, suggesting that the ability to interpolate between parameters is only loosely related to functional similarity.
\end{itemize}

\section{Convex Mode Connectivity}
%
%
\begin{figure}[!t]
\begin{minipage}{.7\linewidth}
    \small
    \begin{algorithm2e}[H]    
        Initialize weights with $W_0$, and seed for data order with $z_0$\;
        Train $k$ steps with algorithm $\mathcal{A}$ to yield $W_k = \mathcal{A}_k(W_0, z_0)$\;
        Sample $P$ new data order seeds $z_1, \ldots, z_P$\;
        Train for $T$ steps: $W_T^1 = \mathcal{A}_T(W_k, z_1), \ldots, W_T^P = \mathcal{A}_T(W_k, z_P)$\;
        \For{$j=1, \ldots , J$}{
            Sample $a^j \sim \textrm{Unif}(\Delta_n)$ and compute $W_\text{conv}^j = \sum_{i=1}^P a_i^j W_T^i$\;
            Find closest endpoint $W_\text{end}^j = \argmin_{W_T^i} \lVert W_\text{conv}^j - W_T^i \rVert$\;
            Sample random perturbation $\epsilon^j \sim \mathcal{N}(0, I)$\;
            Compute perturbed parameter $W_\text{rand}^j = W_\text{end}^j + \frac{\lVert W_\text{end}^j - W_\text{conv}^j \rVert}{\lVert \epsilon^j \rVert} \epsilon^j$\;
            Measure $I_\text{conv}^j = \mathbbm{1}[\mathcal{L}(W_\text{conv}^j) < \mathcal{L}(W_\text{end}^j)]$ and $I_\text{rand}^j = \mathbbm{1}[\mathcal{L}(W_\text{rand}^j) < \mathcal{L}(W_\text{end}^j)]$\;
        }
        Compute $P_\text{conv} = \frac{1}{J}\sum_j I_\text{conv}^j$ and $P_\text{rand} = \frac{1}{J}\sum_j I_\text{rand}^j$\;
    \caption{Convex mode procedure}\label{alg:cmc}
    \end{algorithm2e}
\end{minipage}
\hfil
\begin{minipage}{.24\linewidth}
\tdplotsetmaincoords{30}{0}
\begin{figure}[H]
    \begin{tikzpicture}[xscale=2.5,yscale=1.5, scale=0.4]

        \tikzset{
            dot/.style = {circle, draw=black, fill=black, minimum size=6pt, 
                          very thick, inner sep=0pt, outer sep=0pt},
            point/.style = {circle, draw=black, fill=black, minimum size=3pt, 
                          very thick, inner sep=0pt, outer sep=0pt},
        }
    
        \tikzstyle{connect}=[thick]
        \tdplotsetrotatedcoords{0}{0}{0}
        \begin{scope}[tdplot_rotated_coords]
            \node[dot,label=left:$W_T^1$] (w1) at (-1.0,0) {};
            \node[dot,label={[label distance=-0.1cm]45:$W_T^2$}] (w2) at (-0.25,1) {};
            \node[dot,label=right:$W_T^3$] (w3) at (1.0,0.75) {};
            \node[dot,label=right:$W_T^4$] (w4) at (1.0,-0.75) {};
            \node[dot,label=below right:$W_T^5$] (w5) at (-0.25,-1.0) {};
    
            \node[dot,label=above:$W_k$] (w0) at (0.0,4.0) {};
            \node[point,label=right:$W_\textrm{conv}^\prime$] (wc) at (-0.15,0.0) {};
            \node[point,label=above left:$W_\textrm{rand}^\prime$] (wr) at (-1.25,0.8124) {};

            \path[->] (w0) edge [connect] (w1);
            \path[->] (w0) edge [connect] (w2);
            \path[->] (w0) edge [connect] (w3);
            \path[->] (w0) edge [connect] (w4);
            \path[->] (w0) edge [connect] (w5);
    
            \path (w1) edge [connect] (wc);
            \path (w1) edge [connect] (wr);
    
            \draw[opacity=0.5,fill=black!50,thick](w1.center)--(w2.center)--(w3.center)--(w4.center)--(w5.center)--(w1.center);
    
            \draw[dashed] (w1) circle (0.85);
        \end{scope}
    \end{tikzpicture}
    \label{cmc}
\end{figure}
\end{minipage}
    \caption{The algorithmic procedure and graphic for measuring convex mode connectivity.}
\end{figure}
Linear mode connectivity was defined for a pair of branch trajectories. Instead of a pair, we split into $P$ branches and sample convex combinations of the endpoints to better understand the loss inside the convex hull of these endpoints. Algorithm~\ref{alg:cmc} presents this convex mode connectivity (CMC) procedure, which is a slight modification of the original LMC procedure~\citep{frankle2020linear}.

\begin{figure}[!t]
    \centering
    \begin{subfigure}[b]{0.35\textwidth}
        \centering
        \includegraphics[width=\textwidth]{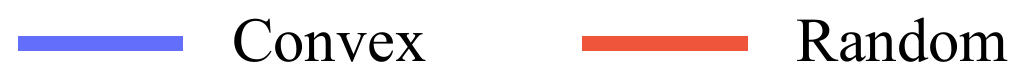}
    \end{subfigure}\\
    \begin{subfigure}[b]{0.24\textwidth}
        \centering
        \includegraphics[width=\textwidth]{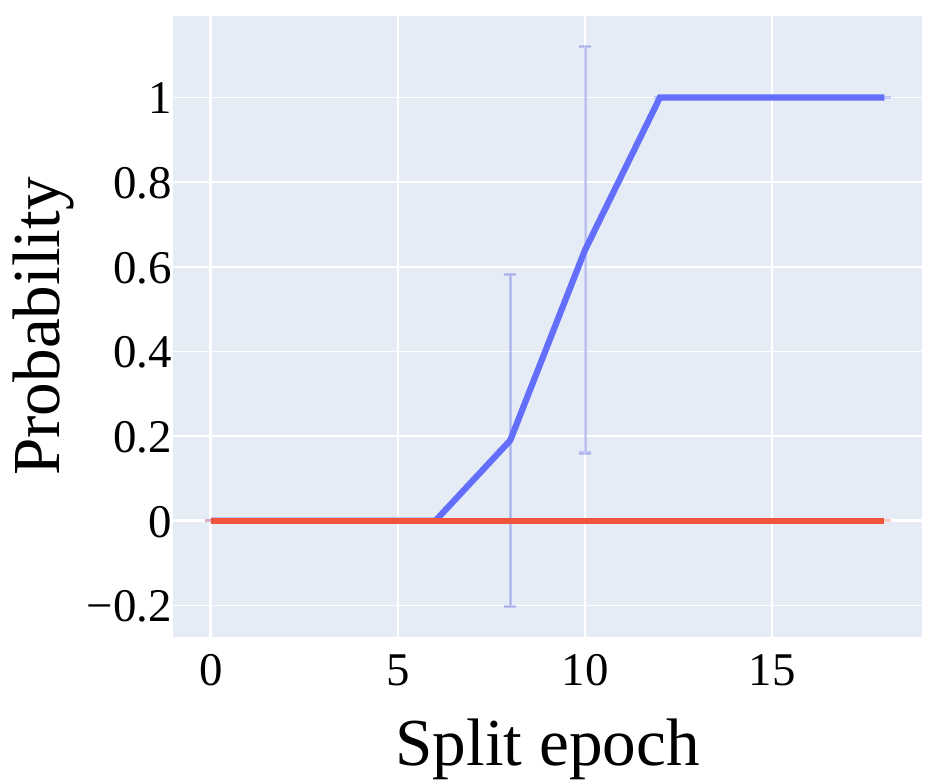}
    \end{subfigure}\hfil
    \begin{subfigure}[b]{0.24\textwidth}
        \centering
        \includegraphics[width=\textwidth]{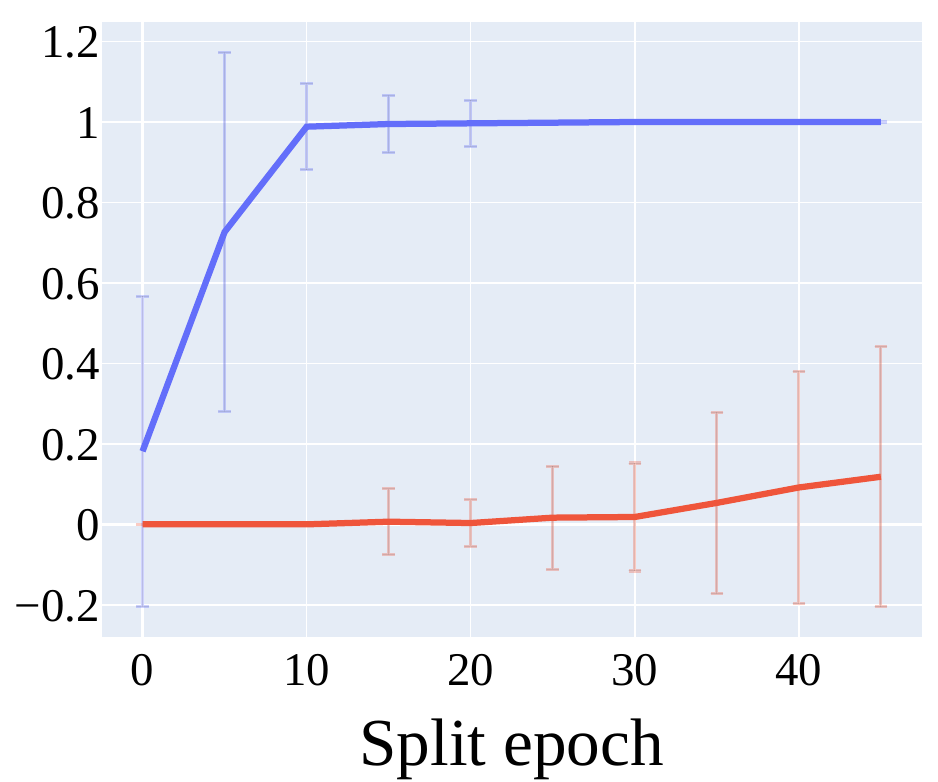}
    \end{subfigure}\hfil
    \begin{subfigure}[b]{0.24\textwidth}
        \centering
        \includegraphics[width=\textwidth]{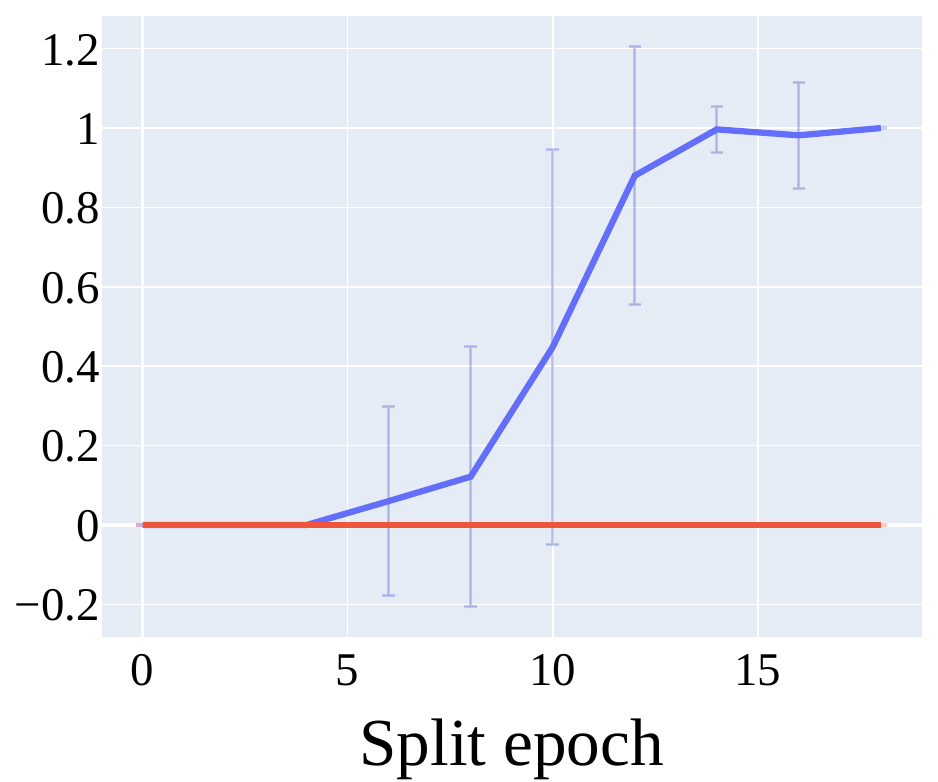}
    \end{subfigure}\hfil
    \begin{subfigure}[b]{0.24\textwidth}
        \centering
        \includegraphics[width=\textwidth]{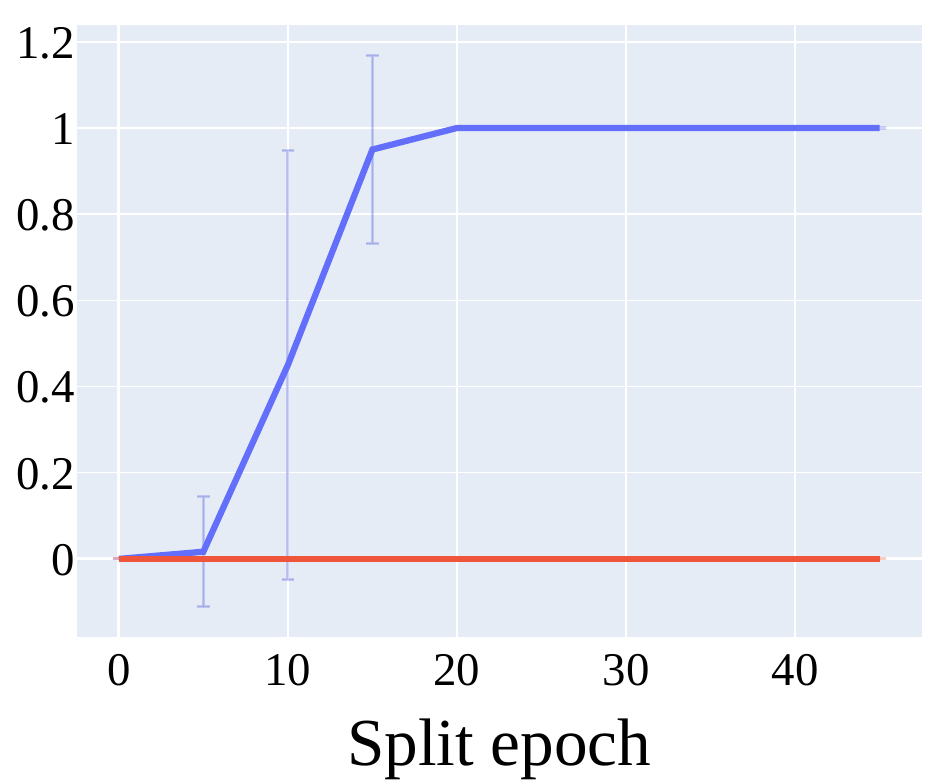}
    \end{subfigure}\hfil
    %
    \vspace{3pt}
    \begin{subfigure}[b]{0.75\textwidth}
        \centering
        \includegraphics[width=\textwidth]{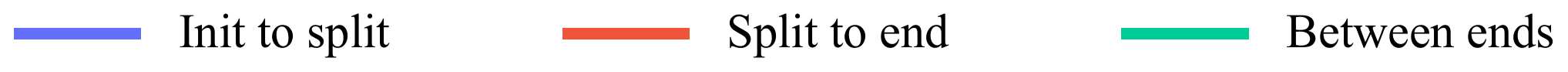}
    \end{subfigure}\\
    \begin{subfigure}[b]{0.24\textwidth}
        \centering
        \includegraphics[width=\textwidth]{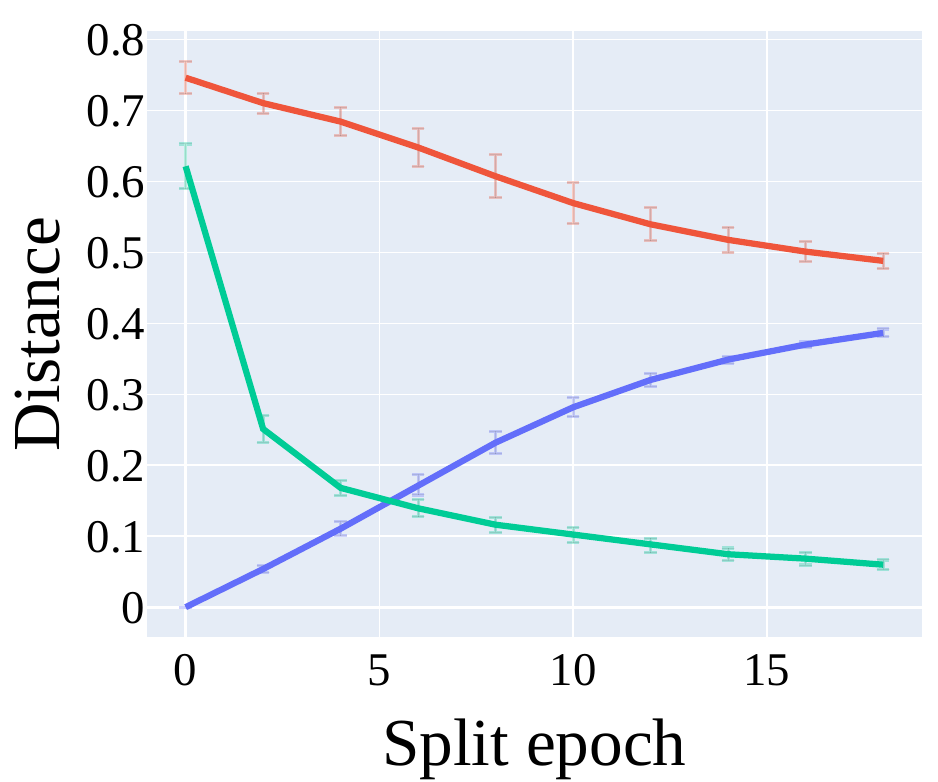}
        \caption{SIREN}
    \end{subfigure}\hfil
    \begin{subfigure}[b]{0.24\textwidth}
        \centering
        \includegraphics[width=\textwidth]{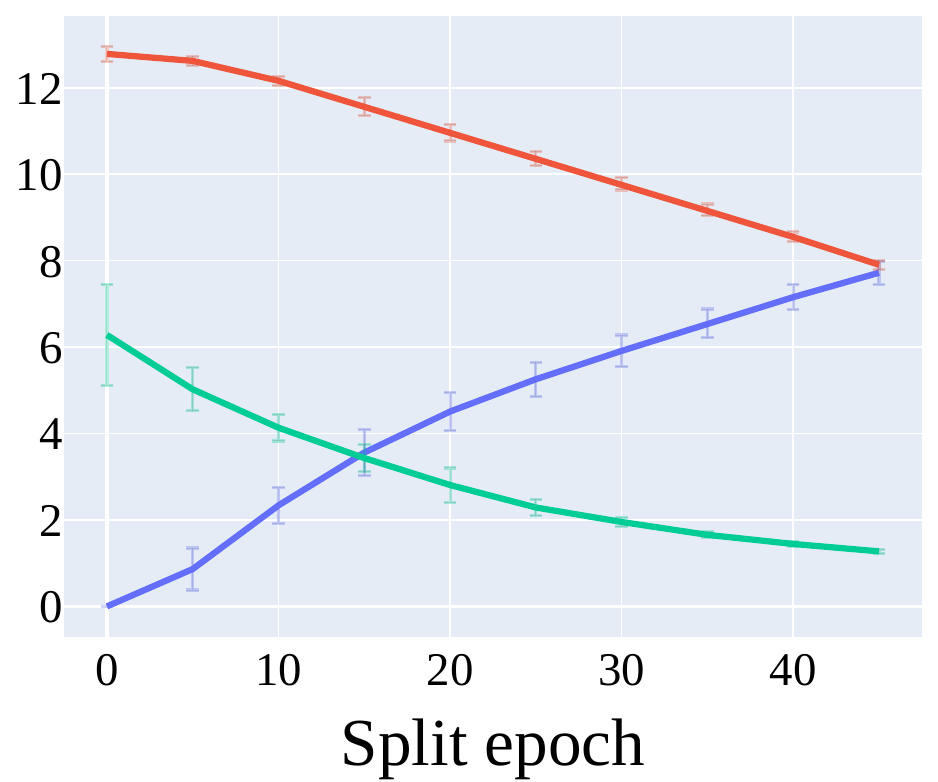}
        \caption{LeNet}
    \end{subfigure}\hfil
    \begin{subfigure}[b]{0.24\textwidth}
        \centering
        \includegraphics[width=\textwidth]{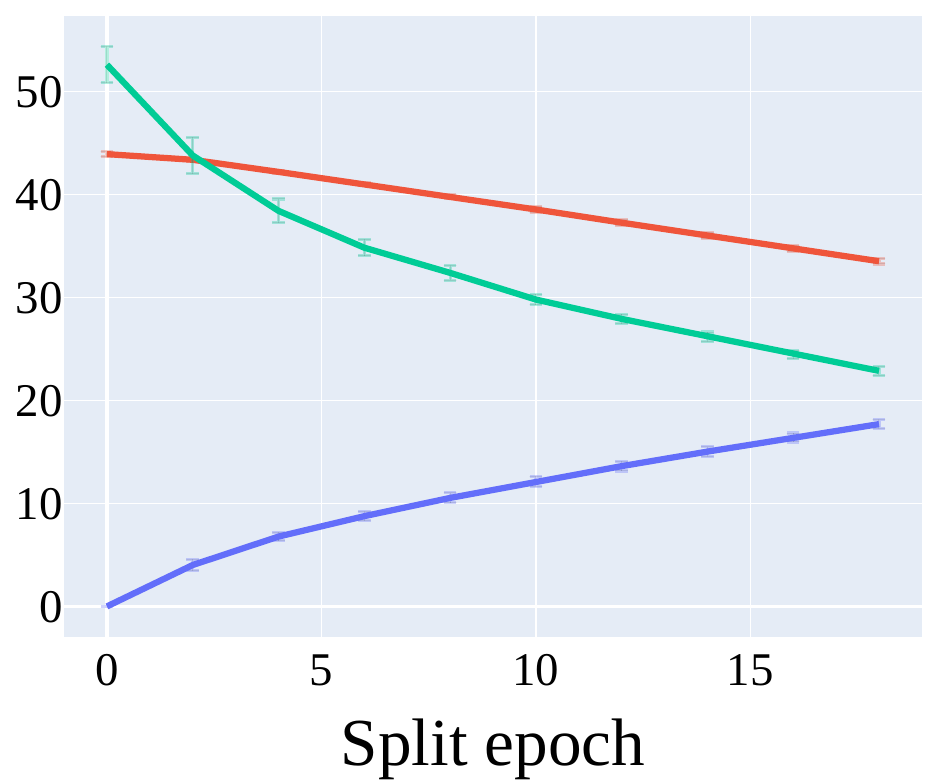}
        \caption{LeNet-m}
    \end{subfigure}\hfil
    \begin{subfigure}[b]{0.24\textwidth}
        \centering
        \includegraphics[width=\textwidth]{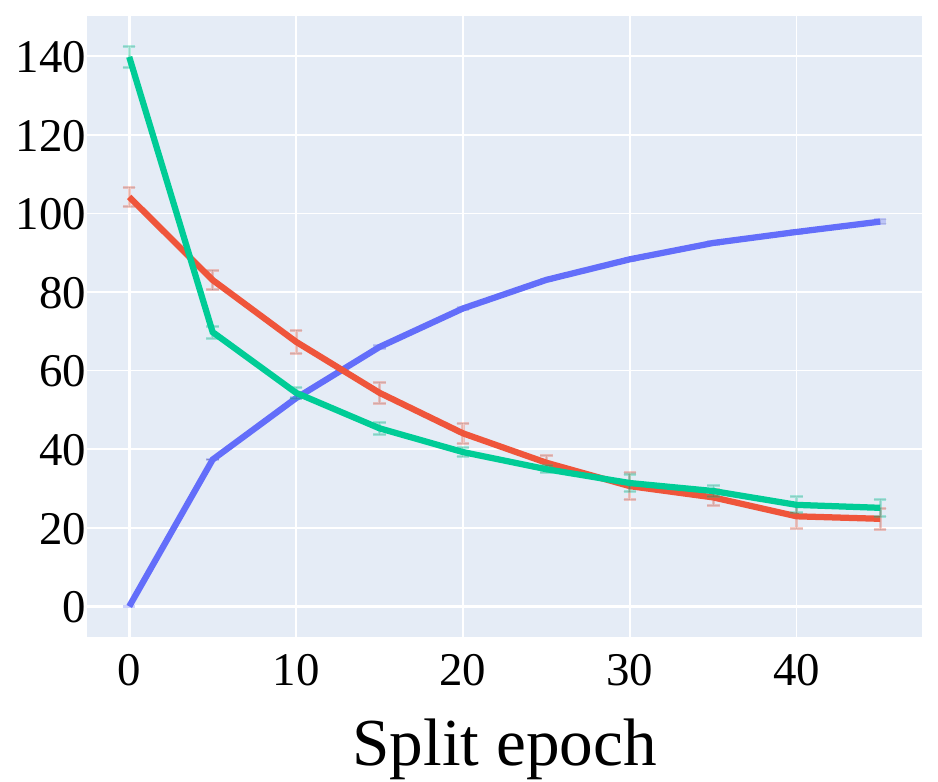}
        \caption{ResNet}
    \end{subfigure}\hfil
    \caption{\textbf{Top row:} The empirical probability $P_\text{conv}$ that a sample inside the convex hull (Convex) or a random perturbation $P_\text{rand}$ around the endpoint (Random) have lower loss than their closest endpoints vs. the split epoch ($k$ in Algorithm~\ref{alg:cmc}) for the many different trajectories. After an initial phase of training, the probability of lower loss inside the hull quickly converges to one, the loss inside the hull is always better than at the endpoints. \textbf{Bottom row:} Parameter distances between pairs of points in the CMC procedure. The distance between pairs of endpoints is of the same order of magnitude as other distances for all tasks. In other words, the apparent scale of convexity is similar to that of the total trajectory. Error bars for probabilities are computed by empirical standard deviations $\text{std}(I_\text{conv}^j)$ and $\text{std}(I_\text{rand}^j)$, hence are not meaningful when values stretch greater than 1 or less than 0.}
    \label{fig:cmc_plot}
\end{figure}

In Figure \ref{fig:cmc_plot}, we plot the probability that convex combinations of hull parameters have better training loss than hull endpoints, and likewise for random perturbations around hull endpoints. Notably, like LMC, after some point in training, random convex combinations become better than endpoint averages, while random perturbations are almost always worse---strong evidence that LMC describes a convex region in the loss surface. We call the time after this first point in training the \textit{LMC regime}. We verify these observations in a regression task (SIREN) and a few different networks trained to classify CIFAR-10. For more details on the task selection, see Appendix~\ref{app:hyps}.

\begin{figure}[!t]
    \centering
    \begin{subfigure}[b]{0.24\textwidth}
        \centering
        \includegraphics[width=\textwidth]{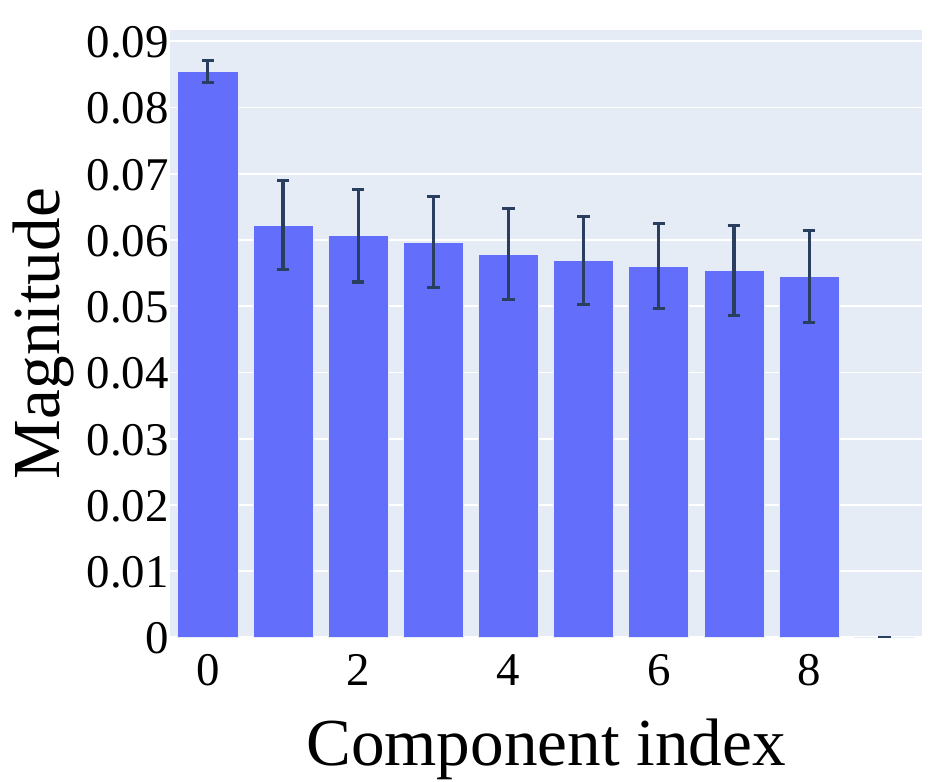}
        \caption{SIREN}
    \end{subfigure}\hfil
    \begin{subfigure}[b]{0.24\textwidth}
        \centering
        \includegraphics[width=\textwidth]{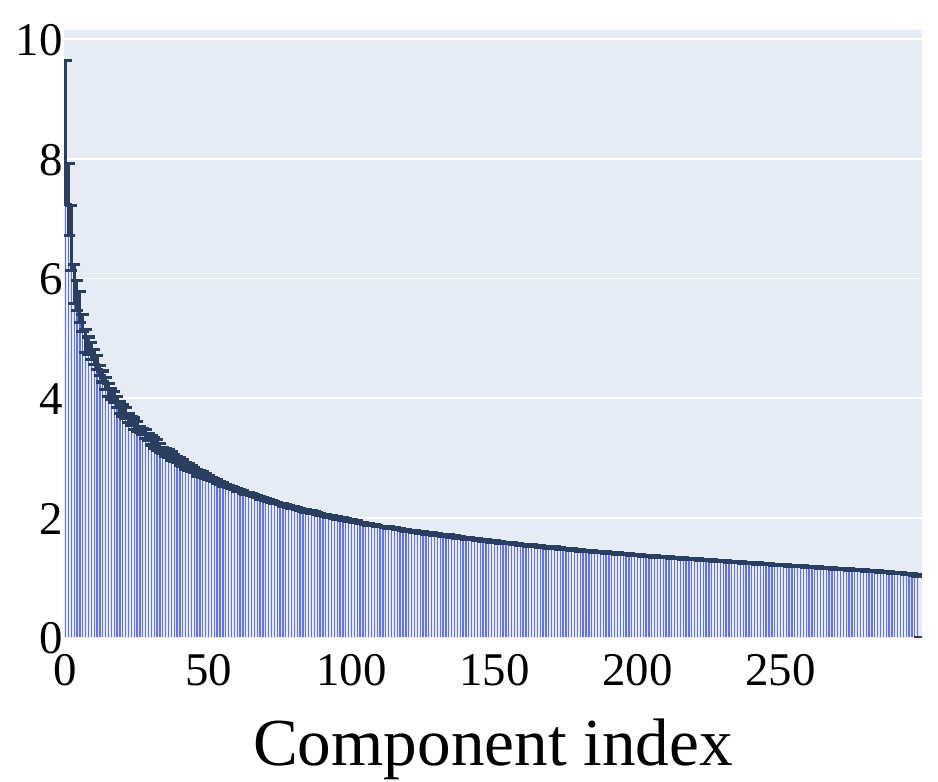}
        \caption{LeNet}
    \end{subfigure}\hfil
    \begin{subfigure}[b]{0.24\textwidth}
        \centering
        \includegraphics[width=\textwidth]{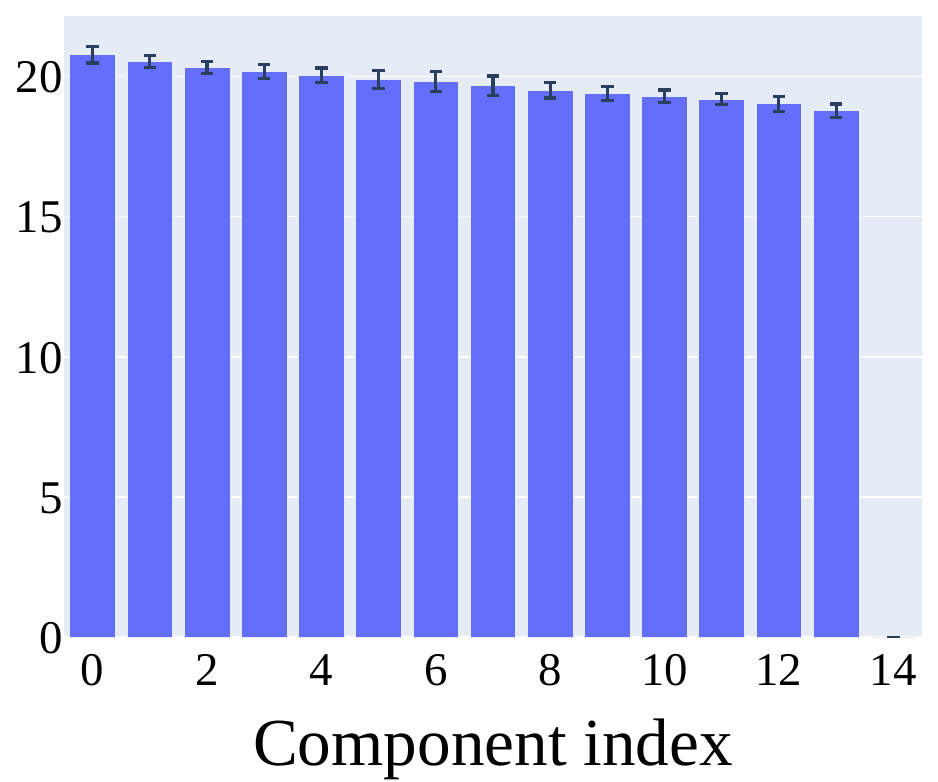}
        \caption{LeNet-m}
    \end{subfigure}\hfil
    \begin{subfigure}[b]{0.24\textwidth}
        \centering
        \includegraphics[width=\textwidth]{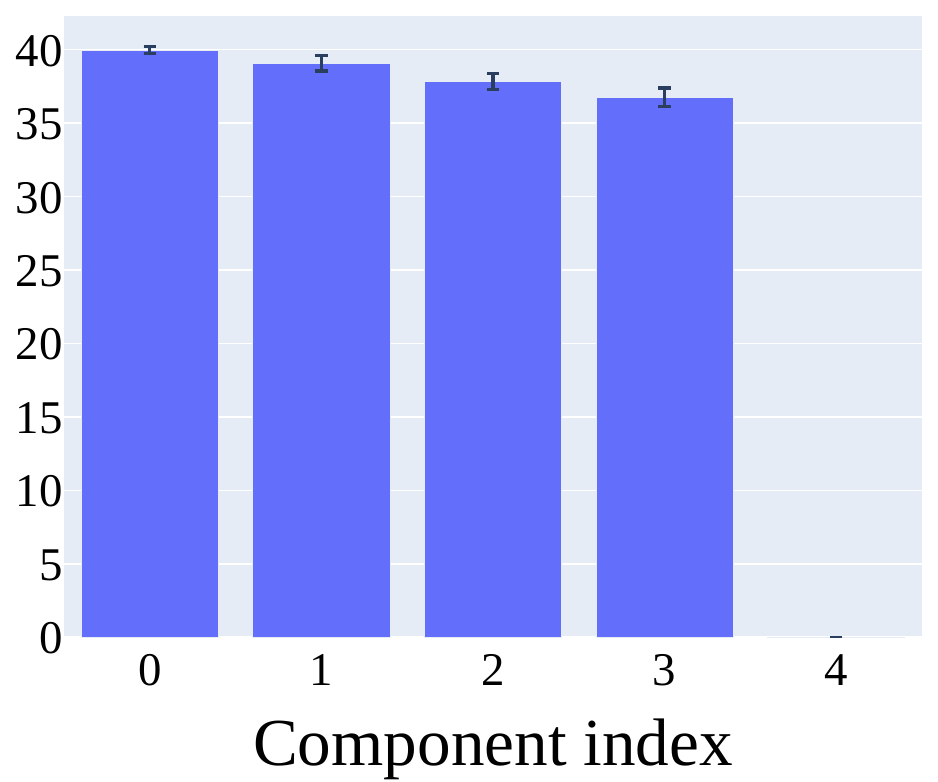}
        \caption{ResNet}
    \end{subfigure}\hfil
    \caption{Singular values for the matrix of stacked endpoints of the mode. We examine endpoints split after the point when the hull center performs better than the endpoints (see Figure~\ref{fig:cmc_plot}). For $P$ points in space, the shape they define can have at most $P-1$ dimensions. We see in all cases that the magnitudes do not decay to zero until component $P-1$. In the case that there are $K$ outputs and $P < K$ (ResNet), we see a roughly uniform mode. In cases with $P > K$ (SIREN, LeNet, and LeNet-m), it does not appear that there is any special fragmenting at $K$. In particular, the LeNet plot is over 300 endpoints, and the tail does not look as if it will decay to zero any time soon. The computational expense of these experiments limits how many endpoints we train.}
    \label{fig:pca}
\end{figure}

In order to understand the space that this mode takes up in parameter space, we stack all the parameters $W_1, \ldots, W_n$ into a matrix $A = [\textrm{flatten}(W_1)^\top \cdots \textrm{flatten}(W_n)^\top]^\top$ and take its SVD. We then plot the singular values of this matrix as a measure of the volume spanned by the endpoints for the different tasks. We see in Figure~\ref{fig:pca}, that for all tasks for which we train sufficient branches, there does not appear to be a sharp cutoff in the magnitude plot corresponding to the number of classes or the hidden dimension. The only cutoffs occur when there are not enough endpoints to see any higher dimension ($P$ points define an object of dimension at most $P-1$). This tells us that we are seeing a non-degenerate convex mode in the loss landscape, and that the particular region does not seem related to the Hessian subspace identified by \citet{gur2018gradient}, or to other simplices of low loss achieved via additional optimization~\citep{fort2019deep, benton2021loss}.

\section{Measures of Convexity}
One natural explanation for the previously demonstrated convex hulls is that, after some point in training, the trajectory enters a convex ``basin'' and the rest of the time is spent inside this region. However, previous work has shown that the linear interpolation between points early on the training trajectory and at the end can exhibit large increases in training loss~\citep{frankle2020revisiting,vlaar2022can}, making such an explanation tenuous.

Still, the loss could locally be behaving more \textit{``convexly''} after some early point in training. In order to test this, we examine the bottom part of the Hessian spectrum and the convexity of the training loss between two possible stochastic updates at a point (see Appendix~\ref{app:calc} for details). We see in Figure~\ref{fig:hess} that the Hessian shows only a small amount of local non-convexity throughout training and that the measurement of convexity between pairs of updates makes training always appear locally convex, making neural network optimization look quite well-behaved. None of the curves show distinguishing features at the LMC point (see Figure \ref{fig:cmc_plot}). Rather than believing that there is a particular point in training at which a difference in SGD updates causes the trajectories to diverge, it seems the eventual convergence to different convex hulls happens much more gradually.

\begin{figure}
    \centering
    %
    \begin{subfigure}[b]{0.24\textwidth}
        \centering
        \includegraphics[width=\textwidth]{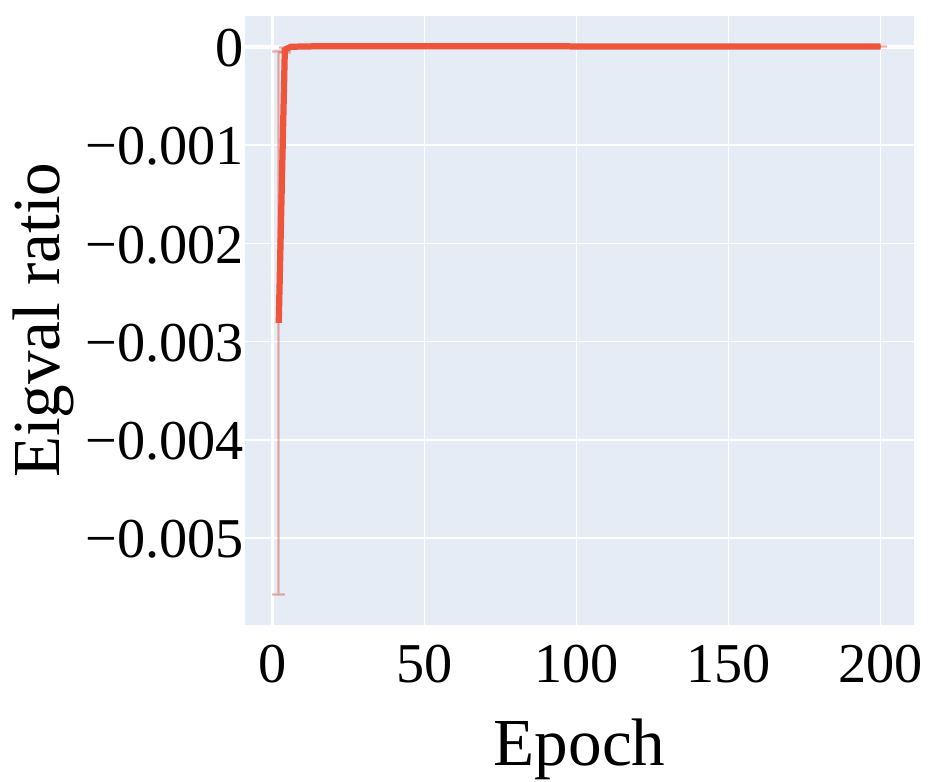}
    \end{subfigure}\hfil
    \begin{subfigure}[b]{0.24\textwidth}
        \centering
        \includegraphics[width=\textwidth]{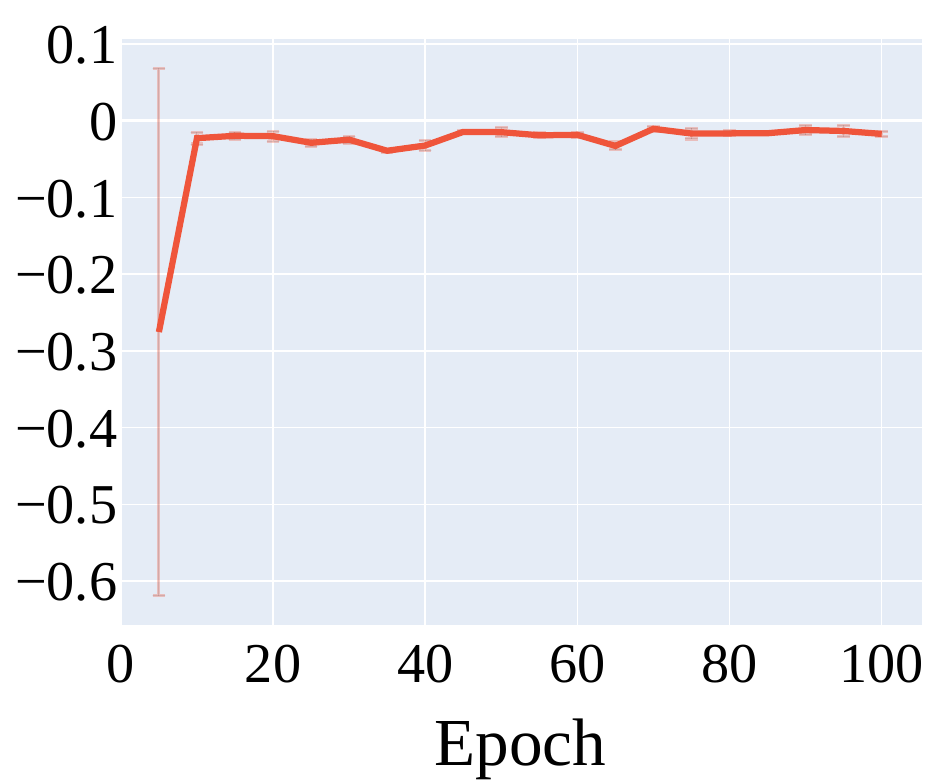}
    \end{subfigure}\hfil
    \begin{subfigure}[b]{0.24\textwidth}
        \centering
        \includegraphics[width=\textwidth]{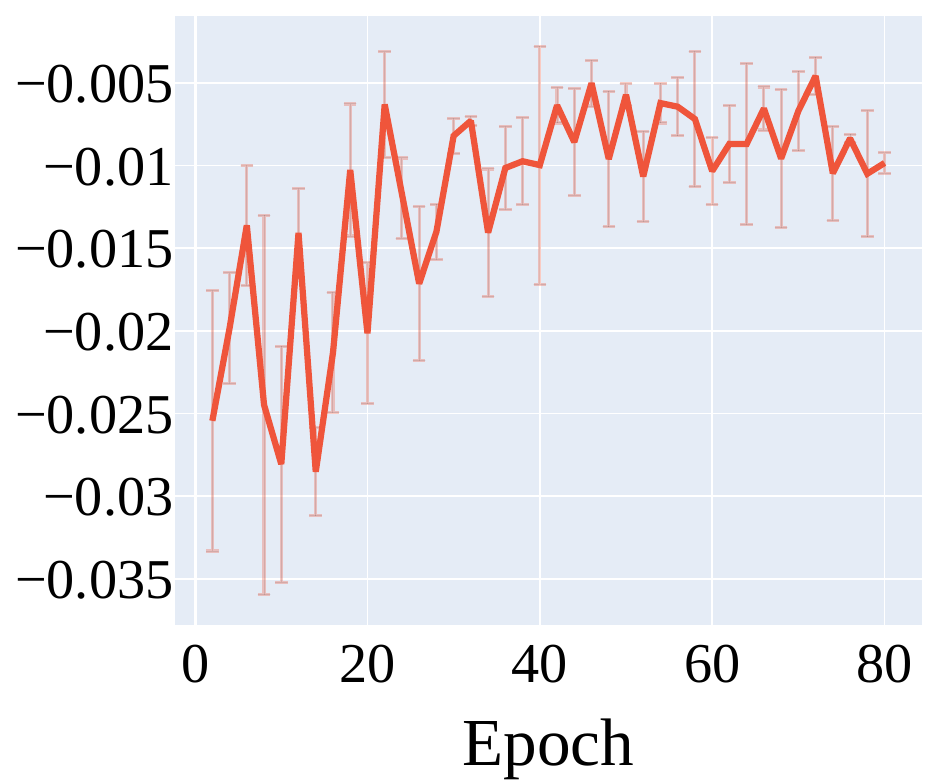}
    \end{subfigure}\hfil
    \begin{subfigure}[b]{0.24\textwidth}
        \centering
        \includegraphics[width=\textwidth]{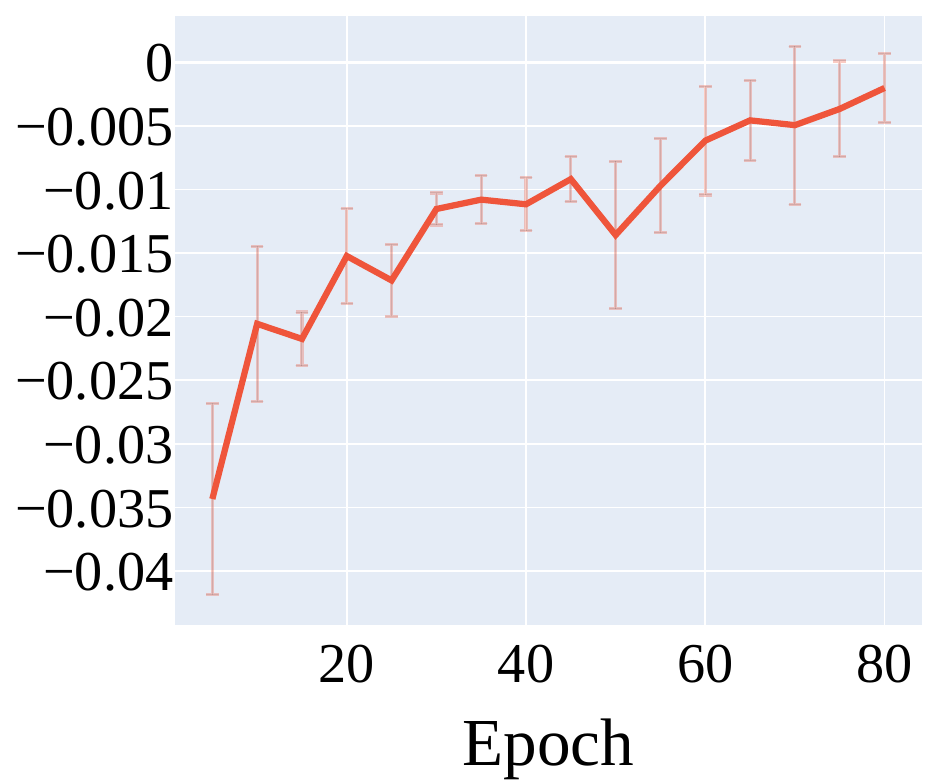}
    \end{subfigure}\hfil
    \\
    %
    \begin{subfigure}[b]{0.24\textwidth}
        \centering
        \includegraphics[width=\textwidth]{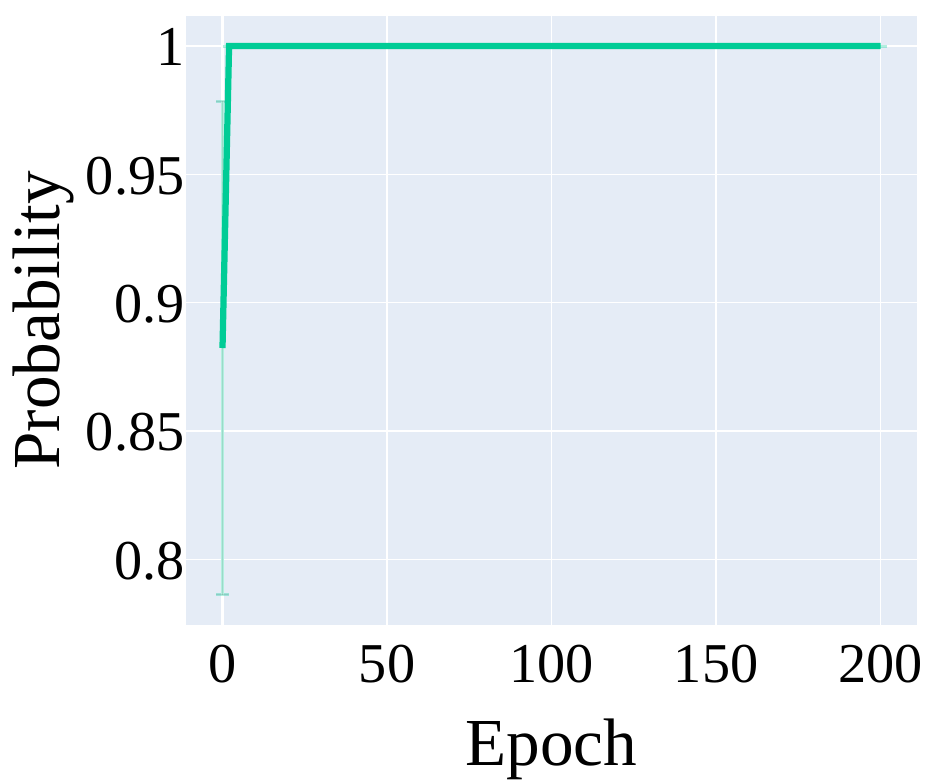}
        \caption{SIREN}
    \end{subfigure}\hfil
    \begin{subfigure}[b]{0.24\textwidth}
        \centering
        \includegraphics[width=\textwidth]{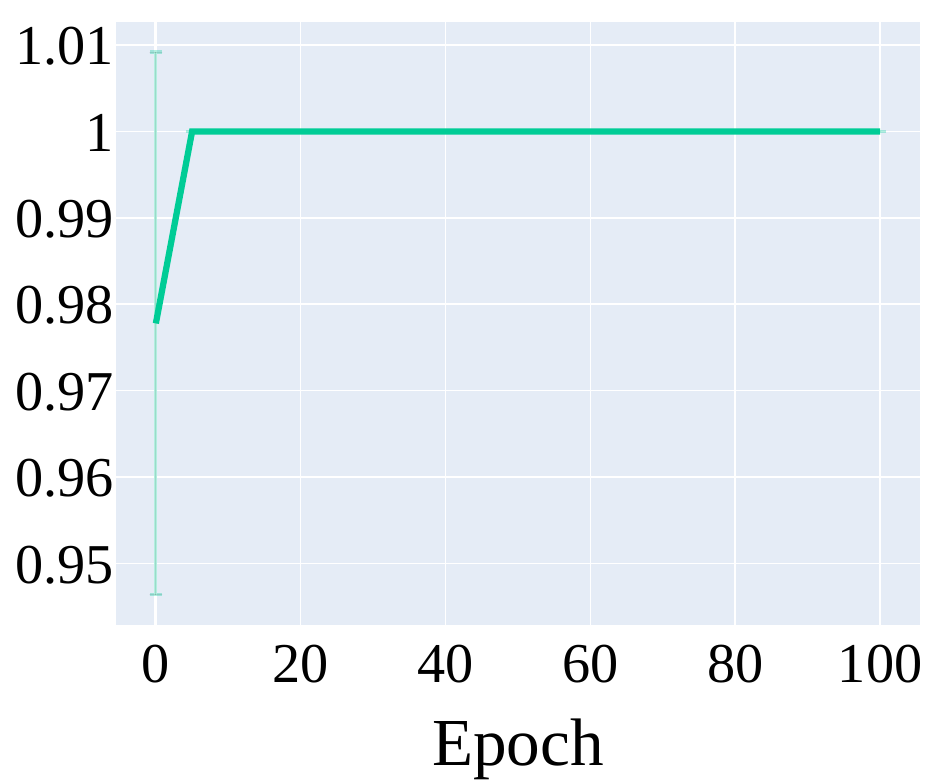}
        \caption{LeNet}
    \end{subfigure}\hfil
    \begin{subfigure}[b]{0.24\textwidth}
        \centering
        \includegraphics[width=\textwidth]{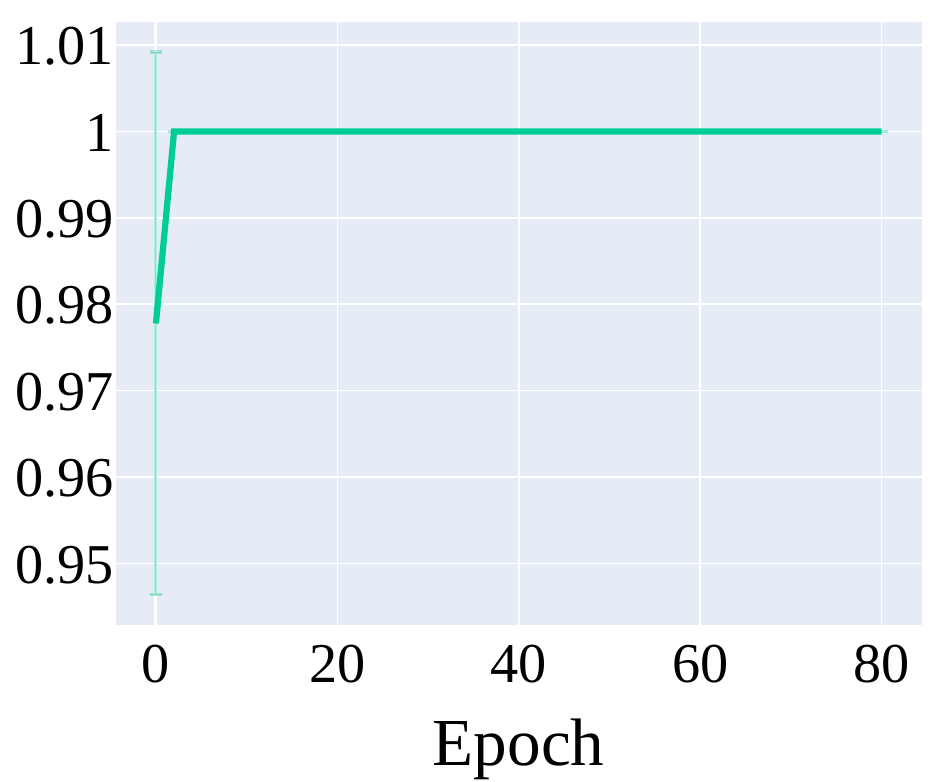}
        \caption{LeNet-m}
    \end{subfigure}\hfil
    \begin{subfigure}[b]{0.24\textwidth}
        \centering
        \includegraphics[width=\textwidth]{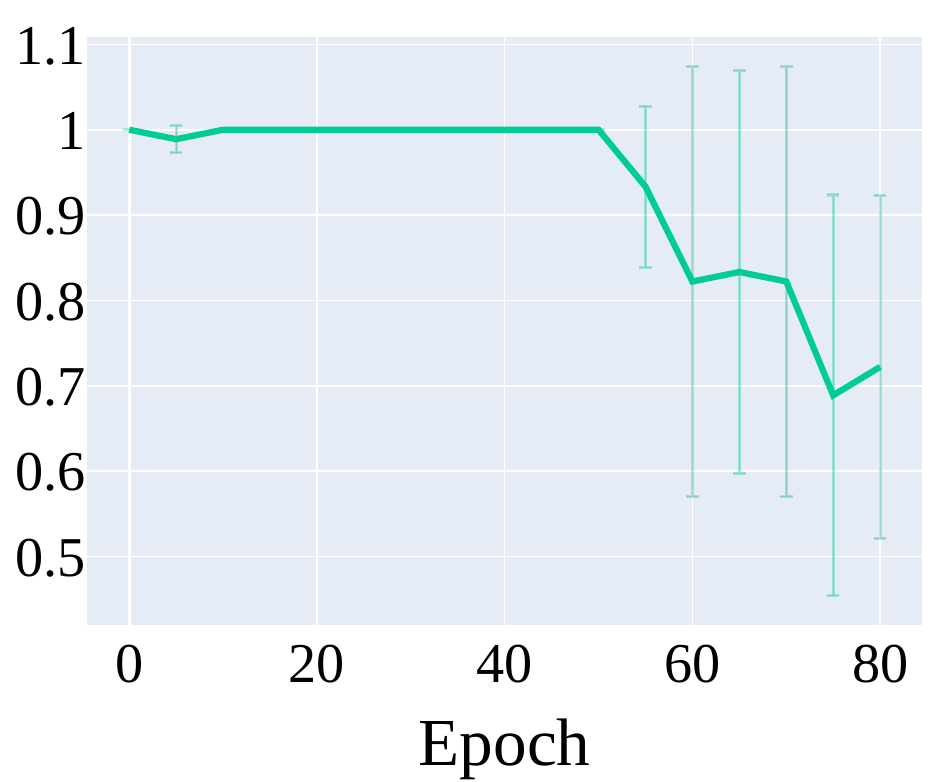}
        \caption{ResNet}
    \end{subfigure}\hfil
    \caption{
    \textbf{Top row:} The ratio of smallest to largest eigenvalue of the Hessian throughout training. This ratio is almost always negative, but quite small, making any local non-convexity appear unimportant. \textbf{Bottom row:} Probability of a convex line of loss between a pair of updates. Notice that this probability is close to 1 for the entire optimization, typically before the LMC regime in Figure \ref{fig:cmc_plot}.} 
    \label{fig:hess}
\end{figure}

\section{Diversity of Endpoints}
Figure \ref{fig:cmc_plot} (bottom) and prior work have shown that endpoints that exhibit LMC are far apart in parameter space with respect to the total trajectory length~\citep{nagarajan2019uniform, frankle2020linear}. A natural question is whether these distances are deceptive and the endpoints are actually \textit{``functionally similar''}. It is well known that with clever scaling, one can transform a set of parameters into another set, and the distance between the two will be large, while the functions they describe will remain the same. Such scaling was employed by~\citet{dinh2017sharp} to make the case that two equivalent functions could have very different sharpness. Another way to preserve functionality is through permutations of rows of weight matrices. \citet{entezari2021role} hypothesize that such permutations can map unconnected endpoints into the same mode and \citet{ainsworth2022git} show this is true for wide networks, but not the case for standard ResNets, which \citet{benzing2022random} concurrently confirm. Still, these works do not characterize the behavior within a mode.

\begin{figure}[!t]
    \centering
    \begin{subfigure}[b]{0.68\textwidth}
        \centering
        \includegraphics[width=\textwidth]{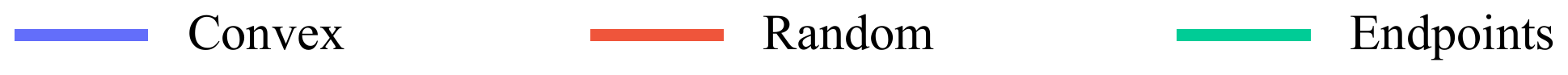}
    \end{subfigure}\\
    \begin{subfigure}[b]{0.33\textwidth}
        \centering
        \includegraphics[width=\textwidth]{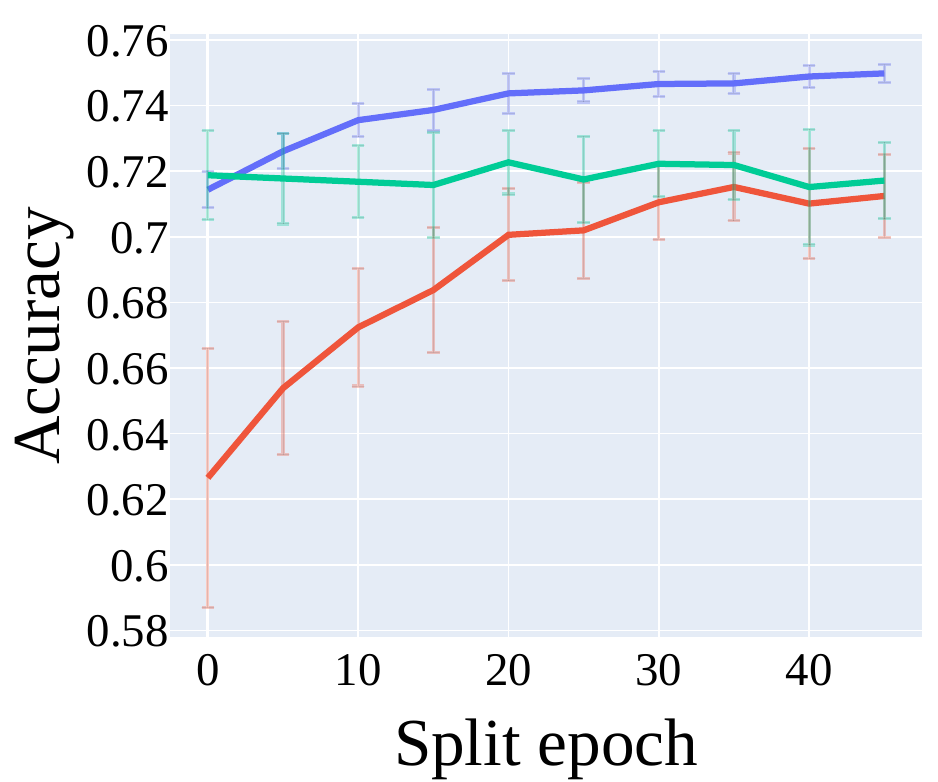}
    \end{subfigure}\hfil
    \begin{subfigure}[b]{0.33\textwidth}
        \centering
        \includegraphics[width=\textwidth]{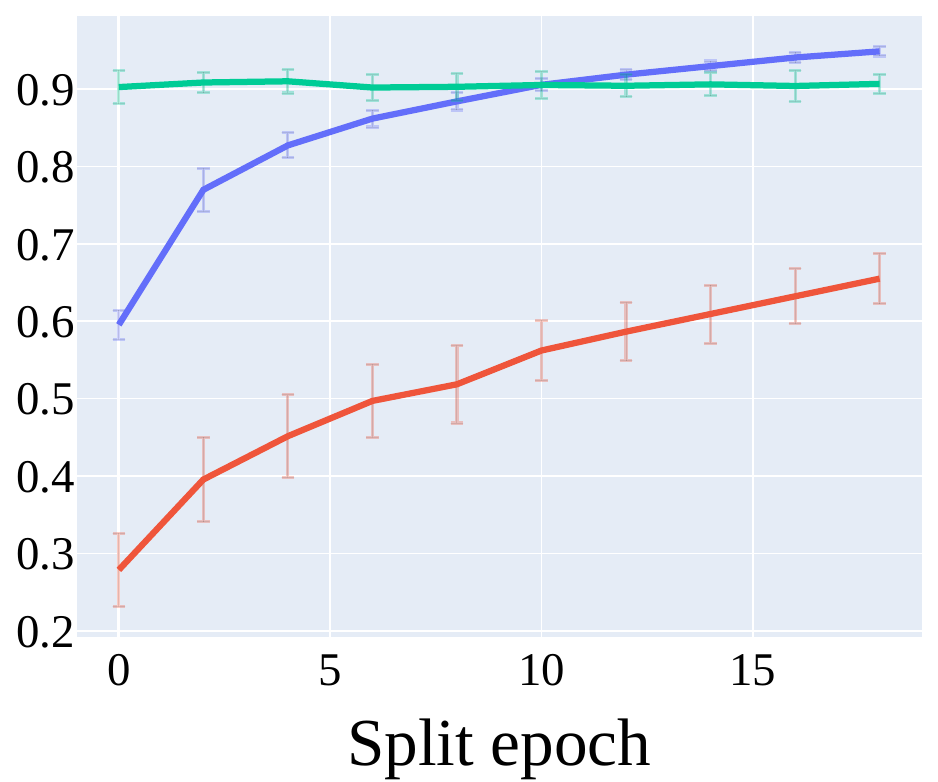}
    \end{subfigure}\hfil
    \begin{subfigure}[b]{0.33\textwidth}
        \centering
        \includegraphics[width=\textwidth]{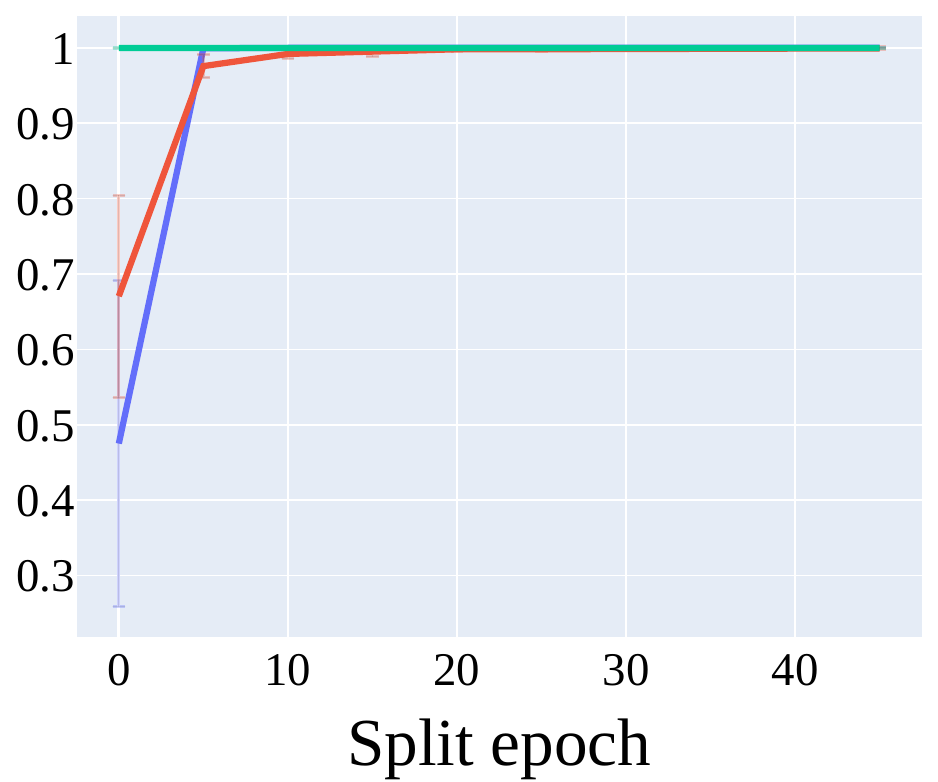}
    \end{subfigure}\hfil
    %
    \begin{subfigure}[b]{0.75\textwidth}
        \centering
        \includegraphics[width=\textwidth]{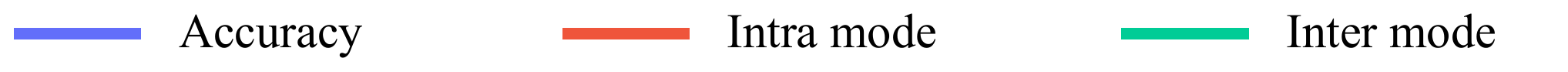}
    \end{subfigure}\\
    \begin{subfigure}[b]{0.33\textwidth}
        \centering
        \includegraphics[width=\textwidth]{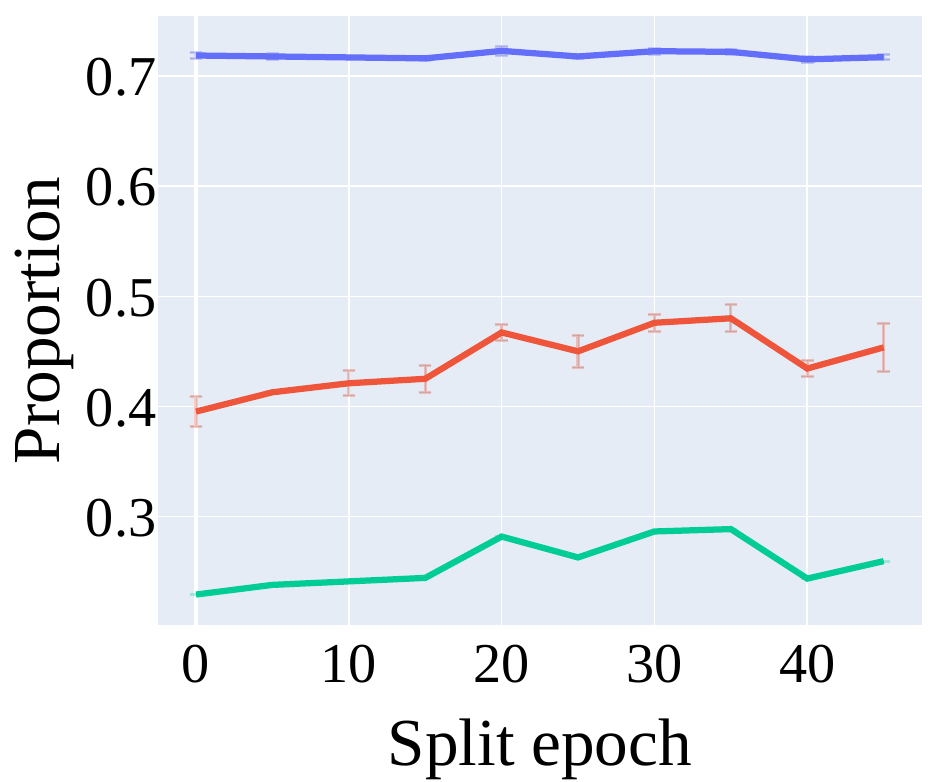}
        \caption{LeNet}
    \end{subfigure}\hfil
    \begin{subfigure}[b]{0.33\textwidth}
        \centering
        \includegraphics[width=\textwidth]{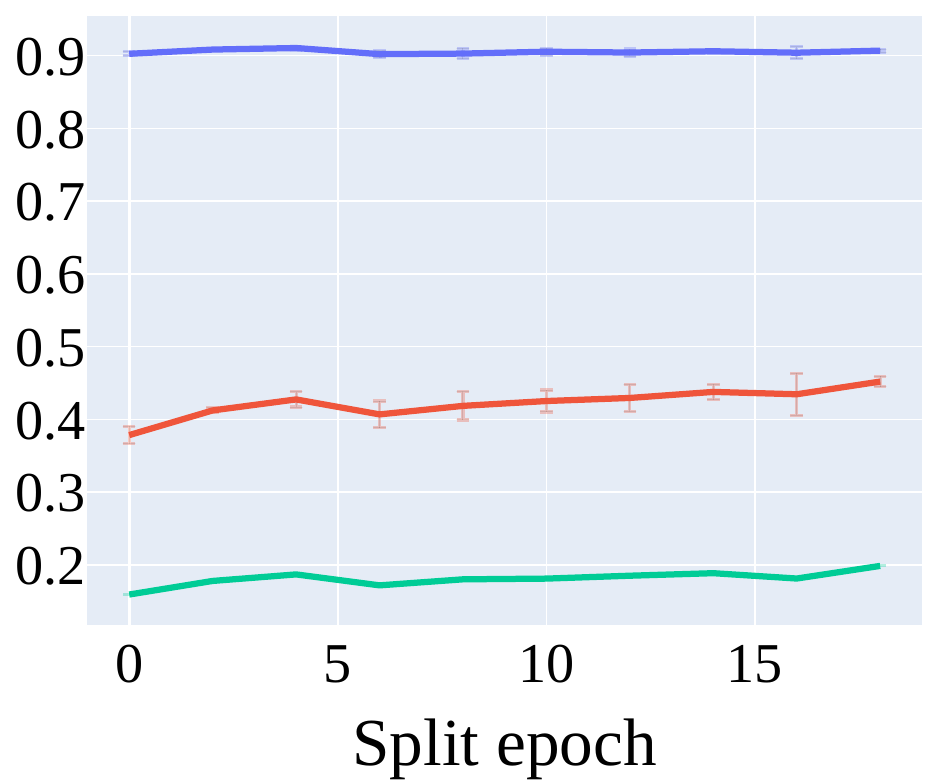}
        \caption{LeNet-m}
    \end{subfigure}\hfil
    \begin{subfigure}[b]{0.33\textwidth}
        \centering
        \includegraphics[width=\textwidth]{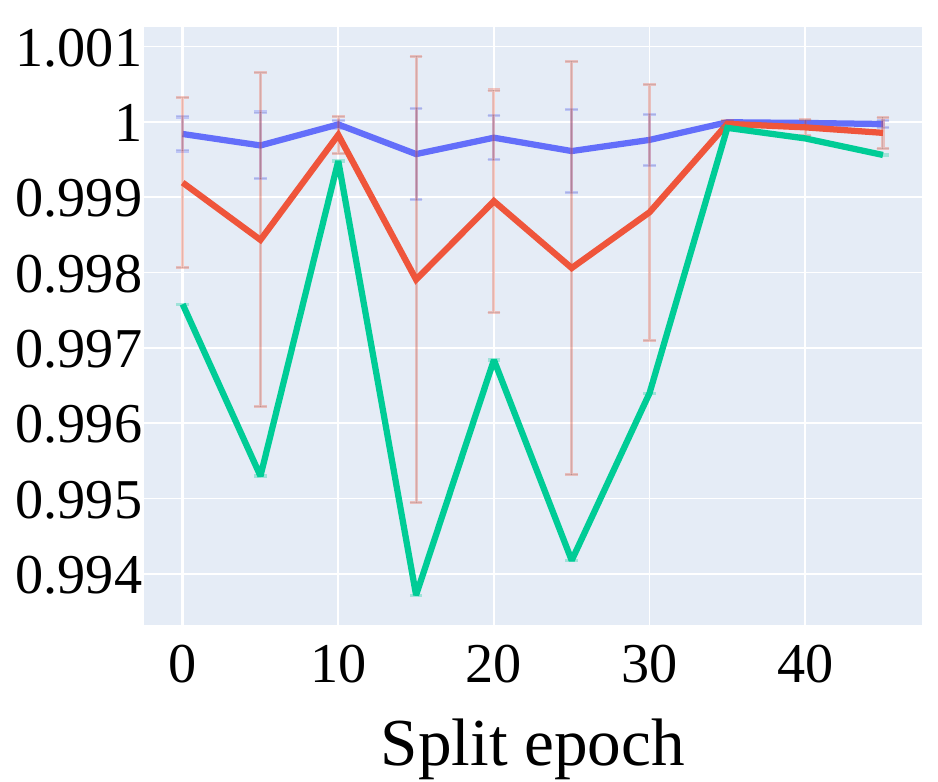}
        \caption{ResNet}
    \end{subfigure}\hfil
    \caption{\textbf{Top row:} Average accuracies at the endpoints of training (Endpoint), inside the convex mode defined by the endpoints (Convex), and at random directions around endpoints (Random). Notice that after a late enough split, samples in the hull are better than the endpoints. \textbf{Bottom row:} Average training accuracy over runs (Accuracy) compared to common points predicted correctly within a mode (Intra mode) and common points predicted correctly between hulls (Inter mode). We see that for models with an imperfect training accuracy, intra mode agreement is substantially lower than accuracy, and inter mode agreement is even lower. This suggests that endpoints of a mode are not similar functions.}
    \label{fig:agreement}
\end{figure}

Let $C_i^j$ be the set of correct predictions for endpoint $W_T^i$ of a convex hull, where $j$ indexes over different hulls. The set of correct predictions on the training set common to all endpoints is $S_j = \bigcap_{i} C_i^j$. The set of correct predictions common between hulls is $S^* = \bigcap_{j} S_j$. Comparing $\lvert S^* \rvert$ to $\mathbb{E}[\lvert S_j\rvert]$ provides a proxy measurement of the similarity between the functions for two different hulls. Figure~\ref{fig:agreement} reveals that there is a substantial gap between $\lvert S^* \rvert$, $\mathbb{E}[\lvert S_j\rvert]$, and $\mathbb{E}[\lvert C_i^j \rvert]$ for functions that do not interpolate the data perfectly, suggesting only a fraction of the predictions within a mode are shared, and only a fraction of those are shared across hulls. We also see that accuracy improves inside the convex hull, so it is not the case that the loss improvement we saw previously is driven by better performance on a common set of points shared across the whole mode. Rather, it seems like there is a large collection of points in the data for which it is possible to flip predictions easily without incurring increases in the loss. Additionally, the poor agreement between hulls suggests that permutations either are not a sufficient explanation for the similarity between hulls, or that they only preserve the function on a relatively small set of predictions.

\section{Discussion}
In this chapter, we present some dichotomous evidence for local convexity in the neural network loss landscape. We show that trajectories that split after a point early in training define a convex hull of low loss in the loss landscape. On one hand, this might be taken as evidence for the loss landscape being (weakly) convex after this early point in training. But on the other, we cannot detect such a point with simple proxies for convexity. In particular, our measures indicate no increase in the convexity of the function before or after the split point that gives rise to LMC. Failure to measure a difference could be due to sparse sampling of our metrics, but it might also indicate that neural network optimization is much more convex than we might expect. Probing the mode, it seems the convexity in training loss and error is not due to a large subset of shared predictions on the training set, but rather a diverse set of functions that are amenable to interpolation.

Prior work has shown that, on the scale of the optimization trajectory, the neural network loss landscape does not appear convex~\citep{li2018visualizing, frankle2020revisiting, vlaar2022can}, yet a large convex hull appears at the end of training. Given that hulls only agree on a small subset of predictions, yet interpolation is possible, studying the behavior of the convex hull while training on the subset of the data corresponding to predictions that are and are not shared may provide more insight.

There may also be a relationship between LMC and neural collapse \citep{papyan2020prevalence}, a recently identified tendency in classifiers to collapse the output feature space to a set of class means over the course of training. It could be the case that interpolating within a given convex hull corresponds to minimal disturbances to class means, but interpolating between convex hulls corresponds to changing the location of class means in feature space. If such a relationship existed, neural collapse might provide an early metric for detecting the first iteration after which a convex hull will exist, which would allow for parallelization of training without the extra computational cost needed to identify the LMC regime.


The deep nature of this phenomenon, and its commonality across settings, suggests that there must be a similarly common explanation in terms of the optimization of neural networks. So the critical question is why? Why do we see this high dimensional convex mode across all neural networks studied? And why is it still the case even given the constant nonconvexity and the functional dissimilarity?

We find a clue to answering the convex mode puzzle through the low-rank dynamics of deep linear models. In particular, \citet{arora2019implicit} show that, under certain technical assumptions, one can derive an equation for the dynamics of singular values and vectors of the product matrix. In effect, larger singular values grow larger faster, so there is an implicit low-rank dynamic with increasing depth.

Thus, we hypothesize that similar low-rank dynamics in nonlinear neural networks are responsible for LMC. Intuitively, the idea is that as large singular values become stable, when trajectories are split in the LMC procedure, the dynamics of those top vectors remain stable. Thus at the end of training, models that can be averaged will share their top singular vectors. The following chapter details this development.






\chapter{Approaching Deep Learning through the Spectral Dynamics of Weights \\ \citet{yunis2024approaching}}\label{chp:spectral_dynamics}
{We study the spectral dynamics of weights---the behavior of singular values and vectors during optimization---showing that they clarify and link many phenomena in deep learning. Through extensive experiments, covering small-scale ``grokking'' to large-scale tasks like image classification with ConvNets, image generation with UNets, speech recognition with LSTMs, and language modeling with Transformers, we identify a consistent bias with three key ingredients. First, singular values evolve unequally leading to rank minimization. As a result, top singular vectors stabilize well before the end of training, and lastly this happens without displaying alignment between neighboring layers used in several theoretical results. We show how this bias tracks the transition to generalization in grokking. We demonstrate more generally that weight decay enhances rank minimization beyond its role as a norm regularizer in practical systems. Moreover, we show that these spectral dynamics distinguish random label training from true labels, offering a novel perspective on this longstanding conundrum. Additionally, these dynamics reveal structure in well-performing sparse subnetworks (lottery tickets) and the shape of the loss surface through linear mode connectivity. Our findings suggest that spectral dynamics provide a coherent view that links the behavior of neural networks across diverse settings.}

\section{Introduction}

\begin{figure}[t]
  \centering
  \begin{subfigure}[b]{0.28\columnwidth}
    \centering
    \includegraphics[width=\columnwidth]{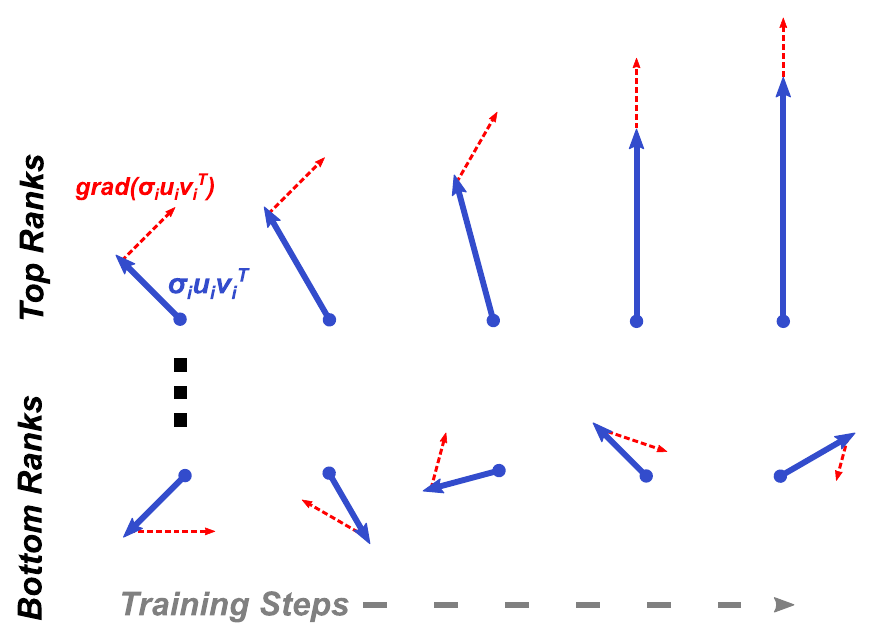}
    \caption{SV Schematic}\label{fig:teaser-a}
  \end{subfigure}
  \begin{subfigure}[b]{0.19\columnwidth}
    \centering
    \includegraphics[width=\columnwidth]{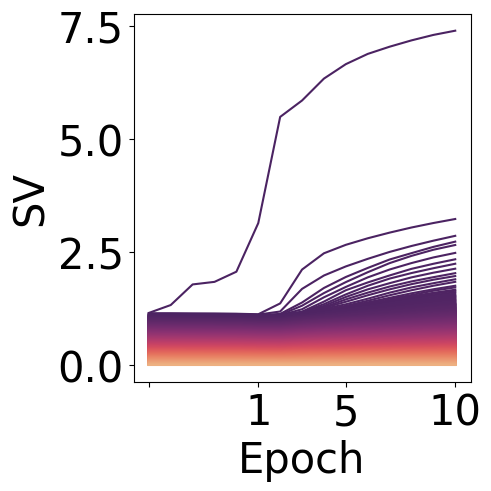}
    \caption{SVs}\label{fig:teaser-b}
  \end{subfigure}
  \begin{subfigure}[b]{0.51\columnwidth}
    \centering
    \includegraphics[width=\columnwidth]{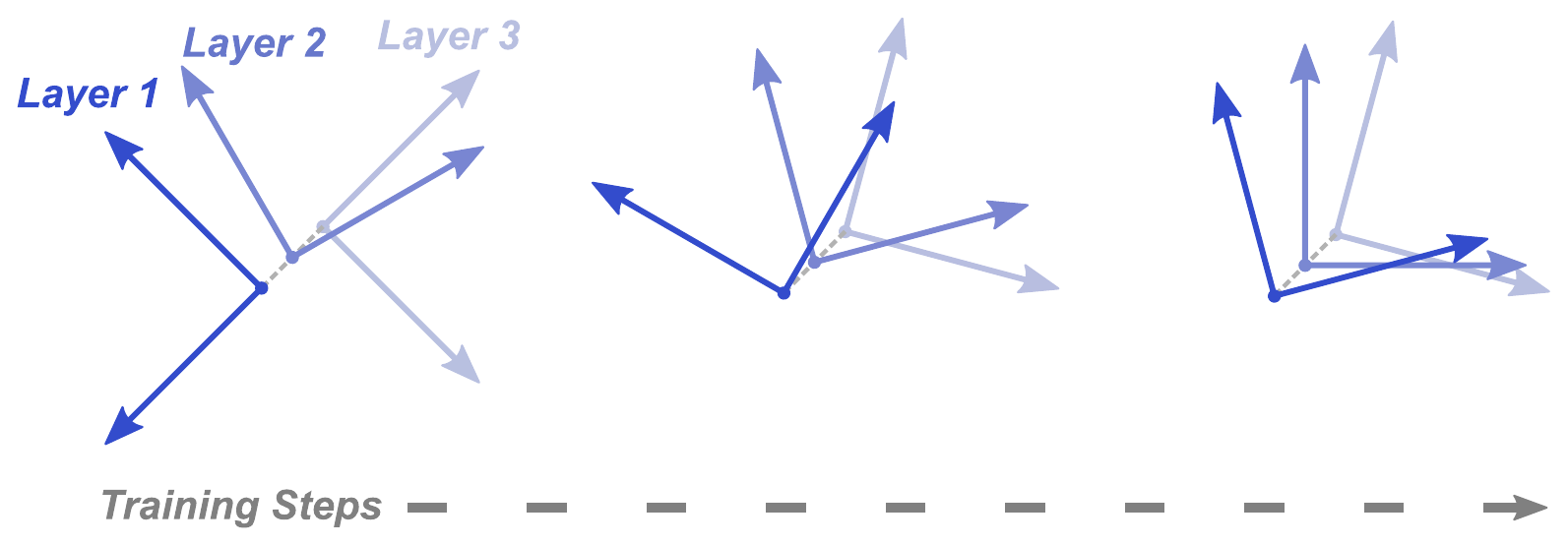}
    \caption{Alignment Schematic}\label{fig:teaser-c}
  \end{subfigure}
  \caption{\subref{fig:teaser-a} Schematic for the spectral dynamics of a weight matrix. As training proceeds, top singular vectors become stable and top singular values grow disproportionately large. \subref{fig:teaser-b} Singular value evolution for a single intermediate matrix parameter in a Transformer, where each line is a single singular value. Darker colors correspond to larger singular values while lighter ones correspond to smaller values. We see a disproportionate trend where large singular values grow larger faster. \subref{fig:teaser-c} Previous works use SV basis alignment between layers to prove similar theoretical results, however actual alignment between consecutive layers is weak, which we describe in Section~\ref{sec:spectral_dynamics}. We explore these spectral dynamics of weights and connect them to generalization, regularization, and seemingly unrelated phenomena like linear mode connectivity.}\label{fig:teaser}
\end{figure}

We find a task- and architecture-agnostic view to link many disparate phenomena in deep learning, simultaneously studying image classification with ConvNets, image generation with UNets, speech recognition with LSTMs and language modeling with Transformers. Through extensive experiments, we examine the dynamics (i.e., evolution over training) of singular values and singular vectors of weight matrices---the spectral dynamics of weights. We observe a few key properties: singular values evolve unequally, with larger ones evolving faster. As a result, top singular vectors stabilize much faster. Finally, for top ranks we see alignment between neighboring layers' singular vectors, though this varies somewhat with the setting. We preview these phenomena in Figure~\ref{fig:teaser}. We are motivated to study these dynamics specifically as optimization is the fundamental process driving deep learning~\citep{nagarajan2019uniform, zhang2021understanding}, neural networks at their core are chained matrices and nonlinearities, and the SVD is fundamental to every matrix. The following paragraphs detail how these properties connect to generalization and the mode connectivity phenomenon.

As a test bed for understanding generalization, \citet{power2022grokking} introduce the ``grokking'' phenomenon, where a small-scale model applied to arithmetic tasks initially achieves essentially zero training loss but performs poorly on validation data, then with much more training suddenly minimizes the validation loss (see Figure~\ref{fig:grokking}). In particular, \citet{nanda2023progress} showed that in modular arithmetic, the feature learning that occurs during grokking could be reverse-engineered entirely from the final weight matrices. Although this description is precise, it is limited to arithmetic tasks. In Section~\ref{sec:grokking}, we notice a task-agnostic view of grokking, 
observing that the drop in validation loss during grokking coincides with the simultaneous discovery of low-rank solutions across all weight matrices in the network, whether it be modular arithmetic or image classification. We also find that this transition relies on weight decay, echoing existing works~\citep{lyu2023dichotomy, liu2023omnigrok} as neither grokking nor low-rank weights occur without sufficient weight decay.

Though the common tie between low-rank weights and generalization in grokking suggests an intriguing explanation for generalization, grokking is typically studied on synthetic tasks with very small-scale models like single-layer Transformers or small MLPs and requires very particular hyperparameter settings~\citep{power2022grokking, nanda2023progress, gromov2023grokking, kumar2023grokking}. If our perspective is to be useful, we should obtain similar results in larger systems. Thus, we turn to common empirical tasks drawn from the literature like image classification, image generation, speech recognition and language modeling as well as varied and larger networks like VGG~\citep{simonyan2014very}, UNet~\citep{ronneberger2015u}, LSTM~\citep{hochreiter1997long} and multi-layer Transformers~\citep{vaswani2017attention}.

In Section~\ref{sec:spectral_dynamics}, we demonstrate that spectral dynamics are biased toward effective rank minimization across various practical neural networks in complex settings. Although this behavior echoes theoretical predictions in the deep linear setting, we find that the behavior of networks disagrees with a common theoretical assumption about low-rank dynamics: the alignment of singular vectors in consecutive layers~\citep{saxe2014exact, arora2018optimization, arora2019implicit, milanesi2021implicit}. Thus, the rank minimization mechanism differs from what the theory describes. Our hyperparameters are drawn from existing literature, thus the trend toward rank minimization coincides with well-generalizing networks across settings, echoing the observations in grokking.

One particularly notable ingredient for grokking was an extreme level of weight decay. In Section~\ref{sec:weight-decay}, we empirically connect rank minimization to weight decay, showing that weight decay promotes rank minimization across architectures and tasks, echoing findings in Section~\ref{sec:grokking}. In addition, in some cases, weight decay also appears to promote singular vector alignment in consecutive weight matrices despite the nonlinearities between layers. Although weight decay explicitly penalizes norm, studying spectral dynamics allows us to observe these effects on rank and alignment. Such effects may lead to better generalization bounds by understanding the reason weight decay is useful, as the most obvious norm-based explanations are insufficient~\citep{andriushchenko2023we}.

To further probe the rank-generalization connection, we turn to the classic memorization experiments of \citet{zhang2021understanding}, who demonstrated that even small networks can memorize random labels, thus any arguments about generalization cannot be capacity-based alone. In Section~\ref{sec:connections} and Appendix~\ref{sec:random_labels}, we show that training with random labels leads to high-rank solutions, while rank with true labels is much lower. We also find that while random labels do not align consecutive layers, true labels do, which is surprising given the nonlinearities between layers. This echoes prior discussion on rank and generalization. Through spectral dynamics, we see a clear distinction between generalizing and memorizing networks, which provides a foothold toward better theoretical understanding.

Our results suggest that viewing neural networks through the lens of spectral dynamics can shed light on several generalization-related phenomena, but there are broader connections. We return to the topic of the previous chapters. Linear mode connectivity (LMC) and its connection to the lottery ticket hypothesis~\citep{frankle2018lottery, nagarajan2019uniform, frankle2020linear, neyshabur2020being}. We find that global magnitude pruning, a standard procedure for finding lottery tickets, preserves top singular vectors and acts like low-rank pruning. We also see that the ability to interpolate between models in LMC strongly correlates with sharing top singular vectors. With these results, we note that the two phenomena can be seen as aspects of the spectral dynamics of weights, bringing them under the umbrella of Chapter~\ref{chp:convex_mode_connectivity}. For detailed discussions, see Section~\ref{sec:connections} and Section~\ref{sec:beyond-generalization}.

To summarize the discussion above, we study spectral dynamics along a few lines:
\begin{itemize}
    \item \textbf{How does generalization interact with rank?} We find grokking, a phenomenon of sudden generalization, is intimately linked to rank minimization (Section~\ref{sec:grokking}).
    \item \textbf{What about generalization and rank in standard settings?} We show rank minimization is a general phenomenon in more complex tasks and architectures (Section~\ref{sec:spectral_dynamics}).
    \item \textbf{If rank and generalization are tightly connected, what about common regularizations?} We find that the extremely common weight decay, a norm regularization, enhances the rank minimization behavior (Section~\ref{sec:weight-decay}), further tying rank to generalization.
    \item \textbf{Given the connection between rank and generalization, what about pure memorization?} We see that memorizing random labels through training yields high-rank, unaligned parameters compared to true-label training (Section~\ref{sec:connections}).
    \item \textbf{If it is so central, how does rank minimization relate to the puzzle of linear mode connectivity?} Top singular vectors are preserved when performing magnitude pruning and while linearly interpolating between connected modes (Section~\ref{sec:connections}), which provides an explanation for the complex convex mode connectivity of Chapter~\ref{chp:convex_mode_connectivity}.
\end{itemize}

All of these phenomena and effects have previously been studied in isolation to varying degrees. By approaching deep learning through spectral dynamics, we aim to inform existing small-scale theoretical and experimental results, and link them to a much broader literature. Our hope is that this will provide a deeper footing for further theory and practice.

\section{Related Work}
\subsection{Singular Value Dynamics}

Prior work on deep linear networks~\citep{arora2019implicit, milanesi2021implicit} suggests that rank minimization may better describe implicit regularization in deep matrix factorization than simple matrix norms. See \cite{arora2018optimization} (Appendix~A) for a detailed argument. However, a critical assumption in these works is ``balanced initialization.'' This means that for consecutive matrices $W_i$ and $W_{i+1}$ in the product matrix $\prod_j W_j$, we have $W^\top_{i+1}W_{i+1} = W_iW_i^\top$ at initialization. Decomposing these matrices with SVDs and leveraging orthogonality leads to matching left and right singular vectors between consecutive matrices. See Appendix~\ref{sec:balancedness} for a detailed explanation. Consequently, the product of the diagonals will evolve in a closed-form manner, with larger singular values growing faster than smaller ones. As shown by~\cite{arora2019implicit}, this translates to rank-minimizing behavior with increasing depth in the matrix products. This formula is also empirically validated for linear matrix factorization problems. Similar results have been derived for tensor products and other structured settings~\citep{saxe2014exact, yaras2023invariant}. \citet{ji2019gradient} show that alignment between layers will happen specifically for deep linear networks with infinite training. Still, there is no reason to believe standard networks obey this balancedness condition under practical initialization procedures. In Section~\ref{sec:spectral_dynamics}, we explore how these conclusions and assumptions hold for much larger, practical neural networks that are far from linear.

\subsection{Low-Rank Properties}\label{sec:low_rank}

Another line of research focuses on more general low-rank biases. Early work explored norms as an implicit bias~\citep{gunasekar2017implicit}. Theoretical analyses reveal that norms or closed-form functions of weights might be insufficient to explain implicit regularization, but they do not necessarily contradict the possibility of rank minimization~\citep{razin2020implicit, vardi2021implicit}. Numerous studies investigate low-rank biases in various matrices, including the Jacobian~\citep{pennington2018emergence}, weight matrices~\citep{le2021training, martin2020heavy, martin2021implicit, frei2022implicit, ongie2022role}, Gram matrix~\citep{huh2022low}, and features~\citep{yu2023compressing, feng2022rank}. Additionally, research suggests that dynamics influence the decay of rank~\citep{li2020towards, chen2023stochastic, wang2023implicit}. Orthogonally, weight decay has a long history as a regularizer explicitly penalizing parameter norm, which can be used for norm-based generalization bounds~\citep{bartlett1996valid}, but these bounds do not seem to explain the success of practical systems~\citep{nagarajan2019uniform, jiang2019fantastic}. Some works establish connections between weight decay and rank minimization in idealized settings~\citep{ziyin2022exact, galanti2022sgd, zangrando2024neural, ergen2023path, parhi2023deep, shenouda2023vector}, which may be connected to generalization~\citep{razin2020implicit}. We are particularly interested in how far these connections extend beyond ideal settings to practice.

Though these prior results are interesting in their own right, our goal is to link them together on a much broader suite of experiments and show further unexplored consequences.

\section{Common Quantities}\label{sec:background}
Throughout the rest of the chapter we will rely on a number of common quantities for our analysis. Inspired by work in the deep linear case~\citep{saxe2014exact, arora2019implicit, milanesi2021implicit, yaras2023law}, we track the evolution of singular values for individual weight matrices. To gain a high-level overview of all matrix parameters, we compute the (normalized) effective rank of a matrix $W$~\citep{roy2007effective} with rank $R$ as
\begin{align}\label{eqn:normalized-effective-rank}
    & \text{EffRank}(W) := -\sum_{i=1}^R \frac{\sigma_i}{\sum_j \sigma_j} \log \frac{\sigma_i}{\sum_j \sigma_j}\enspace,\\
    & \text{NormEffRank}(W) := \frac{\text{EffRank}(W)}{R}\enspace,
\end{align}
where $\sigma_i$'s are the singular values of matrix $W$ and $\text{EffRank}(W)$ is the entropy of the normalized singular value distribution. As the probability mass concentrates, the effective rank decreases. We plot $\text{NormEffRank}(W)$ to compare across layers and time.

\begin{figure*}[!t]
  \centering
  \begin{subfigure}[b]{0.24\linewidth}
    \centering
    \includegraphics[width=\linewidth]{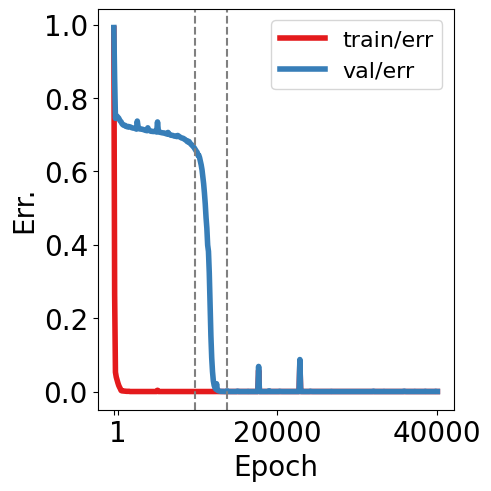}
    \\
    \includegraphics[width=\linewidth]{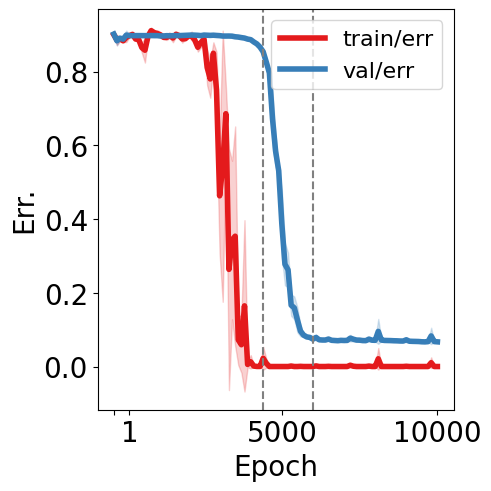}
    \caption{Error}
  \end{subfigure}\hfil
  \begin{subfigure}[b]{0.24\linewidth}
    \centering
    \includegraphics[width=\linewidth]{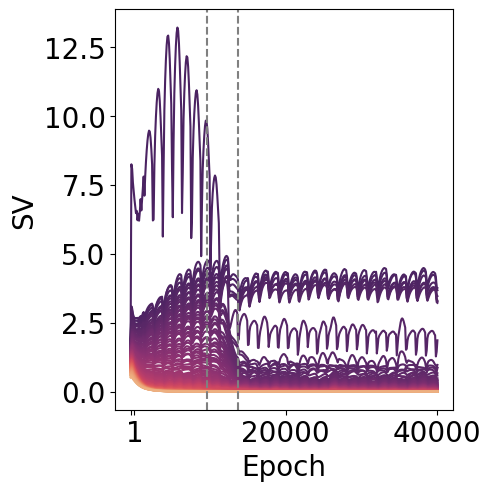}
    \\
    \includegraphics[width=\linewidth]{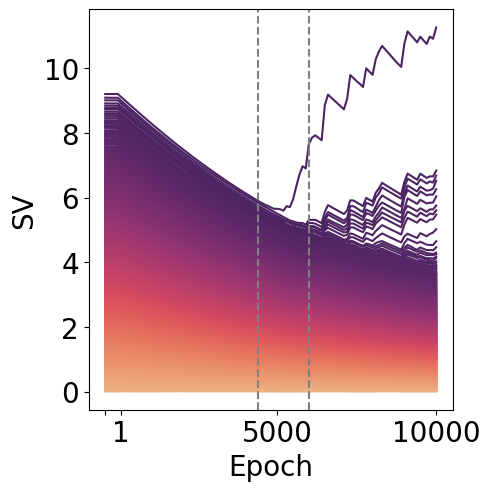}
    \caption{SV Evolution}
  \end{subfigure}\hfil
  \begin{subfigure}[b]{0.24\linewidth}
    \centering
    \includegraphics[width=\linewidth]{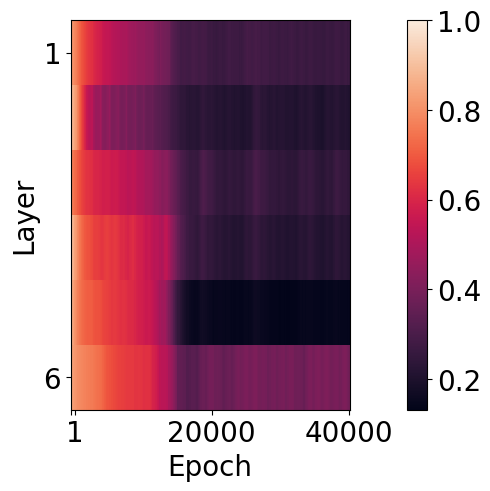}
    \\
    \includegraphics[width=\linewidth]{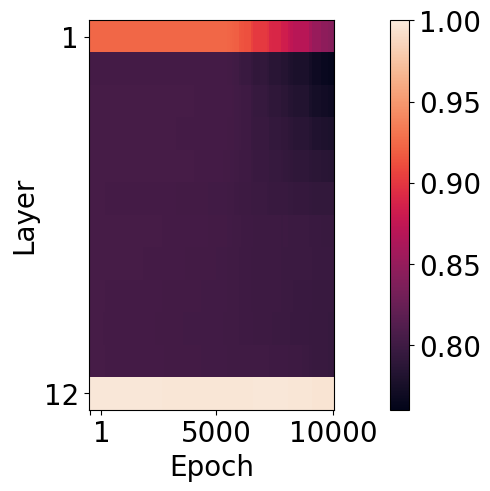}
    \caption{Effective Rank}
  \end{subfigure}\hfil
  \begin{subfigure}[b]{0.24\linewidth}
    \centering
    \includegraphics[width=\linewidth]{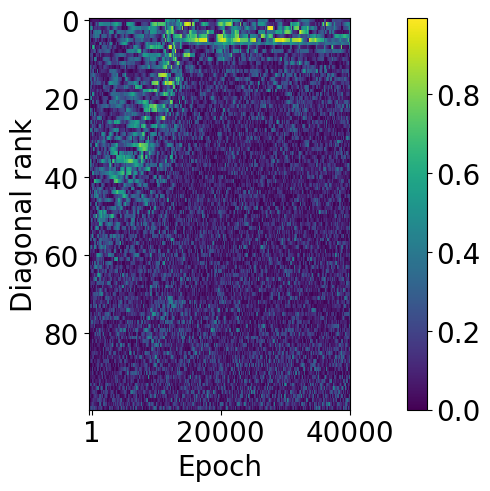}
    \\
    \includegraphics[width=\linewidth]{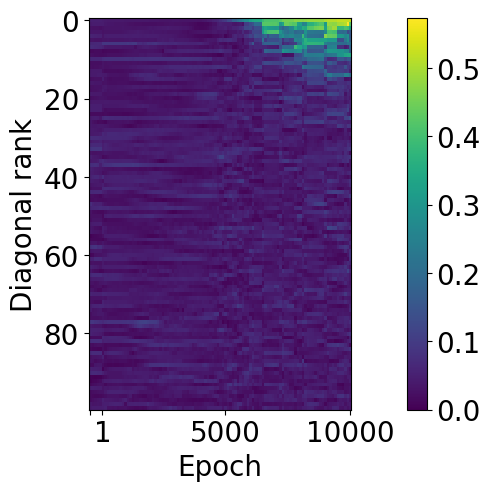}
    \caption{Alignment}
  \end{subfigure}
  \caption{\textbf{Grokking and Spectral Dynamics.} \textbf{Top row:} Grokking for Transformers on modular addition~\citep{nanda2023progress} for a single seed due to different convergence times. See Figure~\ref{fig:grokking-multiseed} for averaging over multiple seeds. \textbf{Bottom row:} Grokking for a 12-layer MLP on MNIST~\citep{fan2024deep}. \textbf{(a)} Training and validation error. Shaded regions correspond to standard deviation, where large deviations are due to different seeds converging at different times. \textbf{(b)} A visualization of singular value evolution for the first attention parameter and the second MLP layer, where each line represents a single singular value and the color represents the rank. \textbf{(c)} Effective rank of all layers (matrix parameters) over time (Eqn.~\ref{eqn:normalized-effective-rank}). \textbf{(d)} A visualization of the alignment (Eqn.~\ref{eqn:alignment-matrix}) between the embedding and the first attention parameter, and the first and second MLP layers, where the $y$-axis corresponds to index $i$ of the diagonal. We see that grokking co-occurs with a transition to low-rank weights. In addition, there is an alignment that begins early in training that evolves up the diagonal. In the image classification case, we see a similar rank transition, though alignment does not evolve up the diagonal (see Figure~\ref{fig:alignment-matrix-evolution}).}
  \label{fig:grokking}
\end{figure*}

In addition, inspired by the assumptions of balancedness made by prior work~\citep{arora2018optimization, arora2019implicit}, we examine the alignment of consecutive weight matrices in the Transformer. To examine and quantify this alignment between SVDs of consecutive matrices in a network at training time $t$, i.e.,
\begin{align*}
    W_i = \sum_{j=1}^R \sigma_j(t) u_j(t) v_j(t)^\top,\qquad 
    W_{i+1} = \sum_{k=1}^R \sigma'_k(t) u'_k(t) {v'_k}(t)^\top\enspace, 
\end{align*}
we compute the inner products between neighboring singular vectors,
\begin{equation} \label{eqn:alignment-matrix}
    A(t)_{jk} = \lvert \langle u_{j}(t), v'_{k}(t) \rangle \rvert\enspace,
\end{equation}
where the absolute value is taken to ignore sign flips in the SVD computation. We then plot the diagonal of this matrix $A(t)_{ii}~\forall~i \leq 100$ over time, which we focus on because we observed the most signal here. See Appendix~\ref{app:alignment-evolution} for more details. These plots give us a sense as to whether simultaneous diagonalization occurs at least in the top ranks. For exact details on how alignment is computed for different architectures and layers more complex than the fully connected case, see Appendix~\ref{app:experimental-details}.

We not only employ the alignment matrix defined in Eqn.~\ref{eqn:alignment-matrix}, but also derive and plot a scalar measure. We focus on the top diagonal entries as we observed they typically contained the most structure (see Figure~\ref{fig:alignment-matrix-evolution} for an example):
\begin{equation} \label{eqn:alignment-measure}
    a(t) = \frac{1}{10} \sum_{i=1}^{10} A(t)_{ii}.
\end{equation}
For specific details on calculating this measure in diverse architectures and complex layers (beyond fully connected layers), please see Appendix~\ref{app:experimental-details}.

\section{Grokking and Rank Minimization}\label{sec:grokking}

\citet{power2022grokking} first noticed a surprising phenomenon they called ``grokking'' where models quickly fit the training data on toy tasks, then after a long period of training, very quickly generalize on the validation data (see Figure~\ref{fig:grokking}). Later, others found that this phenomenon can occur on a logarithmic timescale~\citep{thilak2022slingshot} in simpler models and different datasets~\citep{liu2022towards, gromov2023grokking, kumar2023grokking, xu2023benign}. In addition, weight decay seems to be a critical ingredient~\citep{lyu2023dichotomy, liu2023omnigrok, tan2023understanding}.

Motivated by experimental results that show the importance of weight decay for grokking \citep{power2022grokking, lyu2023dichotomy, liu2023omnigrok}, and theoretical work that connects low-rank weights, generalization and weight decay~\citep{razin2020implicit, galanti2022sgd, timor2023implicit, yaras2023law, zangrando2024neural}, we evaluate the potential connection between parameter rank and grokking in neural networks. Low-rank weights would naturally complement other descriptions such as Fourier decomposition \mbox{\citep{nanda2023progress}}, the simplification of linear decision boundaries~\citep{humayun2024deep}, the connection to double descent~\citep{davies2022unifying}, and the discovery of a sparse solution~\citep{merrill2023tale}.

We replicate grokking in two settings: a single-layer Transformer for modular addition~\citep{nanda2023progress}, and a 12-layer MLP for MNIST image classification~\citep{fan2024deep} (see Appendix~\ref{app:experimental-details} for details). We select these settings as they are somewhat more realistic than other tasks, require more complex architectures, and display grokking on a linear time scale.

In Figure~\ref{fig:grokking}, we see a tight connection: the sudden drop in validation loss coincides precisely with the onset of low-rank behavior in the singular values. Examining inter-layer alignment during training, we observe that the final low-rank solution gradually emerges from the model's middle ranks. As we show later, in the absence of weight decay, no low-rank solution seems to develop (Figure~\ref{fig:grokking-modadd}). Additionally, when using 90\% of the data and no weight decay, generalization still coincides with effective rank minimization. \citet{fan2024deep} noted that in deep MLPs, grokking coincided with a feature rank decrease, which agrees with the parameter rank decrease that we see here. The careful reader will also note that \citet{nanda2023progress} previously showed that the particular solution found in modular addition is a low-rank Fourier decomposition, so our observations on low-rank weights will follow, yet the same structure also applies to the MLP where such reverse-engineering is difficult. In the following sections, we argue that rank minimization is a perspective that can be applied in more complex settings when one does not know what to look for in the weights. It may thus be possible to interpret the neural network via the top ranks~\citep{praggastis2022svd}. Specifically, we will show in Section~\ref{sec:weight-decay} that large amounts of weight decay have a strong rank-regularizing effect, so one way to understand grokking is that the network first memorizes the training data, and with further training the rank regularization of weight decay pushes toward generalization.

\section{Spectral Dynamics Across Tasks}\label{sec:spectral_dynamics}
Inspired by the results on grokking and prior work on deep linear networks that studies the evolution of the SVD of the weight matrices~\citep{saxe2014exact, arora2018optimization, arora2019implicit, milanesi2021implicit, yaras2023invariant}, we apply the same analysis to larger, more practical systems. We show that the trends we saw in the analysis of grokking mostly hold true across networks and tasks at a much larger scale, though our findings do deviate from theoretical settings.

\subsection{Methodology}
Our experiments aim to examine reasonably sized neural networks across a variety of tasks. We select models and tasks that are representative of current applications. Specifically, we consider:
\begin{itemize}[leftmargin=8mm]
    \item \textbf{Image classification} with CNNs (VGG-16~\citep{simonyan2014very}) on CIFAR10~\citep{krizhevsky2009learning};
    \item \textbf{Image generation} through diffusion with UNets~\citep{ronneberger2015u} on MNIST \citep{lecun1998mnist};
    \item \textbf{Speech recognition} with LSTMs~\citep{hochreiter1997long} on LibriSpeech \citep{panayotov2015librispeech}; and
    \item \textbf{Language modeling} with Transformers~\citep{vaswani2017attention} on Wikitext-103~\citep{merity2016pointer}.
\end{itemize}

Training hundreds of runs for each of the settings above is computationally expensive, limiting the scale of models we can explore. We primarily adopt hyperparameters from existing literature, with minor modifications for simplicity. This ensures that any correlations observed are likely a reflection of common practices, and not bias introduced on our part. We also provide evidence with larger language models (up to 3B parameters) in Appendix~\ref{sec:pythia}, from the Pythia suite~\citep{biderman2023pythia}.

The primary evidence in this section comes from computing the SVDs of weight matrices within the models; we disregard 1D bias and normalization parameters in our analysis. Previous research suggests that these parameters are not always crucial for performance~\citep{zhang2018fixup,mohan2019robust,karras2023analyzing}, and many large models do not use them~\citep{raffel2020exploring, grattafiori2024llama}. As the matrices are the vast majority of the parameters, we believe they are the primary object of interest and it is worthwhile focusing on them. Due to the large number of matrices in these models, we present plots of individual layers' matrix parameters and statistics that summarize behavior across layers for conciseness. We generated hundreds of thousands of plots during this study, making it impossible to include them all.  Full experimental details, including the choice of hyperparameters, are available in Appendix~\ref{app:experimental-details}.

\subsection{Effective Rank Minimization}

\begin{figure}[!t]
  \centering
  \begin{subfigure}[b]{0.23\columnwidth}
    \centering
    \includegraphics[width=\columnwidth]{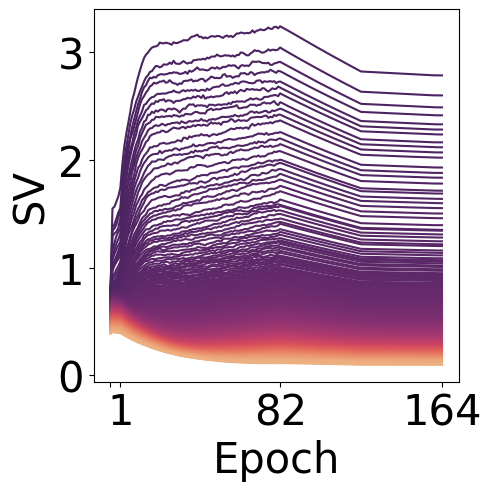}\\\includegraphics[width=\columnwidth]{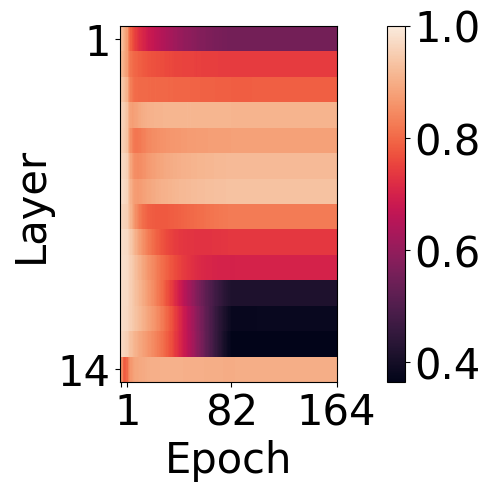}
    \caption{VGG}
  \end{subfigure}
  \begin{subfigure}[b]{0.23\columnwidth}
    \centering
    \includegraphics[width=\columnwidth]{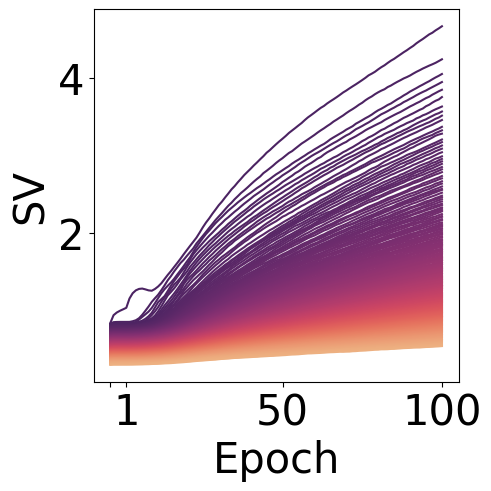}\\\includegraphics[width=\columnwidth]{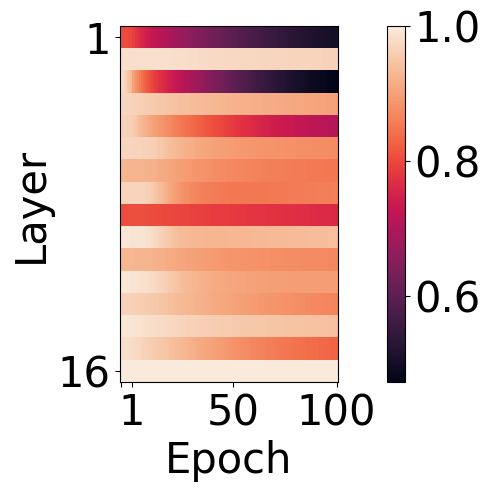}
    \caption{UNet}
  \end{subfigure}
  \begin{subfigure}[b]{0.23\columnwidth}
    \centering
    \includegraphics[width=\columnwidth]{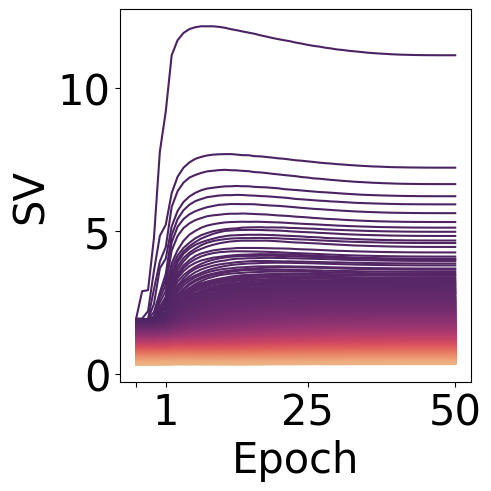}\\\includegraphics[width=\columnwidth]{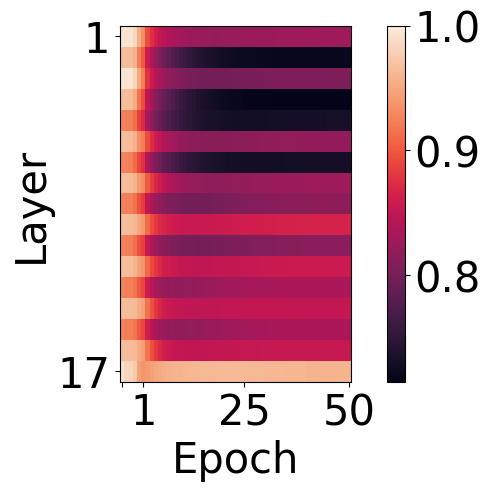}
    \caption{LSTM}
  \end{subfigure}
  \begin{subfigure}[b]{0.23\columnwidth}
    \centering
    \includegraphics[width=\columnwidth]{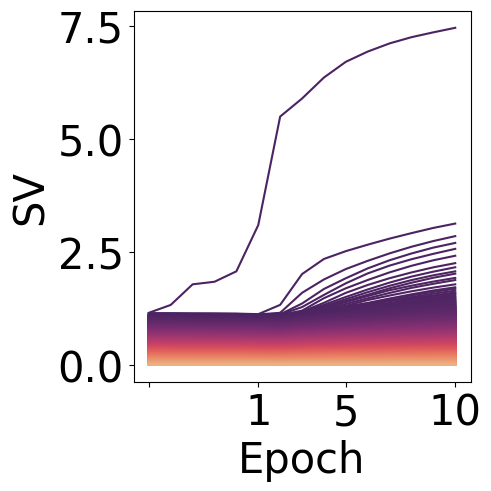}\\\includegraphics[width=\columnwidth]{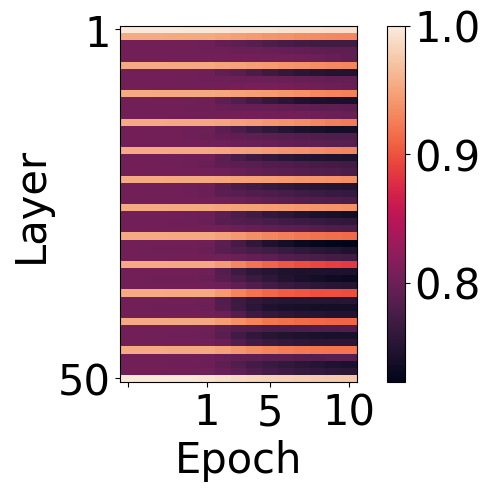}
    \caption{Transformer}
  \end{subfigure}
  \caption{\textbf{Top row:} Singular value evolution for a single matrix in the middle of each model. Each line represents a singular value, darker colors are larger values, while lighter colors are smaller values. Notice the unequal evolution where top singular values grow at a disproportionate rate. \textbf{Bottom row:} Normalized effective rank (Eqn.~\ref{eqn:normalized-effective-rank}) evolution visualized in color for all matrix parameters across architectures and time. As we move down the $y$-axis, the depth of the parameters in the model increases, while the $x$-axis tracks training time. The axis label "Layer" is shorthand for "matrix parameter". For example, in the Transformer we visualize each of the $W_q, W_k, W_v, MLP_1, MLP_2$ parameters for each block. Notice decreasing effective rank with training time across nearly all parameters, though the magnitude differs across layers (indicated by absolute color). The block-like patterns for VGG are likely due to different channel dimension sizes. The banding in the UNet, LSTM, and Transformer is due to the differences between convolutional and linear layers, residual block connections, and attention and fully connected layers, respectively. The sharp transitions through training in VGG are due to 10$\times$ learning rate decays.}
  \label{fig:normalized-effective-rank}
\end{figure}

Building on theoretical~\citep{saxe2014exact, arora2019implicit, milanesi2021implicit, boix2023transformers, yaras2023invariant} and empirical~\citep{dittmer2019singular, martin2020heavy, martin2021implicit, boix2023transformers} findings, we investigate effective rank minimization across larger models and a more diverse array of tasks. Figure~\ref{fig:normalized-effective-rank} reveals a consistent trend: the effective rank of network parameters generally decreases throughout training, regardless of the specific parameter or network architecture. This suggests a progressive ``simplification'' of the network as training progresses.

\begin{figure}[!t]
  \centering
  \begin{subfigure}[b]{0.24\columnwidth}
    \centering
    \includegraphics[width=\columnwidth]{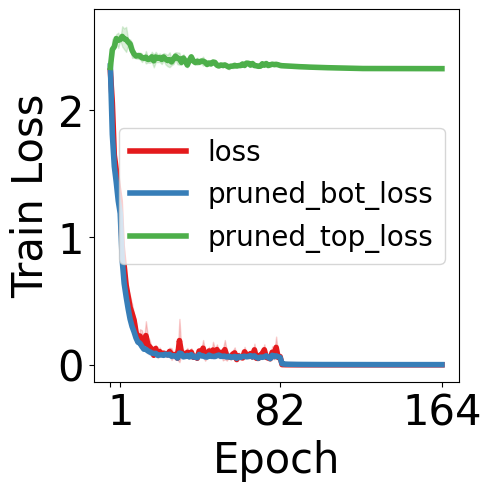}\\\includegraphics[width=\columnwidth]{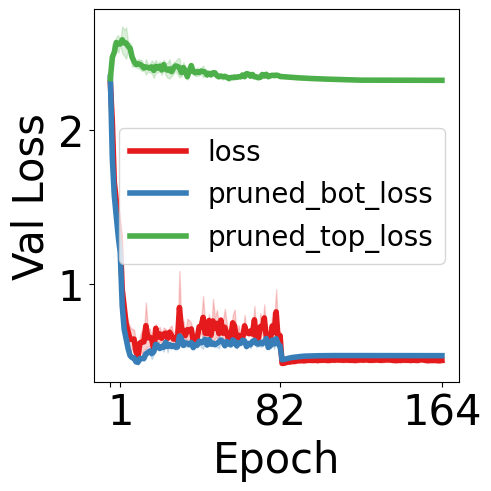}
    \caption{VGG}
  \end{subfigure}
  \begin{subfigure}[b]{0.24\columnwidth}
    \centering
    \includegraphics[width=\columnwidth]{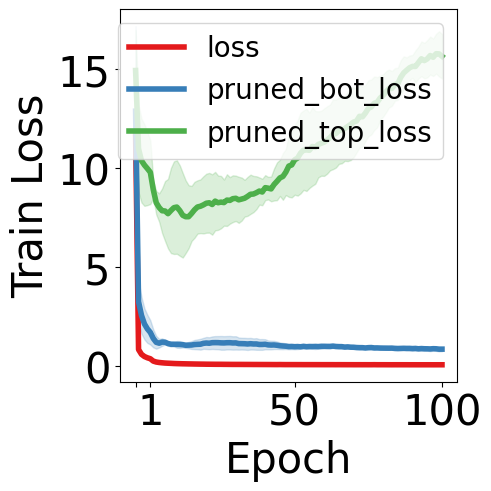}\\\includegraphics[width=\columnwidth]{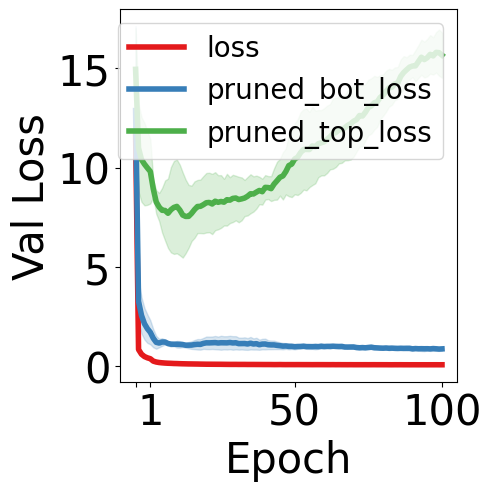}
    \caption{UNet}
  \end{subfigure}
  \begin{subfigure}[b]{0.24\columnwidth}
    \centering
    \includegraphics[width=\columnwidth]{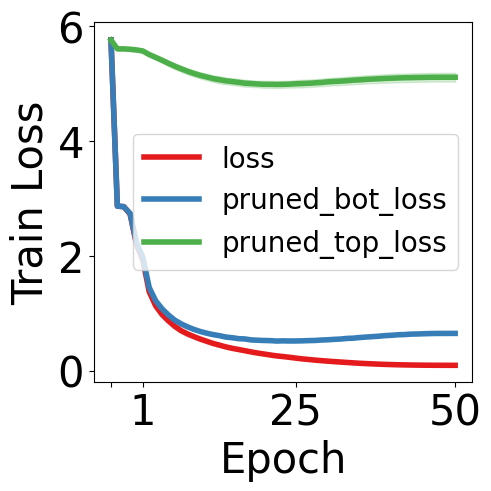}\\\includegraphics[width=\columnwidth]{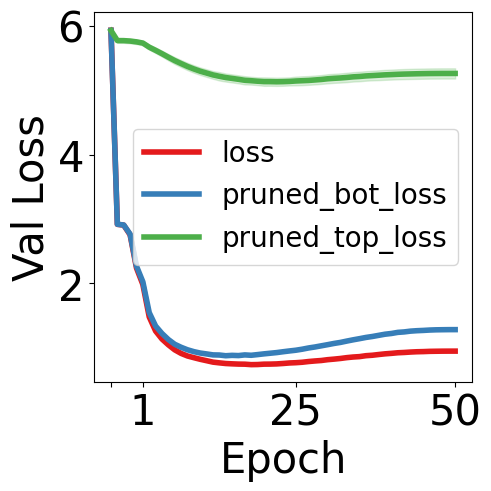}
    \caption{LSTM}
  \end{subfigure}
  \begin{subfigure}[b]{0.24\columnwidth}
    \centering
    \includegraphics[width=\columnwidth]{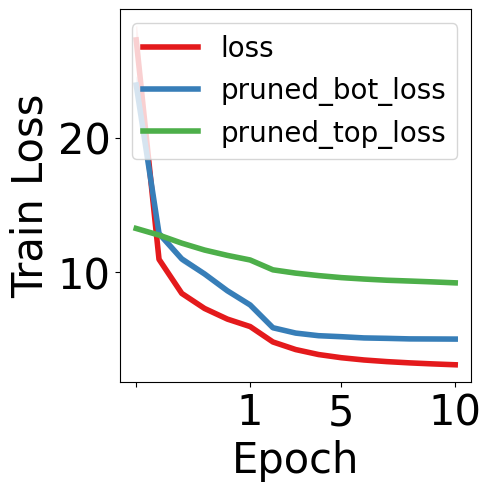}\\\includegraphics[width=\columnwidth]{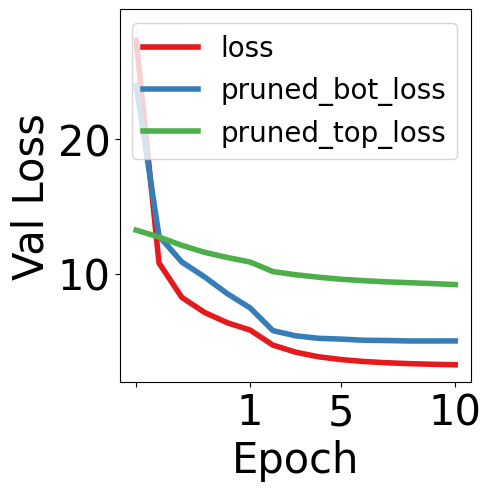}
    \caption{Transformer}
  \end{subfigure}
  \caption{\textbf{Top row:} Training loss. \textbf{Bottom row:} Validation loss. \textcolor{redplot}{Red} is the full model. \textcolor{blueplot}{Blue} is post-training pruning the bottom half of the SVD for every matrix in the model that is not the final layer. \textcolor{greenplot}{Green} is post-training pruning the top half of the SVD. Notice that for all models, keeping the top half of the SVD is close to the full model performance, supporting the idea that the top directions provide a good approximation to the function. We specifically avoid pruning the final layer as for some tasks it is so low-rank that pruning further affects performance in an anomalous way~\citep{frankle2020linear}.}
  \label{fig:pruned-performance}
\end{figure}

We further conduct a singular-value pruning experiment to explore the relationship between low-rank behavior and model performance. We prune either the top or bottom half of the singular values for each weight matrix in the network and then evaluate the pruned model at each training step. Given their importance in $\mathcal{L}^2$ space, we expect the top singular values to capture the information most critical to the network's function. Figure~\ref{fig:pruned-performance} confirms this, demonstrating that the pruned parameters, without further training, can closely approximate the full model's performance. It is not necessarily obvious that pruning would have this effect. In particular, simultaneously pruning lower components across all layers may lead to losing some critical signal that must be passed between layers, or it could be that small-magnitude singular values may provide some important regularizing noise. This result is reminiscent of prior work that uses low-rank approximations for efficiency or performance~\citep{yu2017compressing, sharma2023truth, chen2024truncating}, or work that explicitly optimizes for low-rank networks~\citep{wang2021pufferfish, schotthofer2022low}. Here we point out that the reason this is possible is due to the dynamics of the singular values themselves. In later sections, we will rely on this observation that large singular values are more critical to the function of the network.

\subsection{Alignment of Singular Vectors Between Layers}

Similar to the analysis of grokking, we investigate the alignment between consecutive layers in the larger neural networks considered in this section.

\begin{figure}[!t]
  \centering
  \begin{subfigure}[b]{0.24\columnwidth}
    \centering
    \includegraphics[width=\columnwidth]{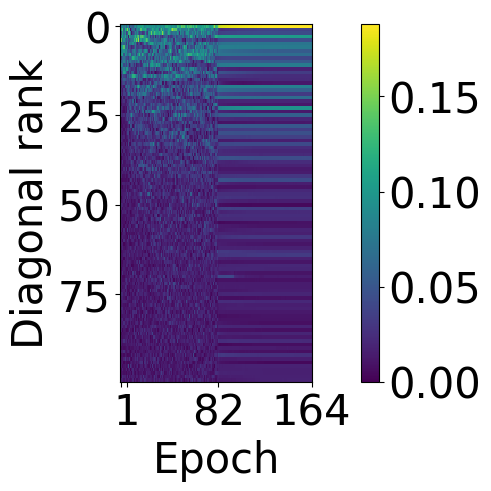}\\\includegraphics[width=\columnwidth]{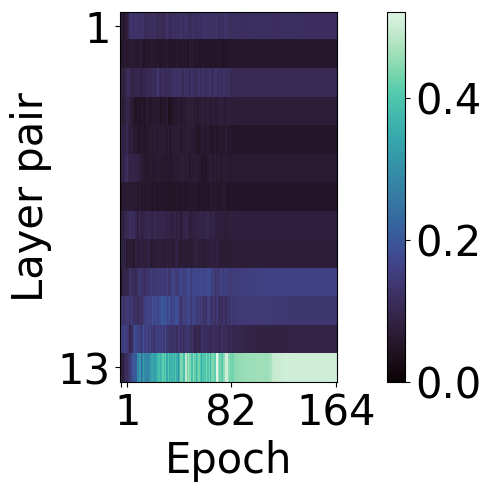}
    \caption{VGG}
  \end{subfigure}
  \begin{subfigure}[b]{0.24\columnwidth}
    \centering
    \includegraphics[width=\columnwidth]{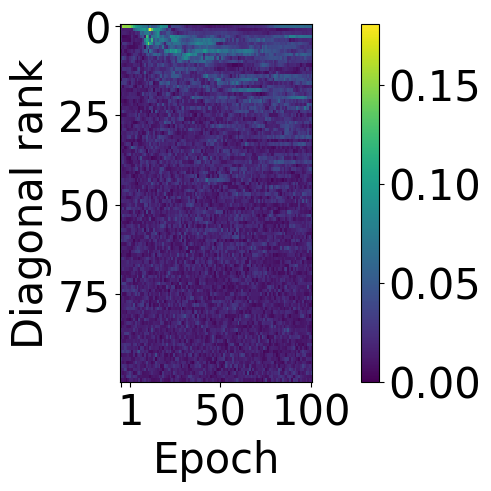}\\\includegraphics[width=\columnwidth]{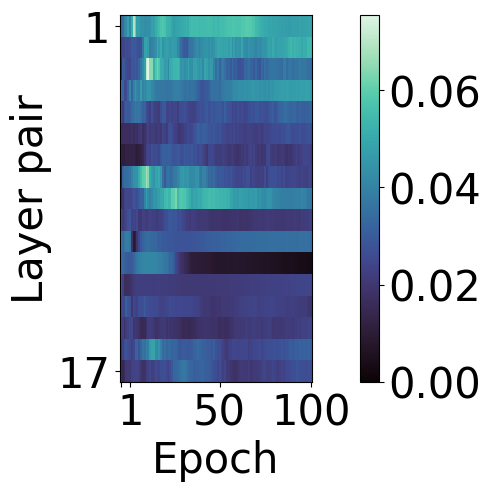}
    \caption{UNet}
  \end{subfigure}
  \begin{subfigure}[b]{0.24\columnwidth}
    \centering
    \includegraphics[width=\columnwidth]{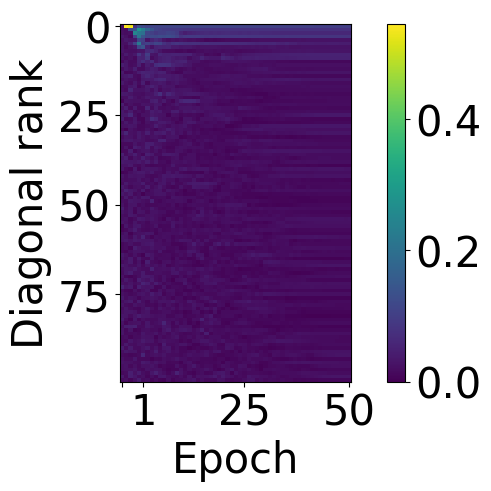}\\\includegraphics[width=\columnwidth]{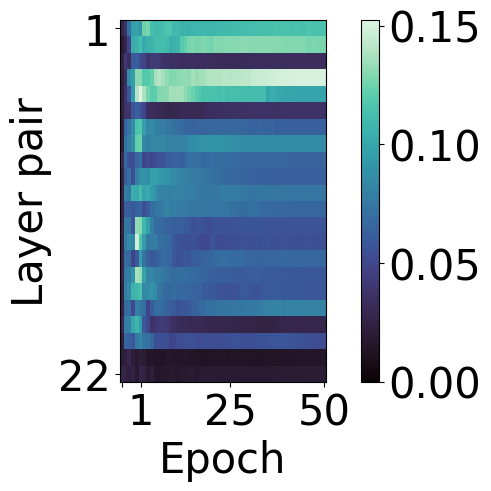}
    \caption{LSTM}
  \end{subfigure}
  \begin{subfigure}[b]{0.24\columnwidth}
    \centering
    \includegraphics[width=\columnwidth]{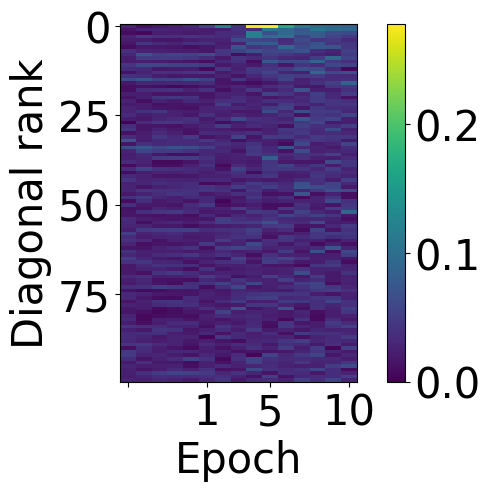}\\\includegraphics[width=\columnwidth]{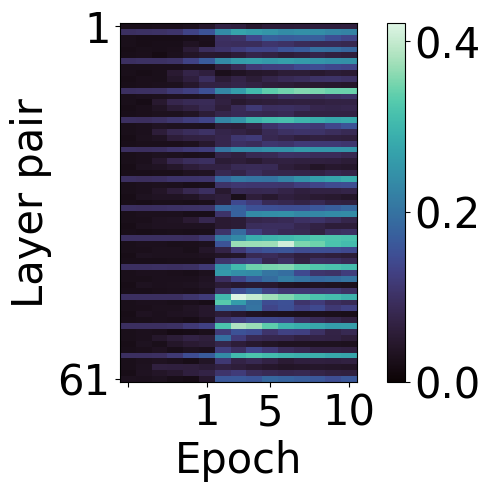}
    \caption{Transformer}
  \end{subfigure}
  \caption{Neighboring layer alignment of singular vectors. \textbf{Top row:} The diagonal of the alignment matrix $A(t)_{ii}$ (Eqn.~\ref{eqn:alignment-matrix}) vs.\ training time for a single pair of matrices in the middle of each model. We see a small amount of alignment in the top ranks between layers shortly after training begins, but this becomes more diffuse over time. \textbf{Bottom row:} Alignment metric (Eqn.~\ref{eqn:alignment-measure}) for pairs of matrices for depth vs.\ training time. It is hard to make out a global trend across models, though the LSTM shows a weak signal around Epoch 1 when the initial alignment occurs, and the Transformer case has a banding pattern with depth due to alignment between the query and key matrices that have no nonlinearity in between.}
  \label{fig:layer-alignment}
\end{figure}

Figure~\ref{fig:layer-alignment} reveals a key finding: the theoretical assumption of \textbf{balanced initialization}, which posits aligned singular value decompositions (SVDs) between weight matrices~\citep{arora2018optimization, saxe2014exact}, does not hold true at the start of training in these larger networks. Additionally, unlike in the linear case discussed in \citet{du2018algorithmic}, the alignment does not appear to remain static throughout training. However, a weak signal of alignment in the top ranks develops and disappears. We are very far from the theoretical settings of prior work~\citep{du2018algorithmic, arora2019implicit, mulayoff2020unique} that use balanced or near-zero initialization, and the strange trends mean that existing theoretical models do not capture the complexities of neural network training.

\section{The Effect of Weight Decay}\label{sec:weight-decay}
In light of the previously observed evolution of singular values, we investigate a proposed effect of weight decay. Though weight decay explicitly penalizes the norm of weights, there is evidence that complicates the connection between the norm and generalization for neural networks~\citep{razin2020implicit, andriushchenko2023we}, meaning that we do not have a full understanding as to why weight decay may be useful. Alternatively, some theoretical~\citep{boix2023transformers, razin2020implicit, yaras2023invariant, timor2023implicit, ongie2022role, galanti2022sgd, zangrando2024neural} and empirical works~\citep{galanti2022sgd, boix2023transformers} propose a connection between weight decay, generalization and the rank of matrices in constrained settings. Still, comprehensive evidence for larger empirical networks is missing.

\begin{figure*}[!t]
  \centering
  \begin{subfigure}[b]{0.24\linewidth}
    \centering
    \includegraphics[width=0.5\linewidth]{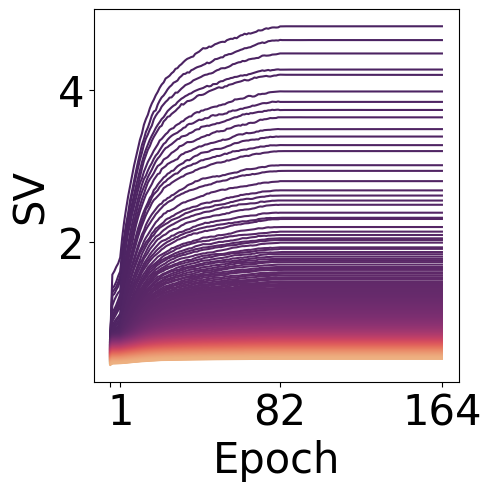}\hfil
    \includegraphics[width=0.5\linewidth]{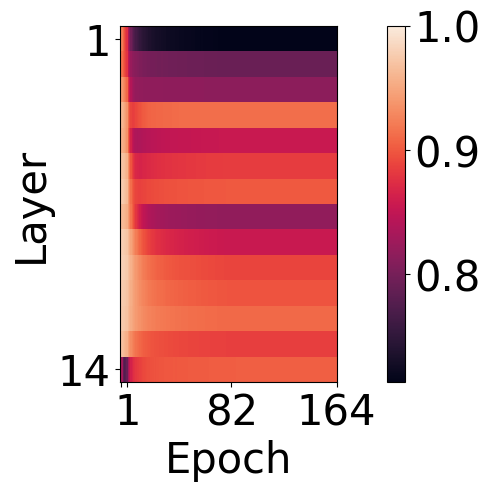}
    \\
    \includegraphics[width=0.5\linewidth]{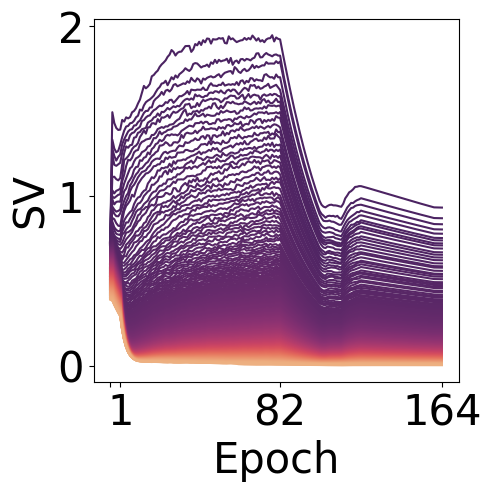}\hfil
    \includegraphics[width=0.5\linewidth]{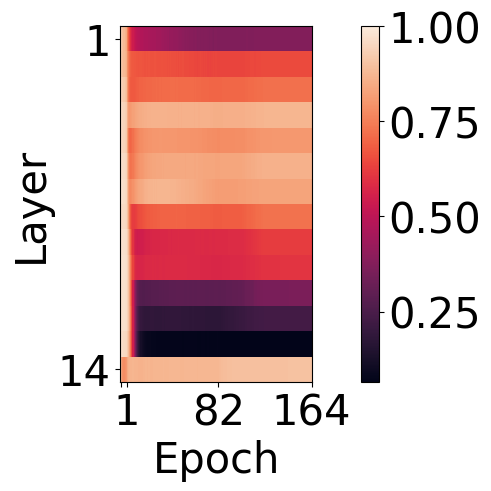}
    \\
    \includegraphics[width=0.5\linewidth]{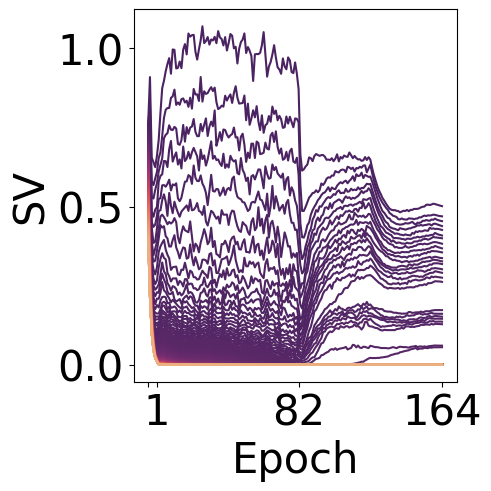}\hfil
    \includegraphics[width=0.5\linewidth]{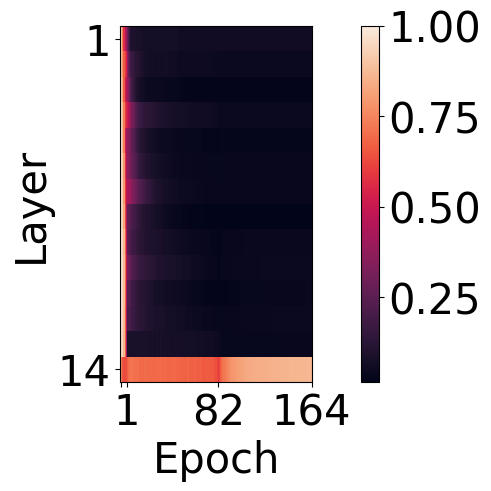}
    \\
    \includegraphics[width=0.5\linewidth]{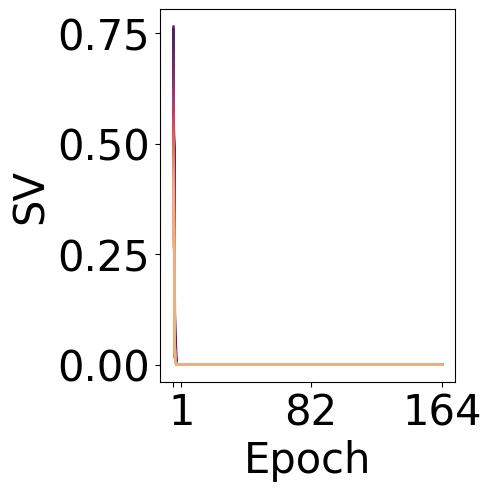}\hfil
    \includegraphics[width=0.5\linewidth]{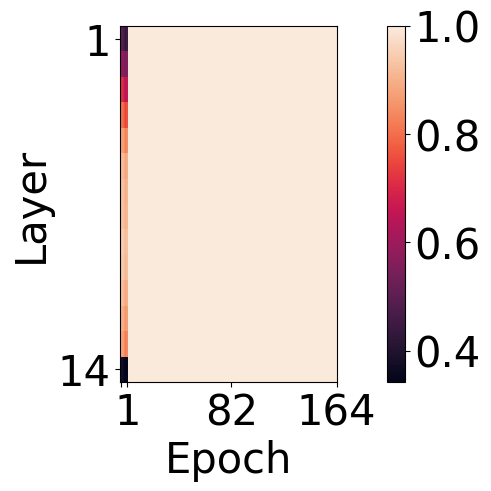}
    \caption{VGG}
  \end{subfigure}
  \begin{subfigure}[b]{0.24\linewidth}
    \centering
    \includegraphics[width=0.5\linewidth]{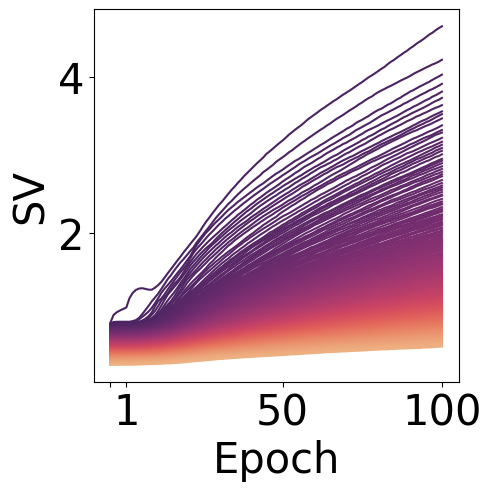}\hfil
    \includegraphics[width=0.5\linewidth]{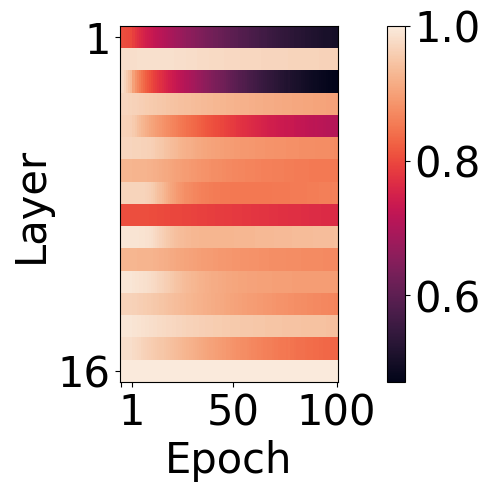}
    \\
    \includegraphics[width=0.5\linewidth]{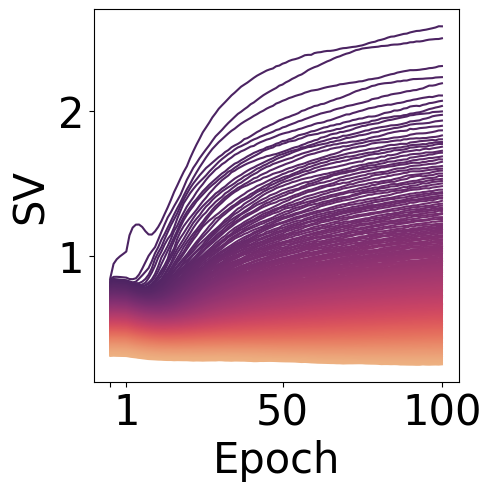}\hfil
    \includegraphics[width=0.5\linewidth]{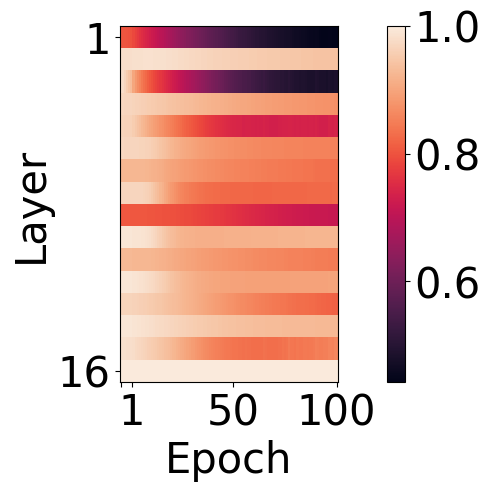}
    \\
    \includegraphics[width=0.5\linewidth]{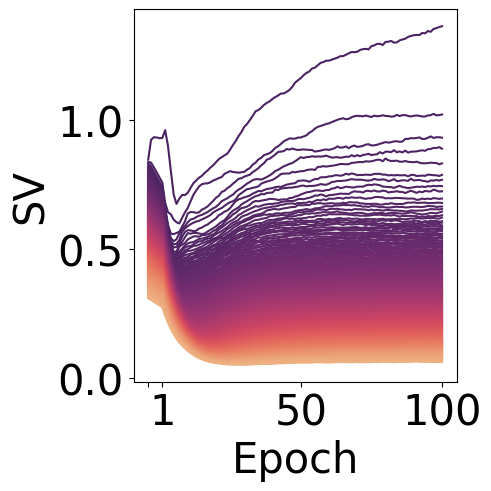}\hfil
    \includegraphics[width=0.5\linewidth]{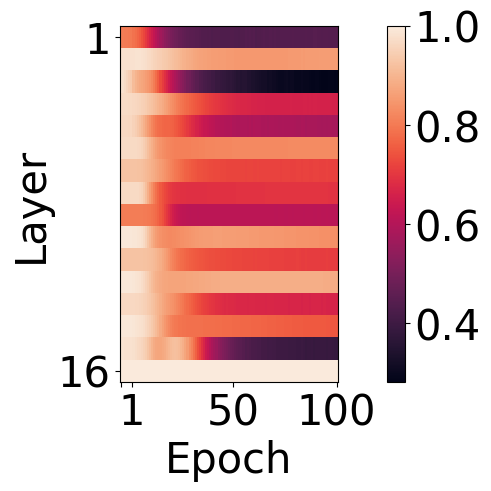}
    \\
    \includegraphics[width=0.5\linewidth]{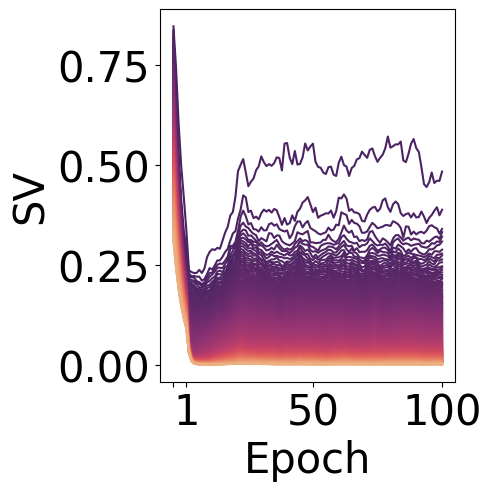}\hfil
    \includegraphics[width=0.5\linewidth]{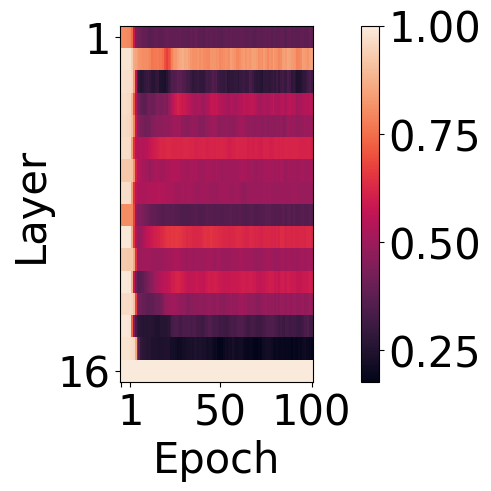}
    \caption{UNet}
  \end{subfigure}
  \begin{subfigure}[b]{0.24\linewidth}
    \centering
    \includegraphics[width=0.5\linewidth]{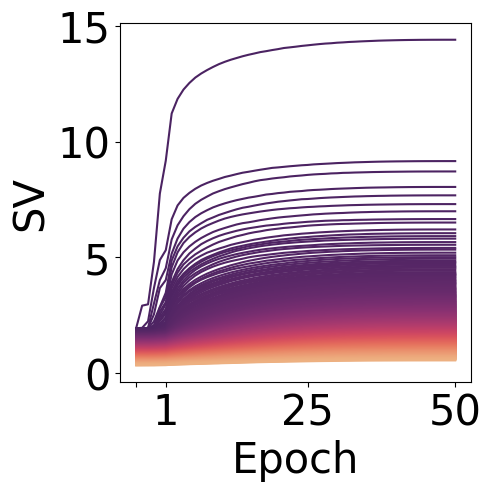}\hfil
    \includegraphics[width=0.5\linewidth]{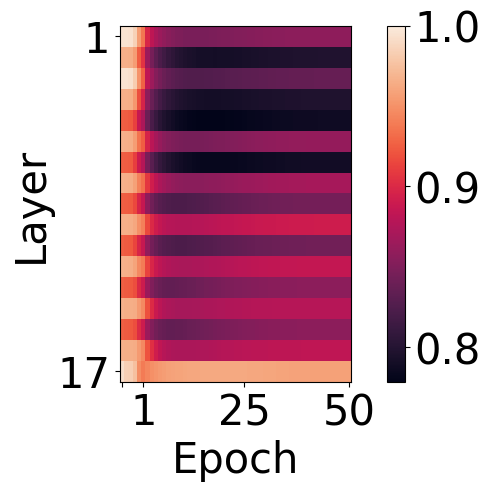}
    \\
    \includegraphics[width=0.5\linewidth]{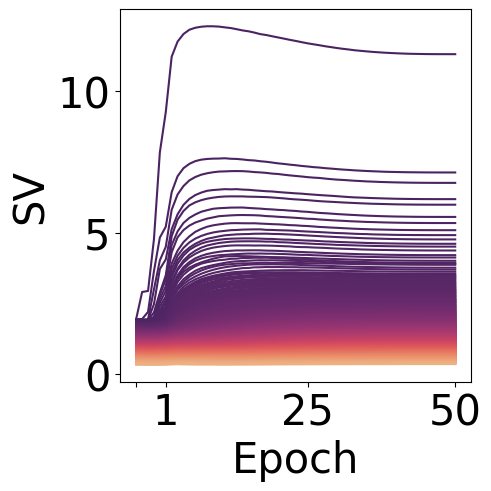}\hfil
    \includegraphics[width=0.5\linewidth]{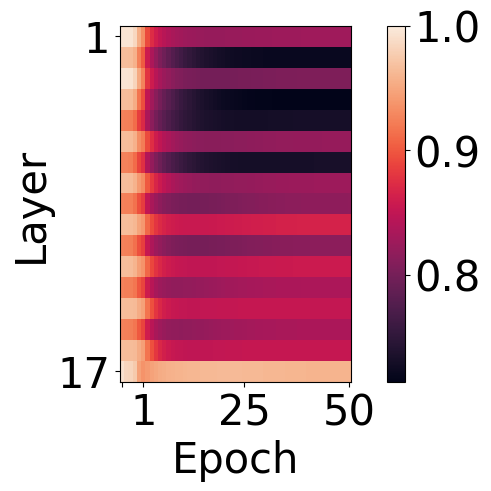}
    \\
        \includegraphics[width=0.5\linewidth]{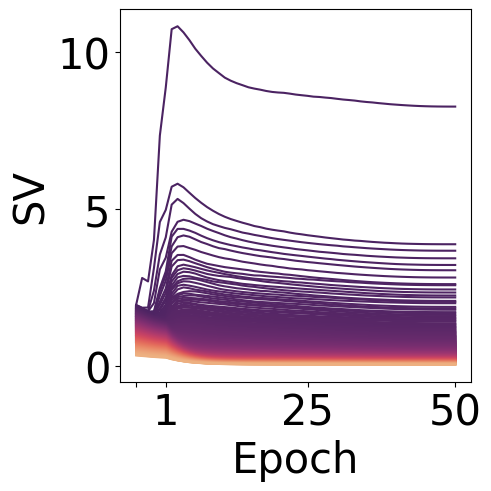}\hfil
    \includegraphics[width=0.5\linewidth]{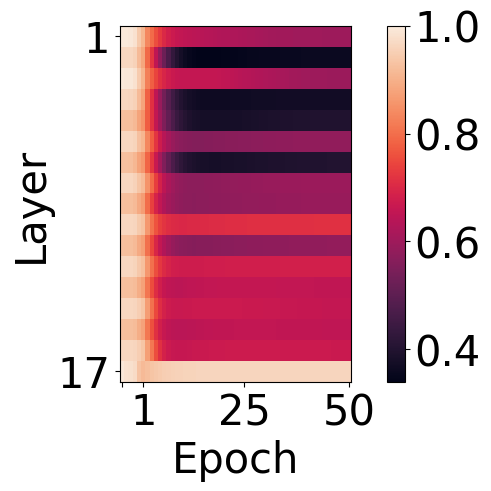}
    \\
    \includegraphics[width=0.5\linewidth]{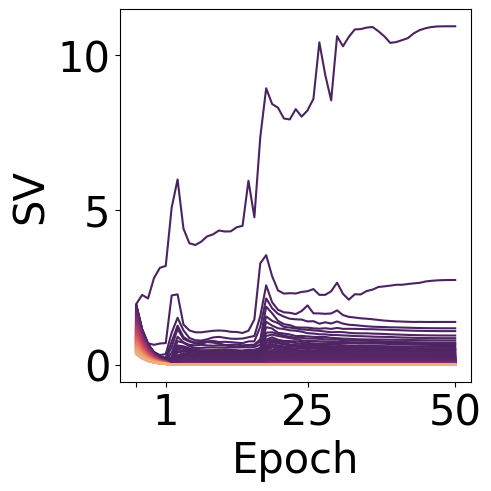}\hfil
    \includegraphics[width=0.5\linewidth]{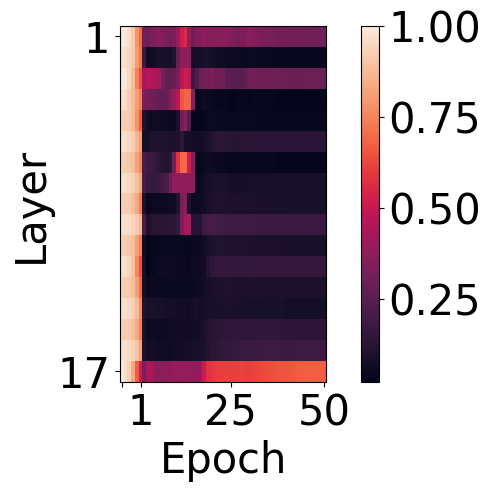}
    \caption{LSTM}
  \end{subfigure}
  \begin{subfigure}[b]{0.24\linewidth}
    \centering
    \includegraphics[width=0.5\linewidth]{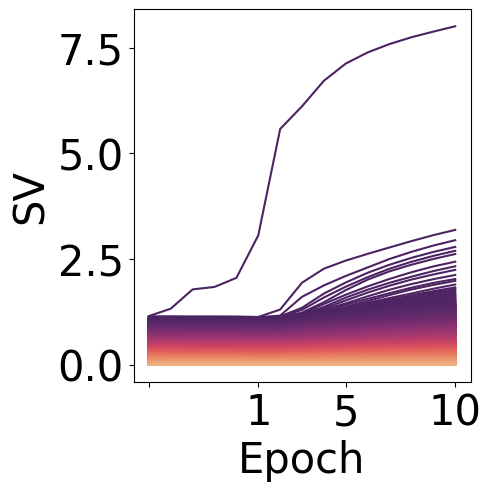}\hfil
    \includegraphics[width=0.5\linewidth]{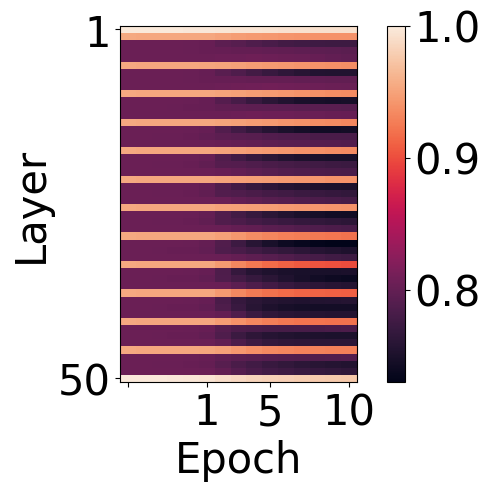}
    \\
    \includegraphics[width=0.5\linewidth]{./spectral_dynamics/figs/tfmr/weight_decay_0.1-sv-model.transformer_encoder.layers.7.linear1.weight_sv.png}\hfil
    \includegraphics[width=0.5\linewidth]{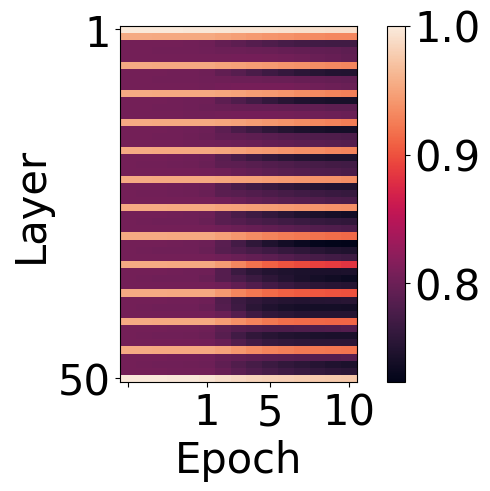}
    \\
    \includegraphics[width=0.5\linewidth]{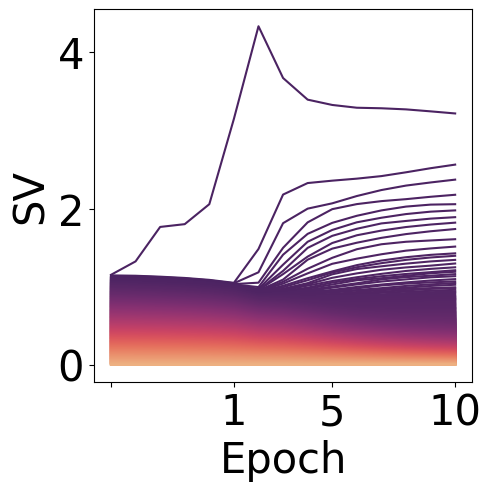}\hfil
    \includegraphics[width=0.5\linewidth]{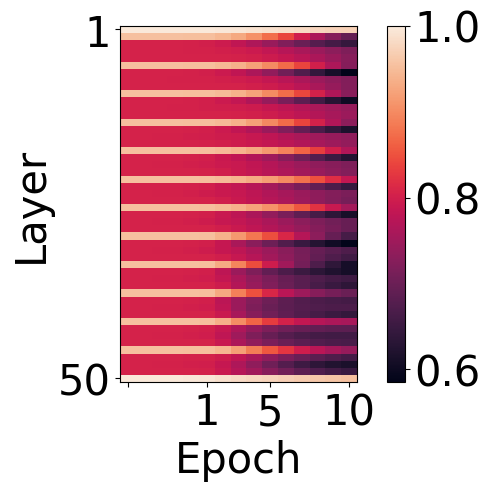}
    \\
    \includegraphics[width=0.5\linewidth]{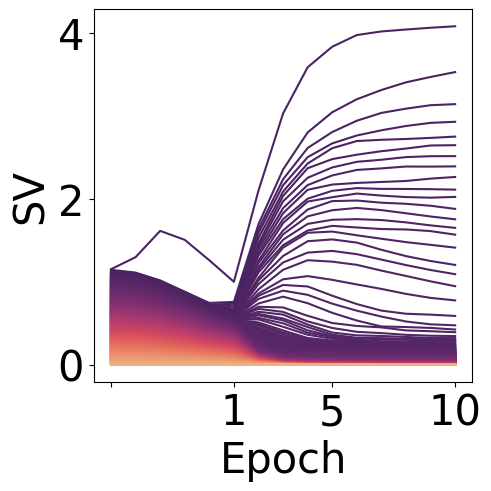}\hfil
    \includegraphics[width=0.5\linewidth]{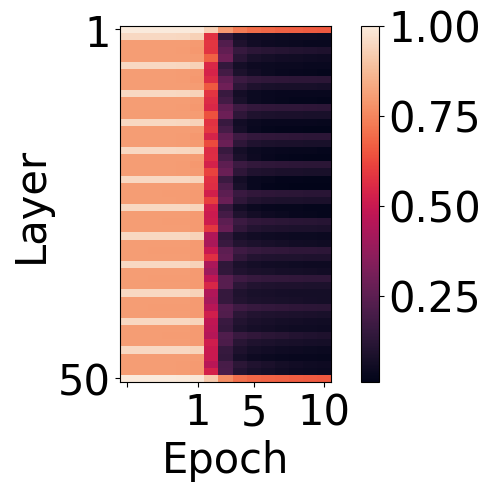}
    \caption{Transformer}
  \end{subfigure}
  \caption{Singular value evolution for a single matrix and normalized effective rank (Eqn.~\ref{eqn:normalized-effective-rank}) across matrices over time, where the rows use differing amounts of weight decay. From top to bottom, for VGG we use coefficients $\{ 0, 0.001, 0.01, 0.1 \}$, while for other networks we use coefficients $\{0, 0.1, 1, 10\}$. Higher weight decay coefficients promote more aggressive rank minimization. VGG uses SGD with momentum, while the rest use AdamW~\citep{loshchilov2018fixing}, which may explain the earlier norm collapse. We provide an additional version with shared color bars in Figure~\ref{fig:wd-effective-rank-shared}.}
  \label{fig:wd-effective-rank}
\end{figure*}

We speculate on the intuition behind the mechanism in more practical settings. In its simplest form, weight decay involves the optimization: $\text{arg min}_{W} \, \mathcal{L}(W) + \lambda \lVert W \rVert_F^2$, where $\lVert W \rVert_F^2 = \sum_{i=1}^R \sigma_i^2$, and $\sigma_i$ are the singular values of weight matrix $W$ with rank $R$. We saw previously that larger singular values of neural networks grow faster (Fig.~\ref{fig:normalized-effective-rank}, top row) and that the top singular vectors are much more useful for minimizing task loss than the bottom ones (Fig.~\ref{fig:pruned-performance}). Thus, with minor weight decay regularization, one straightforward solution for the network may be to minimize the rank of a given weight matrix while preserving the top singular values to minimize $\mathcal{L}(W)$. \citet{timor2023implicit} argue for a similar effect in simple systems: if all singular values are less than one, the norm of activations will shrink with depth, so as depth grows very large, any input signal will converge to zero, making it impossible to learn. Thus, it is better for a few singular values to be sufficiently large while the rest can be very small.

Figure~\ref{fig:wd-effective-rank} shows that adding weight decay produces this exact low-rank behavior, while too much weight decay leads to complete norm collapse. The exact choice of ``too much'' varies across architectures and tasks. Despite the low-rank regularization, we do not see particularly tight alignment (Eqn.~\ref{eqn:alignment-matrix}) in the top singular vectors, except in the highest weight decay Transformer (see Figure~\ref{fig:wd-alignment-score} in the Appendix). The alignment in this case is reminiscent of the balancedness condition~\citep{arora2018optimization, arora2019implicit, du2018algorithmic}, though the Transformer considered here has nonlinearities and a much more complex structure.

Orthogonally, we provide additional evidence in Appendix~\ref{app:experimental-details}, where Figure~\ref{fig:weight-decay-performance} shows that the solutions with moderate weight decay can perform better than without, while with very high weight decay models still generalize, even though they are much lower rank. To a very rough degree, lower rank leads to better generalization and a smaller training-validation gap, though when rank is pushed too low the performance collapses as the model can no longer use any capacity. In general, different tasks may have different minimal ranks necessary. Language modeling may require some form of memorization for long-tail words and thus a low-rank solution may be impossible, while 10-class image classification may not have this property. Still, we note that the role of weight decay in improving generalization is tied up with its function as a rank regularizer. In addition, although we lack precise tools to interpret complex models entirely, when there are only a few ranks per matrix, it may become possible to extend analysis efforts~\citep{nanda2023progress, praggastis2022svd} to more complex domains.

\section{Additional Connections}\label{sec:connections}
Here we briefly preview some connections between spectral dynamics and additional phenomena.

\begin{figure*}[!t]
  \centering
  \begin{subfigure}[b]{0.18\linewidth}
    \centering
    \includegraphics[width=\linewidth]{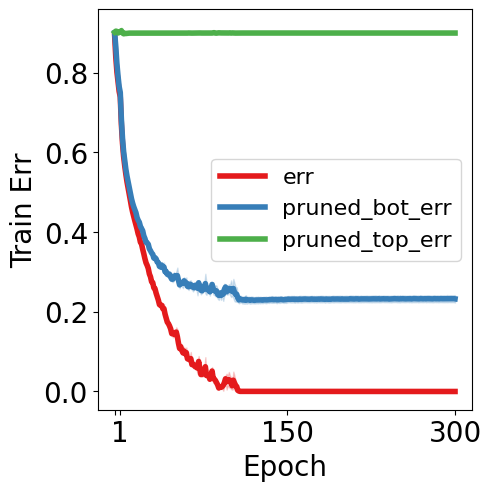}
    \\
    \includegraphics[width=\linewidth]{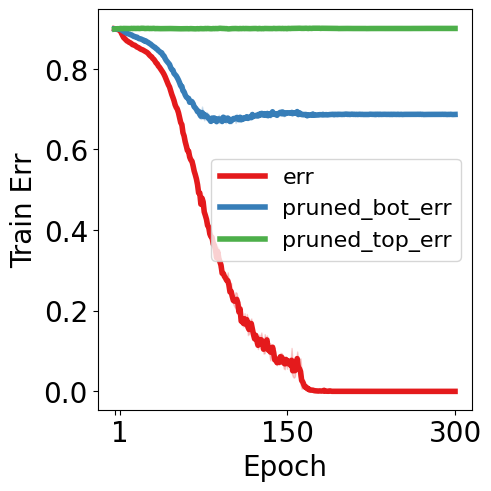}
    \caption{Train Err.}
  \end{subfigure}
  \begin{subfigure}[b]{0.18\linewidth}
    \centering
    \includegraphics[width=\linewidth]{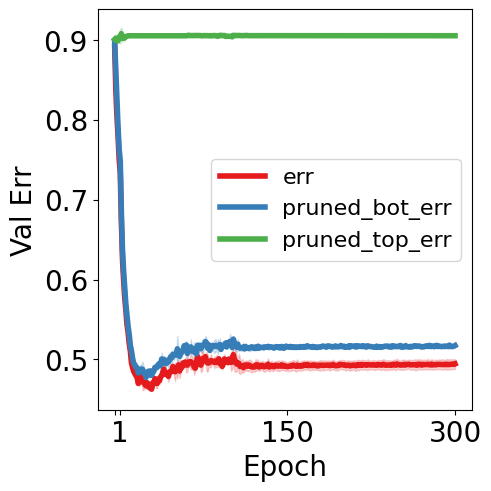}
    \\
    \includegraphics[width=\linewidth]{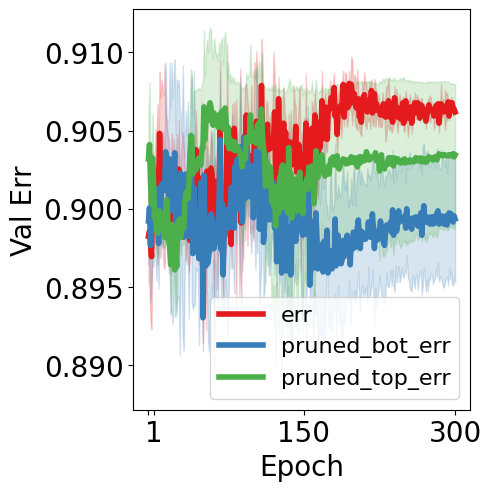}
    \caption{Val. Err.}
  \end{subfigure}
  \begin{subfigure}[b]{0.18\linewidth}
    \centering
    \includegraphics[width=\linewidth]{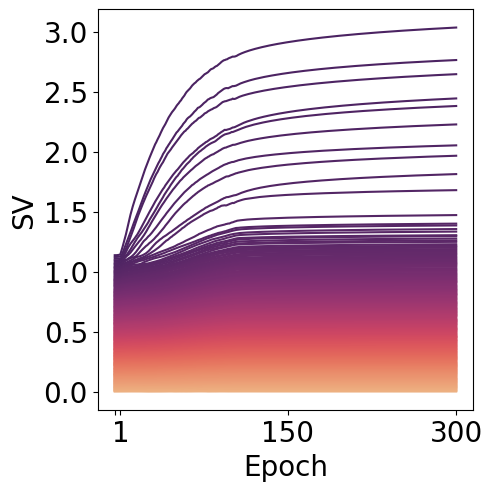}
    \\
    \includegraphics[width=\linewidth]{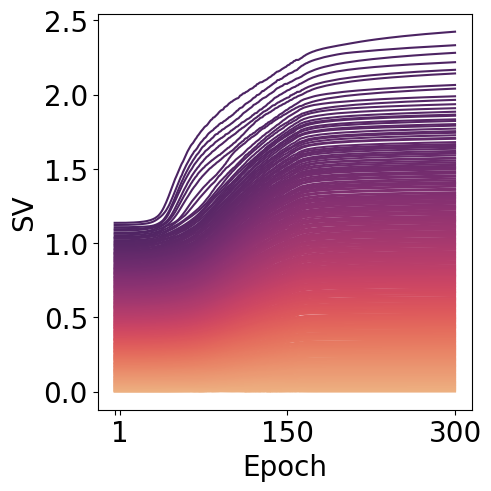}
    \caption{SVs}
  \end{subfigure}
  \begin{subfigure}[b]{0.18\linewidth}
    \centering
    \includegraphics[width=\linewidth]{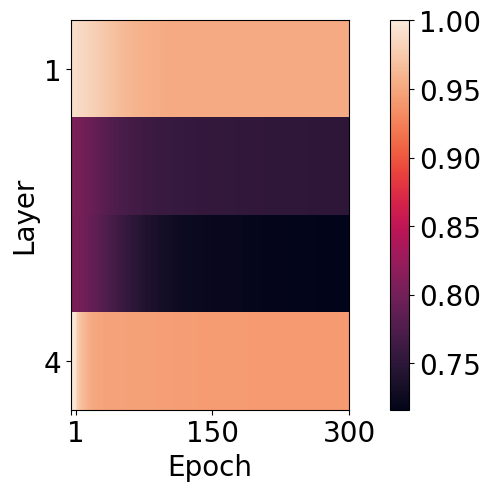}
    \\
    \includegraphics[width=\linewidth]{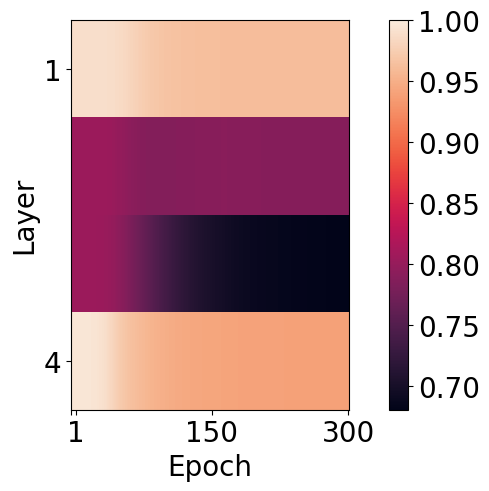}
    \caption{Eff. Rank}
  \end{subfigure}
  \begin{subfigure}[b]{0.18\linewidth}
    \centering
    \includegraphics[width=\linewidth]{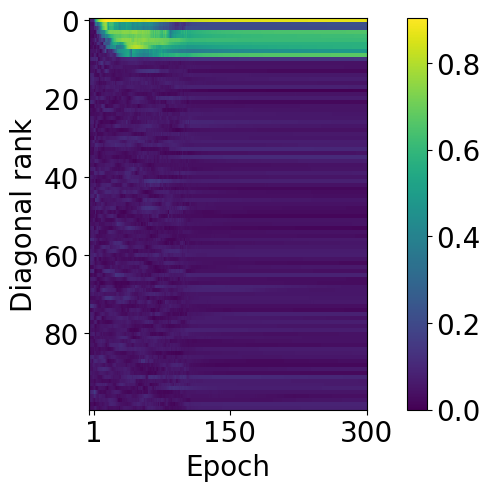}
    \\
    \includegraphics[width=\linewidth]{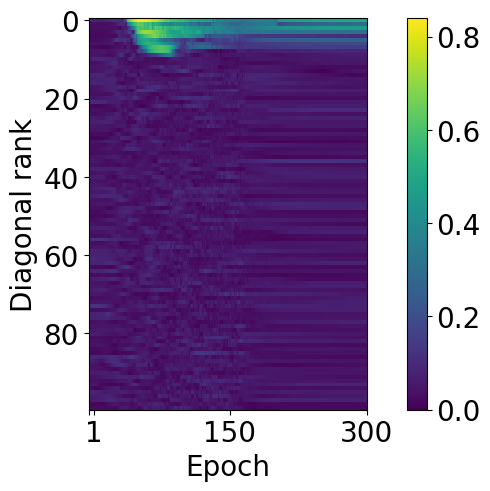}
    \caption{Alignment}
  \end{subfigure}
  \caption{\textbf{Top row:} results with true labels. \textbf{Bottom row:} results with random labels. We see that the middle layers have a lower effective rank (Eqn.~\ref{eqn:normalized-effective-rank}) when using true labels and that alignment (Eqn.~\ref{eqn:alignment-matrix}) in the middle layers persists throughout training, unlike in the random label case. We emphasize this alignment occurs despite the nonlinearities.}
  \label{fig:random-labels}
\end{figure*}
\textbf{Memorization vs.\ Generalization:} In Figure~\ref{fig:random-labels}, we replicate the core memorization experiment of \citet{zhang2021understanding}, which highlighted the ability of modern neural networks to memorize even random labels perfectly. We find that when training with random labels, we obtain networks with higher-rank final parameters as opposed to when training with true labels. Thus, the spectral dynamics can distinguish between memorization (of random labels) and generalization in this simple experiment. We also see an alignment (Eqn.~\ref{eqn:alignment-matrix}) structure between the middle layers that disappears with random labels, perhaps as it is necessary in order for the network to pass signals from input to output. This echoes the pattern of grokking in Section~\ref{sec:grokking}, where generalization came with a transition to low-rank and alignment in middle layers. We expand this experiment to more settings in Appendix~\ref{sec:random_labels} with additional figures.

\textbf{Lottery Tickets:} On very small networks, \citet{frankle2018lottery} found the existence of sparse sub-networks via magnitude pruning, keeping only the top $p$\% of weights globally by magnitude, that could train to similar performance as the full network. This was particularly surprising as they could find such networks with $1-10$\% of the weights, indicating much of the capacity was unnecessary. For larger image classification networks, \citet{frankle2020linear} observed that in order to find such sparse subnetworks, it was necessary to train until the end in order to acquire the pruning mask that could be used retroactively in training. This curious observation still lacks a compelling explanation. We show that such global magnitude pruning functions similarly to low-rank pruning, thus the lottery ticket masks found by rewinding~\citep{frankle2020linear} are effectively low-rank masks for the singular components that will become important at the end of training. Training the masked network leads to similar dynamics in these components. However, taking masks from too early in training leads to poor approximation of these final components and simultaneously stunts training. We provide detail on this discussion in Appendix~\ref{sec:beyond-generalization}.

\begin{figure}[!t]
  \centering
  \begin{subfigure}[b]{0.24\linewidth}
    \centering
    \includegraphics[width=\linewidth]{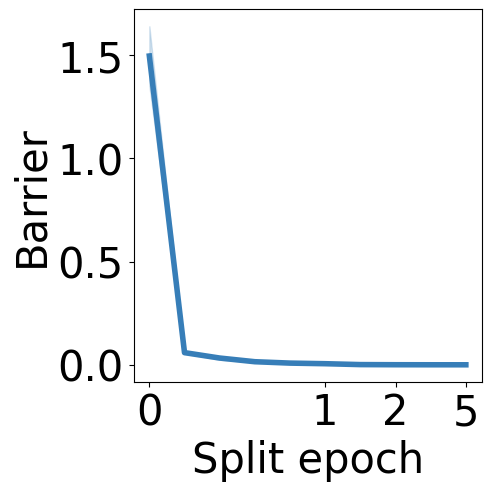}
    \\
    \includegraphics[width=\linewidth]{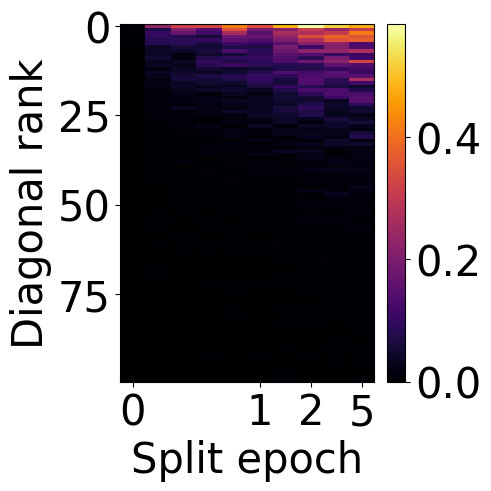}
    \\
    \includegraphics[width=\linewidth]{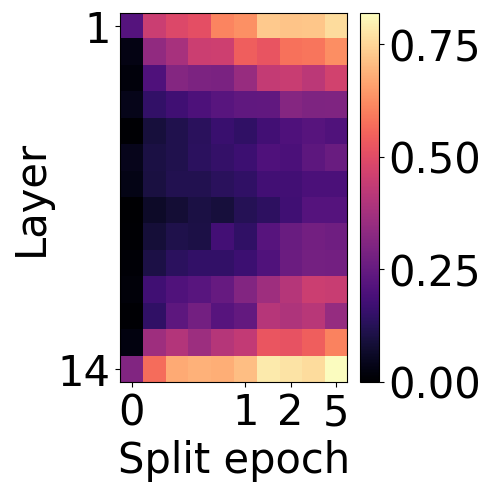}
    \caption{VGG}
  \end{subfigure}
  \begin{subfigure}[b]{0.24\linewidth}
    \centering
    \includegraphics[width=\linewidth]{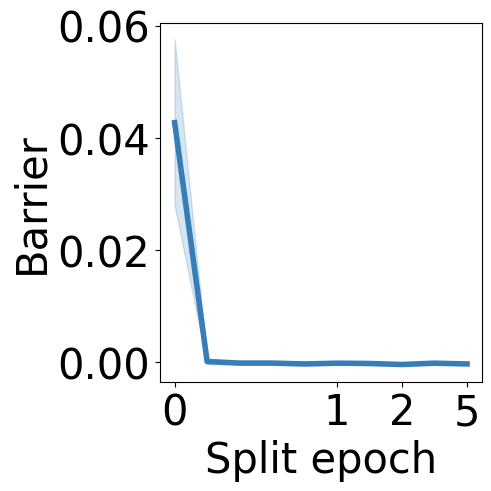}
    \\
    \includegraphics[width=\linewidth]{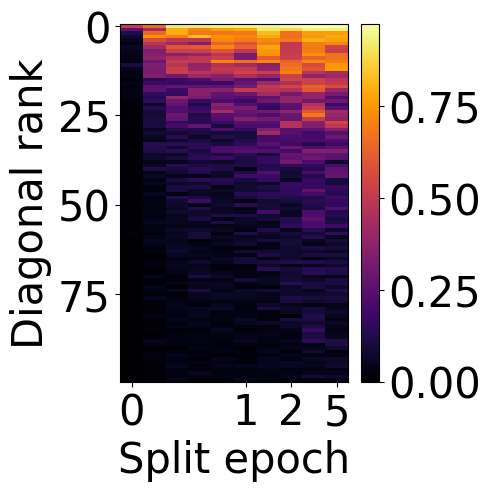}
    \\
    \includegraphics[width=\linewidth]{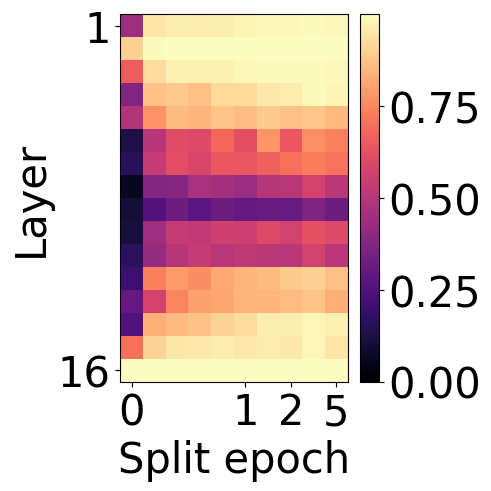}
    \caption{UNet}
  \end{subfigure}
  \begin{subfigure}[b]{0.24\linewidth}
    \centering
    \includegraphics[width=\linewidth]{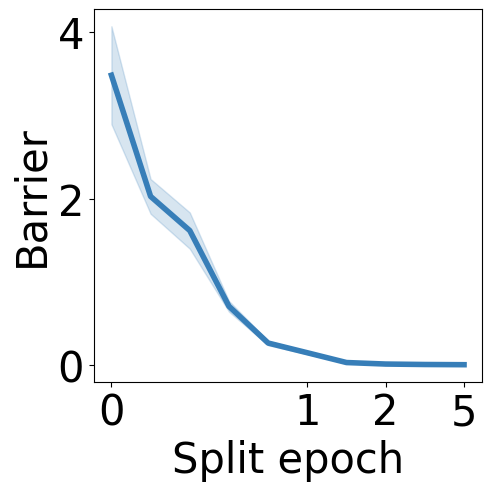}
    \\
    \includegraphics[width=\linewidth]{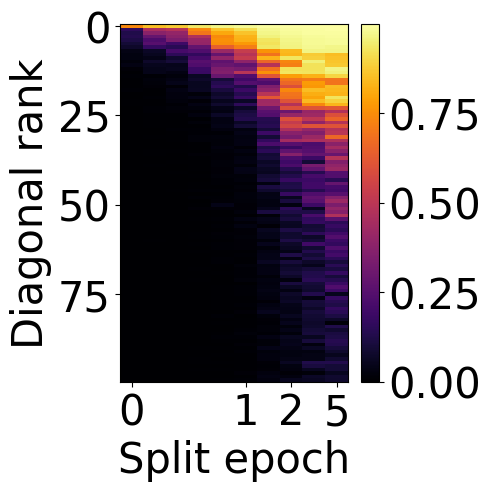}
    \\
    \includegraphics[width=\linewidth]{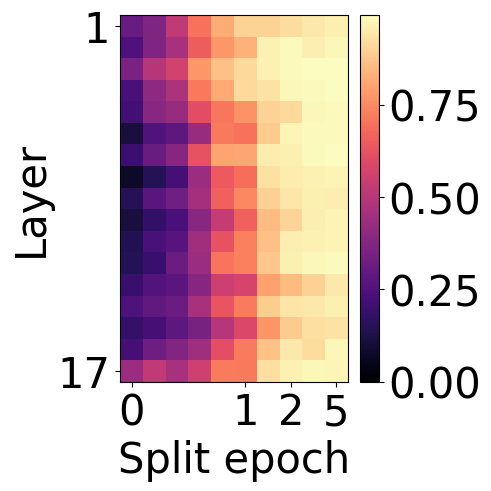}
    \caption{LSTM}
  \end{subfigure}
  \begin{subfigure}[b]{0.24\linewidth}
    \centering
    \includegraphics[width=\linewidth]{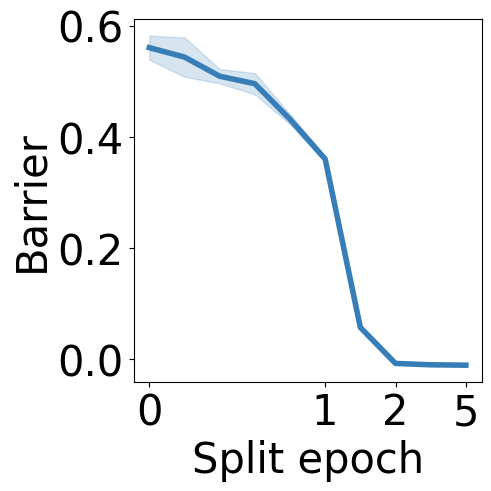}
    \\
    \includegraphics[width=\linewidth]{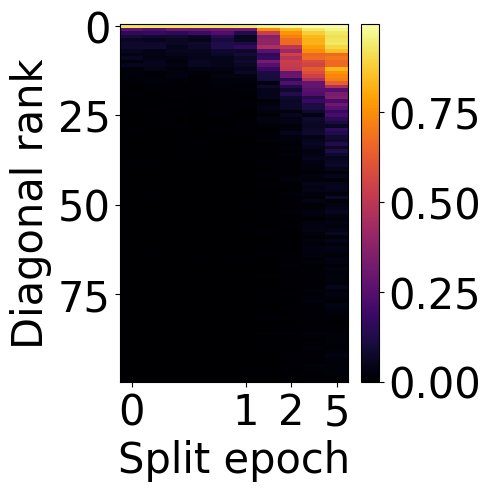}
    \\
    \includegraphics[width=\linewidth]{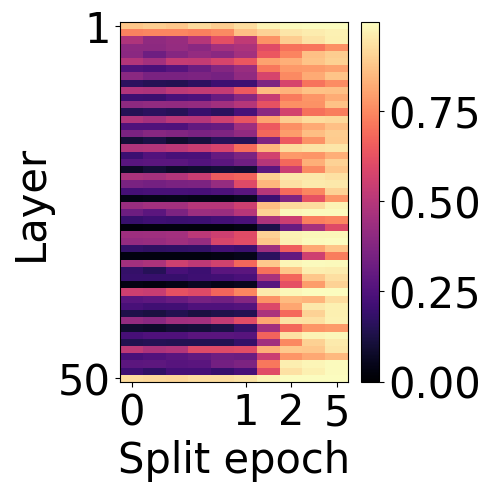}
    \caption{Transformer}
  \end{subfigure}
  \caption{\textbf{Top row:} Barrier size (Eqn.~\ref{eqn:lmc-barrier}) vs.\ split step. \textbf{Middle row:} singular vector agreement for a single matrix parameter between branch endpoints that share a common trunk. \textbf{Bottom row:} summary statistic for singular vector agreement across layers vs.\ split step. We see that as models exhibit LMC, they also share a small number of top singular vectors. We only show agreement for 100 dimensions as the rest are near zero.}
  \label{fig:lmc-agreement}
\end{figure}

\textbf{Linear Mode Connectivity (LMC):} Linear Mode Connectivity~\citep{nagarajan2019uniform, frankle2020linear} refers to the property that models that share a portion of the training trajectory can be averaged in weight-space to yield a stronger model~\citep{wortsman2022model, ramesh2022hierarchical}. This phenomenon indicates that, after some training, the loss surface is quite convex in a subspace, even though the optimization problem is theoretically highly non-convex. As prior work has found, fine-tuning from pre-trained models stays in this convex space~\citep{neyshabur2020being, li2022branch, sadrtdinov2023stay}, an explanation for what underlies LMC could help to clarify the role of pre-training, and may lead to faster fine-tuning. We show that LMC is tied with singular vector sharing. In particular, when models display LMC, they share top singular vectors between weights (see Figure~\ref{fig:lmc-agreement}), and when they do not they also do not share parameters (Appendix~\ref{sec:lmc}). We argue this is an outcome of the early stability of top singular vectors, which arises due to the unequal evolution of singular values. It is also straightforward to explain the large Euclidean distance between checkpoints~\citep{frankle2020linear, yunis2022convexity} that can be averaged, as they only share a very small portion of the parameter space. Thus, LMC, and by extension, model-averaging, is deeply intertwined with the dynamics of singular values that we explore in Section~\ref{sec:spectral_dynamics}. The full discussion is deferred to Appendix~\ref{sec:beyond-generalization}.

\section{Spectral Dynamics with Random Labels}\label{sec:random_labels}
\begin{figure*}[!t]
  \centering
  \begin{subfigure}[b]{0.18\linewidth}
    \centering
    \includegraphics[width=\linewidth]{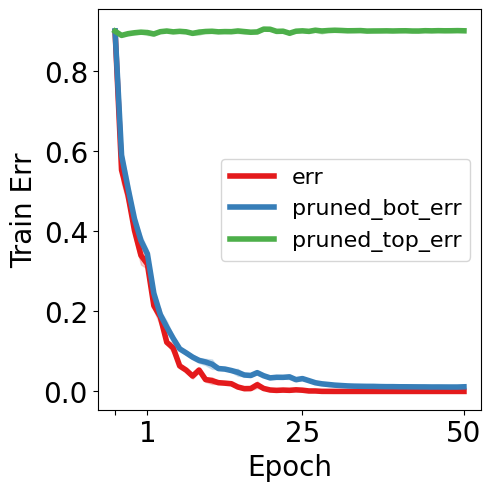}
    \\
    \includegraphics[width=\linewidth]{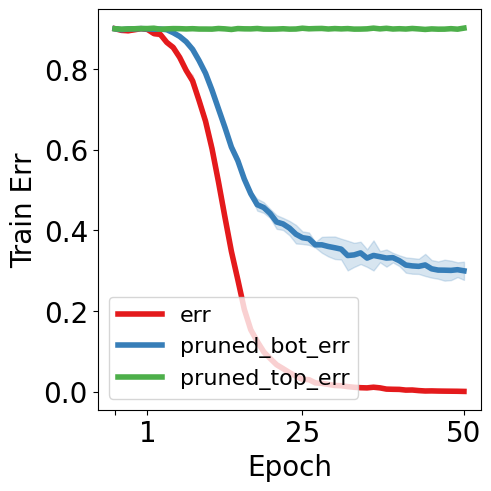}
    \caption{Train Err.}
  \end{subfigure}
  \begin{subfigure}[b]{0.18\linewidth}
    \centering
    \includegraphics[width=\linewidth]{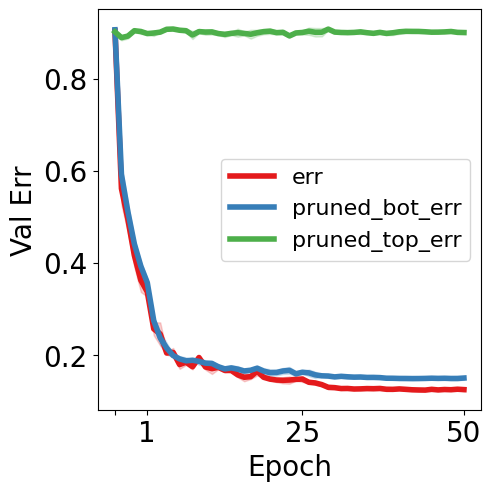}
    \\
    \includegraphics[width=\linewidth]{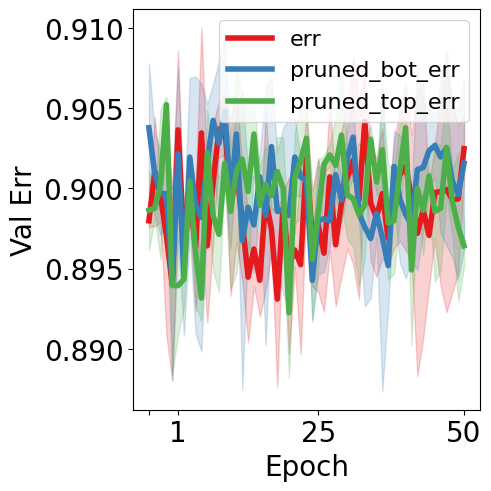}
    \caption{Val. Err.}
  \end{subfigure}
  \begin{subfigure}[b]{0.18\linewidth}
    \centering
    \includegraphics[width=\linewidth]{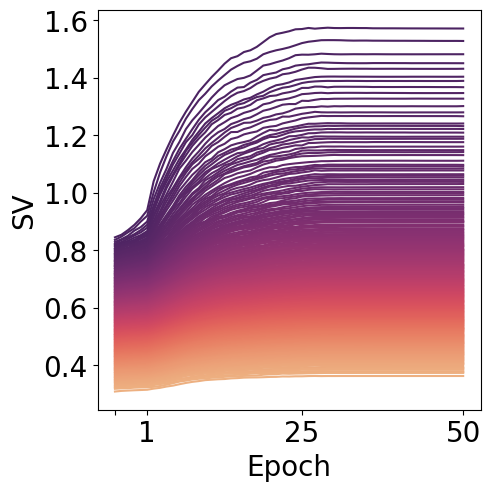}
    \\
    \includegraphics[width=\linewidth]{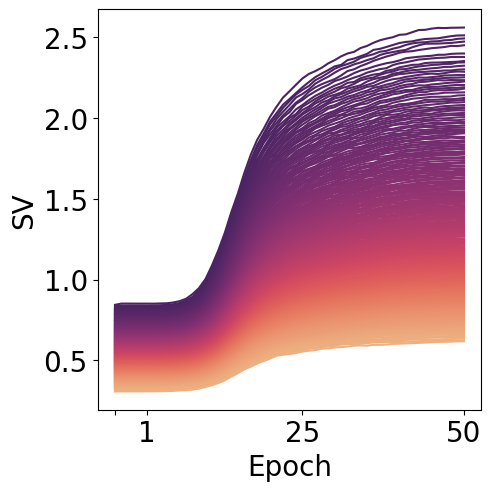}
    \caption{SVs}
  \end{subfigure}
  \begin{subfigure}[b]{0.18\linewidth}
    \centering
    \includegraphics[width=\linewidth]{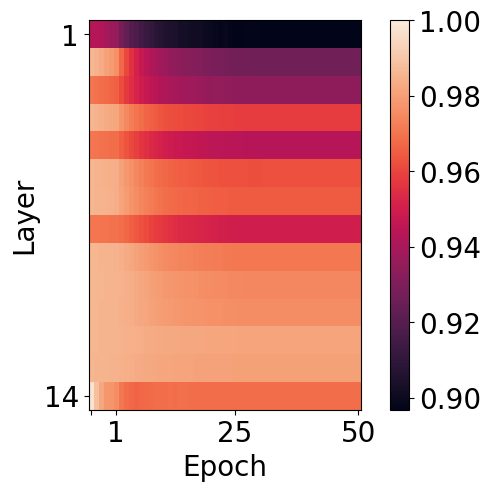}
    \\
    \includegraphics[width=\linewidth]{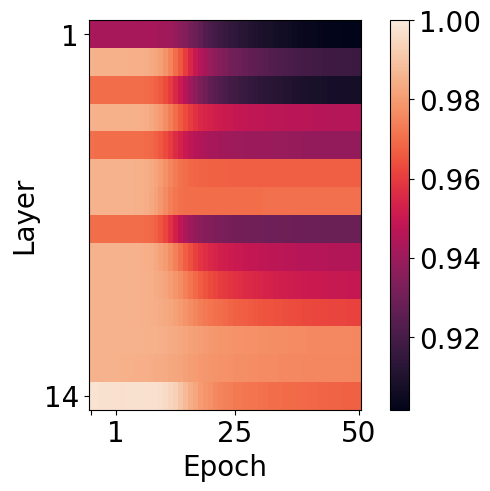}
    \caption{Eff. Rank}
  \end{subfigure}
  \begin{subfigure}[b]{0.18\linewidth}
    \centering
    \includegraphics[width=\linewidth]{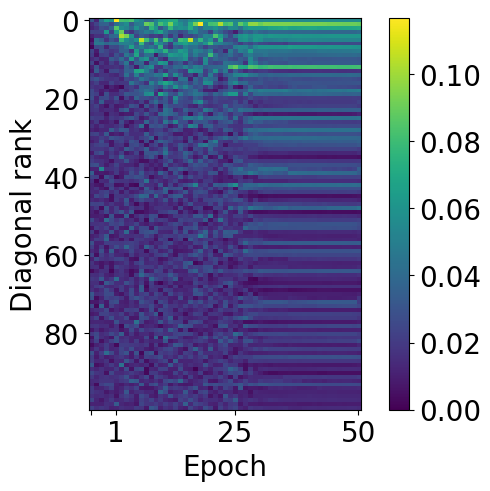}
    \\
    \includegraphics[width=\linewidth]{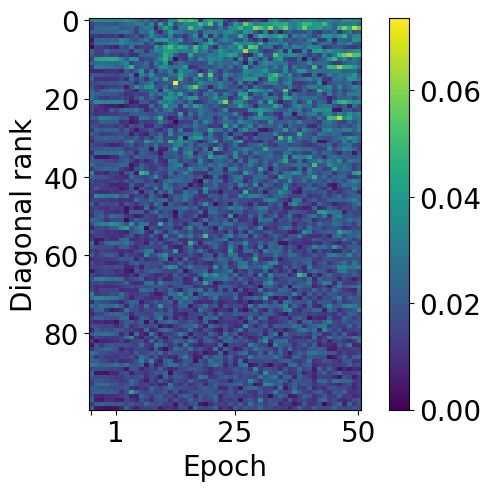}
    \caption{Alignment}
  \end{subfigure}
  \caption{Dynamics with random labels for VGG. \textbf{Top row:} results with true labels. \textbf{Bottom row:} results with random labels. We see that the middle layers have a lower effective rank (Eqn.~\ref{eqn:normalized-effective-rank}) when using true labels and that alignment (Eqn.~\ref{eqn:alignment-matrix}) in the middle layers persists throughout training. The results are less stark in the VGG case, but similar to the MLP.}
  \label{fig:random-labels-vgg}
\end{figure*}
\begin{figure*}[!t]
  \centering
  \begin{subfigure}[b]{0.18\linewidth}
    \centering
    \includegraphics[width=\linewidth]{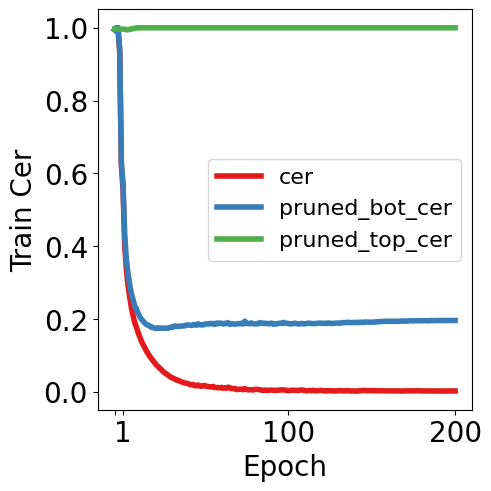}
    \\
    \includegraphics[width=\linewidth]{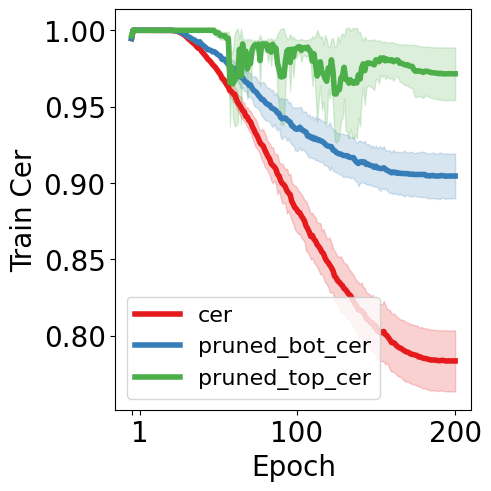}
    \caption{Train Err.}
  \end{subfigure}
  \begin{subfigure}[b]{0.18\linewidth}
    \centering
    \includegraphics[width=\linewidth]{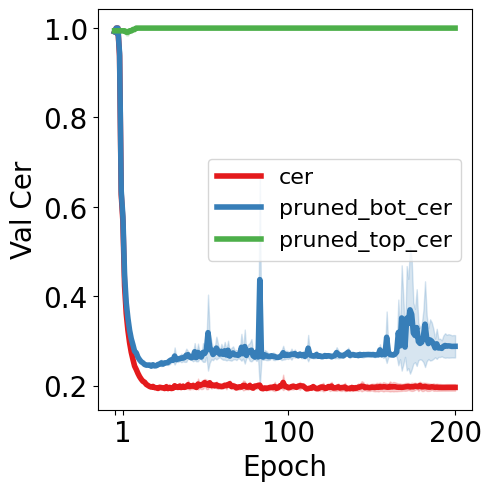}
    \\
    \includegraphics[width=\linewidth]{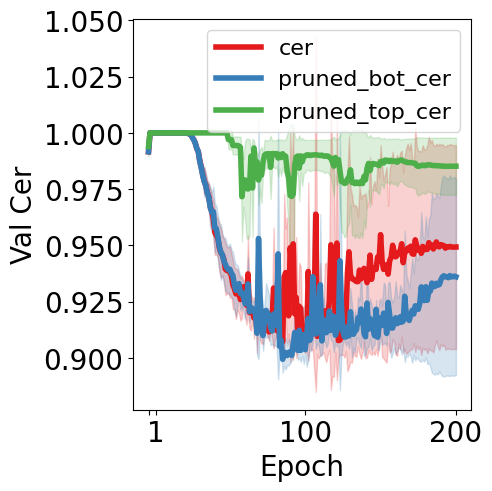}
    \caption{Val. Err.}
  \end{subfigure}
  \begin{subfigure}[b]{0.18\linewidth}
    \centering
    \includegraphics[width=\linewidth]{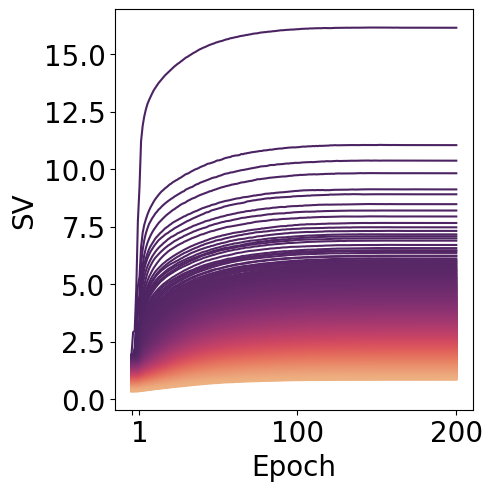}
    \\
    \includegraphics[width=\linewidth]{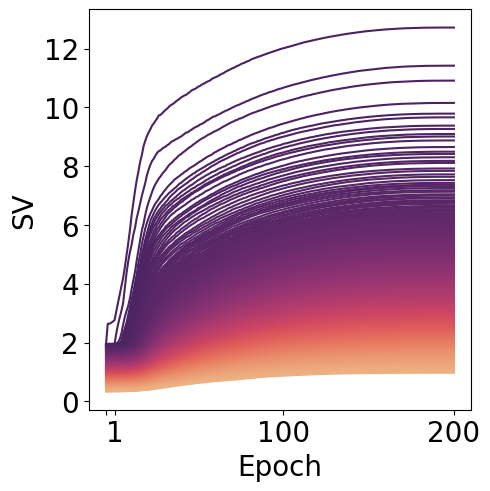}
    \caption{SVs}
  \end{subfigure}
  \begin{subfigure}[b]{0.18\linewidth}
    \centering
    \includegraphics[width=\linewidth]{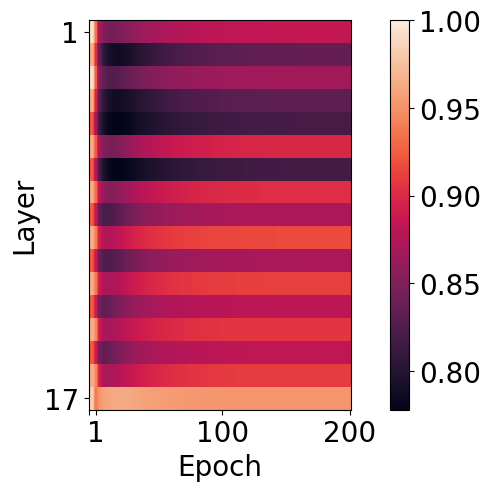}
    \\
    \includegraphics[width=\linewidth]{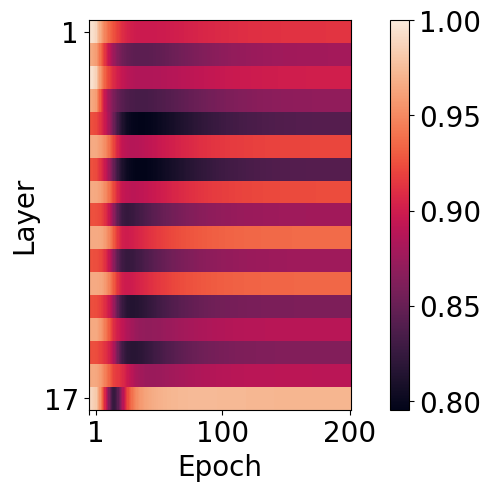}
    \caption{Eff. Rank}
  \end{subfigure}
  \begin{subfigure}[b]{0.18\linewidth}
    \centering
    \includegraphics[width=\linewidth]{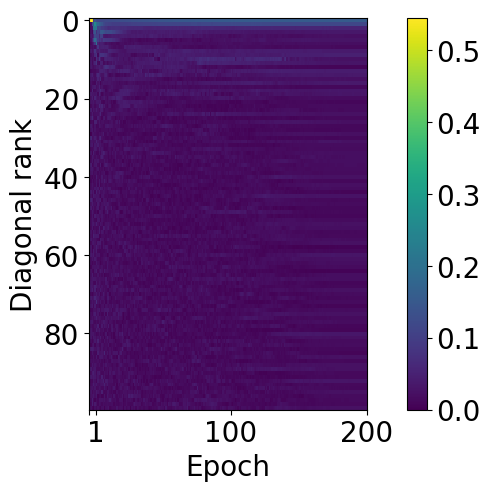}
    \\
    \includegraphics[width=\linewidth]{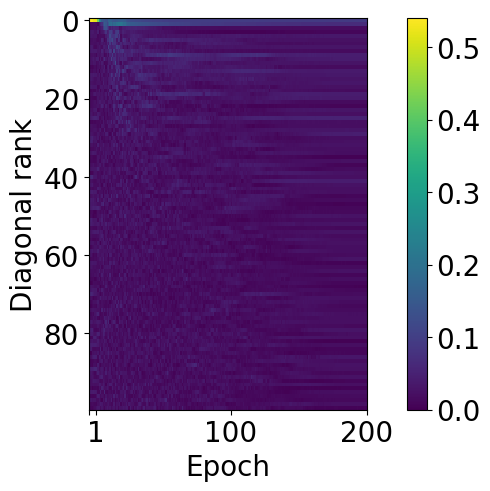}
    \caption{Alignment}
  \end{subfigure}
  \caption{Dynamics with random labels for LSTM. \textbf{Top row:} results with true labels. \textbf{Bottom row:} results with random labels. We see that the middle layers have a lower effective rank (Eqn.~\ref{eqn:normalized-effective-rank}) when using true labels and that alignment (Eqn.~\ref{eqn:alignment-matrix}) in the middle layers persists throughout training. Though the LSTM does not fit the random labels perfectly, the results are qualitatively similar to the other cases.}
  \label{fig:random-labels-lstm}
\end{figure*}
Given the observations connecting generalization and rank thus far, and the enlightening view on the implicit effects of weight decay, we are interested in seeing whether the perspective developed sheds any light on the classic random label memorization experiments of \citet{zhang2021understanding}.

Similar to \citet{zhang2021understanding}, we train an MLP, a VGG, and an LSTM to fit random or true labels. Please see Appendix~\ref{app:experimental-details} for the details regarding the experimental setup. \citet{zhang2021understanding} decay the learning rate to zero, and the random label experiments only converge late in training. Consequently, we use a constant learning rate to avoid confounding factors. We see in Figure~\ref{fig:random-labels} that both cases for the MLP are able to achieve zero error, though with different singular value evolution and alignment (Eqn.~\ref{eqn:alignment-matrix}) in the middle layer.

Surprisingly in Figure~\ref{fig:random-labels}, we see that with true labels the inner layers are low rank, while with random labels they are much higher rank. This may be explained by the shared structure in the true classes of the dataset, which manifests in the parameters. Even more surprisingly, we find here that even without weight decay, inner layers align with true labels, while with random labels, this alignment occurs and then disappears with more training. This is particularly intriguing as there are non-linearities that could theoretically separate the network from the linear case, and yet strong alignment occurs despite that. Such alignment has not yet been leveraged by existing theory, and might provide structured assumptions for new understanding. Results on the VGG (Figure~\ref{fig:random-labels-vgg}) and LSTM (Figure~\ref{fig:random-labels-lstm}) are qualitatively quite similar, though weaker in alignment. In summary, these results suggest that viewing generalization through the lens of rank and alignment may be fruitful, deepening the connection previously revealed.

\section{Beyond Generalization}\label{sec:beyond-generalization}
We have seen over the course of many experiments that deep models are biased toward low rank, and that there is a tempting connection between rank minimization and generalization. Still, spectral dynamics may have broader connections. In the following subsections, we explore lottery tickets~\citep{frankle2018lottery} and linear mode connectivity~\citep{frankle2020linear} in greater detail. Beyond shedding further light on neural networks, these phenomena have implications for more efficient inference and storage, as well as understanding the importance of pretraining~\citep{neyshabur2020being}. We find that lottery tickets are a sparse approximation of final-checkpoint top singular vectors. The ability to linearly interpolate between faraway checkpoints and improve performance coincides strongly with top singular vector sharing between checkpoints. Such observations may form a foundation for a better understanding compression and model averaging~\citep{wortsman2022model, ilharco2022editing}.

\subsection{Top Singular Vectors Become Stable Earlier}

\begin{figure}
  \centering
  \begin{subfigure}[b]{0.24\columnwidth}
    \centering
    \includegraphics[width=\columnwidth]{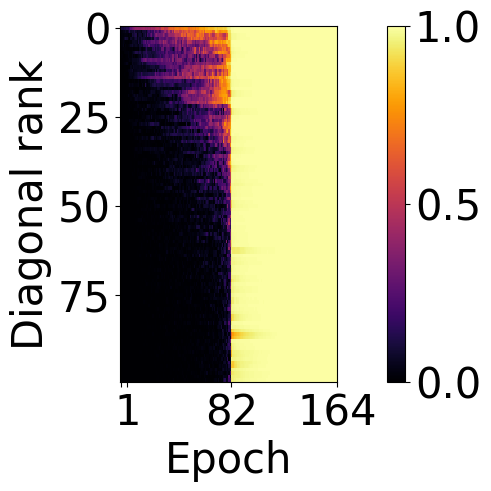}
    \\
    \includegraphics[width=\columnwidth]{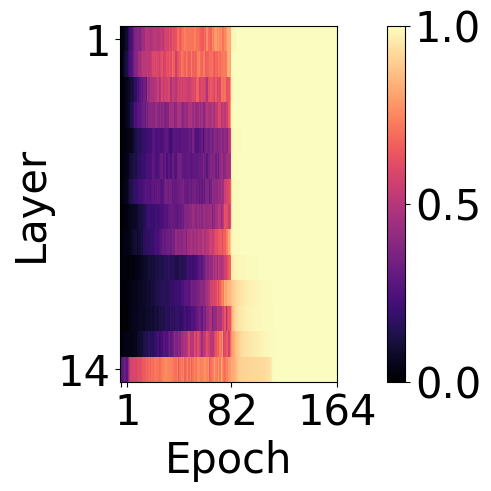}
    \caption{VGG}
  \end{subfigure}
  \begin{subfigure}[b]{0.24\columnwidth}
    \centering
    \includegraphics[width=\columnwidth]{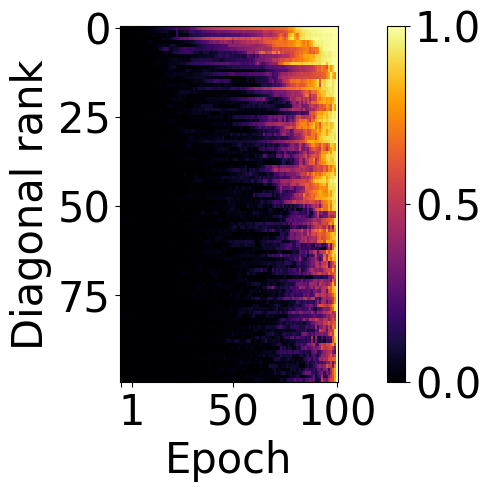}
    \\
    \includegraphics[width=\columnwidth]{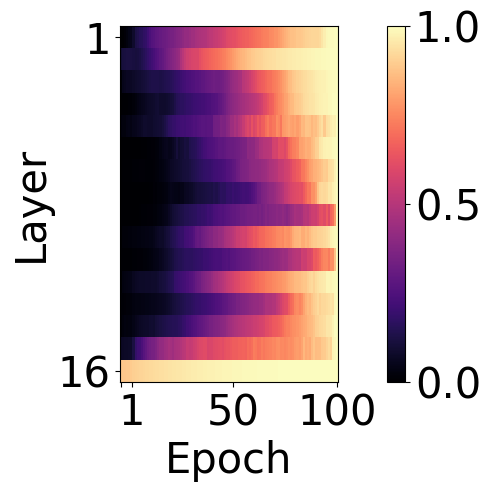}
    \caption{UNet}
  \end{subfigure}
  \begin{subfigure}[b]{0.24\columnwidth}
    \centering
    \includegraphics[width=\columnwidth]{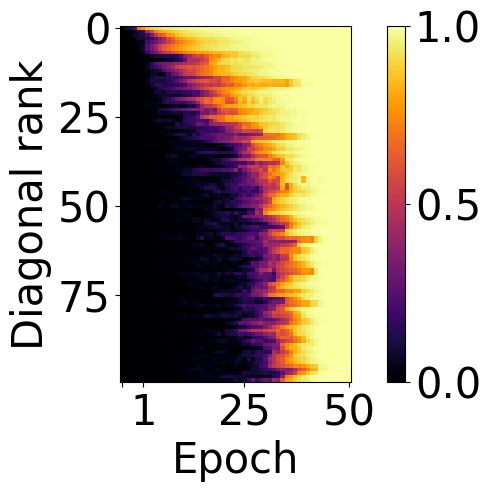}
    \\
    \includegraphics[width=\columnwidth]{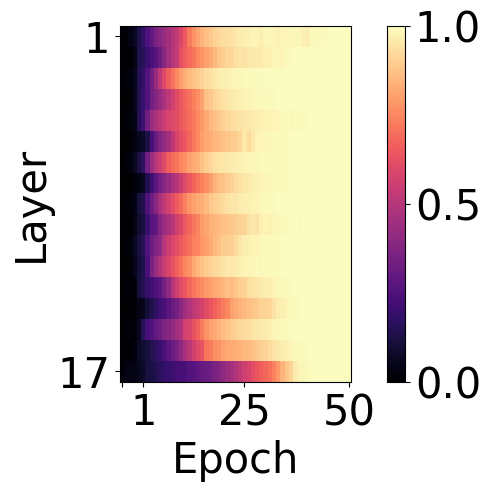}
    \caption{LSTM}
  \end{subfigure}
  \begin{subfigure}[b]{0.24\columnwidth}
    \centering
    \includegraphics[width=\columnwidth]{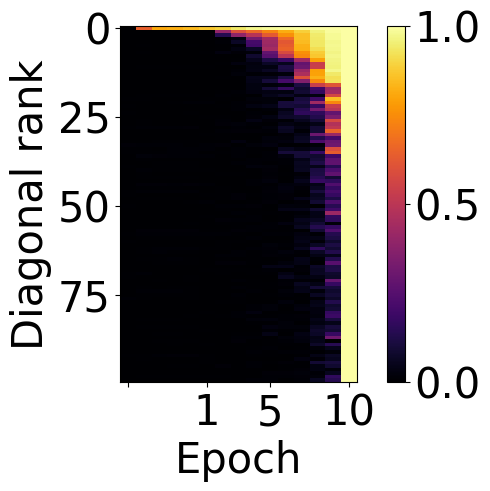}
    \\
    \includegraphics[width=\columnwidth]{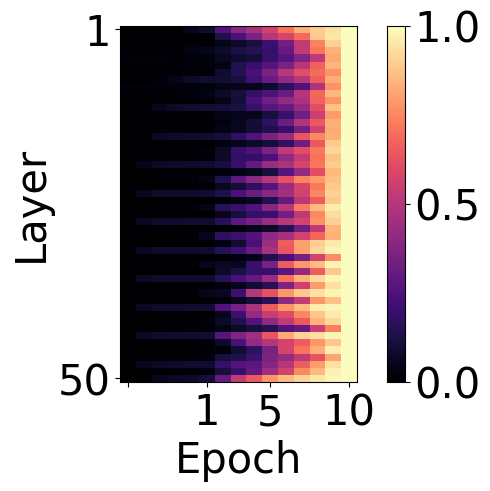}
    \caption{Transformer}
  \end{subfigure}
  \caption{\textbf{Top row:} Singular vector agreement for a single matrix in the middle of each model (diagonal of Eqn.~\ref{eqn:sva-matrix}). Notice top singular vectors become stable in direction earlier. \textbf{Bottom row:} Summary score for each matrix across architectures. As we move down the $y$-axis, the depth of the parameters in the model increases, while the $x$-axis tracks training time. The sharp transition midway through training in the VGG case is likely due to a 10x learning rate decay.}
  \label{fig:singular-vector-agreement}
\end{figure}

Before we explore the phenomena, we first make another observation that will be helpful. As top singular values grow disproportionately large, it would be natural that top singular vectors become stable in direction as the gradients remain small. To demonstrate this, for a given matrix in the network $W_i(t) = \sum_{j=1}^R \sigma_j(t) u_j(t) v_j(t)^\top$ at training time t, we compute
\begin{equation} \label{eqn:sva-matrix}
    S(t)_{jk} = \lvert \langle u_j(t)v_j(t)^\top, u_k(T)v_k(T)^\top \rangle \rvert,
\end{equation}
where $T$ is the final step of training, and the absolute value is taken to ignore sign flips in the SVD computation. We then plot the diagonal of this matrix $S(t)_{ii}~\forall~i \leq 100$ over time. We also use a scalar measure of the diagonal to summarize like in the alignment case: $s(t) = \frac{1}{10} \sum_i S(t)_{ii}$. In Figure~\ref{fig:singular-vector-agreement}, we see that top singular vectors converge in direction earlier than bottom vectors.

\subsection{Lottery Tickets Preserve Final Top Singular Vectors}

As large singular vectors will become stable late in training, we wonder about the connection to magnitude pruning and the lottery ticket hypothesis. \citet{frankle2018lottery} first showed evidence for the lottery ticket hypothesis, the idea that there exist sparse subnetworks of neural networks that can be trained to a comparable performance as the full network, where the sparse mask is computed from the largest magnitude weights of the network at the end of training. \citet{frankle2020linear} build further on this hypothesis and notice that, for larger networks, the masking cannot begin at initialization, but rather at some point early in training. Still, the mask must come from the end of training.

The reason for this particular choice of mask may be connected to the dynamics we previously observed. Specifically, at the end of training large singular values are disproportionately larger, so high-magnitude weights may correspond closely to weights in the top singular vectors. If magnitude masks were computed at the beginning, the directions that would become the top singular vectors might be prematurely masked as they have not yet stabilized, which may prevent learning on the task.

\begin{figure*}[!t]
  \centering
   \begin{subfigure}[b]{0.18\linewidth}
    \centering
    \includegraphics[width=\linewidth]{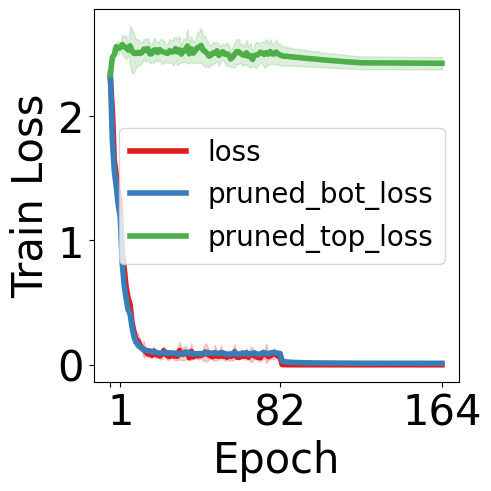}
    \\
    \includegraphics[width=\linewidth]{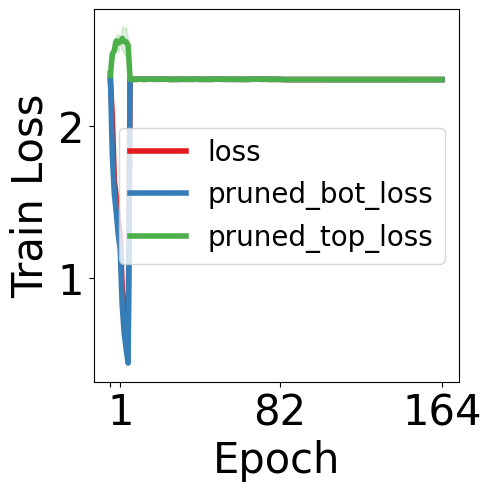}
    \caption{Loss}
  \end{subfigure}
  \begin{subfigure}[b]{0.18\linewidth}
    \centering
    \includegraphics[width=\linewidth]{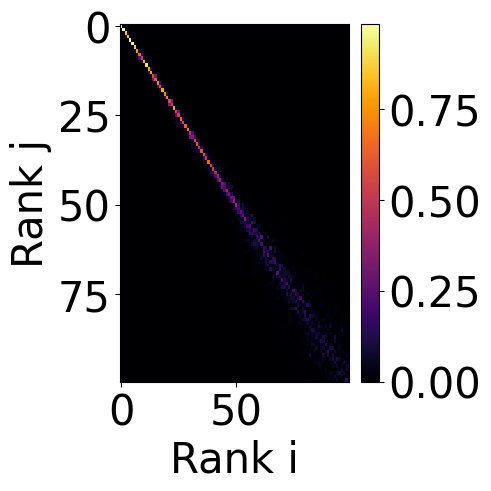}
    \\
    \includegraphics[width=\linewidth]{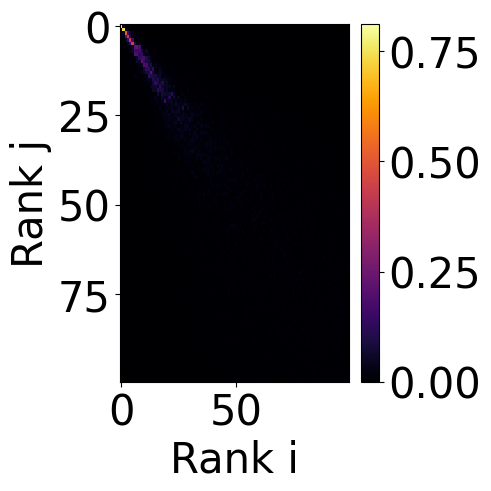}
    \caption{Pruned SVA}
  \end{subfigure}
  \begin{subfigure}[b]{0.18\linewidth}
    \centering
    \includegraphics[width=\linewidth]{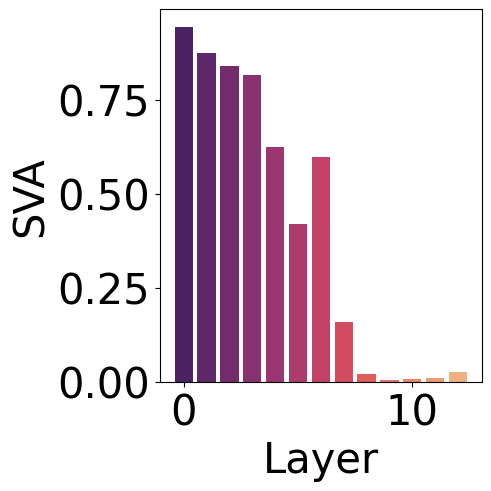}
    \\
     \includegraphics[width=\linewidth]{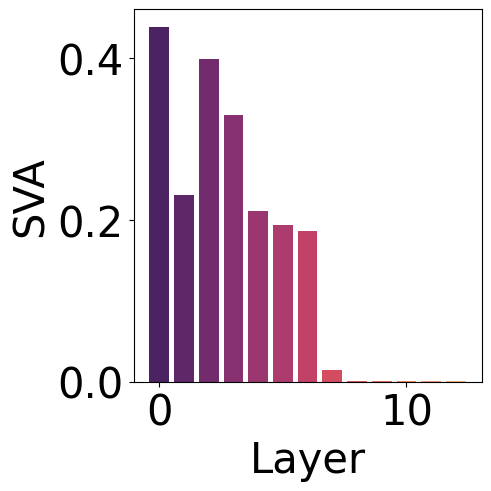}
    \caption{All Layers}
  \end{subfigure}
  \begin{subfigure}[b]{0.18\linewidth}
    \centering
    \includegraphics[width=\linewidth]{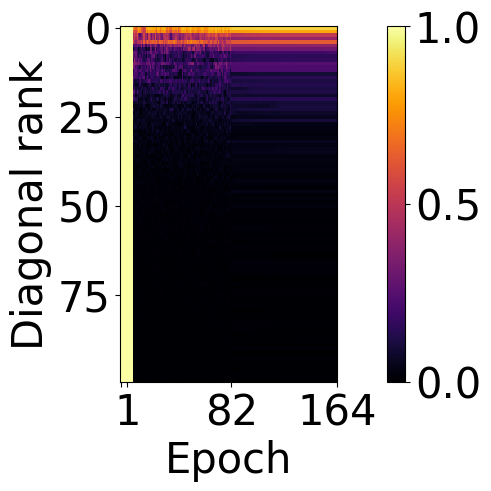}
    \\
    \includegraphics[width=\linewidth]{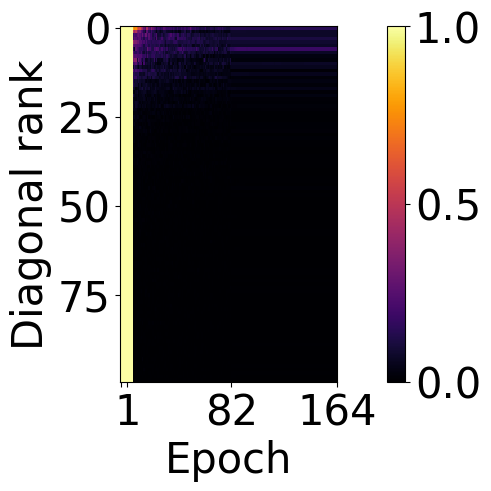}
    \caption{SVA evol.}
  \end{subfigure}
  \begin{subfigure}[b]{0.18\linewidth}
    \centering
    \includegraphics[width=\linewidth]{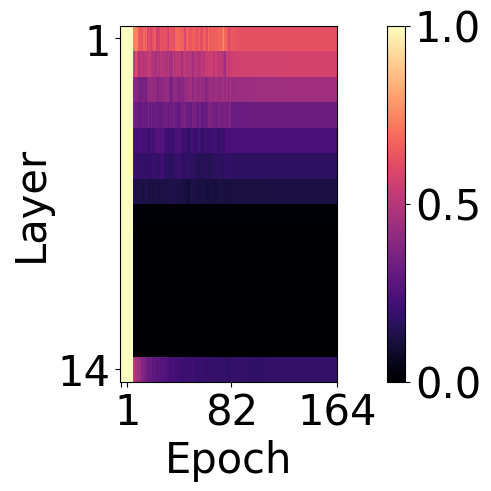}
    \\
    \includegraphics[width=\linewidth]{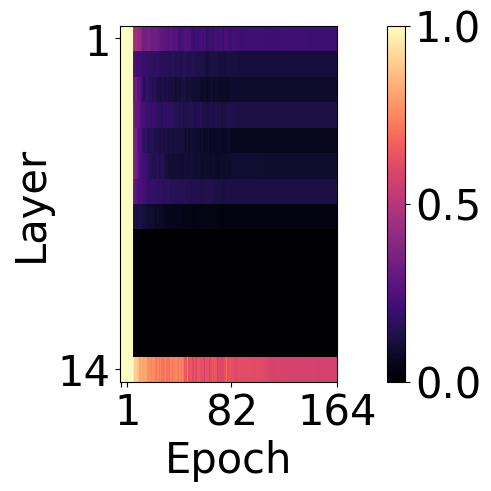}
    \caption{All Layers}
  \end{subfigure}
  \caption{Pruning results for VGG. \textbf{Top row:} Magnitude pruning. \textbf{Bottom row:} random pruning. \textbf{(a):} Training loss. We see that at 5\% sparsity magnitude pruning is significantly better than random pruning of the same layerwise sparsity. \textbf{(b):} Singular vector alignment pre- and post-pruning at the end of training for a single layer (the 3rd convolution). We see that magnitude pruning approximates many more top singular vectors than random pruning at the same sparsity. \textbf{(c):} Singular vector alignment score pre- and post-pruning across all layers. Agreement is higher across all layers for magnitude pruning, though curiously deeper layers do not agree, likely as later layers are wider so weights are lower magnitude and more will be pruned by the unstructured process. \textbf{(d):} Singular vector alignment between the pruned and unpruned models along the training trajectory. We see that the magnitude pruning still has similar dynamics in its top singular vectors, while random pruning does not. \textbf{(e):} Singular vector alignment score between pruned and unpruned models across layers and time. Again evolution is similar for early layers with magnitude pruning, and completely different for random pruning.}
  \label{fig:lottery-tickets-vgg}
\end{figure*}

\begin{figure*}[!t]
  \centering
   \begin{subfigure}[b]{0.18\linewidth}
    \centering
    \includegraphics[width=\linewidth]{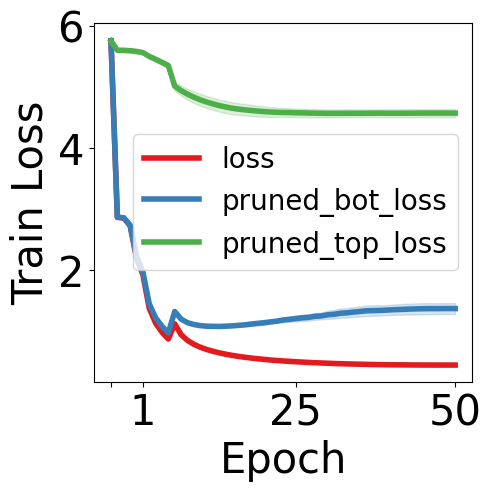}
    \\
    \includegraphics[width=\linewidth]{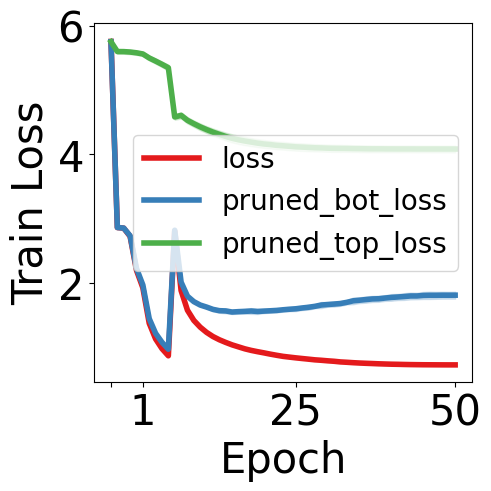}
    \caption{Loss}
  \end{subfigure}
  \begin{subfigure}[b]{0.18\linewidth}
    \centering
    \includegraphics[width=\linewidth]{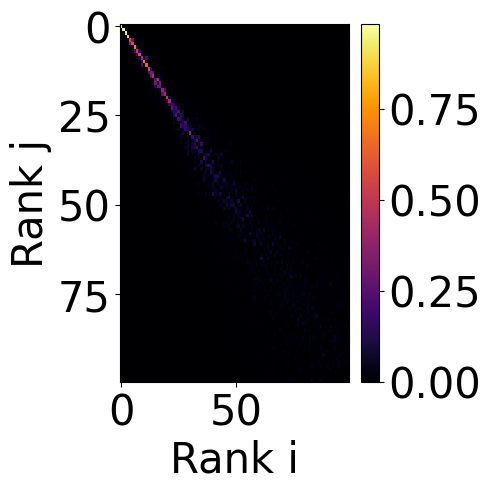}
    \\
    \includegraphics[width=\linewidth]{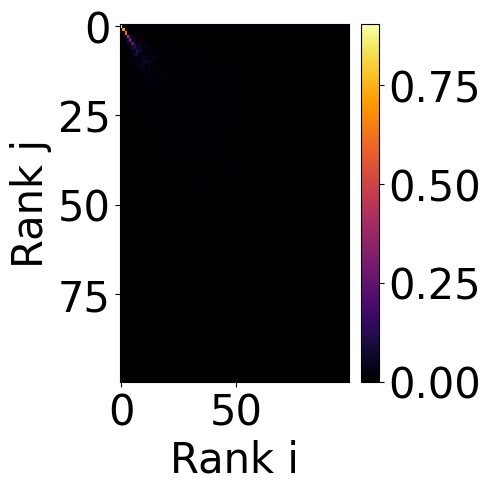}
    \caption{Pruned SVA}
  \end{subfigure}
  \begin{subfigure}[b]{0.18\linewidth}
    \centering
    \includegraphics[width=\linewidth]{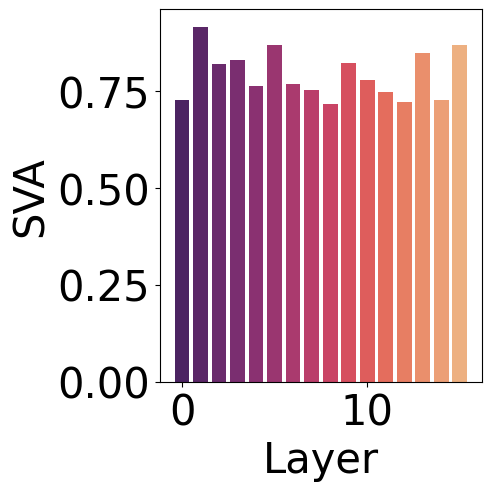}
    \\
     \includegraphics[width=\linewidth]{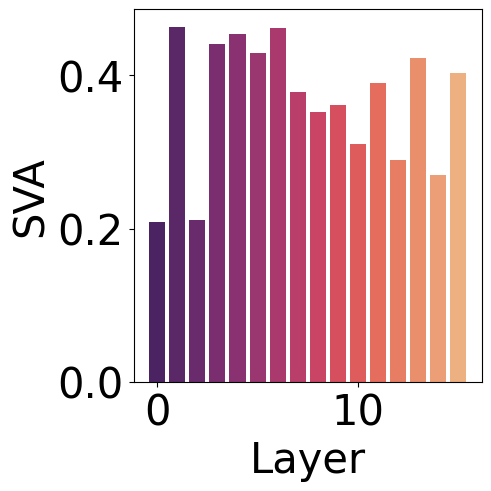}
    \caption{All Layers}
  \end{subfigure}
  \begin{subfigure}[b]{0.18\linewidth}
    \centering
    \includegraphics[width=\linewidth]{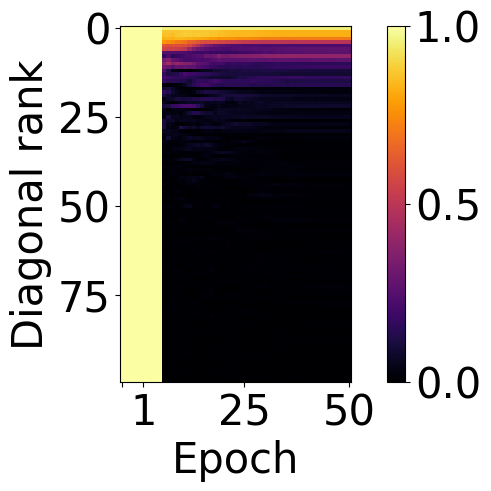}
    \\
    \includegraphics[width=\linewidth]{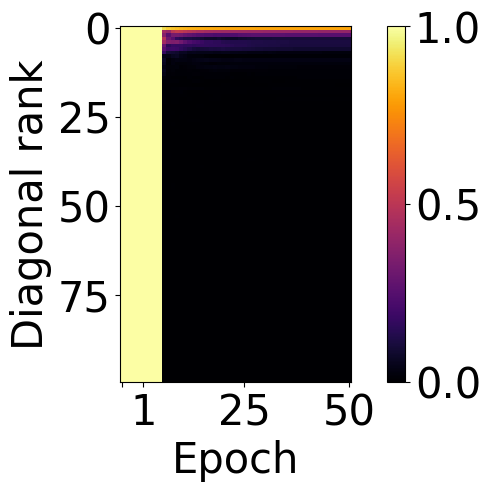}
    \caption{SVA evol.}
  \end{subfigure}
  \begin{subfigure}[b]{0.18\linewidth}
    \centering
    \includegraphics[width=\linewidth]{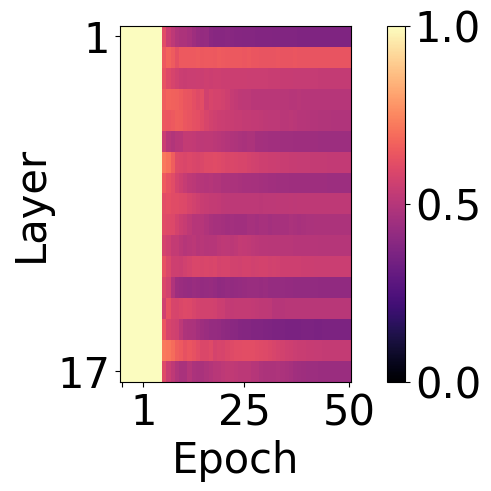}
    \\
    \includegraphics[width=\linewidth]{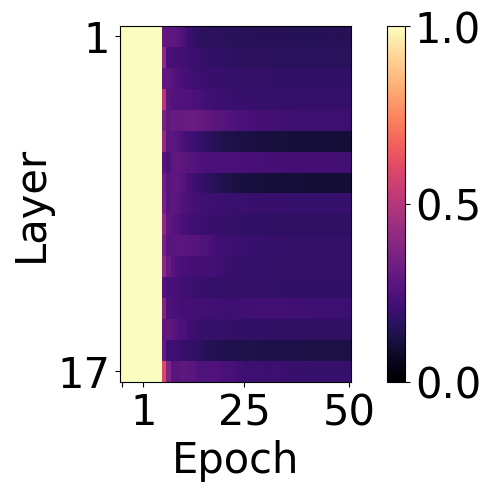}
    \caption{All Layers}
  \end{subfigure}
  \caption{Pruning results for LSTM. \textbf{Top row:} Magnitude pruning. \textbf{Bottom row:} random pruning. See Figure~\ref{fig:lottery-tickets-vgg} for details. Results are quite similar for the LSTM at 25\% pruning as the VGG in Figure~\ref{fig:lottery-tickets-vgg}.}
  \label{fig:lottery-tickets-lstm}
\end{figure*}

Here we train an unmasked VGG-16~\citep{simonyan2014very} on CIFAR10, then compute either a random mask, or a global magnitude mask from the end of training, and rewind to an early point~\citep{frankle2020linear} to start sparse retraining. We also do the same with an LSTM~\citep{hochreiter1997long} on LibriSpeech~\citep{panayotov2015librispeech}. Please see Appendix~\ref{app:experimental-details} for details. In Figures~\ref{fig:lottery-tickets-vgg} and \ref{fig:lottery-tickets-lstm}, we plot the singular vector agreement (SVA, Eqn.~\ref{eqn:sva-matrix}) between the final model, masked and unmasked, where we see exactly that magnitude masks preserve the top singular vectors of parameters, and with increasing sparsity fewer directions are preserved. Even though prior work has remarked that it is possible to use low-rank approximations for neural networks~\citep{yu2017compressing}, and others have explicitly optimized for low-rank lottery tickets~\citep{wang2021pufferfish, schotthofer2022low}, we rather are pointing out that the magnitude pruning procedure seems to recover a low-rank approximation though it was intended to be "unstructured."

We also compute the singular vector agreement (SVA) between the masked model trajectory and the original unmasked model trajectory (diagonal of Eqn.~\ref{eqn:sva-matrix}). We see in Figures~\ref{fig:lottery-tickets-vgg} and \ref{fig:lottery-tickets-lstm} that there is no agreement between the bottom singular vectors at all, but there is still loose agreement in the top singular vectors. Thus, it seems the mask allows the dynamics of only the top singular vectors to remain similar, which we know are most important from the pruning analysis in Figure~\ref{fig:pruned-performance}.

Preserving top singular vectors by pruning seems like a natural outcome of large matrices regardless of the mask, so as a control, we follow exactly the same protocol except we generate the mask randomly with the same layerwise sparsity. We can see in Figures~\ref{fig:lottery-tickets-vgg} and \ref{fig:lottery-tickets-lstm} that this results in much lower preservation of top singular vector dynamics, and also performs worse, as in \citep{frankle2020linear}. It would not be surprising that random pruning is worse if simply evaluated at the end of training, but masking is applied quite early in training at epoch 4 of 164 long before convergence, so it's striking that the network now fails to learn further even though it is far from convergence. We interpret this as evidence that the mask has somehow cut signal flow between layers, so it is now impossible for the network to learn further, while magnitude pruning and rewinding still allows signals to pass that eventually become important. Examining the SVD dynamics of neural network weights allows us to see all of this structure that was hidden previously.

\subsection{Spectral Dynamics and Linear Mode Connectivity}\label{sec:lmc}

We come to the final phenomenon that we seek to describe: linear mode connectivity. Linear mode connectivity (LMC) is the property that one can interpolate linearly between two different minima in weight space and every parameter set along that path performs well, which gives the impression that the loss surface of neural networks is somehow convex despite its theoretical nonconvexity. This was first demonstrated in small networks with the same initialization~\citep{nagarajan2019uniform}, then expanded to larger networks and connected to lottery tickets~\citep{frankle2020linear, paul2022unmasking}. \citet{entezari2021role} first conjecture that two arbitrary minima show LMC up to permutation, and demonstrate it in simple models. Others expanded this to wide models~\citep{ainsworth2022git, jordan2022repair, qu2024rethink}, and permutation connectivity can be proven in various ways~\citep{kuditipudi2019explaining, brea2019weight, simsek2021geometry, ferbach2023proving}. Still, these conditions and results do not hold for standard models~\citep{qu2024rethink}. LMC has also been exploited for model-averaging and performance gains~\citep{wortsman2022model, ilharco2022editing, rame2022diverse}. Still despite all of this work, we lack a description for why LMC occurs. In particular: why is there a convex, high dimensional~\citep{yunis2022convexity} basin that models find shortly in training~\citep{frankle2020linear}, or after pretraining~\citep{neyshabur2020being, sadrtdinov2023stay}? We do not answer this question in full, but find a deep connection through the dynamics of singular vectors.

\subsubsection{Linear Mode Connectivity Correlates with Top Singular Vector Agreement}

We saw in Figure~\ref{fig:singular-vector-agreement} that top singular vectors converge in direction earlier. We also know that for models to display LMC, they need to share an early part of the training trajectory. Perhaps the top singular vectors become stable after this early stage, so we might expect mode-connected solutions to share these components. To examine this, we plot agreement between the singular vectors of the weight matrices at either endpoint of branches: 
\begin{align*}
 W^{(1)}(T) &= \sum_{j}^R \sigma_j(T) u_j(T) v_j(T)^\top\enspace,\\
 W^{(2)}(T) &= \sum_{k}^R \sigma'_k(T) u'_k(T) {v'_k}(T)^\top\enspace,   
\end{align*}
spawned from the same initialization in training. If the branches are split from an initialization on a trunk trajectory $W(t)$, we call $t$ the split point or epoch. We visualize the diagonal of $\lvert \langle u_j(T) v_j(T)^\top, u'_k(T) {v'_k}(T)^\top \rangle \rvert_{jk}$ vs.\ split epoch, where the absolute value is taken to ignore sign flips in SVD computation.

\begin{figure}[!t]
  \centering
  \begin{subfigure}[b]{0.24\linewidth}
    \centering
    \includegraphics[width=\linewidth]{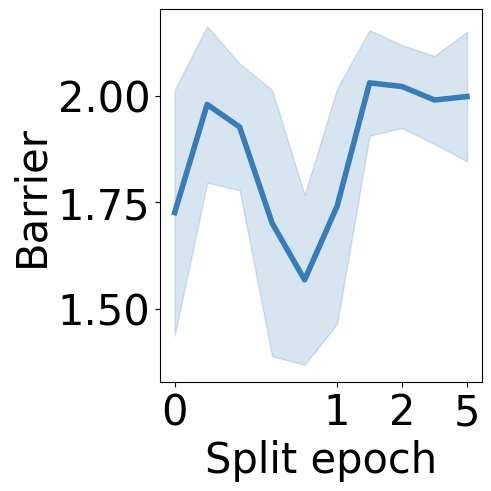}
    \\
    \includegraphics[width=\linewidth]{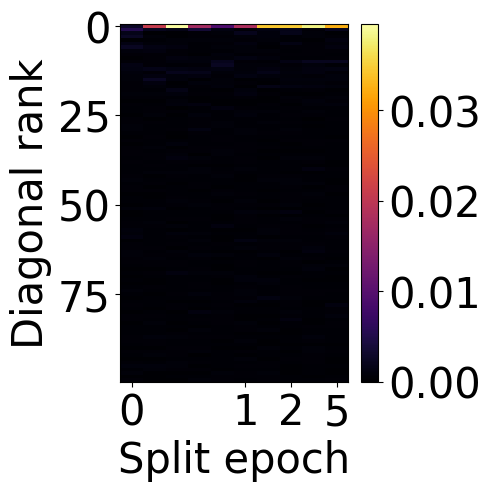}
    \\
    \includegraphics[width=\linewidth]{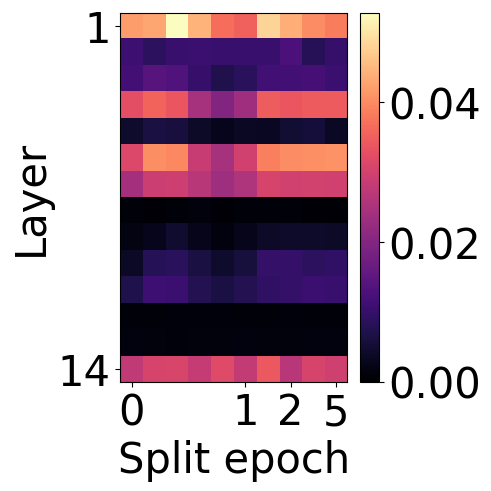}
    \caption{VGG}
  \end{subfigure}
  \begin{subfigure}[b]{0.24\linewidth}
    \centering
    \includegraphics[width=\linewidth]{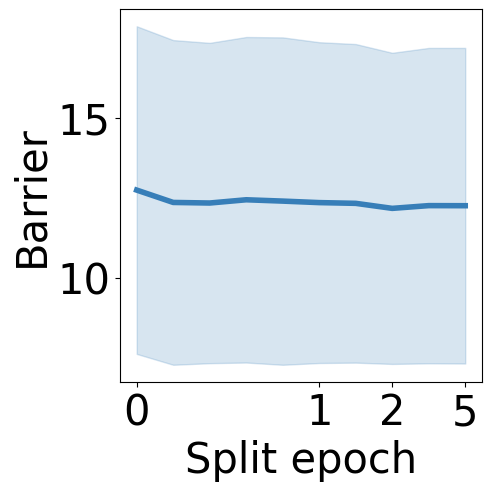}
    \\
    \includegraphics[width=\linewidth]{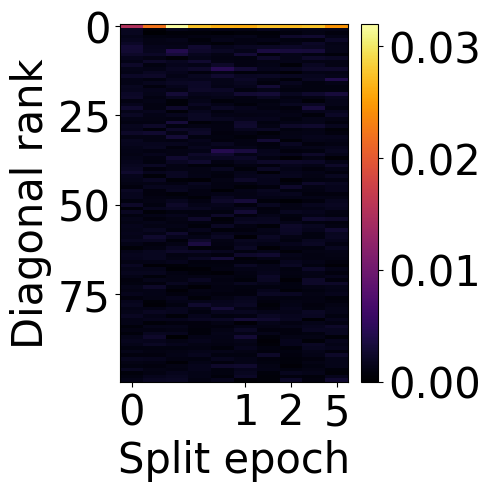}
    \\
    \includegraphics[width=\linewidth]{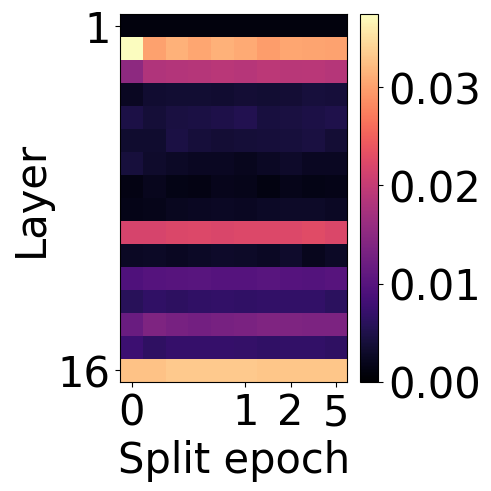}
    \caption{UNet}
  \end{subfigure}
  \begin{subfigure}[b]{0.24\linewidth}
    \centering
    \includegraphics[width=\linewidth]{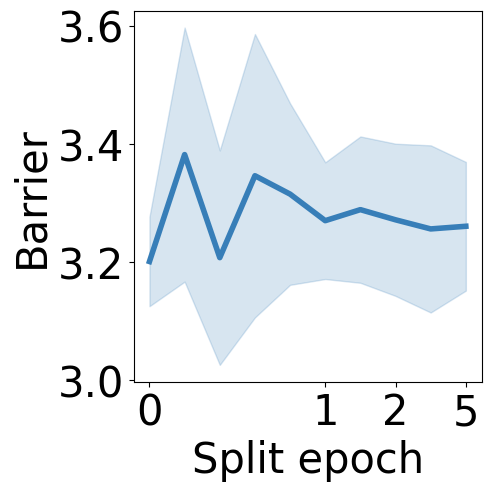}
    \\
    \includegraphics[width=\linewidth]{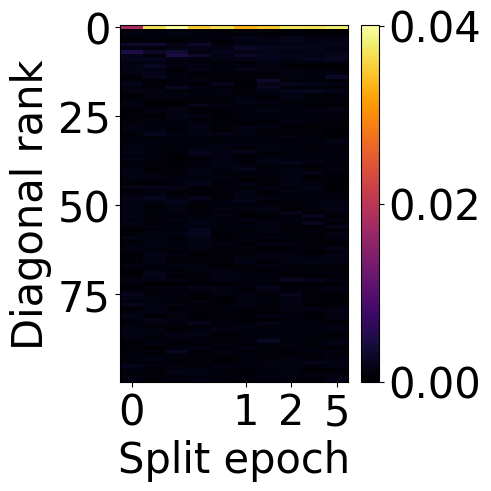}
    \\
    \includegraphics[width=\linewidth]{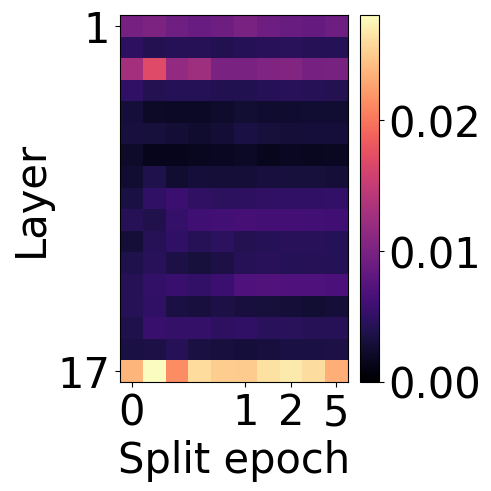}
    \caption{LSTM}
  \end{subfigure}
  \begin{subfigure}[b]{0.24\linewidth}
    \centering
    \includegraphics[width=\linewidth]{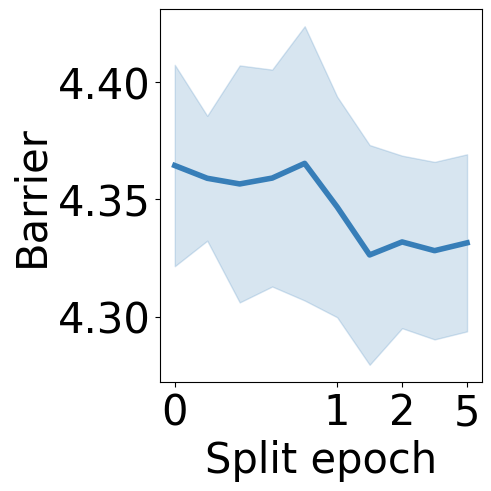}
    \\
    \includegraphics[width=\linewidth]{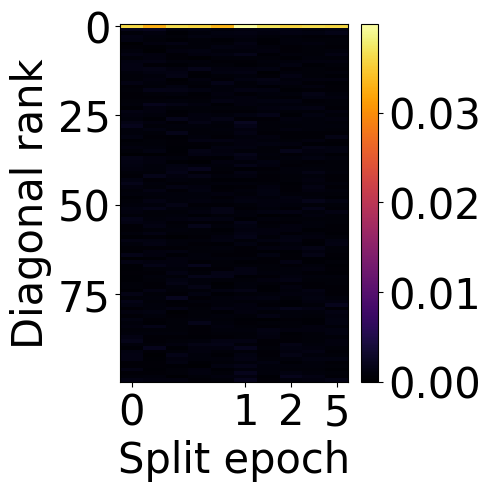}
    \\
    \includegraphics[width=\linewidth]{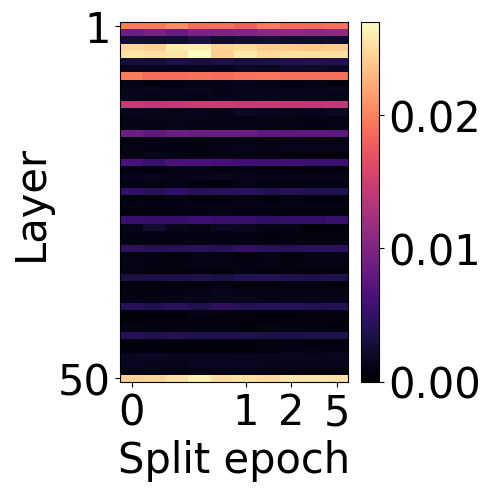}
    \caption{Transformer}
  \end{subfigure}
  \caption{\textbf{Top row:} Barrier size vs.\ split step. \textbf{Middle row:} singular vector agreement for a single matrix parameter between branch endpoints that do not share a common trunk, but do share split time and branch data order. \textbf{Bottom row:} summary statistic for singular vector agreement across layers. We see that when branches do not share a common trunk, there is neither LMC nor singular vector agreement, even though the optimization is otherwise the same.}
  \label{fig:lmc-cross-agreement}
\end{figure}

The technical definition of LMC requires measuring the bump, or barrier, in the loss surface along the linear interpolation between final checkpoints~\citep{frankle2020linear}. To measure this precisely, we use the definition from \citet{neyshabur2020being}, which is the maximum deviation from a linear interpolation in the loss, an empirical measure for convexity in this linear direction. When this deviation is 0, we consider the checkpoints to exhibit LMC. Please see Appendix~\ref{app:lmc-details} for details on the calculation. Given evidence in Figure~\ref{fig:pruned-performance} that top components are the most important for prediction, and that top components become stable before training has finished, it is plausible that LMC is connected to the stability of top singular vectors in the later portion of training.

This would mean that checkpoints that do not exhibit the LMC property should not share top singular vectors, while checkpoints that do exhibit the LMC property should share top singular vectors. We see in Figure~\ref{fig:lmc-agreement} that this is the case across models and tasks, where the alignment between endpoints is much stronger in top singular vectors. We also see no LMC and poor agreement in top components between branches that have initializations from different trunk trajectories, but with the same split epoch $t$ and the same branch data order in Figure~\ref{fig:lmc-cross-agreement}. Thus, these top directions are not a unique property of the architecture and data, but rather are dependent on initialization. It is notable that concurrent work~\citep{ito2024analysis} arrives at a similar conclusion: permutation solvers between optima match top singular vectors. Though the conclusions are similar, their experiments are primarily conducted on smaller scale settings, and only for permutation matching at the end of training. Here we connect these observations to the optimization behavior of SVDs throughout training, tying our previous observations on generalization in with weight averaging.

\subsubsection{Perturbing Breaks Linear Mode Connectivity and Singular Vector Agreement Simultaneously}

\begin{figure}[!t]
  \centering
  \begin{subfigure}[b]{0.24\linewidth}
    \centering
    \includegraphics[width=\linewidth]{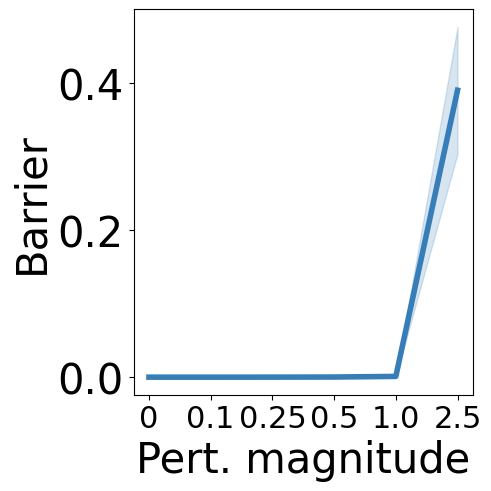}
    \\
    \includegraphics[width=\linewidth]{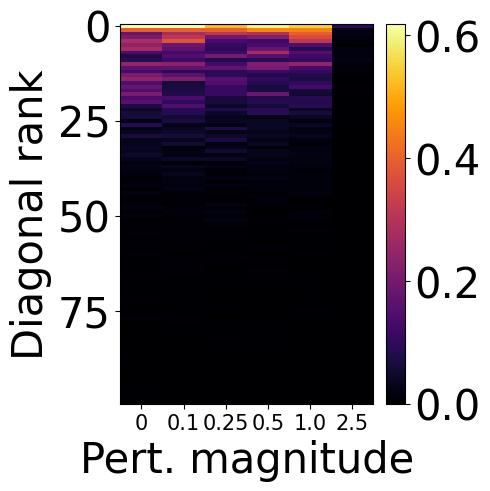}
    \\
    \includegraphics[width=\linewidth]{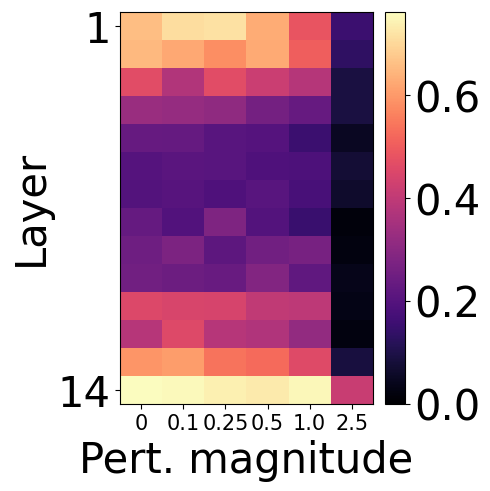}
    \caption{VGG}
  \end{subfigure}
  \begin{subfigure}[b]{0.24\linewidth}
    \centering
    \includegraphics[width=\linewidth]{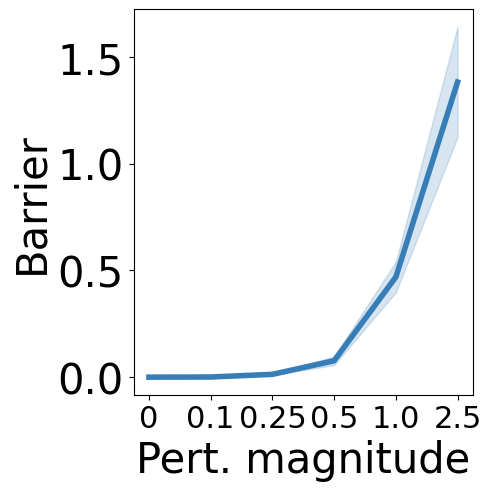}
    \\
    \includegraphics[width=\linewidth]{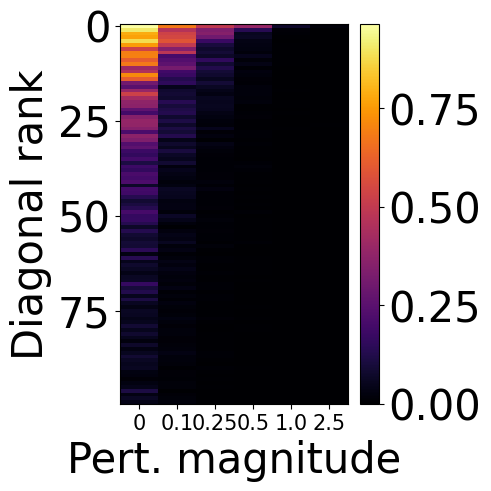}
    \\
    \includegraphics[width=\linewidth]{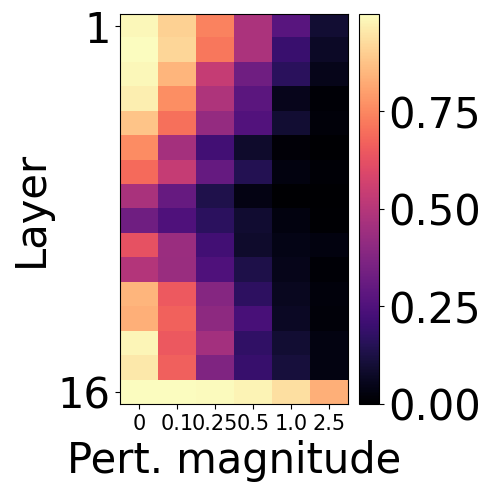}
    \caption{UNet}
  \end{subfigure}
  \begin{subfigure}[b]{0.24\linewidth}
    \centering
    \includegraphics[width=\linewidth]{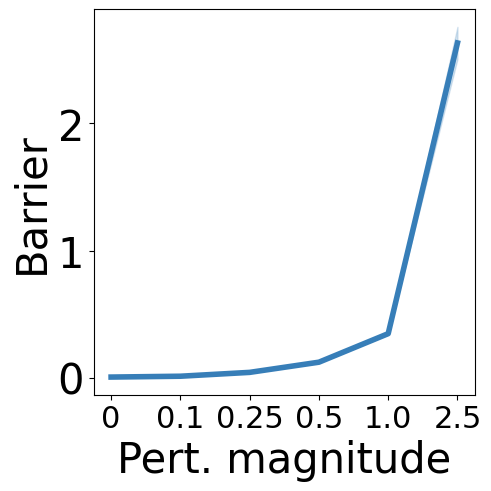}
    \\
    \includegraphics[width=\linewidth]{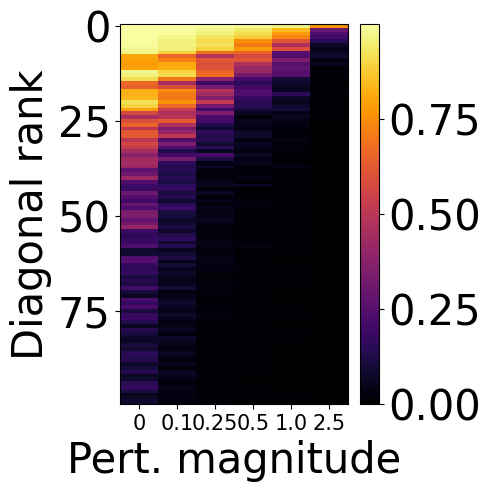}
    \\
    \includegraphics[width=\linewidth]{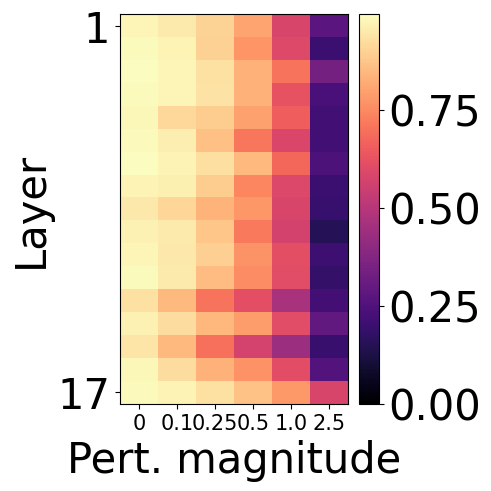}
    \caption{LSTM}
  \end{subfigure}
  \begin{subfigure}[b]{0.24\linewidth}
    \centering
    \includegraphics[width=\linewidth]{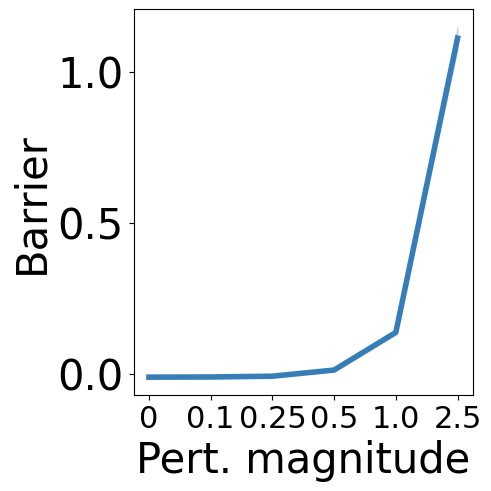}
    \\
    \includegraphics[width=\linewidth]{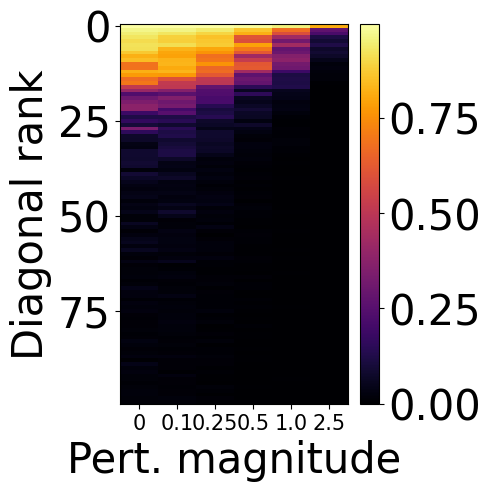}
    \\
    \includegraphics[width=\linewidth]{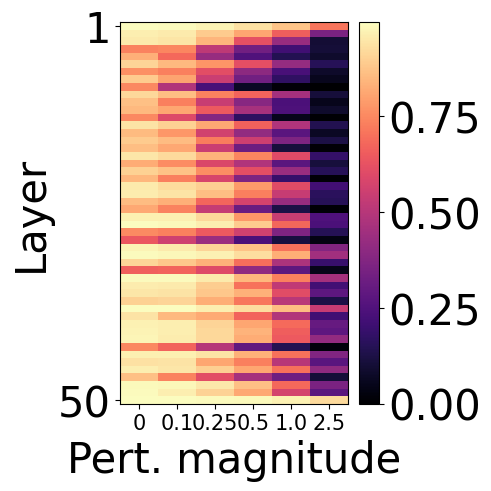}
    \caption{Transformer}
  \end{subfigure}
  \caption{\textbf{Top row:} Barrier size vs.\ perturbation magnitude. \textbf{Middle row:} singular vector agreement for a single matrix parameter between branch endpoints vs.\ perturbation magnitude. \textbf{Bottom row:} summary statistic for singular vector agreement across layers with perturbation magnitude. We see that whereas without perturbation models would exhibit LMC after training, with increasing perturbations the LMC property disappears simultaneously with the agreement in top singular vectors.}
  \label{fig:lmc-pert-agreement}
\end{figure}

To make the connection between top singular vectors and LMC even tighter, we intervene in the training process. If we add random perturbations to destabilize the components that will become the top components long before they have converged, and if singular vector agreement is tied to LMC, we would like to see that final models no longer exhibit the LMC property. Indeed this is the case. In Figure~\ref{fig:lmc-pert-agreement}, when increasingly large random perturbations are applied, the barrier between final checkpoints increases and the LMC behavior disappears. Please see Appendix~\ref{app:experimental-details} for details. In addition, the previously-strong singular vector agreement disappears simultaneously. Thus it seems this agreement is tied to linear mode connectivity.

We speculate that, due to the results in Figure~\ref{fig:pruned-performance} that show the top half of the SVDs are much more critical for performance, if these components are shared then interpolating will not affect performance much. Rather, interpolation will eliminate the orthogonal bottom components which may only make a minor impact on performance. If however the top components are not shared, then interpolating between two models will remove these components, leading to poor performance in between. Such observations may help in explaining the utility of pretraining~\citep{neyshabur2020being}, weight averaging~\citep{rame2022diverse, wortsman2022model, ilharco2022editing} or the use of LoRA~\citep{huh2022low} to replace full finetuning. Studying spectral dynamics allows us to see this deep structure.

\section{Discussion}
We provide an empirical perspective to understand deep learning through SVD dynamics. We first note a tendency toward rank minimization on a small scale in grokking, then expand these findings to practical networks and tasks. In addition, we find that weight decay, though it explicitly penalizes norm, implicitly promotes this low-rank bias. We also show that training with random or true labels differs in the rank and alignment of solutions found by optimization, echoing the rank-generalization connection during grokking. We go beyond remarks on generalization and show that magnitude pruning for lottery tickets acts similarly to low-rank pruning, and LMC coincides with the sharing of top singular vectors between checkpoints.

A comprehensive theory for all these results remains elusive. Our goal in this work is to provide these observations as a platform for a deeper understanding of deep learning. Notably, the observed spectral dynamics appear consistent across diverse settings, even without restrictive assumptions like balanced initialization, linearity, or small weight scales. This suggests a common underlying mechanism that further theoretical work may be able to model. One limitation of our study is the focus on linear alignment between neighboring layers. In linear systems, this alignment is a proxy for the network's ability to pass signals from input to output, but once nonlinearities are introduced, it is unclear how this proxy applies. We would guess there is still alignment in larger systems, but it may need to be viewed post-activation for particular examples. This is a promising direction for future exploration.

On the empirical side, several interesting problems present themselves. Interpretability of neural networks is a growing area of research~\citep{nanda2023progress}, and there already exist efforts to interpret singular vectors of convolutional weights~\citep{praggastis2022svd}. There may also be connections to other unexplained phenomena such as double descent~\citep{belkin2019reconciling, nakkiran2021deep, davies2022unifying} or adversarial examples~\citep{szegedy2013intriguing, ilyas2019adversarial, hendrycks2021natural}. For example, is it actually the case that low-rank models are more robust? The answers to these questions may help design better optimizers or diagnose deployment risks in the wild. There are also concerns of safety~\citep{bai2022constitutional, mazeika2024harmbench} that a better understanding of neural networks can alleviate~\citep{burns2023discovering, park2024linear}. In particular with low-rank models, interpretability may become easier as there is much less model to study. We present this large empirical exploration in an effort to deepen scientific understanding, provide fertile ground for new theoretical assumptions, and offer inspiration for better engineering. We believe the developed perspective will be useful in bridging gaps between many different attempts to explain neural networks.

So we have seen that low rank dynamics are tightly related to the development of the convex mode and ties nicely with prior work on lottery tickets~\citep{frankle2020linear}. Given their central role in neural network optimization, we believe that such low-rank dynamics can be understood as the primary feature learning that happens during optimization. This fits nicely with previous work leveraging low-rank pruning for compression and performance~\citep{yu2017compressing, sharma2024truth}, and other studies that show different modes (defined by different low-rank cores) learn different functions~\citep{lubana2023mechanistic}. We now turn to using these spectral dynamics to overcome simplicity bias. The key idea is that the low-rank bias can be used as a definition for simplicity bias, meaning by manipulating the spectral dynamics we can overcome spurious correlations. The following chapter details this approach.

\chapter{Overcoming Spurious Correlations through Diverse Feature Learning}\label{chp:diverse_features}
{We pursue a generic method to learn more diverse features from existing data, without task-specific augmentations or reliance on external information. Due to simplicity bias, such feature learning is often blocked in standard optimization. Taking inspiration from spectral dynamics as a definition for simplicity bias, we demonstrate that projected gradient descent off of top singular vectors suffices to learn complex features in an agnostic way in a difficult testbed. In more complex cases, we show large benefits for diversity losses, without the drawbacks or assumptions of prior work. We also show that such benefits extend into language modeling, beyond the commonly-studied classification setting.}
\section{Introduction}
Neural networks show simplicity bias~\citep{rahaman2019spectral, huh2021low, pezeshki2021gradient}, which on a large scale can be responsible for issues of class bias and adversarial hacking of features~\citep{zou2023universal, rumbelow2023solidgoldmagikarp}.

One way to understand this is that the optimization objective encourages neural networks to find a set of features to solve a given task, but the dynamics of optimization lead only to the minimal set of features~\citep{pezeshki2021gradient}. Intuitively, gradients are very small for examples that are already well learned, so there is no incentive to learn more features for those examples. A classic demonstration of this is networks learning to classify cows by the presence of grass in the background~\citep{beery2018recognition}. If grass is always present, why pay attention to the cow?

Many methods have been proposed to diversify feature learning. One general approach upsamples misclassified examples~\citep{liu2021just, zhang2022rich}, but such methods ignore further information from examples already correctly classified. Other methods try to enforce some notion of disagreement using information about the test distribution~\citep{pagliardini2022agree, lee2022diversify, to2025diverse}, but this requires a particular test distribution which may not be known ahead of time. \citet{zhang2023learning} show that the simple ensembles of parallel training runs have significant diversity in their representations, when compared to single runs of larger models, but such a method still falls prey to simplicity biases for any individual run.

As far as we know, there is no generic solution to the problem of simplicity bias. \citet{pezeshki2021gradient} suggest a regularizer that makes competing correlations less imbalanced, but this requires a dataset design where a single simple feature is not perfectly predictive. \citet{shah2020pitfalls} demonstrate that increased model size, adversarial training, and standard regularization techniques are also insufficient. For any particular data distribution, it might be possible to design specific augmentations that address spurious correlations, for example masking backgrounds in the case of cows on grass, but it may not be clear how to do this in general tasks like language modeling.

At the end of the day, what we are left with is the ability to collect more and more data and grow the training distribution to encompass all possible combinations of features, hoping that we learn diverse features through brute force scaling~\citep{kaplan2020scaling}. This is an accurate reflection of current practice in training large language models, where diverse and large data appears to be the most critical component~\citep{brown2020language, achiam2023gpt}.

Instead we propose a method to overcome simplicity bias generically in neural networks. First, we posit that the most important features can be found in the top ranks of neural network weights given extensive historical evidence of concentration during optimization and performance during pruning (see Chapter~\ref{chp:spectral_dynamics}). Next we perform an iterative process, training a neural network, collecting its top features (the top ranks of its weights) and then running subsequent rounds of training with projected gradient descent off of those directions so as to force the neural network to learn new features. Though naive in nature, we find that this iterative process is sufficient to discover more diverse features in a case of extreme simplicity bias. We then explore scaling of this method to more complex tasks, developing deeper understanding along the way.

In particular, we ask and answer the following questions:
\begin{itemize}
\item \textbf{Can we use our understanding of spectral dynamics to address spurious correlations?} We find that a projected gradient descent method based on our understanding of spectral dynamics can force a neural network to extract more information from the data without relying on any task-specific information;
\item \textbf{Does this method scale to larger networks?} We then scale this method to larger image classification networks, and find that it fails, but a related method takes its place, showing substantial improvement over prior methods;
\item \textbf{Can we apply this to even more complex tasks?} We then move to language modeling, where we find the task is too difficult for the assumptions of our prior methods. Even though the rich feature structure exists in language modeling, we demonstrate that optimization makes it very difficult to take advantage of this structure. Through an extensive raft of experiments we explore this space, and make prescriptions for the future.
\end{itemize}


\section{Simplicity Bias and Spectral Dynamics}

\begin{figure}[t]
\centering
\includegraphics[width=0.8\textwidth]{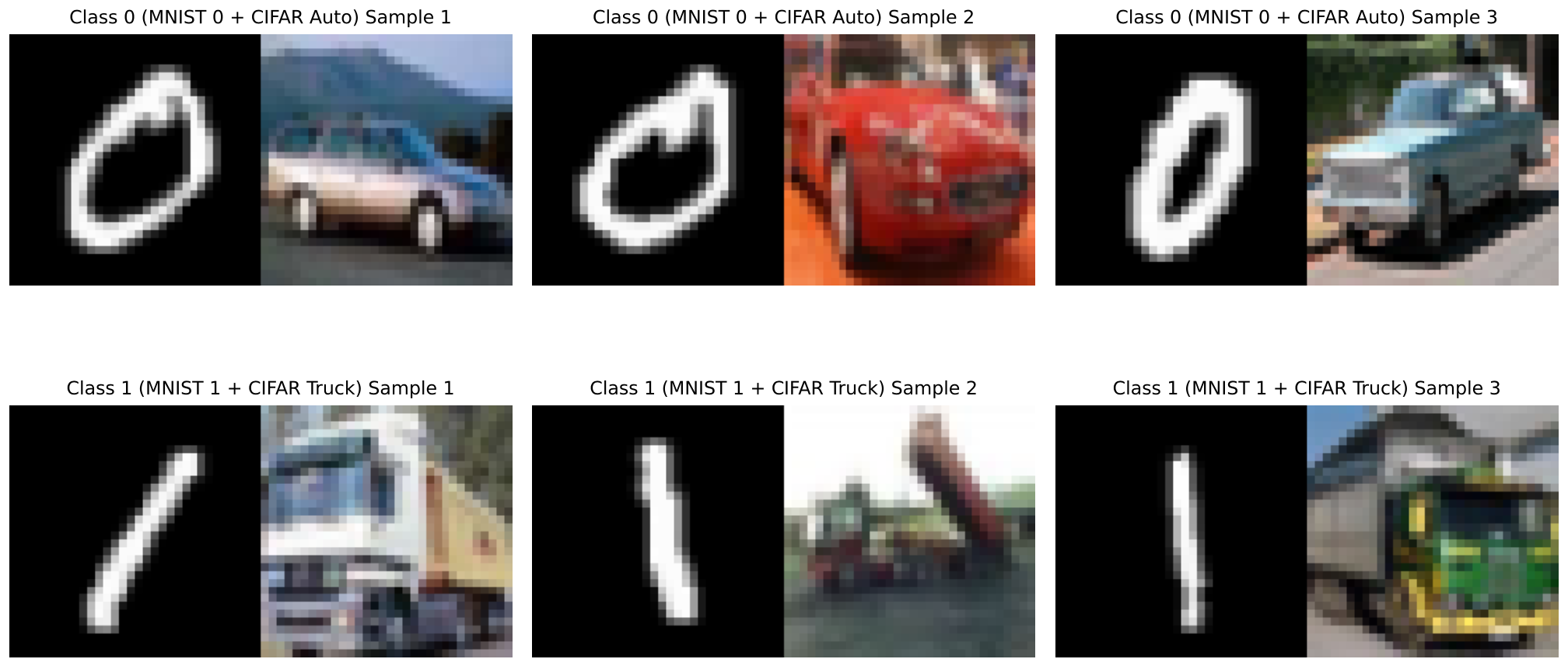}
\caption{Examples from the MNIST-CIFAR dataset in \citet{shah2020pitfalls}.}
\label{fig:mnist_cifar}
\end{figure}

We begin with a simple model system, drawn from prior work. \citet{shah2020pitfalls} showed, through carefully crafted datasets, that neural networks exploit the simplest features in a dataset, even if a more complex feature may hold useful information.

In particular, they introduced MNIST-CIFAR, a binary classification dataset with a controlled structure. Each example combines two images, one from MNIST, and one from CIFAR. Class 1 corresponds to a concatenation of an MNIST 0 and a CIFAR car, while class 2 corresponds to an MNIST 1 and a CIFAR truck. Both the MNIST and CIFAR classes are perfectly correlated with the label (Figure~\ref{fig:mnist_cifar}).

\citet{shah2020pitfalls} trained a variety of networks on this dataset, and showed that even with scale, regularization, adversarial training, and ensembling, no model paid attention to the CIFAR half of images. The predictions of the trained models were entirely dependent on the MNIST half. Large perturbations or substitutions in the CIFAR half did not change the result.

We use this MNIST-CIFAR task to illustrate the benefit of spectral dynamics as feature learning. By taking the top ranks as the most important features, we can find neural networks generically successfully pay attention to CIFAR.

\subsection{Projected Gradient Descent for Diverse Feature Learning}

Our method consists of training successive rounds of neural networks with modified optimization in each round. After a round, we accumulate a set of orthogonal directions and run projected gradient descent off of those directions for all parameters in the network in the next round. Roughly what this means is we prevent a new network from using the directions, which we believe correspond to features, of the previous rounds. Thus, the network will learn more complex and diverse features on the same data.

We train N rounds to convergence. In the first round we train normally. At the end of each round, for each parameter matrix we compute the SVD  $W_l = \sum_{i=1}^R \sigma_i u_i v_i^\top$, then we stack the top $k$ ranks on the input side into a matrix:

\begin{equation}
    \tilde{V}^\top = \begin{bmatrix} 
        v_1^\top \\
        ... \\
        v_k^\top
    \end{bmatrix}.
\end{equation}

After constructing this stack we reinitialize each $W_l$ and its corresponding optimization parameters, and start a new training run. With every gradient step we project the input and space of the gradients off of our previously collected $\tilde{V}$ like:

\begin{equation}
    \nabla W_l' \leftarrow \nabla W_l (I - \tilde{V}\tilde{V}^\top)
\end{equation}
and the same for the parameters because of possible momentum and bias issues.

When we reach the end of a training round, we append the top $k$ directions of the new parameters to our $\tilde{V}$ buffer, growing an ever larger list of orthogonal directions to avoid during training. This means that optimization is impossible if the number of rounds times $k$ is greater than the rank of $W_l$, but practically speaking that happens after many many rounds of training.

For the experiments, we typically use $k=3$ and 2000 epochs of training so that models reach convergence before computing SVDs. This method looks quite similar to prior work in continual learning~\citep{saha2021gradient}, with a few key differences. First, we run repeated iterations on the same data, not on unrelated classes. Second, we project using the SVD of the weight matrix, not the SVD of collected activations, making our method substantially more memory efficient.

\subsection{Experiments}

We use the previously described MNIST-CIFAR~\citep{shah2020pitfalls} dataset (see Figure~\ref{fig:mnist_cifar} for examples). We train a simple 4-layer MLP for binary classification. See Appendix~\ref{app:mnist_cifar_details} for more details.

\begin{figure}[t]
\centering
\includegraphics[width=\textwidth]{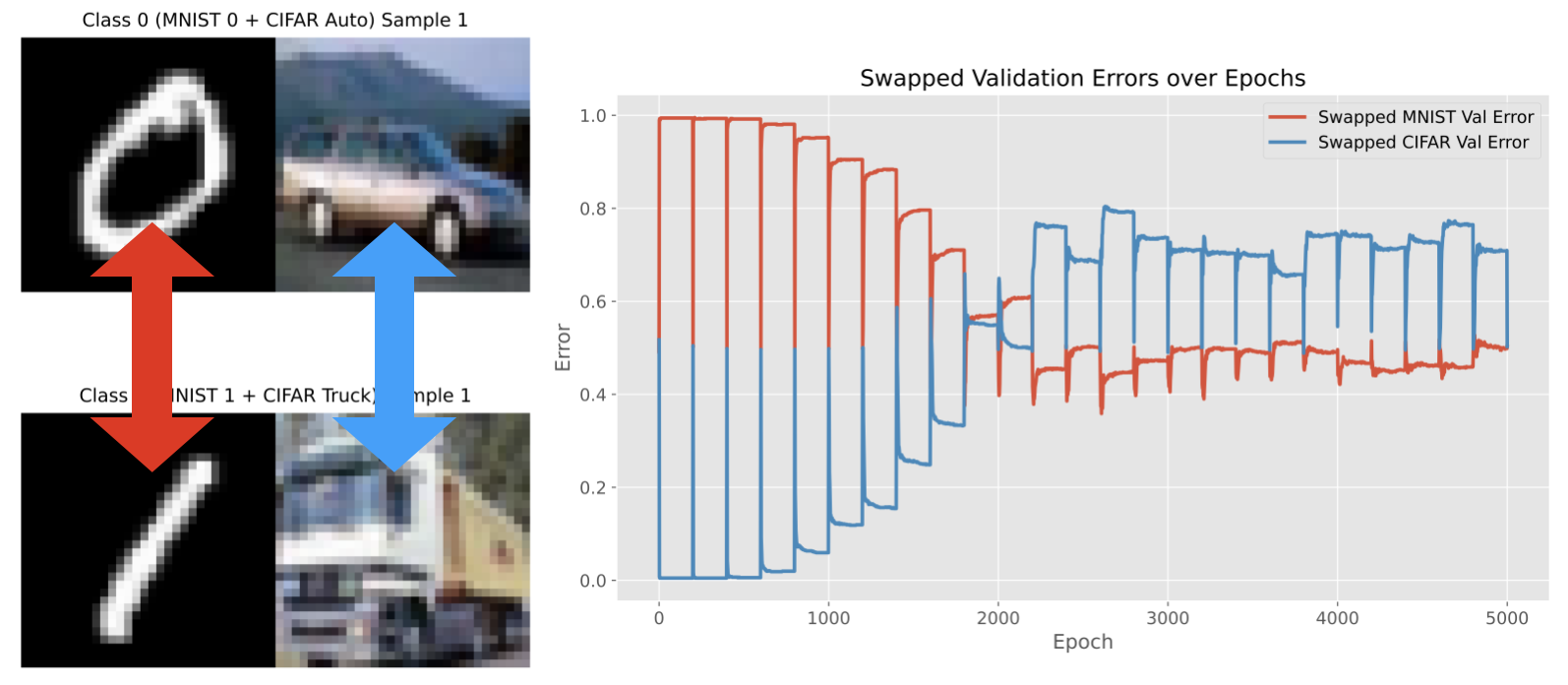}
\caption{Evaluation protocol for experiments. After training on MNIST-CIFAR, we swap the classes of either the MNIST half, or the CIFAR half of the image and evaluate on that setting. When models rely entirely on MNIST, error on the swapped task degrades to 100\%, and similarly for CIFAR. On the right we plot the errors over successive rounds of PGD. We see that models start to learn CIFAR after the early rounds of PGD, coming to rely quite heavily on CIFAR in later rounds. Spikes correspond to restarting a new round of training.}
\label{fig:swap_protocol}
\end{figure}

We evaluate training error and validation error as usual. We also include a few different evaluations to diagnose whether the model learns any CIFAR features, namely: 1) we evaluate swapping the classes of the MNIST portions of the input while keeping the label fixed, and 2) we evaluate swapping the CIFAR portions of the input while keeping the label fixed. If the model relies entirely on MNIST, we should see that swapping the MNIST images should flip the predictions entirely, while swapping the CIFAR portion should result in no change at all. See Figure~\ref{fig:swap_protocol} for an illustration.

\begin{figure}[t]
\centering
\begin{subfigure}[b]{0.48\columnwidth}
\centering
\includegraphics[width=\textwidth]{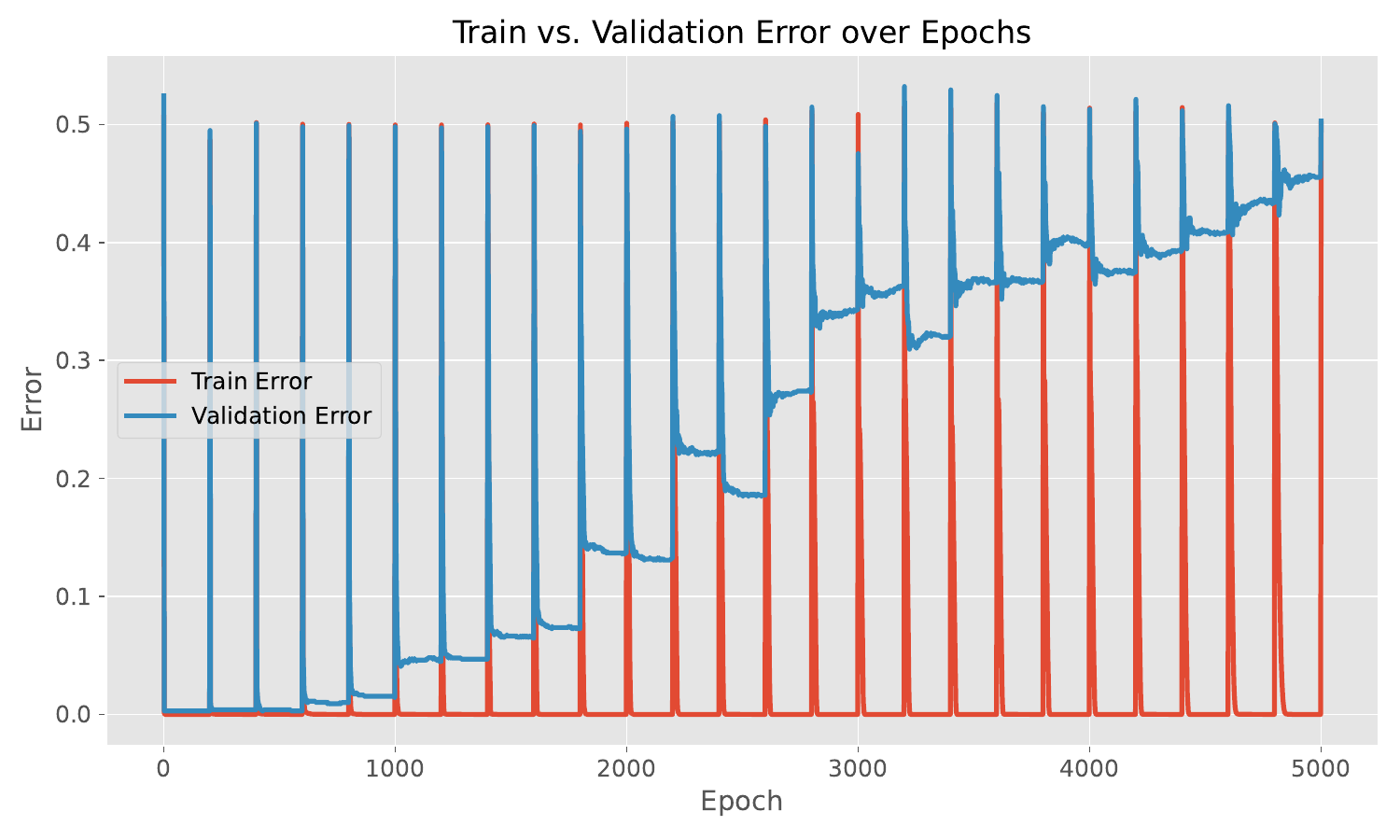}
\caption{Normal}
\end{subfigure}
\hfil
\begin{subfigure}[b]{0.48\columnwidth}
\centering
\includegraphics[width=\textwidth]{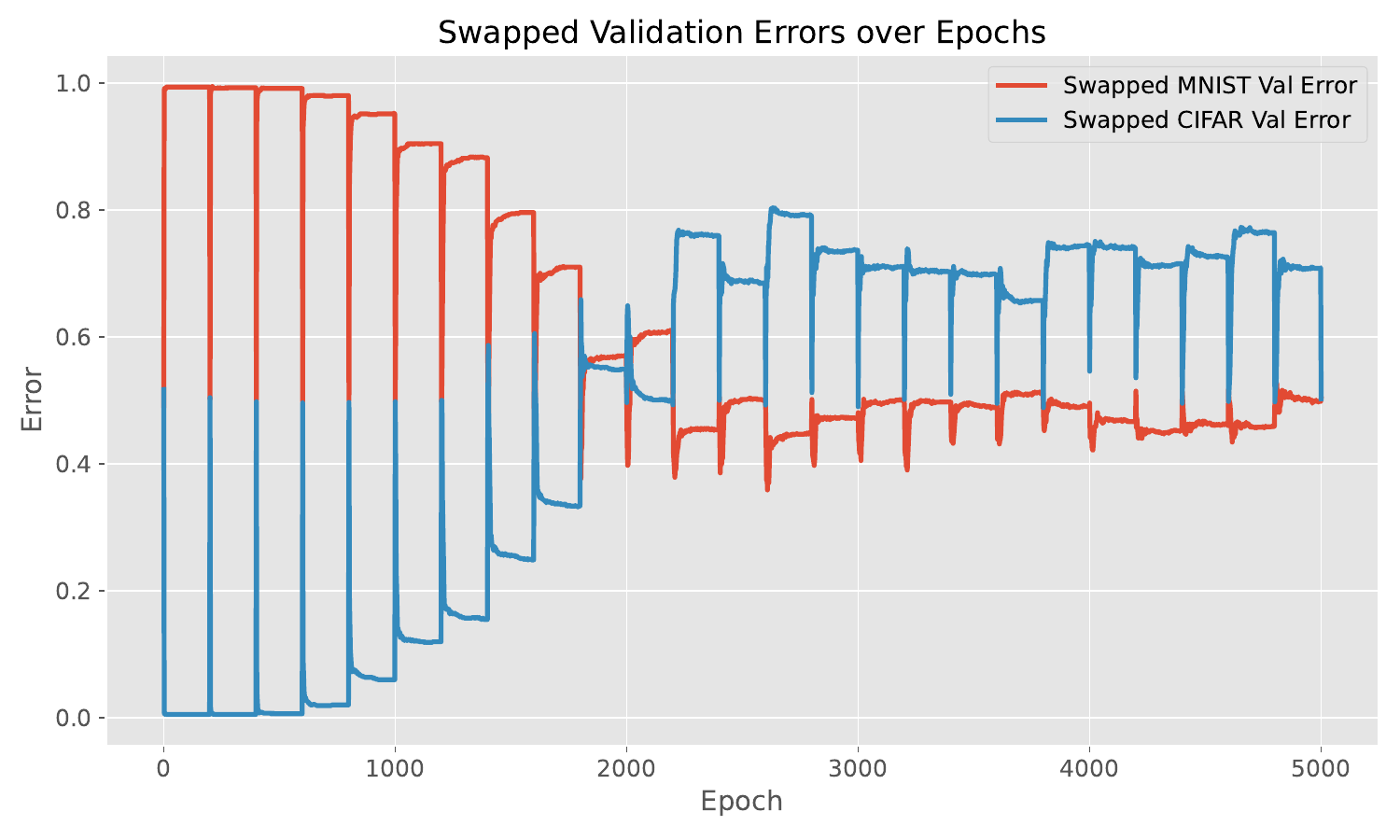}
\caption{Swapped-class}
\end{subfigure}
\caption{\textbf{Left:} Training and validation error for successive rounds with our proposed PGD method on top singular vectors. \textbf{Right:} Swapped-class errors throughout rounds. We see that as the round number increases, validation error decreases due to the complexity of the task increasing. In addition, the model increasingly relies more on CIFAR, as the swapped CIFAR error goes from near 0 to 60\% and above. This indicates the model is learning to extract more complex signals which are less predictive but still useful.}
\label{fig:mnist_cifar_pgd}
\end{figure}

We see the results for our PGD run in Figure~\ref{fig:mnist_cifar_pgd}. Notably, training error is near 0 for every round of training, but validation error climbs due to the reliance on more complex CIFAR features. We see that swapping the MNIST digits in early rounds of training yields nearly 100\% error, demonstrating a reliance entirely on MNIST. As PGD proceeds, this decreases to less than random chance, or near the total validation error. Conversely, swapping CIFAR images initially results in no change in predictions when compared to validation error, but in later rounds, swapping CIFAR yields errors above 70\%. This shows that PGD successfully incentivizes later members of the ensemble to extract diverse features, exactly in line with our hopes.

\begin{figure}[t]
\centering
\includegraphics[width=\textwidth]{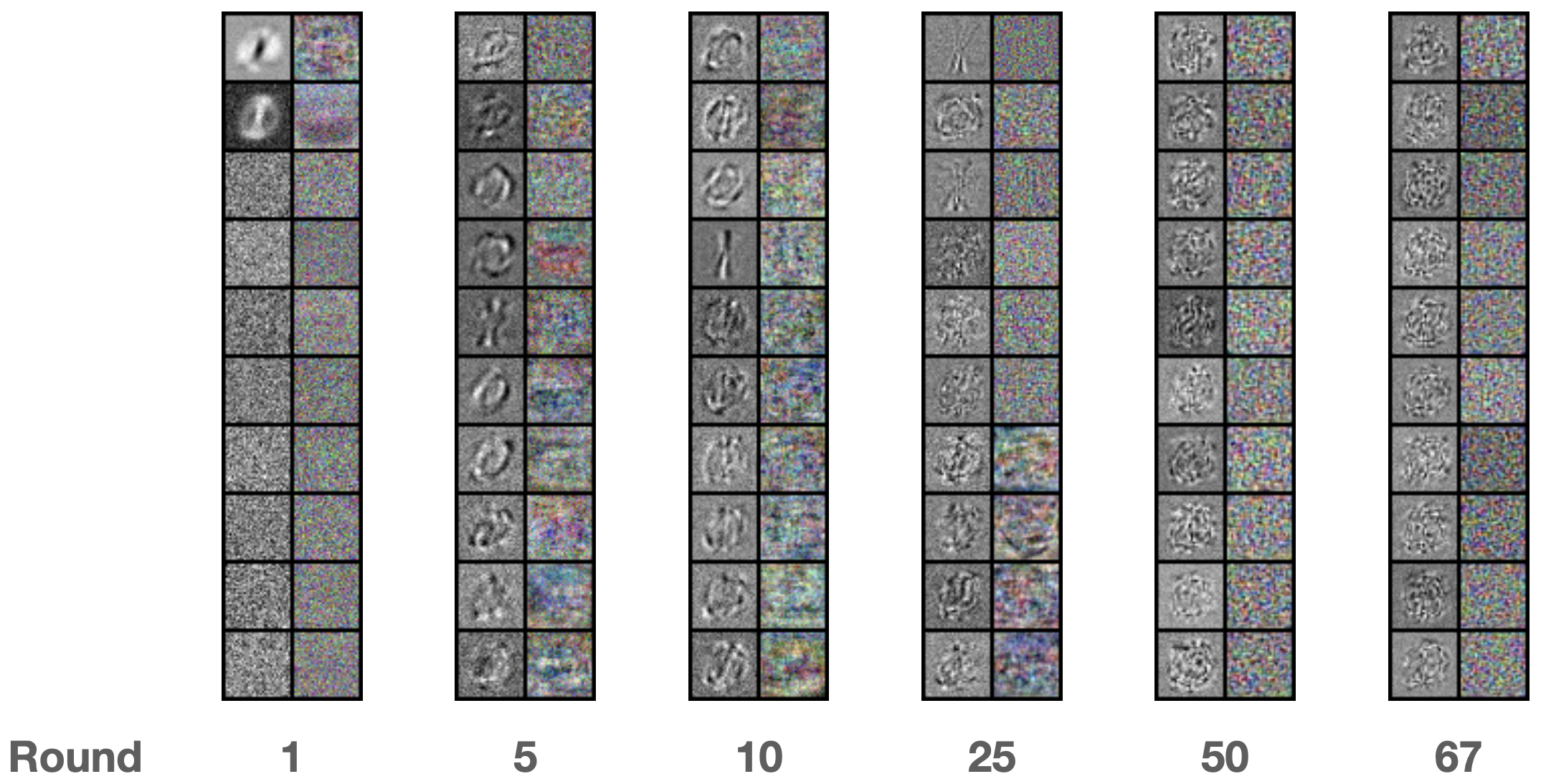}
\caption{First layer singular vectors over rounds of PGD training. Each set of columns corresponds to the MNIST and CIFAR halves of the reshaped singular vector. Descending down the rows corresponds to increasing rank index when sorted by singular value. We see initially that the first layer singular vectors form digit classifiers exclusively. In middle rounds the network focuses on higher and higher frequency signals in MNIST. In late rounds, all but the high frequency signals are present in MNIST, hence the model starts to leverage the CIFAR half of the image.}
\label{fig:mnist_cifar_singvec_evol}
\end{figure}

In Figure~\ref{fig:mnist_cifar_singvec_evol}, we show the evolution of the top 10 singular vectors of the first layer in the model across different rounds of training. In the first round, the top two singular vectors correspond exactly to MNIST digit detectors, while there is little signal in the CIFAR portion. Later rounds have finer and finer grained curve detectors, and a growing sensitivity in the CIFAR portion. At some point the model is forced to rely on CIFAR due to the destruction of the signal in MNIST. These visualizations concretely demonstrate that PGD is successfully blocking the simple features in MNIST-CIFAR, forcing a simple MLP to overcome its own simplicity bias.

\begin{figure}[t]
\centering
\begin{subfigure}[b]{0.48\columnwidth}
\centering
\includegraphics[width=\textwidth]{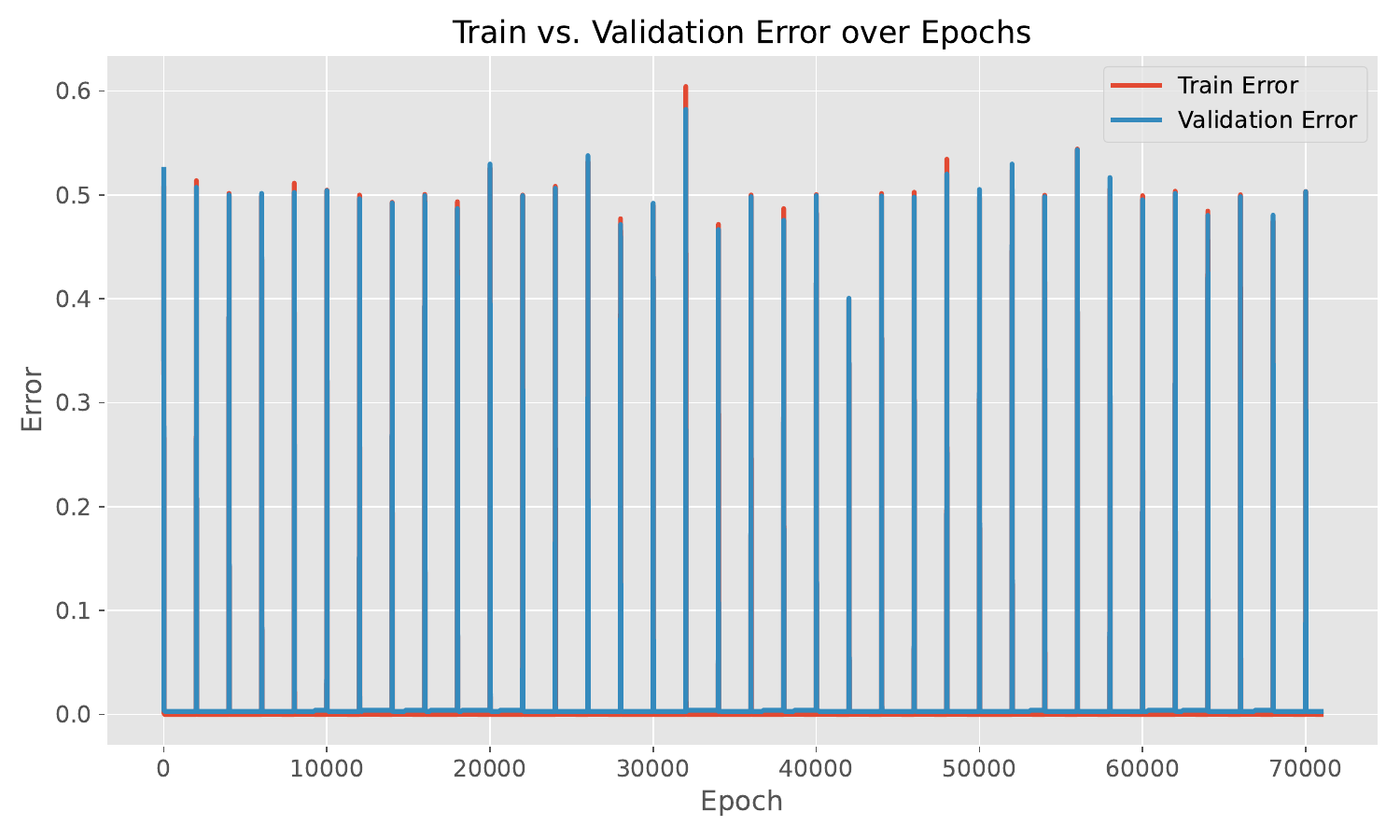}
\caption{Normal}
\end{subfigure}
\hfil
\begin{subfigure}[b]{0.48\columnwidth}
\centering
\includegraphics[width=\textwidth]{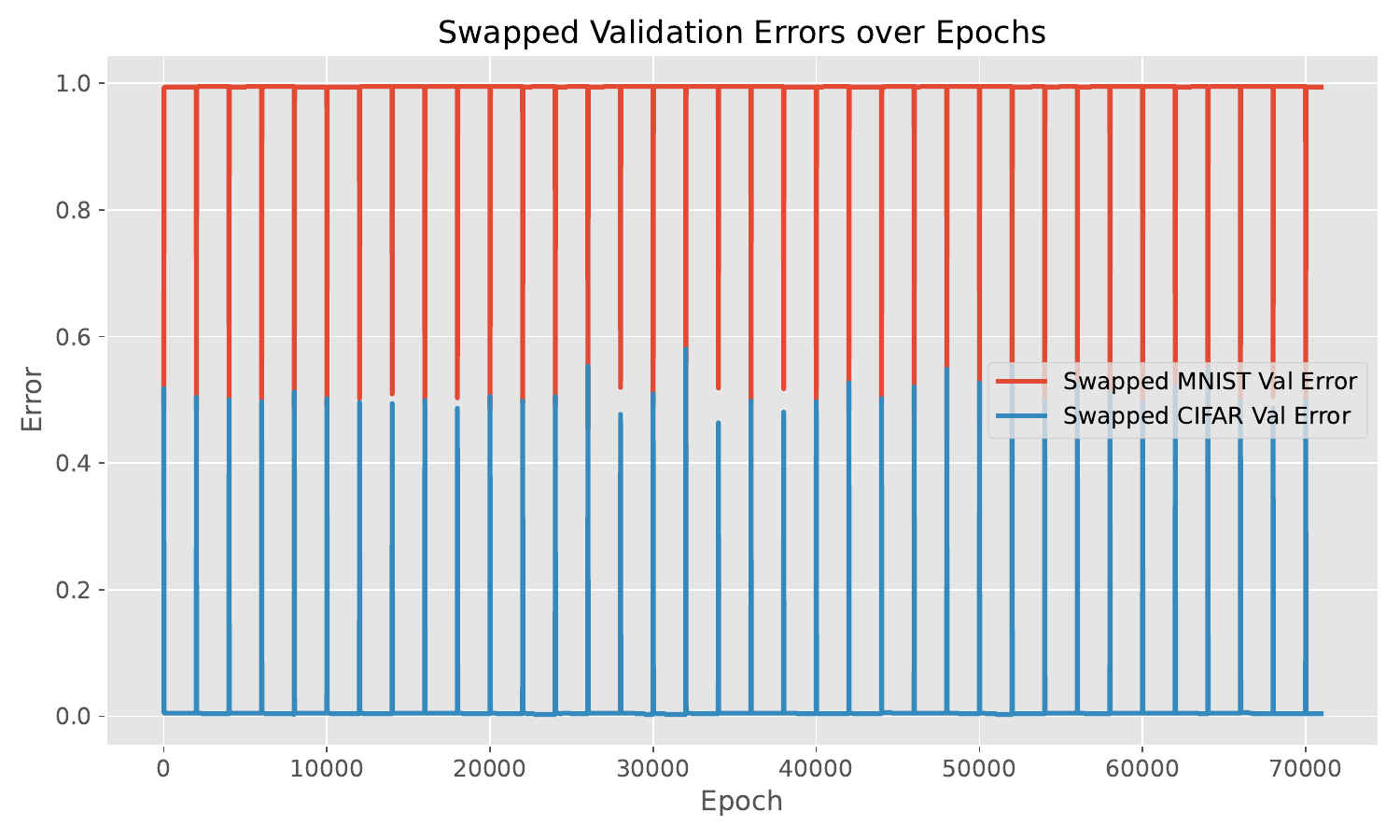}
\caption{Swapped-class}
\end{subfigure}
\caption{\textbf{Left:} Training and validation error for successive rounds with PGD where we flatten the gradient and singular vectors before projection. \textbf{Right:} Swapped-class errors throughout rounds. We see that no new features are learned when ignoring the matrix structure of the parameters, so the spectral design is critical.}
\label{fig:mnist_cifar_pgdflat}
\end{figure}

\begin{figure}[t]
\centering
\begin{subfigure}[b]{0.48\columnwidth}
\centering
\includegraphics[width=\textwidth]{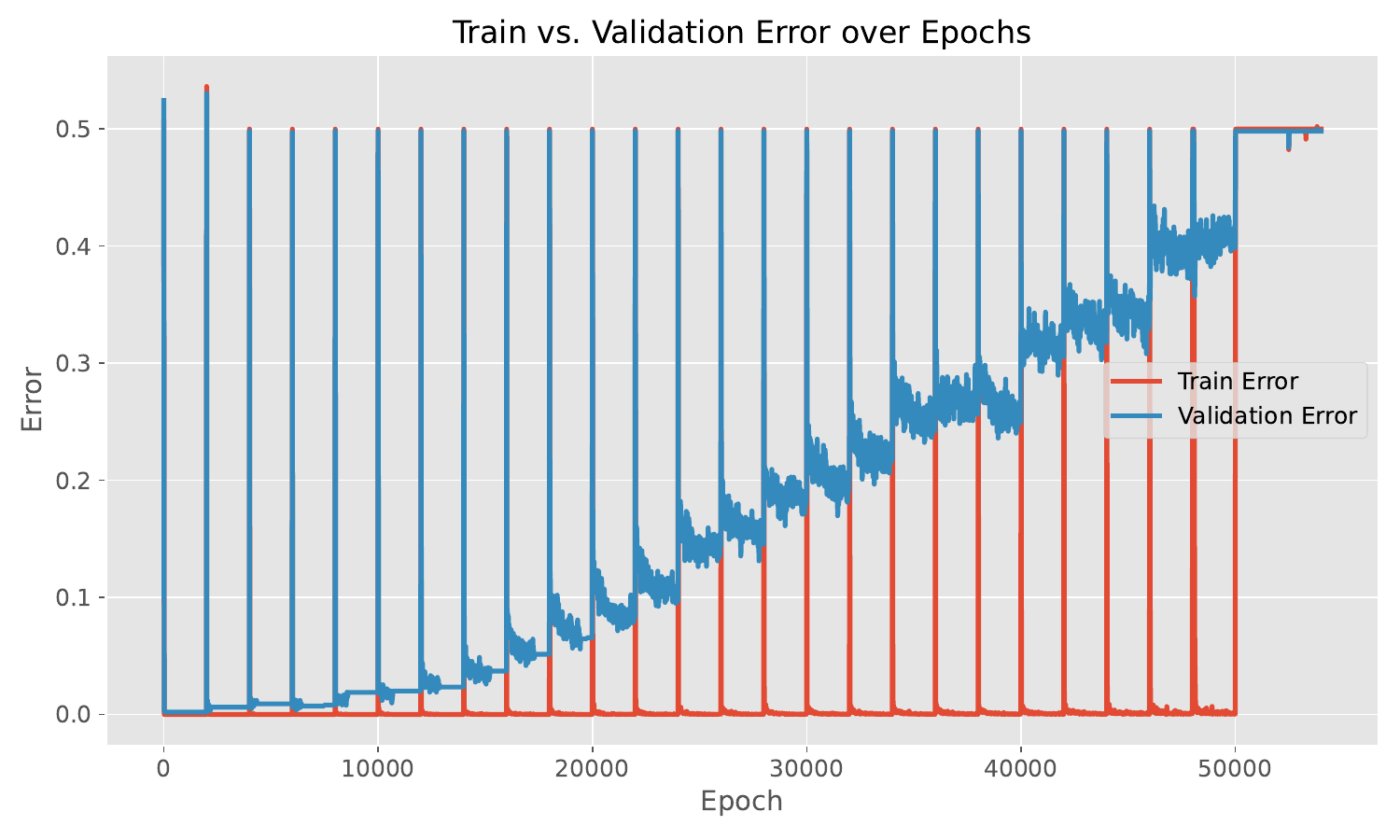}
\caption{Normal}
\end{subfigure}
\hfil
\begin{subfigure}[b]{0.48\columnwidth}
\centering
\includegraphics[width=\textwidth]{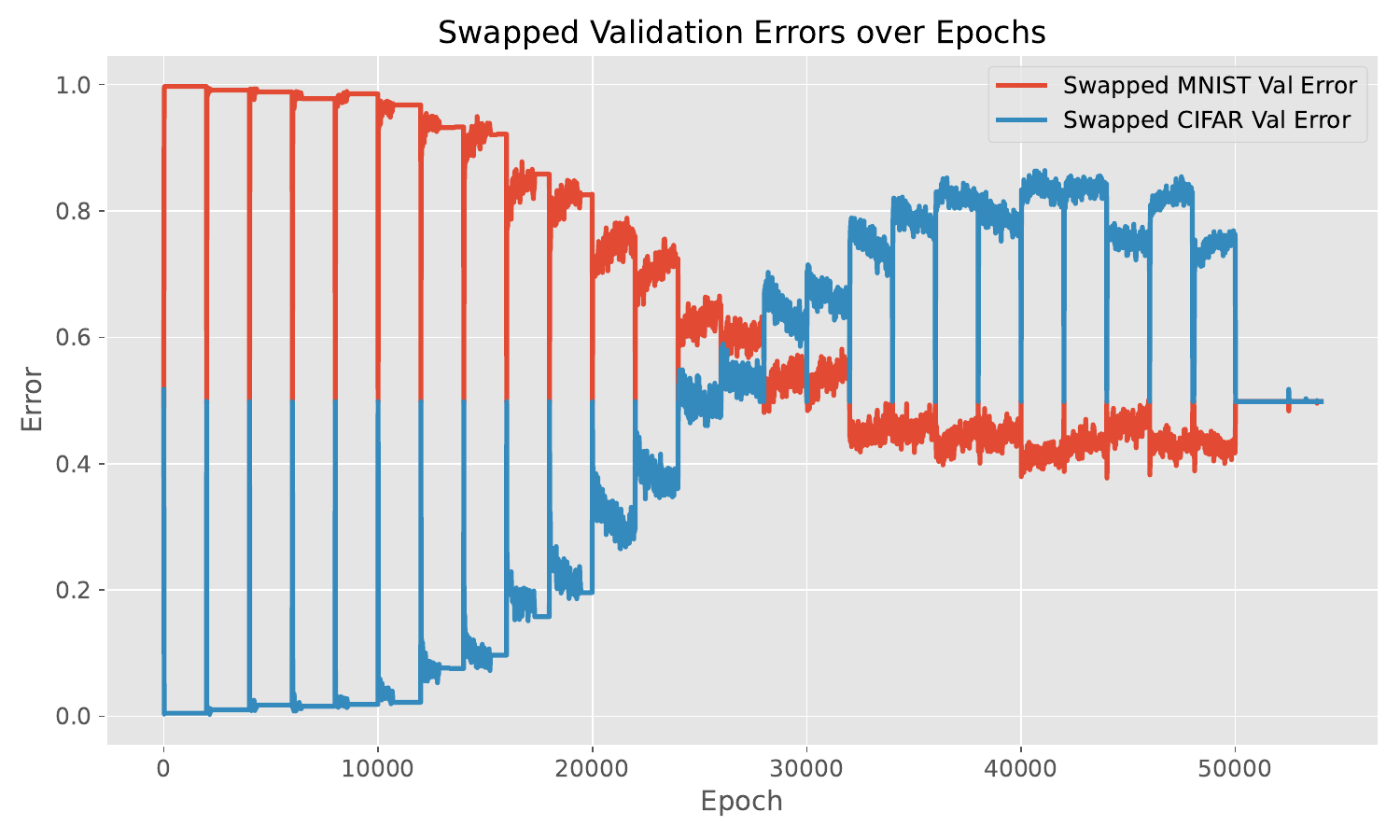}
\caption{Swapped-class}
\end{subfigure}
\caption{\textbf{Left:} Training and validation error for successive rounds with Muon. \textbf{Right:} Swapped-class errors throughout rounds. Even though Muon is incentivized to give relatively more gradient to less prominent features (through higher-rank upweighting), it is insufficient for picking up CIFAR.}
\label{fig:mnist_cifar_muon}
\end{figure}

In Figure~\ref{fig:mnist_cifar_pgdflat}, we examine the performance of a similar method that orthogonalizes the gradient-as-a-vector (similar to OrthoGrad~\citep{prieto2025grokking}). We see that this method does not result in significant learning of CIFAR. This is likely because orthogonality in such a high dimensional space is easy to satisfy trivially. Hence it is critical to take into account the matrix structure of the model. In addition we see in Figure~\ref{fig:mnist_cifar_muon} that Muon~\citep{jordan2024muon}, an optimizer designed to take advantage of features downspectrum, is also not able to take advantage of the CIFAR signal in the dataset. Thus, even more modern tricks and optimizer designs are unable to overcome the foundational issue of simplicity bias.

\begin{table}[htbp]
\centering
\caption{Probing accuracy for CIFAR on middle layers for an ensemble of 5 models drawn from early or late rounds of PGD. We see that early rounds are substantially worse at extracting CIFAR information from the middle of the model, reinforcing the argument that PGD forces the model to learn more.}
\label{tab:mnist_cifar_probe}
\begin{tabular}{lcc}
\toprule
\textbf{Model} & \textbf{Accuracy} \\
\midrule
Early & $0.651 \pm 0.005$ \\
Late & $0.693 \pm 0.039$ \\
\bottomrule
\end{tabular}
\end{table}

As our evaluations are based on swapping inputs and relying on the output predictions of the final model, it is possible that we are missing some encoding of CIFAR that is present in middle layers of the model. In Table~\ref{tab:mnist_cifar_probe} we concatenate all representations from middle layers and train a linear probe to classify the CIFAR car vs. truck using an ensemble of models drawn from early rounds of PGD, and drawn from late ones. More details on the experimental design in Appendix~\ref{app:mnist_cifar_details}. We see that later models have a significantly higher probing performance than earlier ones, reinforcing the conclusion that PGD extracts diverse information in successive rounds. The absolute numbers here are likely high as the probe is near the capacity of the original model due to its size.

All of the preceding experiments were performed on a very simple model system, but we hope they give the reader a sense of the potential issue on larger scale systems. Next we turn to larger systems, and show that very similar issues still plague neural network training. We find that our naive design of PGD fails in these larger systems, so we introduce a modified method.


\section{Diverse Feature Learning in Representation Space}
Much ink has been spilled about rich or diverse feature learning in neural networks. \citet{zhang2023learning} demonstrate the conundrum clearly at larger scale. They show that a larger neural network performs similarly to a parameter-matched ensemble of narrower networks in-distribution when all networks are trained on the same data. However, when linear probing the ensemble out of distribution on a related task, the ensemble performs substantially better. This result was shown not only for image classification, but also for self-supervised objectives and across architectures. \citet{yang2025these} expanded the conclusions to point out that, when matched for computation, the larger network still underperforms the narrower networks out of distribution.

Such a result tells us that any individual training run likely misses some information in the data, even though the model may be larger. Where it gains slightly in distribution, its features do not generalize. In a world where the deployment domain may be significantly different than the training distribution, we believe we should optimize for these generalizable features. In particular, \citet{zhang2023learning} suggest quite strongly that current dogma for scaling individual models~\citep{kaplan2020scaling} misses a critical mass of features underneath the proverbial ice.

In the following sections, we borrow a similar experimental protocol to show that incentivizing diverse feature learning results in substantial improvements over simple random ensembles in larger scale settings. In particular we compare ensembles of 10 smaller models to a single larger model that is 4 times as wide (hence roughly 16 times the parameters as well as more training compute). See Appendix~\ref{app:imgclass_ensembles} for more details.

\subsection{A Problem with Projected Gradient Descent}

Our first instinct is to replace the random ensembles in \citet{zhang2023learning} with ensembles of models acquired via repeated rounds of PGD, similar to the toy case. However, we encounter an issue in more complex systems: PGD enforces orthogonality, but this does not mean the information content needs to be distinct.

For example, if most of the entries of a vector are zero, with some large values, one can find a pathologically orthogonal vector with the same large values by pushing a small negative value into the rest of the positions. This likely would represent the same semantic content, but would satisfy PGD.

\begin{table}[htbp]
\centering
\caption{OOD linear probing accuracy between different methods. We observe the same effect as \citet{zhang2023learning}, that random ensembles (CAT) are much stronger than a large model that is parameter advantaged, but PGD ensembles do not appear to improve over random.}
\label{tab:img_cat_pgd}
\begin{tabular}{lcc}
\toprule
\textbf{Model} & \textbf{Accuracy} \\
\midrule
4x wide & $0.251 \pm 0.006$ \\
CAT & $0.298 \pm 0.004$ \\
PGD & $0.304 \pm 0.003$ \\
\bottomrule
\end{tabular}
\end{table}

\begin{figure}[t]
\centering
\begin{subfigure}[b]{0.48\columnwidth}
\centering
\includegraphics[width=\textwidth]{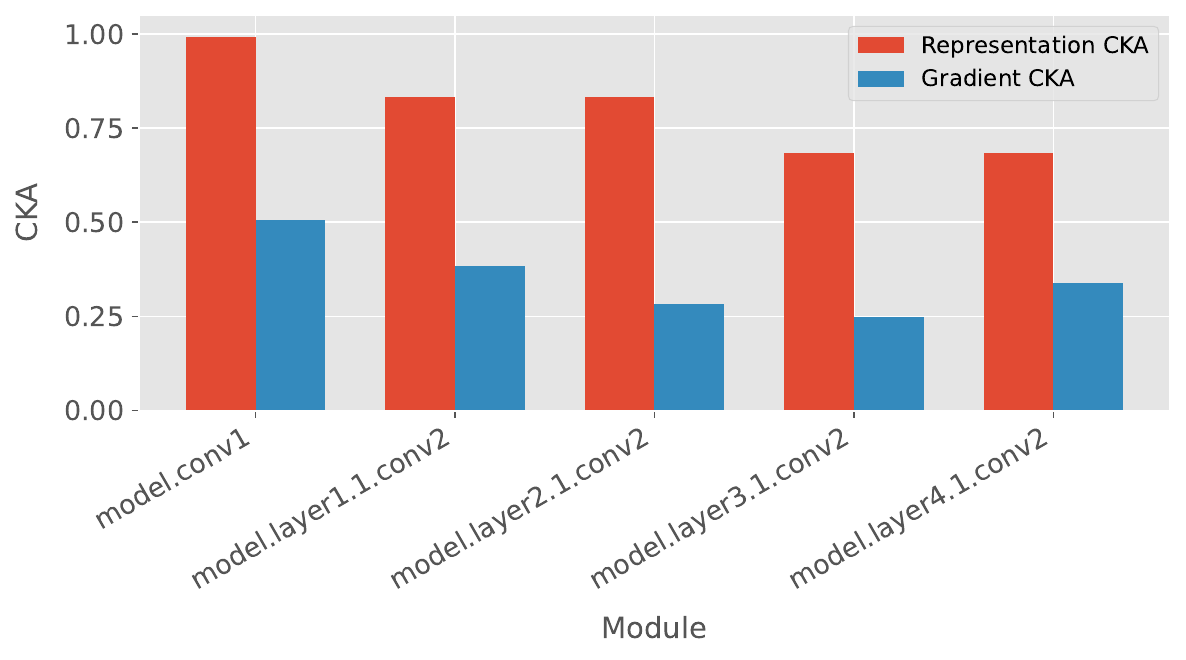}
\caption{CAT}
\end{subfigure}
\hfil
\begin{subfigure}[b]{0.48\columnwidth}
\centering
\includegraphics[width=\textwidth]{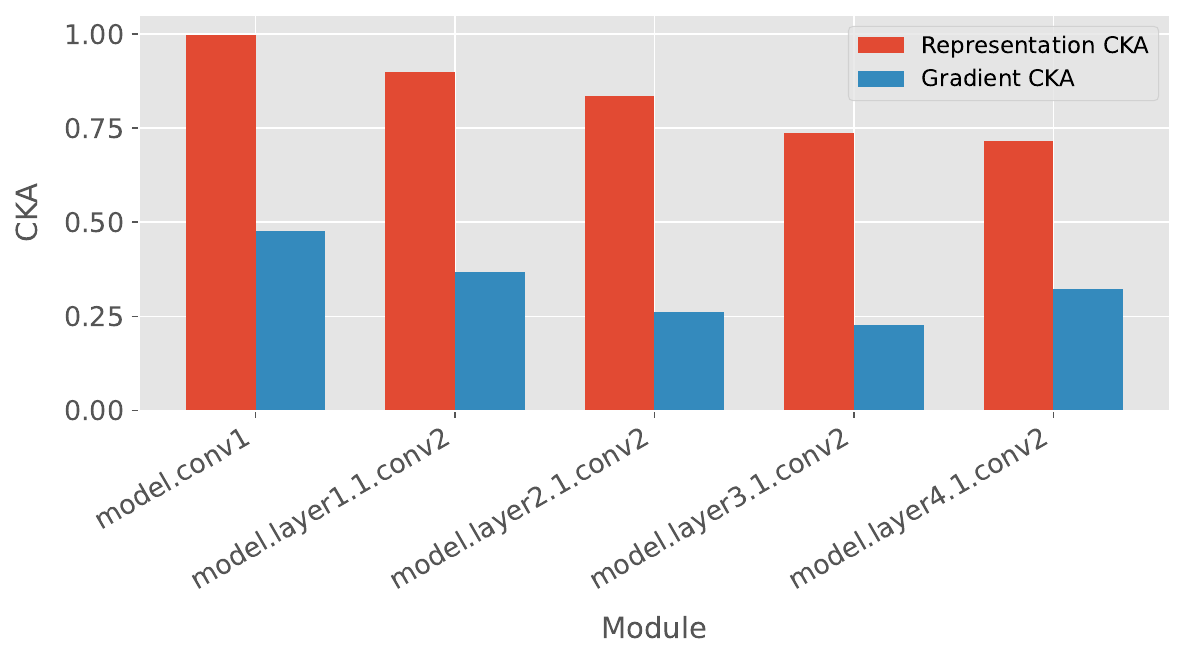}
\caption{PGD}
\end{subfigure}
\caption{Linear CKA~\citep{kornblith2019similarity} between layers for random ensembles (CAT) and PGD ensembles. The values are quite similar, suggesting that PGD is managing to find similar solutions as simple random reruns, possibly permutations of the original weights. Hence, PGD does not capture every notion of diversity.}
\label{fig:cka_random_pgd}
\end{figure}

In addition, one might be able to find permutations of a vector that are orthogonal, but the network could learn to remap those permutations in the next layer. We believe we encounter such an issue for the following reason: when comparing random ensembles (CAT for short) using different initializations with PGD ensembles that used the same initialization, we find the OOD performance between the random and PGD ensembles to be close (Table~\ref{tab:img_cat_pgd}). In addition we measure similarity between representations of different members of each ensemble using CKA~\citep{kornblith2019similarity}, and find the intra-ensemble representational similarity to be near the same magnitude for CAT and PGD (Figure~\ref{fig:cka_random_pgd}). As CKA is permutation invariant, this means that PGD ensembles are no more diverse than random ensembles, so it is very likely that PGD is rediscovering permutations of the same features.

However, we would like the functional behavior for different members of the ensemble to be different. Given the success of CKA~\citep{kornblith2019similarity} in gauging representational similarity, we propose to use this metric directly as an optimization objective. Minimizing CKA corresponds precisely to finding features that represent the data differently. CKA is also desirable as it avoids the permutation sensitivity of PGD. We describe our approach below.

\subsection{Inter-Example Similarity as a Measure of Functional Diversity}

Linear CKA~\citep{kornblith2019similarity} is a measure of the similarity between gram matrices. In order to repurpose this for maximizing diversity, at each timestep, we sample a random previously trained member of the ensemble, run a batch of examples through that network, then collect representations at middle layers. Next, we run the same batch through the model we are currently training, and collect representations at the same middle layers. Finally, we compute the average linear CKA between the two sets of features at each layer, and add that with a scalar coefficient to the training loss. As we finish training a model, we add that to the buffer of previously trained models.

To be precise, suppose we are given a batch of inner-layer representations from model $i$ in an ensemble at some layer $l \in \{0, 1, ..., L\}$, given by $X^i_l \in \mathbb{R}^{B \times d}$, where linear CKA is computed by

\begin{equation}
\text{CKA}(X, Y) = \frac{\lVert XY^\top \rVert_F^2}{\lVert XX^\top \rVert_F \lVert YY^
\top \rVert_F}
\end{equation}
for two centered matrices $X, Y \in \mathbb{R}^{B \times d}$.

For model 0, we train as usual. For model $i$, at every batch we sample $j$ uniformly randomly from $\{0, 1, ..., i-1\}$, then compute

\begin{equation}
\mathcal{L}_\text{CKA}(f_i) = \frac{1}{N} \sum_{l=1}^N \text{CKA}(X^i_l, X^j_l)
\end{equation}

for a selection of $N$ layers taken as a hyperparameter. The final loss we optimize for ensemble member $f_i$ becomes

\begin{equation}
\mathcal{L}(i, x, y) = \mathcal{L}_\text{task}(f_i, x, y) + \gamma \mathbb{I}[i > 0] \mathcal{L}_\text{CKA}(f_i, f_j).
\end{equation}

with an additional coefficient for optimization. In all our experiments we tune the choice of layer as well as the coefficient $\gamma$. In an ideal world we would not sample a previous model, instead running CKA against all prior models simultaneously. Unfortunately, this is prohibitive either from a disk space perspective, or from a memory perspective as it requires storing intermediate representations for all prior models. Hence we rely on sampling.

\subsubsection{Image Classification}


To test our hypothesis, we begin with ImageNet to CIFAR transfer, studied in \citet{zhang2023learning}. For computational reasons we train on ImageNet64x64~\citep{chrabaszcz2017downsampled}, and transfer via linear probing to CIFAR100~\citep{krizhevsky2009learning}. We use a ResNet18~\citep{he2016deep} as our base model, train a wider model with 4x the width multiple, and train ensembles that are width-matched (4 members) and larger (10 members) to approach parameter matching. The wider model has 181m parameters, while the original model has 11m parameters. We tune learning rates for all model sizes independently.

For CKA ensembles, we select a variety of middle layers that feed into the residual stream and compute CKA loss on average across these layers. We report the best setting over hyperparameters. More details are available in Appendix~\ref{app:imgclass_ensembles}.

\begin{table}[htbp]
\centering
\caption{Top-1 OOD accuracy for random ensembles (CAT) v.s. CKA ensembles. We see that
CKA improves by a substantial margin over CAT in the linear probing setup.}
\label{tab:imgclass_ensembles}
\begin{tabular}{lcc}
\toprule
\textbf{Model} & \textbf{Accuracy} \\
\midrule
CAT & $0.298 \pm 0.004$ \\
CKA & $0.357 \pm 0.003$ \\
\bottomrule
\end{tabular}
\end{table}

We see in Table~\ref{tab:imgclass_ensembles} that, like prior work~\citep{zhang2023learning, yang2025these}, random ensembles show a significant benefit in OOD linear probing. We also find that CKA improves further upon these random ensembles by a substantial margin. By incentivizing diversity, even completely agnostic to the problem, we can find substantial gains.

\subsubsection{Language Modeling}

One possible criticism of the above experimental protocol that extends to all prior work is the focus on a discriminative objective. So we turn to the task that is currently most important in the deployment of machine learning: language modeling. This task has been previously ignored by the literature on diverse feature learning, so it's unclear whether there is something special about discriminative objectives in the prior results.

We follow the experimental protocol with slight differences. We train a Transformer \citep{vaswani2017attention} language model on BabiStories~\citep{zhang2025memory}. We use a model 3x the width vs. an ensemble of 6 models with 1x width as they are similar compute. For in-distribution evaluation, we measure loss on the validation dataset, while for out of distribution, we measure loss on a related dataset, Simple English Wikipedia\citep{coster2011simple}. We choose this dataset as there is a high degree of lexical overlap, due to the simple language, but the topic distribution is quite different, thus any improvements must be due to learning the general structure of language. The benefit of language modeling is we don't require the linear probing protocol as it is possible to measure out-of-distribution without training any additional parameters. This also closely matches how language models are used in the wild with novel queries.

\begin{table}[htbp]
\centering
\caption{OOD cross entropy for random ensembles (CAT) v.s. CKA ensembles under many
different variations. Unlike the image classification case, we are unable to
find a version of CKA that outperforms random ensembles.}
\label{tab:language_ensembles}
\begin{tabular}{lcc}
\toprule
\textbf{Model} & \textbf{Cross Entropy} \\
\midrule
CAT & $ 7.68 \pm 0.02 $ \\
CKA & $ 7.69 \pm 0.03 $ \\
CKA max-in-depth & $ 7.71 \pm 0.02 $ \\
CKA token pool & $ 7.72 \pm 0.03 $ \\
CKA token cloud & $ 7.69 \pm 0.01 $ \\
CKA random proj. & $ 7.71 \pm 0.05 $ \\
\bottomrule
\end{tabular}
\end{table}

In Table~\ref{tab:language_ensembles}, we find similarly to image classification, that simple random ensembles perform better out of distribution than the larger model. This is striking given the in distribution performance is slightly worse. However, surprisingly, the prior CKA method does not perform well. In all cases we present the best results over hyperparameter tuning (see Appendix~\ref{app:language_ensembles}).

There are numerous reasons this might be the case: we may not enable CKA on the correct modules, hence we perform a sweep over different sections (a full sweep would be combinatoric and is difficult due to computational constraints). It might also be that, due to the residual structure of transformer language models, optimizing CKA would lead to learning the same features in different layers. To narrow this down, we try a version of our objective where we encourage diversity over CKA for only the maximum pairwise similarity over many layers in the model (CKA max-in-depth). However, this still is insufficient to drive improvements over random ensembles (see Table~\ref{tab:language_ensembles}).

Another possibility for the failure is that, as stated, our CKA works in high dimensional space, trying to make the sequence of representations different, which to be precise, is computing a $64\times 64$ matrix of inner products from $256 \times 768$ dimensional vectors. There are so few samples for the size of this vector that there may be a trivial solution for the language model to find by permuting information among the tokens, so instead we consider CKA over the pooled representations from a sequence (CKA token pool), or computing a $(64\cdot 256)\times (64\cdot 256)$ gram matrix (CKA token cloud) instead where we ask the token representations to be different. Neither of these are successful, as shown in Table~\ref{tab:language_ensembles}.

There might still be a problem with the dimensionality of our objective, so orthogonally we try random matrix projections that optimize CKA in a low dimensional space achieved by projecting with a different random matrix at every time. This means that the model cannot just fool the objective by picking a particular set of coordinates in which to push orthogonality, still we see in Table~\ref{tab:language_ensembles} that this is insufficient to beat random ensembles.

We are left with the question then: why does CKA fail in language modeling where it previously succeeded.

\subsubsection{A Problem with Optimizing CKA}

\begin{figure}[t]
\centering
\begin{subfigure}[b]{0.48\columnwidth}
\centering
\includegraphics[width=\textwidth]{./fig/diverse_features/grad_ckas/img_cat_val_cka.pdf}
\caption{CAT}
\end{subfigure}
\hfil
\begin{subfigure}[b]{0.48\columnwidth}
\centering
\includegraphics[width=\textwidth]{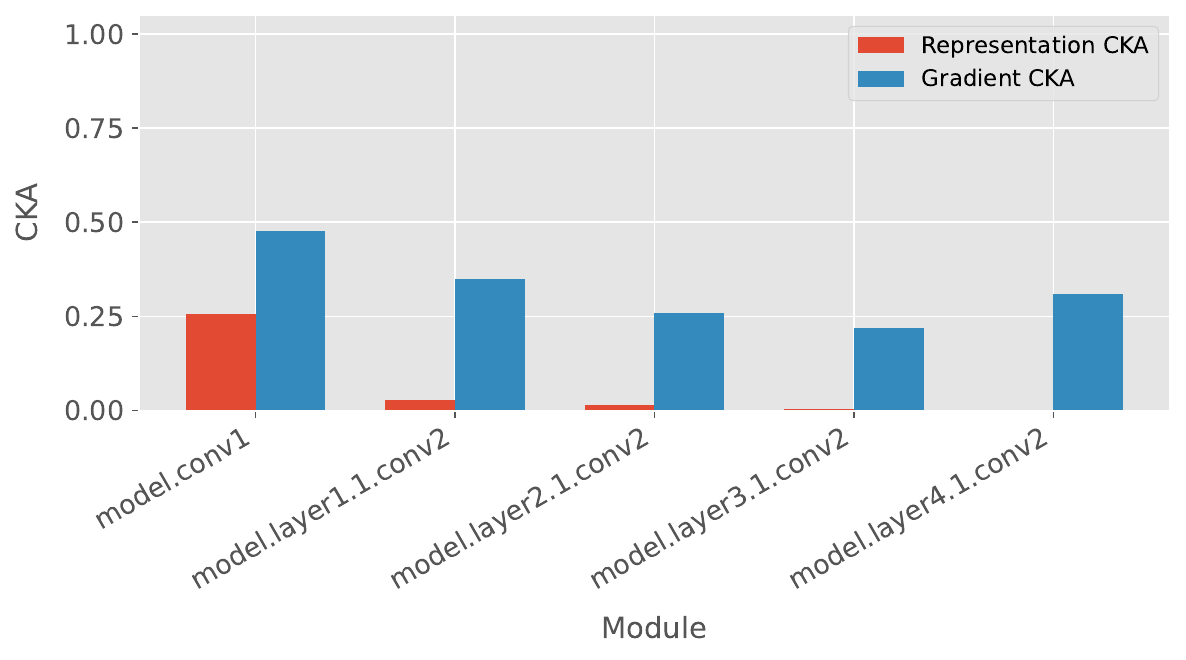}
\caption{CKA}
\end{subfigure}
\caption{CKA for representation vectors and gradient vectors for the task loss alone at different layers for image classification. \textbf{Left:} We see that representation and gradient CKA are reasonably high for CAT. \textbf{Right:} The same, but for the CKA-trained models. We see that using CKA greatly affects the representation CKA as it optimizes for this, and it also affects the task gradient CKA.}
\label{fig:imgclass_grad_cka}
\end{figure}

\begin{figure}[t]
\centering
\begin{subfigure}[b]{0.48\columnwidth}
\centering
\includegraphics[width=\textwidth]{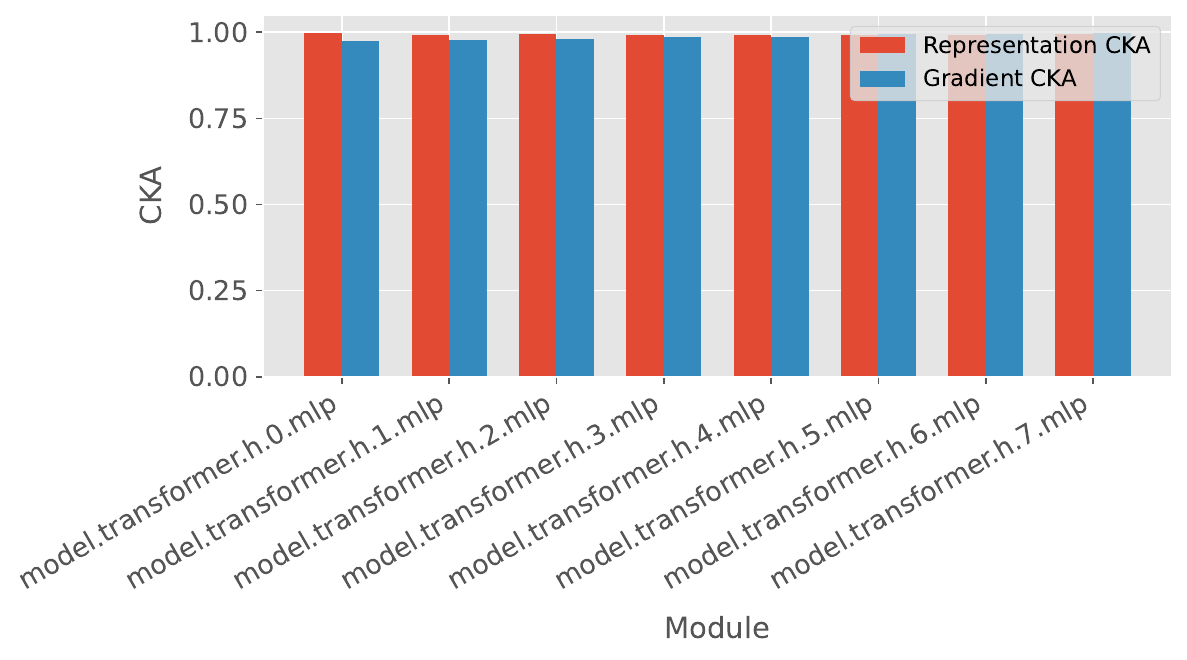}
\caption{CAT}
\end{subfigure}
\hfil
\begin{subfigure}[b]{0.48\columnwidth}
\centering
\includegraphics[width=\textwidth]{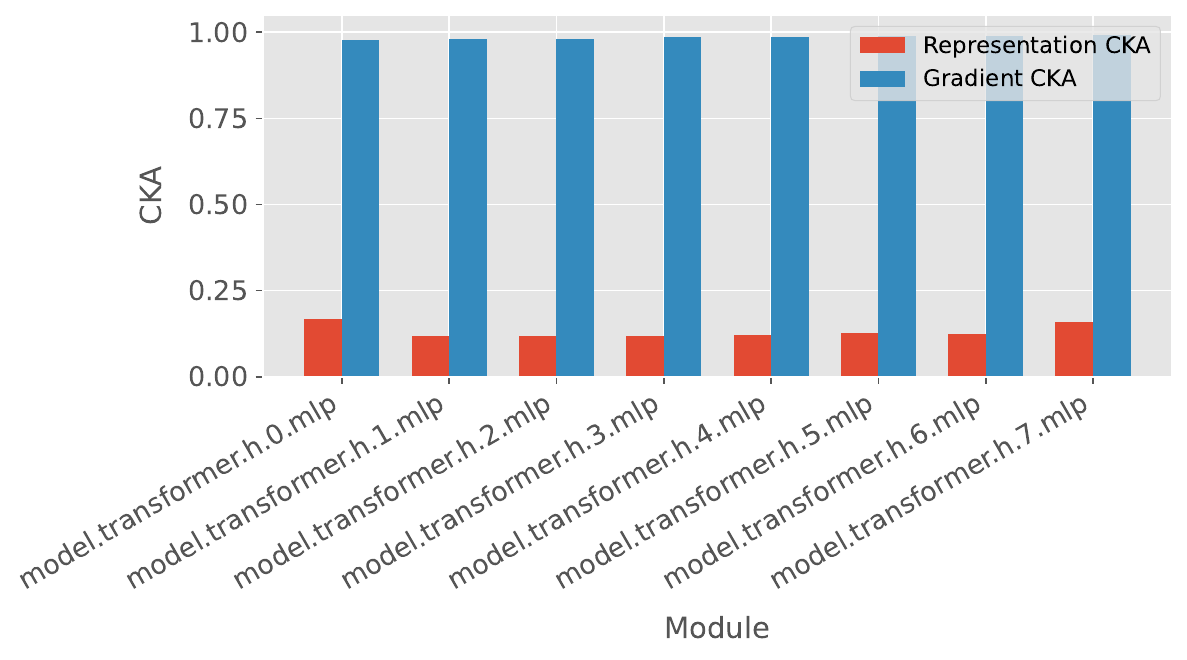}
\caption{CKA}
\end{subfigure}
\caption{CKA between the gradient vectors for the task loss alone at different layers for language modeling. \textbf{Left:} We see CKA between representations and between gradients, which are both high for the CAT models. \textbf{Right:} The same, but for the CKA-trained models. We see that while representation CKA is low, gradient CKA is unaffected by our CKA objective, unlike image classification.}
\label{fig:langmod_grad_cka}
\end{figure}

We believe the answer to this question has to do with the effective difficulty of the solution. In image classification there were relatively few classes, which means that, especially toward the output of the model~\cite{papyan2020prevalence}, it is possible to make the inter-example similarity low as there is a fair amount of room to move. In the case of language models, however, much more of the capacity is required for the task (it is impossible to achieve 0 training error), thus it is not easy to achieve the desired semantic dissimilarity and instead the network finds a pathological solution. The reason we know the solution is pathological is that we find that the CKA is high between the gradient vectors of the next token prediction task (Figures~\ref{fig:imgclass_grad_cka} and~\ref{fig:langmod_grad_cka}).  This means that the model has learned to fool the CKA objective, while not actually altering its downstream predictions. Where representations have low CKA between models, the actual output behavior is very similar.

The core issue with the CKA objective, and the previous PGD objective, is that it relies on the model's implementation of a function, and conflates the structure in high dimensional space with the model's actual function, but it's impossible to measure the actual functional change through CKA. In order to get an absolute sense on the functional difference, we have to look at the output space of the model instead.


\section{Diverse Feature Learning in Output Space}


We have made a roundabout journey to land on a much simpler definition of diversity, employed many times in prior literature. We want the outputs of
different members of the ensemble to look different, because we know that if the members make different predictions, they must by definition be different
functions.

\subsection{Negative Correlation Learning}

One classic way to enforce diversity in this way is negative correlation learning (NCL)~\citep{liu1999ensemble}. This was first defined for regression tasks. At its core
NCL aims to learn an ensemble whose ensembled predictions are good at the task, while each member is distinct. See \citet{buschjager2020generalized} for a recent treatment with much more context. NCL corresponds to a general structure:

\begin{equation}
\mathcal{L}_\text{NCL}(f_0, ..., f_m, x, y) = \mathcal{L}_\text{task}(f_\text{ens}, x, y) - \gamma \mathcal{L}_\text{div} (f_0, ..., f_m, x, y)
\end{equation}

where $f_\text{ens}$ is the ensembled model from the individuals $f_0, ..., f_m$ by some weighting that need not be uniform. The task loss is already defined by
whatever we wish to do (language modeling in this case), so it remains to discuss the diversity loss which we want to maximize, hence the minus sign. Typically it is phrased as:

\begin{equation}
\mathcal{L}_\text{div}(f_0, ..., f_m, x, y) = \frac{1}{m} \sum_i l_\text{diff}(f_\text{ens}, f_i, x, y),
\end{equation}

which signifies that we want to maximize the disagreement between the ensemble and any individual member. As we are working with language models, we experiment with both Jensen-Shannon Divergence (JSD) and simple mean squared error (MSE) for $l_\text{diff}$.

Usually NCL is for a set of models to be optimized jointly online. Prior work has shown that, especially in high dimensional deep learning, joint optimization can easily collapse to poor solutions~\citep{jeffares2023joint}, hence we train models sequentially as in the case of our previous PGD and CKA experiments.

In addition, we face a similar memory and disk constraint with computing $\mathcal{L}_\text{div}$, so we make the same sampling approximation as before. When training model $i$:

\begin{equation}
\hat{\mathcal{L}_\text{div}}(f_0, ..., f_i, x, y) = l_\text{diff}(\frac{1}{2}(f_i + f_j), f_i, x, y)
\end{equation}

where the index $j$ is sampled uniformly randomly from $\{0, ..., i-1\}$.

\begin{table}[htbp]
\centering
\caption{OOD cross entropy for random ensembles v.s. NCL ensembles
We see no significant difference to random ensembles. NCL JSD corresponds to
using Jensen-Shannon divergence for difference. MSE corresponds to MSE for the
probability distribution, and Top-1 corresponds to only considering the
difference between probabilities of the correct class.}
\label{tab:ncl}
\begin{tabular}{lcc}
\toprule
\textbf{Model} & \textbf{Cross Entropy} \\
\midrule
Random & $ 7.68 \pm 0.02 $ \\
NCL JSD & $ 7.69 \pm 0.02 $ \\
NCL MSE & $ 7.69 \pm 0.01 $ \\
NCL correct only & $ 7.66 \pm 0.02 $ \\
\bottomrule
\end{tabular}
\end{table}

We see in Table~\ref{tab:ncl} that our straightforward attempts at NCL are insufficient (see Appendix~\ref{app:language_ensembles} for tuning). In particular we hypothesize that one issue with NCL is similar to prior experiences in CKA and PGD, whereby the high dimensional probability distribution can find pathological solutions for diversity, so we restrict to only the top-1 of the output probability distribution (NCL correct only). Still we see no major improvement for the best case of each.

\subsection{Boosting Incorrect Predictions}

Another large issue with negative correlation learning is that it incentivizes models to disagree even when the label is correct. One might instead consider an even simpler boosting-like~\citep{freund1999short} method whereby we train models sequentially, and later members need to compensate for the error of previous models. This is similar in spirit to the Bonsai Method suggested by \citet{zhang2022rich} on simpler image classification tasks. However, in their work, the training set was continuously pruned such that later members of the ensemble would have shrinking datasets, which leads exactly to the kind of spurious correlation issue we are trying to avoid. Rather, we believe that there are other, more complex ways to represent \textit{the same} examples, hence we wish to preserve the loss on the entire distribution, but only slightly upweight for regions not covered by previous training runs.

We implement this by training successive models such that later models have their loss upweighted by the error of the ensemble, sometimes called focal loss~\citep{lin2017focal}. In particular, at every optimization step we sample a previously trained model, then add a weighting term to the cross entropy loss corresponding to the incorrectness of the previous model, thus errors are boosted in the loss. However, unlike \citet{zhang2022rich}, we preserve the original loss by the use of a small coefficient on the focal term. Suppose the output of model $i$ for token t is probability vector $p^i \in \mathbb{R}^{C}$ for a vocabulary of $C$ classes, then the loss for model $i$ at token $t$ is.

\begin{equation}\label{loss_weighting}
\mathcal{L}(f_i, f_j, x, y) = \hat{w(f_j, x, y)} * \mathcal{L}_\text{task} (f_i, x, y)
\end{equation}

with weight

\begin{equation}
w(f_j, x, y) = (1 + \gamma * (1 - p^j_\text{correct}))
\end{equation}

where $j$ is sampled randomly and uniformly from $\{0, ..., i-1\}$, $\gamma$ is a hyperparameter, and $\hat{w(f_j, x, y)}$ corresponds to the normalized weight in a sequence such that the weights sum to 1. We normalize so that the learning rate does not need to be tuned jointly. Thus, we shape the loss for subsequent rounds by the mistakes of previous rounds, while always giving some weight to the original task loss to avoid the overfitting issues of a shrinking sample. We normalize the weight so that we do not need to retune learning rates due to different gradient magnitudes.

\begin{table}[htbp]
\centering
\caption{OOD cross entropy for random ensembles v.s. Focal scaling ensembles
We see no significant difference to random ensembles. Top-p corresponds to
scaling the top-p of the tokens by sequence by a constant additive factor.}
\label{tab:focal}
\begin{tabular}{lcc}
\toprule
\textbf{Model} & \textbf{Cross Entropy} \\
\midrule
Random & $ 7.68 \pm 0.02 $ \\
Focal & $ 7.66 \pm 0.01 $  \\
Focal top-p & $ 7.65 \pm 0.02 $  \\
\bottomrule
\end{tabular}
\end{table}

We also consider a version where the loss at a token is simply scaled by a flat factor $\gamma$ as long as $1 - p^j_\text{correct}$ is in the top P portion of incorrect tokens in a sequence. This is so that we do not upweight errors that are large but unlearnable. We find in Table~\ref{tab:focal} that both of these variants are still insufficient to outperform simple random ensembles even when searching over $\gamma$ and the proportion of top p. Find details on tuning in Appendix~\ref{app:language_ensembles}.

\subsection{Boosting Low-Entropy Predictions}

We follow a parallel path, exploring entropy-based uncertainty instead of incorrectness, which has seen success in other areas~\citep{wang2026beyond}. To be precise, following the general formulation of equation~\ref{loss_weighting} and the normalization of the last section, we design a weight based on entropy as

\begin{equation}
w(f_i, x, y) = (1 + \gamma * H(p^j)).
\end{equation}

We also experiment with a threshold version as in the previous section, where we scale the loss by $\gamma$ when the entropy of the token distribution under a previous model is within the top-p entropies for the sequence.

\begin{table}[htbp]
\centering
\caption{OOD cross entropy for random ensembles v.s. entropy scaling ensembles
We see no significant difference to random ensembles. Top-p corresponds to
scaling the top-p of the tokens by sequence by a constant additive factor.}
\label{tab:entropy}
\begin{tabular}{lcc}
\toprule
\textbf{Model} & \textbf{Cross Entropy} \\
\midrule
Random & $ 7.68 \pm 0.02 $ \\
Entropy & $ 7.65 \pm 0.03 $  \\
Entropy top-p & $ 7.65 \pm 0.02 $  \\
\bottomrule
\end{tabular}
\end{table}

We find in Table~\ref{tab:entropy} similarly to focal loss that we are unable to improve upon random ensembles.  Similar arguments may apply as in the case of focal-loss: some
unconfident predictions are very difficult to learn, and forcing the model to learn them may sacrifice other areas of the space.

\subsection{Ensemble Weighting}

As mentioned in the CKA section, we train multiple language models and ensemble their predictions to test the ensemble out of distribution. Throughout the prior experiments we have used uniform ensemble weighting as a straightforward baseline, taking the simple average of the probability distributions in order to compute downstream metrics, but especially in the case of entropy and focal loss, it's very possible that the models are differently calibrated, so even weighting over the distributions would not yield an ideal outcome.

As a result, we test a variety of heuristic per-token ensembling methods. Given a relevance function for a function and token $r(f_i, x_t)$, we compute a softmax distribution over all members of the ensemble $f_i$ with temperature $T$. To be precise, we compute the weight of $f_i$ in ensemble as

\begin{equation}
w(f_i, x_t) = \text{softmax}\left ( \frac{r(f_i, x_t)}{T} \right )
\end{equation}

As relevance metrics we choose entropy and KL distance from the ensemble distribution. To be precise, $r(f_i, x_t) = H(f_i(x_t))$ or $r(f_i, x_t) = KL(f_i(x_t), f_\text{ens}(x_t))$.

\begin{table}[htbp]
\centering
\caption{OOD cross entropy for different ensemble weighting methods on the 
Entropy top-p runs. We use these as the base as we expect there may be
nonuniformity compared to the Random ensembles.}
\label{tab:ensemble_weighting}
\begin{tabular}{lcc}
\toprule
\textbf{Model} & \textbf{Cross Entropy} \\
\midrule
Even & $ 7.68 \pm 0.02 $ \\
Entropy & $ 7.69 \pm 0.03 $  \\
KL disagreement & $ 7.67 \pm 0.02 $ \\
\bottomrule
\end{tabular}
\end{table}

As seen in Table~\ref{tab:ensemble_weighting}, we are unable to make gains over random ensembling in all of these cases. We also combine linearly with uniform weighting in the hopes of smoothing out spiky predictions due to overfitting, but find none of these methods superior to even weighting. See Appendix~\ref{app:ensemble_weighting} for more details on tuning.

%
%
%

We are unsure as to this perplexing observation that it is difficult to beat random ensembling. We believe that the root issue here is that we are again unable to tell whether low entropy or large disagreement from the ensemble is due to a correct but different prediction, or an incorrect but different prediction. The small degree to which the exact OOD evaluation numbers seem to drift suggests both effects are evenly at play. Still, this suggests that there may be room to grow along this exact axis, possibly experimenting with holdout sets for global weighting.

It is also possible that these results suggest that high-entropy or incorrect predictions primarily define the same task as the entire dataset, as cross entropy is magnified on the most difficult examples. In some sense, the difficult examples form the support-vectors of the training set, so entropy weighting recovers similar solutions. Prior work in dataset distillation indicates this may be possible~\citep{wang2018dataset}. Then the difficulty is to find structured subsets that yield different solutions than the entire dataset, without heuristics and while avoiding possible newly-introduced spurious correlations.




\section{Related Work}


The themes of the prior chapter: simplicity bias, diversity in features and ensembling, are already well-represented in the literature. We present a non-exhaustive description of prior work, and articulate our contribution below.

\subsection{Leveraging Downstream Information}

One class of solutions to find robust neural networks is to leverage information about the downstream domain. For example, one might know broad statistics about the task distribution, or have some tiny sample to quickly adapt. For example, \citet{wen2025elastic} design a low rank penalty on the last layer for OOD prediction which requires group labels~\citep{hu2018does, sagawa2019distributionally}.  \citet{pagliardini2022agree} propose a training method to enforce diverse representations on OOD data, which requires a sample.
\citet{lee2022diversify, chen2023project} both require test data.  \citet{kirichenko2022last} make a similar point that retraining only the final layer of a pretrained classifier on test data is very strong compared to bespoke methods. We consider access to test data to be an undesirable assumption, as we would like to extract all the information available in a dataset, agnostic to the downstream distribution.

\subsection{Relying on Pretrained Models}

A large class of methods relies on pretrained models as feature learners, evading the question of dynamics entirely. \citet{gontijo2021no} show that ensembling features from various diverse pretrained models outperforms narrow ensembles, where the diversity is the key. \citet{teney2022evading} train a set of linear heads simultaneously on top of a fixed extractor, enforcing orthogonality among the heads, but such a method fails without the pretrained model (as we saw). \citet{tiwari2023overcoming, joshi2025mitigating} make sure that the final classification depends on features throughout a pretrained model. \citet{niu2022roadblocks} use an autoencoder based on a pretrained model's features. \citet{izmailov2022feature} make the point that retraining the last layer of a pretrained model can achieve SOTA compared to more complex group robustness methods, where the architecture, hyperparameters and data of the pretrained feature extractor are the most important levers. We agree strongly with this conclusion, which is why we are interested in altering the learning dynamics of networks from scratch, instead of relying on an existing model.

\subsection{Architectural Modifications}

Another sensible approach would be to design architectures around diversity. One of the clearest examples would be mixture of experts (MoE) layers~\citep{shazeer2017outrageously, zhou2022mixture}. These were initially designed to make it possible to train much wider networks on less hardware through sparse routing, but can be interpreted as an ensemble inside a model. Still \citet{dai2024deepseekmoe} show in Section 4.3 Table 2 that a dense model as big as an MoE is slightly better, so even if experts specialize to different data, we should not expect that the features learned are more diverse. MoE alone is not enough.  This is further supported by the fact that each expert sees less data than the
whole, so they have a larger incentive to pick up spurious correlations. On a different and much smaller-scale note, \citet{addepalli2023feature} design a two-pronged architecture to attack the MNIST-CIFAR task, but this is too task-specific to scale. \citet{to2025diverse} force a classifier to specialize to different subsets of the original training data with constraints on the decoding. Still we believe, given the generality of the results in this document, that neural networks are subject to a common paradigm. Though architectural improvements may be helpful, they do not attack the issue underlying spurious correlations.

\subsection{Objective and Optimization Approaches}

As one might modify the architecture, researchers have also taken the approach of introducing new objectives. \citet{sinha2021dibs} introduce a method with multiple decoder heads that are optimized with a GAN-like contrastive loss to produce diverse representations. We believe this has similar high dimensional issues as we encountered with CKA on more complex tasks.  \citet{morwani2023simplicity} prove that 1 hidden-layer models depend on a linear subspace, then design an optimization method OrthoP which projects the input off of the low-rank solution of the previous run, like our PGD method.  However, as it is naive projection without the SVD basis, it is quite easy to fool in high dimensions. \citet{saha2021gradient} design a PGD method quite like what we converged on for applications in continual learning, though only in input space. We are more interested in these approaches than previous buckets, but in many cases they encounter the same pitfalls that we did in prior sections.

\subsection{Ensembling}

Lastly there are many generic results for the benefit of ensembling, hence our focus in this area. \citet{zhang2023learning} make the clear observation that an ensemble of small networks beats a single larger network OOD by a substantial margin, even though their in distribution performances are similar. Thus we are missing features in optimization. \citet{yang2025these} expands this result to the step-matched setting, finding the same conclusions. \citet{abe2022deep} argue that OOD performance in deep ensembles is primarily driven by ID performance. Such an argument is supported by \citet{mandal2026q0}, where better optimization for individual members yields stronger results than naive scaling of the ensemble~\citep{kondratyuk2020ensembling, lobacheva2020power, kim2026pre}. One very clear way to encourage diversity is to upsample the mistakes of previous models~\citep{liu2021just, zhang2022rich}, but in cases like MNIST-CIFAR where there are no mistakes, or where the mistake set grows ever smaller, this is likely to end in further spurious correlations. We believe the problem of generically finding a way to incentivize diversity in individual members is still quite open. The frustration of the prior chapter shows this is not a trivial task.


\section{Discussion}
Diverse feature learning means getting more out of less data. What does a world look like in which we don't need the whole internet to train language models?  We believe the key to this problem is overcoming the greedy process whereby neural networks learn the simplest features first and never revisit those choices. Thus, we attack spurious correlations head on.

We begin with a simple testbed of spurious correlations, the MNIST-CIFAR task~\citep{shah2020pitfalls}. We show that we can use our previously-developed spectral dynamics (Chapter~\ref{chp:spectral_dynamics}) as a definition for simplicity bias, which leads to a method using projected gradient descent on singular vectors. This method suffices to overcome simplicity bias without the task-specific assumptions of prior work.

We then scale our method to more complex architectures and tasks, where we find that the task difficulty renders such a simple method ineffective. After finding that our method hides model-wise similarity, we turn to using that similarity metric directly as a diversity loss. We find that in larger image classification networks we are able to beat simple random ensembles, a surprisingly strong baseline that is already stronger than large models~\citep{zhang2023learning}.

Next, we generalize this experimental protocol to language models, where we demonstrate that ensembles of small models perform better than larger models out-of-distribution, extending the observations of \citet{zhang2023learning, yang2025these}. However, our previous diversity losses do not appear to help for language models, which we elucidate through a large number of ablations.

We believe the missing piece is a way to isolate structured subsets of the data so as to find diverse solutions. Merely restricting to difficult examples either finds the same solutions as random ensembles, or encounters spurious correlations. The core difficulty is finding such structured subsets without heuristics engineered to fit the problem at hand. 

Though they are still incomplete, these results suggest that there may be a new path forward for scaling computation. The basic observation that ensembles of smaller models perform better than larger ones OOD can be extended to any domain where neural networks are used, including robotics where domain shifts are common and expected. Across domains, there are many features hiding beneath the ice, waiting to be discovered.

\chapter{Conclusion}\label{chp:conclusion}
It is very clear in this moment that neural networks have changed and will continue to change technology and our world. Yet we have a very poor understanding of how they work. The more embedded neural networks become, the more pressing the gap. Moreso as neural networks take root in economic, social and political questions, far outside of our domain as computer scientists.

This thesis strikes at understanding neural networks. We begin with a strange phenomenon, and carry this thread all the way through to improving neural networks in the low-data regime: the critical problem that scaling has been unable to overcome by definition. Briefly, we recap the contributions of this thesis.

Chapter~\ref{chp:convex_mode_connectivity}, on Convex Mode Connectivity, shows that neural network optima occupy a high dimensional simplex in the loss landscape, that this is
true across a wide variety of neural networks, and that trivial mathematical explanations for this phenomenon do not apply. At the time of presentation, such complexity and generality was unexpected. A body of theoretical literature had presented simple explanations, and prior model classes do not have such nontrivial properties. Moreover, the generality and high dimensionality was genuinely perplexing as if there were so many solutions, how come our optimization process did not find them?

Chapter~\ref{chp:spectral_dynamics}, on Spectral Dynamics, demonstrates across data, architectures, optimizers, and hyperparameters, there is a similar bias in optimization toward lower rank matrices. Such a bias might be expected in simple theoretical settings, but again the assumptions of prior work did not hold in these more complex cases, which muddied the water as to the ultimate cause. This work was a sign that something common explains the behavior of all neural networks. Moreover these spectral dynamics allow us to explain the previously stated conundrum of Convex Mode Connectivity, tying together our understanding about neural networks.

Chapter~\ref{chp:diverse_features}, on Diverse Feature Learning, used the lens of spectral dynamics to attack the problem of learning on less data, a problem for which the field currently has very few tools. In particular, the bias toward low rank matrices provided a clue as to the actual mechanics behind spurious correlations and allowed us to find a task-agnostic method for improving learning in a toy case, which we then scaled to more complex cases. The full application of this understanding still eludes us, but we are firm in our belief that this is the next frontier of neural networks.

Lastly, a note to my readers without access to immense resources. Many well-funded academics and companies would have you believe that the key to improving neural networks is simply scaling~\citep{hestness2017deep, kaplan2020scaling}. All one needs is a sufficient recipe. Then collecting more data with more parameters will provide sufficient ammunition for solving the problem. This is a powerful scaling hypothesis, attractive in its simplicity, and indeed it seems to hold when looking with a broad view across the field. From AlexNet~\citep{krizhevsky2012imagenet}, to later models like VGG \citep{simonyan2014very}, scale has almost always improved performance. AlphaGo~\citep{silver2017mastering} attacked the previously unimaginable game of Go, and shortly after AlphaZero~\citep{silver2018general} showed scale was the primary ingredient. Through the generations from GPT~\citep{radford2018improving} to GPT2~\citep{radford2019language} to GPT3~\citep{brown2020language} to GPT4~\citep{achiam2023gpt} it was quite clear the improvement in the model quality and output. Scale was and is a powerful lever.

However, it is a very blunt lever. VGG~\citep{simonyan2014very} was much larger than AlexNet~\citep{krizhevsky2012imagenet}, but ResNet~\citep{he2016deep} was a fraction of the size and more performant. MLPs can approximate ConvNets in theory, but we are unaware of empirical examples where this occurred without guidance. We do not talk to our linear regressions, rather advances in architectures have led to our current language models. Optimizers failed past a certain scale due to loss spikes~\citep{wortsman2023stable}, which led to new optimizers that required a deeper understanding~\citep{jordan2024muon}. I do believe in the bitter lesson~\citep{sutton2019bitter}, but only when the initial conditions are set correctly. We may reach the edge of existing pretraining data scaling in language models~\citep{villalobos2024position}. That recent advances in the field are due to downstream techniques like RL is evidence that scaling has to be thought through. RL is yielding impressive results in specific domains~\citep{guo2025deepseek}, but not in full generality~\citep{ouyang2022training, chen2026does}. Even now there are many examples of more compute-efficient improvements than pure scale~\citep{saunshi2025reasoning}, including our own results in Chapter~\ref{chp:diverse_features}. And even if it were not so blunt, scale is an extremely expensive lever to pull given its diminishing returns.

This thesis charts a different path. I was and am curious if we can do more with less. It is my hope that this work is helpful to myself and others toward this goal. And it was my pleasure to skate on the edge of uncertainty in pursuit of this understanding.

\clearpage
\bibliography{./main.bib}

@string{iclr ={Proceedings of the International Conference on Learning Representations (ICLR)}}

@string{cvpr="Proceedings of the IEEE/CVF Conference on Computer Vision and Pattern Recognition (CVPR)"}

@string{iccv="Proceedings of the International Conference on Computer Vision (ICCV)"}

@string{eccv="Proceedings of the European Conference on Computer Vision (ECCV)"}

@string{aaai="Proceedings of the National Conference on Artificial Intelligence (AAAI)"}

@string{icml="Proceedings of the International Conference on Machine Learning (ICML)"}

@string{jmlr="Journal of Machine Learning Research"}

@string{neurips="Advances in Neural Information Processing Systems (NeurIPS)"}

@string{icassp="Proceedings of the IEEE International Conference on Acoustics, Speech and Signal Processing (ICASSP)"}

@inproceedings{fort2019large,
  title={Large scale structure of neural network loss landscapes},
  author={Fort, Stanislav and Jastrzebski, Stanislaw},
  booktitle=neurips,
  year={2019}
}

@article{fort2019deep,
  title={Deep ensembles: {A} loss landscape perspective},
  author={Fort, Stanislav and Hu, Huiyi and Lakshminarayanan, Balaji},
  journal={arXiv preprint arXiv:1912.02757},
  year={2019}
}

@inproceedings{frankle2020linear,
  title={Linear mode connectivity and the lottery ticket hypothesis},
  author={Frankle, Jonathan and Dziugaite, Gintare Karolina and Roy, Daniel and Carbin, Michael},
  booktitle=icml,
  pages={3259--3269},
  year={2020},
  OPTorganization={PMLR}
}

@inproceedings{draxler2018essentially,
  title={Essentially no barriers in neural network energy landscape},
  author={Draxler, Felix and Veschgini, Kambis and Salmhofer, Manfred and Hamprecht, Fred},
  booktitle=icml,
  pages={1309--1318},
  year={2018},
  OPTorganization={PMLR}
}

@article{entezari2021role,
  title={The role of permutation invariance in linear mode connectivity of neural networks},
  author={Entezari, Rahim and Sedghi, Hanie and Saukh, Olga and Neyshabur, Behnam},
  journal={arXiv preprint arXiv:2110.06296},
  year={2021}
}

@inproceedings{neyshabur2020being,
  title={What is being transferred in transfer learning?},
  author={Neyshabur, Behnam and Sedghi, Hanie and Zhang, Chiyuan},
  booktitle=neurips,
  OPTvolume={33},
  pages={512--523},
  year={2020}
}

@inproceedings{kuditipudi2019explaining,
  title={Explaining landscape connectivity of low-cost solutions for multilayer nets},
  author={Kuditipudi, Rohith and Wang, Xiang and Lee, Holden and Zhang, Yi and Li, Zhiyuan and Hu, Wei and Ge, Rong and Arora, Sanjeev},
  booktitle=neurips,
  OPTvolume={32},
  year={2019}
}

@article{brea2019weight,
  title={Weight-space symmetry in deep networks gives rise to permutation saddles, connected by equal-loss valleys across the loss landscape},
  author={Brea, Johanni and Simsek, Berfin and Illing, Bernd and Gerstner, Wulfram},
  journal={arXiv preprint arXiv:1907.02911},
  year={2019}
}

@article{keskar2016large,
  title={On large-batch training for deep learning: {G}eneralization gap and sharp minima},
  author={Keskar, Nitish Shirish and Mudigere, Dheevatsa and Nocedal, Jorge and Smelyanskiy, Mikhail and Tang, Ping Tak Peter},
  journal={arXiv preprint arXiv:1609.04836},
  year={2016}
}

@inproceedings{dinh2017sharp,
  title={Sharp minima can generalize for deep nets},
  author={Dinh, Laurent and Pascanu, Razvan and Bengio, Samy and Bengio, Yoshua},
  booktitle=icml,
  pages={1019--1028},
  year={2017},
  OPTorganization={PMLR}
}

@inproceedings{li2018visualizing,
  title={Visualizing the loss landscape of neural nets},
  author={Li, Hao and Xu, Zheng and Taylor, Gavin and Studer, Christoph and Goldstein, Tom},
  booktitle=neurips,
  OPTvolume={31},
  year={2018}
}

@article{gur2018gradient,
  title={Gradient descent happens in a tiny subspace},
  author={Gur-Ari, Guy and Roberts, Daniel A and Dyer, Ethan},
  journal={arXiv preprint arXiv:1812.04754},
  year={2018}
}

@inproceedings{fort2020deep,
  title={Deep learning versus kernel learning: an empirical study of loss landscape geometry and the time evolution of the neural tangent kernel},
  author={Fort, Stanislav and Dziugaite, Gintare Karolina and Paul, Mansheej and Kharaghani, Sepideh and Roy, Daniel M and Ganguli, Surya},
  booktitle=neurips,
  OPTvolume={33},
  pages={5850--5861},
  year={2020}
}

@article{jiang2019fantastic,
  title={Fantastic generalization measures and where to find them},
  author={Jiang, Yiding and Neyshabur, Behnam and Mobahi, Hossein and Krishnan, Dilip and Bengio, Samy},
  journal={arXiv preprint arXiv:1912.02178},
  year={2019}
}

@article{zhang2021understanding,
  title={Understanding deep learning (still) requires rethinking generalization},
  author={Zhang, Chiyuan and Bengio, Samy and Hardt, Moritz and Recht, Benjamin and Vinyals, Oriol},
  journal={Communications of the ACM},
  volume={64},
  number={3},
  pages={107--115},
  year={2021},
  OPTpublisher={ACM New York, NY, USA}
}

@inproceedings{he2016deep,
  title={Deep residual learning for image recognition},
  author={He, Kaiming and Zhang, Xiangyu and Ren, Shaoqing and Sun, Jian},
  booktitle=cvpr,
  pages={770--778},
  year={2016}
}

@inproceedings{sitzmann2020implicit,
  title={Implicit neural representations with periodic activation functions},
  author={Sitzmann, Vincent and Martel, Julien and Bergman, Alexander and Lindell, David and Wetzstein, Gordon},
  booktitle=neurips,
  OPTvolume={33},
  pages={7462--7473},
  year={2020}
}

@mastersthesis{krizhevsky2009learning,
  title={Learning multiple layers of features from tiny images},
  author={Krizhevsky, Alex},
  school={University of Toronto},
  year={2009},
}

@inproceedings{panayotov2015librispeech,
  title={Librispeech: {A}n {ASR} corpus based on public domain audio books},
  author={Panayotov, Vassil and Chen, Guoguo and Povey, Daniel and Khudanpur, Sanjeev},
  booktitle=icassp,
  pages={5206--5210},
  year={2015},
  OPTorganization={IEEE}
}

@inproceedings{vlaar2022can,
  title={What can linear interpolation of neural network loss landscapes tell us?},
  author={Vlaar, Tiffany J and Frankle, Jonathan},
  booktitle=icml,
  pages={22325--22341},
  year={2022},
  OPTorganization={PMLR}
}

@inproceedings{nagarajan2019uniform,
  title={Uniform convergence may be unable to explain generalization in deep learning},
  author={Nagarajan, Vaishnavh and Kolter, J Zico},
  booktitle=neurips,
  OPTvolume={32},
  year={2019}
}

@inproceedings{wortsman2022model,
  title={Model soups: averaging weights of multiple fine-tuned models improves accuracy without increasing inference time},
  author={Wortsman, Mitchell and Ilharco, Gabriel and Gadre, Samir Ya and Roelofs, Rebecca and Gontijo-Lopes, Raphael and Morcos, Ari S and Namkoong, Hongseok and Farhadi, Ali and Carmon, Yair and Kornblith, Simon and others},
  booktitle=icml,
  pages={23965--23998},
  year={2022},
  OPTorganization={PMLR}
}

@article{li2022branch,
  title={{Branch-Train-Merge}: {E}mbarrassingly Parallel Training of Expert Language Models},
  author={Li, Margaret and Gururangan, Suchin and Dettmers, Tim and Lewis, Mike and Althoff, Tim and Smith, Noah A and Zettlemoyer, Luke},
  journal={arXiv preprint arXiv:2208.03306},
  year={2022}
}

@article{ainsworth2022git,
  OPTdoi = {10.48550/ARXIV.2209.04836},
  OPTurl = {https://arxiv.org/abs/2209.04836},
  author = {Ainsworth, Samuel K. and Hayase, Jonathan and Srinivasa, Siddhartha},
  title = {Git Re-Basin: {M}erging Models modulo Permutation Symmetries},
  journal={arXiv preprint arXiv:2209.04836},
  year = {2022}
}

@article{papyan2020prevalence,
  title={Prevalence of neural collapse during the terminal phase of deep learning training},
  author={Papyan, Vardan and Han, XY and Donoho, David L},
  journal={Proceedings of the National Academy of Sciences},
  volume={117},
  number={40},
  pages={24652--24663},
  year={2020},
  OPTpublisher={National Acad Sciences}
}

@article{belkin2019reconciling,
  title={Reconciling modern machine-learning practice and the classical bias--variance trade-off},
  author={Belkin, Mikhail and Hsu, Daniel and Ma, Siyuan and Mandal, Soumik},
  journal={Proceedings of the National Academy of Sciences},
  volume={116},
  number={32},
  pages={15849--15854},
  year={2019},
  OPTpublisher={National Acad Sciences}
}

@article{nakkiran2021deep,
  title={Deep double descent: {W}here bigger models and more data hurt},
  author={Nakkiran, Preetum and Kaplun, Gal and Bansal, Yamini and Yang, Tristan and Barak, Boaz and Sutskever, Ilya},
  journal={Journal of Statistical Mechanics: Theory and Experiment},
  volume={2021},
  number={12},
  OPTpages={124003},
  year={2021},
  pOPTublisher={IOP Publishing}
}

@article{frankle2020revisiting,
  title={Revisiting ``{Q}ualitatively Characterizing Neural Network Optimization Problems''},
  author={Frankle, Jonathan},
  journal={arXiv preprint arXiv:2012.06898},
  year={2020}
}

@article{lecun1998gradient,
  title={Gradient-based learning applied to document recognition},
  author={LeCun, Yann and Bottou, L{\'e}on and Bengio, Yoshua and Haffner, Patrick},
  journal={Proceedings of the IEEE},
  volume={86},
  number={11},
  pages={2278--2324},
  year={1998},
  publisher={Ieee}
}

@inproceedings{benton2021loss,
  title={Loss surface simplexes for mode connecting volumes and fast ensembling},
  author={Benton, Gregory and Maddox, Wesley and Lotfi, Sanae and Wilson, Andrew Gordon},
  booktitle=icml,
  pages={769--779},
  year={2021},
  OPTorganization={PMLR}
}

@article{benzing2022random,
  title={Random initialisations performing above chance and how to find them},
  author={Benzing, Frederik and Schug, Simon and Meier, Robert and von Oswald, Johannes and Akram, Yassir and Zucchet, Nicolas and Aitchison, Laurence and Steger, Angelika},
  journal={arXiv preprint arXiv:2209.07509},
  year={2022}
}

@misc{jordan2024muon,
  author       = {Keller Jordan and Yuchen Jin and Vlado Boza and Jiacheng You and
                  Franz Cesista and Laker Newhouse and Jeremy Bernstein},
  title        = {Muon: An optimizer for hidden layers in neural networks},
  year         = {2024},
  url          = {https://kellerjordan.github.io/posts/muon/}
}

@article{yunis2024approaching,
  title={Approaching deep learning through the spectral dynamics of weights},
  author={Yunis, David and Patel, Kumar Kshitij and Wheeler, Samuel and Savarese, Pedro and Vardi, Gal and Livescu, Karen and Maire, Michael and Walter, Matthew R},
  journal={arXiv preprint arXiv:2408.11804},
  year={2024}
}

@article{shah2020pitfalls,
  title={The pitfalls of simplicity bias in neural networks},
  author={Shah, Harshay and Tamuly, Kaustav and Raghunathan, Aditi and Jain, Prateek and Netrapalli, Praneeth},
  journal={Advances in Neural Information Processing Systems},
  volume={33},
  pages={9573--9585},
  year={2020}
}

@article{pezeshki2021gradient,
  title={Gradient starvation: A learning proclivity in neural networks},
  author={Pezeshki, Mohammad and Kaba, Oumar and Bengio, Yoshua and Courville, Aaron C and Precup, Doina and Lajoie, Guillaume},
  journal={Advances in Neural Information Processing Systems},
  volume={34},
  pages={1256--1272},
  year={2021}
}

@inproceedings{sharma2024truth,
  title={The Truth is in There: Improving Reasoning in Language Models with Layer-Selective Rank Reduction},
  author={Sharma, Pratyusha and Ash, Jordan T and Misra, Dipendra},
  booktitle={The Twelfth International Conference on Learning Representations},
  year={2024}
}

@inproceedings{chen2024truncating,
  title={How Truncating Weights Improves Reasoning in Language Models},
  author={Chen, Lei and Bruna, Joan and Bietti, Alberto},
  booktitle={High-dimensional Learning Dynamics 2024: The Emergence of Structure and Reasoning},
  year={2024}
}

@inproceedings{zhang2023learning,
  title={Learning useful representations for shifting tasks and distributions},
  author={Zhang, Jianyu and Bottou, L{\'e}on},
  booktitle={International Conference on Machine Learning},
  pages={40830--40850},
  year={2023},
  organization={PMLR}
}

@article{gontijo2021no,
  title={No one representation to rule them all: Overlapping features of training methods},
  author={Gontijo-Lopes, Raphael and Dauphin, Yann and Cubuk, Ekin D},
  journal={arXiv preprint arXiv:2110.12899},
  year={2021}
}

@article{wen2025elastic,
  title={Elastic Representation: Mitigating Spurious Correlations for Group Robustness},
  author={Wen, Tao and Wang, Zihan and Zhang, Quan and Lei, Qi},
  journal={arXiv preprint arXiv:2502.09850},
  year={2025}
}

@article{pagliardini2022agree,
  title={Agree to disagree: Diversity through disagreement for better transferability},
  author={Pagliardini, Matteo and Jaggi, Martin and Fleuret, Fran{\c{c}}ois and Karimireddy, Sai Praneeth},
  journal={arXiv preprint arXiv:2202.04414},
  year={2022}
}

@article{lee2022diversify,
  title={Diversify and disambiguate: Learning from underspecified data},
  author={Lee, Yoonho and Yao, Huaxiu and Finn, Chelsea},
  journal={arXiv preprint arXiv:2202.03418},
  year={2022}
}

@article{kirichenko2022last,
  title={Last layer re-training is sufficient for robustness to spurious correlations},
  author={Kirichenko, Polina and Izmailov, Pavel and Wilson, Andrew Gordon},
  journal={arXiv preprint arXiv:2204.02937},
  year={2022}
}

@article{izmailov2022feature,
  title={On feature learning in the presence of spurious correlations},
  author={Izmailov, Pavel and Kirichenko, Polina and Gruver, Nate and Wilson, Andrew G},
  journal={Advances in Neural Information Processing Systems},
  volume={35},
  pages={38516--38532},
  year={2022}
}

@article{to2025diverse,
  title={Diverse Prototypical Ensembles Improve Robustness to Subpopulation Shift},
  author={To, Minh Nguyen Nhat and RWilson, Paul F and Nguyen, Viet and Harmanani, Mohamed and Cooper, Michael and Fooladgar, Fahimeh and Abolmaesumi, Purang and Mousavi, Parvin and Krishnan, Rahul G},
  journal={arXiv preprint arXiv:2505.23027},
  year={2025}
}

@inproceedings{sinha2021dibs,
  title={Dibs: Diversity inducing information bottleneck in model ensembles},
  author={Sinha, Samarth and Bharadhwaj, Homanga and Goyal, Anirudh and Larochelle, Hugo and Garg, Animesh and Shkurti, Florian},
  booktitle={Proceedings of the AAAI Conference on Artificial Intelligence},
  volume={35},
  number={11},
  pages={9666--9674},
  year={2021}
}

@article{huh2021low,
  title={The low-rank simplicity bias in deep networks},
  author={Huh, Minyoung and Mobahi, Hossein and Zhang, Richard and Cheung, Brian and Agrawal, Pulkit and Isola, Phillip},
  journal={arXiv preprint arXiv:2103.10427},
  year={2021}
}

@inproceedings{rahaman2019spectral,
  title={On the spectral bias of neural networks},
  author={Rahaman, Nasim and Baratin, Aristide and Arpit, Devansh and Draxler, Felix and Lin, Min and Hamprecht, Fred and Bengio, Yoshua and Courville, Aaron},
  booktitle={International conference on machine learning},
  pages={5301--5310},
  year={2019},
  organization={PMLR}
}

@inproceedings{liu2021just,
  title={Just train twice: Improving group robustness without training group information},
  author={Liu, Evan Z and Haghgoo, Behzad and Chen, Annie S and Raghunathan, Aditi and Koh, Pang Wei and Sagawa, Shiori and Liang, Percy and Finn, Chelsea},
  booktitle={International Conference on Machine Learning},
  pages={6781--6792},
  year={2021},
  organization={PMLR}
}

@inproceedings{zhang2022rich,
  title={Rich feature construction for the optimization-generalization dilemma},
  author={Zhang, Jianyu and Lopez-Paz, David and Bottou, L{\'e}on},
  booktitle={International Conference on Machine Learning},
  pages={26397--26411},
  year={2022},
  organization={PMLR}
}

@article{morwani2023simplicity,
  title={Simplicity bias in 1-hidden layer neural networks},
  author={Morwani, Depen and Batra, Jatin and Jain, Prateek and Netrapalli, Praneeth},
  journal={Advances in Neural Information Processing Systems},
  volume={36},
  pages={8048--8075},
  year={2023}
}

@article{chen2023project,
  title={Project and probe: Sample-efficient domain adaptation by interpolating orthogonal features},
  author={Chen, Annie S and Lee, Yoonho and Setlur, Amrith and Levine, Sergey and Finn, Chelsea},
  journal={arXiv preprint arXiv:2302.05441},
  year={2023}
}

@inproceedings{teney2022evading,
  title={Evading the simplicity bias: Training a diverse set of models discovers solutions with superior ood generalization},
  author={Teney, Damien and Abbasnejad, Ehsan and Lucey, Simon and Van den Hengel, Anton},
  booktitle={Proceedings of the IEEE/CVF conference on computer vision and pattern recognition},
  pages={16761--16772},
  year={2022}
}

@inproceedings{tiwari2023overcoming,
  title={Overcoming simplicity bias in deep networks using a feature sieve},
  author={Tiwari, Rishabh and Shenoy, Pradeep},
  booktitle={International Conference on Machine Learning},
  pages={34330--34343},
  year={2023},
  organization={PMLR}
}

@inproceedings{addepalli2023feature,
  title={Feature reconstruction from outputs can mitigate simplicity bias in neural networks},
  author={Addepalli, Sravanti and Nasery, Anshul and Radhakrishnan, Venkatesh Babu and Netrapalli, Praneeth and Jain, Prateek},
  booktitle={The Eleventh International Conference on Learning Representations},
  year={2023}
}

@inproceedings{joshi2025mitigating,
  title={Mitigating Simplicity Bias in Neural Networks: A Feature Sieve Modification, Regularization, and Self-Supervised Augmentation Approach},
  author={Joshi, Gaurav and Shah, Parth Narendra and Verma, Rachit},
  booktitle={Workshop on Spurious Correlation and Shortcut Learning: Foundations and Solutions},
  year={2025}
}

@article{niu2022roadblocks,
  title={Roadblocks for temporarily disabling shortcuts and learning new knowledge},
  author={Niu, Hongjing and Li, Hanting and Zhao, Feng and Li, Bin},
  journal={Advances in Neural Information Processing Systems},
  volume={35},
  pages={29064--29075},
  year={2022}
}

@article{prieto2025grokking,
  title={Grokking at the edge of numerical stability},
  author={Prieto, Lucas and Barsbey, Melih and Mediano, Pedro AM and Birdal, Tolga},
  journal={arXiv preprint arXiv:2501.04697},
  year={2025}
}

@article{dai2024deepseekmoe,
  title={Deepseekmoe: Towards ultimate expert specialization in mixture-of-experts language models},
  author={Dai, Damai and Deng, Chengqi and Zhao, Chenggang and Xu, RX and Gao, Huazuo and Chen, Deli and Li, Jiashi and Zeng, Wangding and Yu, Xingkai and Wu, Yu and others},
  journal={arXiv preprint arXiv:2401.06066},
  year={2024}
}

@article{yang2025these,
  title={These are Not All the Features You are Looking For: A Fundamental Bottleneck In Supervised Pretraining},
  author={Yang, Xingyu Alice and Zhang, Jianyu and Bottou, L{\'e}on},
  journal={arXiv preprint arXiv:2506.18221},
  year={2025}
}

@string{iclr = {Proceedings of the International Conference on Learning Representations (ICLR)}}

@string{cvpr = {Proceedings of the IEEE/CVF Conference on Computer Vision and Pattern Recognition (CVPR)}}

@string{iccv = {Proceedings of the International Conference on Computer Vision (ICCV)}}

@string{eccv = {Proceedings of the European Conference on Computer Vision (ECCV)}}

@string{aaai = {Proceedings of the National Conference on Artificial Intelligence (AAAI)}}

@string{icml = {Proceedings of the International Conference on Machine Learning (ICML)}}

@string{jmlr = {Journal of Machine Learning Research}}

@string{neurips = {Advances in Neural Information Processing Systems (NeurIPS)}}

@string{icassp = {Proceedings of the IEEE International Conference on Acoustics, Speech and Signal Processing (ICASSP)}}

@string{aistats = {Proceedings of the International Conference on Artificial Intelligence and Statistics (AISTATS)}}

@string{tmlr = {Transactions on Machine Learning Research}}

@string{colt = {Proceedings of the Annual Conference on Learning Theory (COLT)}}

@string{facct = {{Proceedings of the ACM Conference on Fairness, Accountability, and Transparency (FAccT)},}}

@article{abe2022deep,
  author = {Abe, Taiga and Buchanan, Estefany Kelly and Pleiss, Geoff and Zemel, Richard and Cunningham, John P},
  journal = {Advances in Neural Information Processing Systems},
  pages = {33646--33660},
  title = {{Deep ensembles work, but are they necessary?}},
  volume = {35},
  year = {2022}
}

@article{achiam2023gpt,
  author = {Achiam, Josh and Adler, Steven and Agarwal, Sandhini and Ahmad, Lama and Akkaya, Ilge and Aleman, Florencia Leoni and Almeida, Diogo and Altenschmidt, Janko and Altman, Sam and Anadkat, Shyamal and others},
  journal = {arXiv preprint arXiv:2303.08774},
  title = {{GPT-4 Technical Report}},
  year = {2023}
}

@article{alet2025skillful,
  author = {Alet, Ferran and Price, Ilan and El-Kadi, Andrew and Masters, Dominic and Markou, Stratis and Andersson, Tom R and Stott, Jacklynn and Lam, Remi and Willson, Matthew and Sanchez-Gonzalez, Alvaro and others},
  journal = {arXiv preprint arXiv:2506.10772},
  title = {{Skillful joint probabilistic weather forecasting from marginals}},
  year = {2025}
}

@article{andriushchenko2023we,
  author = {Andriushchenko, Maksym and D'Angelo, Francesco and Varre, Aditya and Flammarion, Nicolas},
  journal = {arXiv preprint arXiv:2310.04415},
  title = {{Why Do We Need Weight Decay in Modern Deep Learning?}},
  year = {2023}
}

@inproceedings{arora2018optimization,
  author = {Arora, Sanjeev and Cohen, Nadav and Hazan, Elad},
  booktitle = icml,
  optorganization = {PMLR},
  optpages = {244--253},
  title = {{On the optimization of deep networks: Implicit acceleration by overparameterization}},
  year = {2018}
}

@inproceedings{arora2019implicit,
  author = {Arora, Sanjeev and Cohen, Nadav and Hu, Wei and Luo, Yuping},
  booktitle = neurips,
  optvolume = {32},
  title = {{Implicit Regularization in Deep Matrix Factorization}},
  year = {2019}
}

@article{bai2022constitutional,
  author = {Yuntao Bai and Saurav Kadavath and Sandipan Kundu and Amanda Askell and Jackson Kernion and Andy Jones and Anna Chen and Anna Goldie and Azalia Mirhoseini and Cameron McKinnon and Carol Chen and Catherine Olsson and Christopher Olah and Danny Hernandez and Dawn Drain and Deep Ganguli and Dustin Li and Eli Tran-Johnson and Ethan Perez and Jamie Kerr and Jared Mueller and Jeffrey Ladish and Joshua Landau and Kamal Ndousse and Kamile Lukosuite and Liane Lovitt and Michael Sellitto and Nelson Elhage and Nicholas Schiefer and Noemi Mercado and Nova DasSarma and Robert Lasenby and Robin Larson and Sam Ringer and Scott Johnston and Shauna Kravec and Sheer El Showk and Stanislav Fort and Tamera Lanham and Timothy Telleen-Lawton and Tom Conerly and Tom Henighan and Tristan Hume and Samuel R. Bowman and Zac Hatfield-Dodds and Ben Mann and Dario Amodei and Nicholas Joseph and Sam McCandlish and Tom Brown and Jared Kaplan},
  journal = {arXiv preprint arXiv:2212.08073},
  title = {{Constitutional {AI}: {H}armlessness from ai feedback}},
  year = {2022}
}

@inproceedings{barak2022hidden,
  author = {Barak, Boaz and Edelman, Benjamin and Goel, Surbhi and Kakade, Sham and Malach, Eran and Zhang, Cyril},
  booktitle = neurips,
  optpages = {21750--21764},
  optvolume = {35},
  title = {{Hidden progress in deep learning: {SGD} learns parities near the computational limit}},
  year = {2022}
}

@article{bartlett1996valid,
  author = {Bartlett, Peter},
  journal = neurips,
  title = {{For valid generalization the size of the weights is more important than the size of the network}},
  volume = {9},
  year = {1996}
}

@inproceedings{beery2018recognition,
  author = {Beery, Sara and Van Horn, Grant and Perona, Pietro},
  booktitle = {Proceedings of the European conference on computer vision (ECCV)},
  pages = {456--473},
  title = {{Recognition in terra incognita}},
  year = {2018}
}

@inproceedings{bender2021dangers,
  author = {Bender, Emily M and Gebru, Timnit and McMillan-Major, Angelina and Shmitchell, Shmargaret},
  booktitle = {Proceedings ACM Conference on Fairness, Accountability, and Transparency (FAccT)},
  pages = {610--623},
  title = {{On the dangers of stochastic parrots: {C}an language models be too big?}},
  year = {2021}
}

@inproceedings{biderman2023pythia,
  author = {Biderman, Stella and Schoelkopf, Hailey and Anthony, Quentin Gregory and Bradley, Herbie and O’Brien, Kyle and Hallahan, Eric and Khan, Mohammad Aflah and Purohit, Shivanshu and Prashanth, USVSN Sai and Raff, Edward and others},
  booktitle = icml,
  optorganization = {PMLR},
  optpages = {2397--2430},
  title = {{Pythia: A suite for analyzing large language models across training and scaling}},
  year = {2023}
}

@misc{biewald2020experiment,
  author = {Biewald, Lukas},
  note = {Software available from wandb.com},
  title = {{Experiment Tracking with Weights and Biases}},
  url = {https://www.wandb.com/},
  year = {2020}
}

@inproceedings{black2022gpt,
  author = {Black, Sidney and Biderman, Stella and Hallahan, Eric and Anthony, Quentin and Gao, Leo and Golding, Laurence and He, Horace and Leahy, Connor and McDonell, Kyle and Phang, Jason and others},
  booktitle = {Proceedings of BigScience Episode\# 5--Workshop on Challenges \& Perspectives in Creating Large Language Models},
  pages = {95--136},
  title = {{Gpt-neox-20b: An open-source autoregressive language model}},
  year = {2022}
}

@article{blum1988training,
  author = {Blum, Avrim and Rivest, Ronald},
  journal = neurips,
  title = {{Training a 3-node neural network is NP-complete}},
  volume = {1},
  year = {1988}
}

@inproceedings{boix2023transformers,
  author = {Boix-Adser{\`a}, Enric and Littwin, Etai and Abbe, Emmanuel and Bengio, Samy and Susskind, Joshua M},
  booktitle = neurips,
  title = {{Transformers learn through gradual rank increase}},
  year = {2023}
}

@article{bommasani2021opportunities,
  author = {Bommasani, Rishi and Hudson, Drew A. and Adeli, Ehsan and Altman, Russ and Arora, Simran and von Arx, Sydney and Bernstein, Michael S. and Bohg, Jeannette and Bosselut, Antoine and Brunskill, Emma and others},
  journal = {arXiv prepring arXiv:2108.07258},
  title = {{On the Opportunities and Risks of Foundation Models}},
  year = {2021}
}

@article{brown2020language,
  author = {Brown, Tom and Mann, Benjamin and Ryder, Nick and Subbiah, Melanie and Kaplan, Jared D and Dhariwal, Prafulla and Neelakantan, Arvind and Shyam, Pranav and Sastry, Girish and Askell, Amanda and others},
  journal = neurips,
  pages = {1877--1901},
  title = {{Language models are few-shot learners}},
  volume = {33},
  year = {2020}
}

@inproceedings{burns2023discovering,
  author = {Burns, Collin and Ye, Haotian and Klein, Dan and Steinhardt, Jacob},
  booktitle = iclr,
  title = {{Discovering Latent Knowledge in Language Models Without Supervision}},
  year = {2023}
}

@article{buschjager2020generalized,
  author = {Buschj{\"a}ger, Sebastian and Pfahler, Lukas and Morik, Katharina},
  journal = {arXiv preprint arXiv:2011.02952},
  title = {{Generalized negative correlation learning for deep ensembling}},
  year = {2020}
}

@article{chen2021evaluating,
  archiveprefix = {arXiv},
  author = {Mark Chen and Jerry Tworek and Heewoo Jun and Qiming Yuan and Henrique Ponde de Oliveira Pinto and Jared Kaplan and Harri Edwards and Yuri Burda and Nicholas Joseph and Greg Brockman and Alex Ray and Raul Puri and Gretchen Krueger and Michael Petrov and Heidy Khlaaf and Girish Sastry and Pamela Mishkin and Prafulla Dhariwal and Scott Gray and Ryan Lowe and Alec Radford and Jeffrey Wu and Wojciech Zaremba},
  journal = {arXiv:2107.03374},
  doi = {10.48550/arXiv.2107.03374},
  eprint = {2107.03374},
  primaryclass = {cs.LG},
  title = {{Evaluating Large Language Models Trained on Code}},
  year = {2021}
}

@inproceedings{chen2023stochastic,
  author = {Chen, Feng and Kunin, Daniel and Yamamura, Atsushi and Ganguli, Surya},
  booktitle = neurips,
  optvolume = {36},
  title = {{Stochastic collapse: How gradient noise attracts {SGD} dynamics towards simpler subnetworks}},
  year = {2023}
}

@article{chen2026does,
  author = {Chen, Zhiqi and Lu, Rui and Zhao, Andrew and Wang, Zhaokai and Yue, Yang and Song, Shiji and Huang, Gao},
  journal = neurips,
  pages = {57654--57689},
  title = {{Does reinforcement learning really incentivize reasoning capacity in llms beyond the base model?}},
  volume = {38},
  year = {2026}
}

@inproceedings{choromanska2015loss,
  author = {Choromanska, Anna and Henaff, Mikael and Mathieu, Michael and Arous, G{\'e}rard Ben and LeCun, Yann},
  booktitle = {Artificial intelligence and statistics},
  organization = {PMLR},
  pages = {192--204},
  title = {{The loss surfaces of multilayer networks}},
  year = {2015}
}

@article{chrabaszcz2017downsampled,
  author = {Chrabaszcz, Patryk and Loshchilov, Ilya and Hutter, Frank},
  journal = {arXiv preprint arXiv:1707.08819},
  title = {{A downsampled variant of imagenet as an alternative to the cifar datasets}},
  year = {2017}
}

@inproceedings{coster2011simple,
  author = {Coster, William and Kauchak, David},
  booktitle = {Proceedings of the 49th Annual Meeting of the Association for Computational Linguistics: Human Language Technologies},
  pages = {665--669},
  title = {{Simple English Wikipedia: a new text simplification task}},
  year = {2011}
}

@article{cybenko1989approximation,
  author = {Cybenko, George},
  journal = {Mathematics of control, signals and systems},
  number = {4},
  pages = {303--314},
  publisher = {Springer},
  title = {{Approximation by superpositions of a sigmoidal function}},
  volume = {2},
  year = {1989}
}

@article{d2022underspecification,
  author = {D'Amour, Alexander and Heller, Katherine and Moldovan, Dan and Adlam, Ben and Alipanahi, Babak and Beutel, Alex and Chen, Christina and Deaton, Jonathan and Eisenstein, Jacob and Hoffman, Matthew D and others},
  journal = {Journal of Machine Learning Research},
  number = {226},
  pages = {1--61},
  title = {{Underspecification presents challenges for credibility in modern machine learning}},
  volume = {23},
  year = {2022}
}

@inproceedings{davies2022unifying,
  author = {Davies, Xander and Langosco, Lauro and Krueger, David},
  booktitle = {NeurIPS ML Safety Workshop},
  title = {{Unifying Grokking and Double Descent}},
  year = {2022}
}

@article{dittmer2019singular,
  author = {Dittmer, S{\"o}ren and King, Emily J and Maass, Peter},
  journal = {IEEE Transactions on Neural Networks and Learning Systems},
  number = {9},
  optpublisher = {IEEE},
  pages = {3594--3605},
  title = {{Singular values for {ReLU} layers}},
  volume = {31},
  year = {2019}
}

@inproceedings{du2018algorithmic,
  author = {Du, Simon S and Hu, Wei and Lee, Jason D},
  booktitle = neurips,
  optvolume = {31},
  title = {{Algorithmic regularization in learning deep homogeneous models: Layers are automatically balanced}},
  year = {2018}
}

@inproceedings{ergen2023path,
  author = {Ergen, Tolga and Pilanci, Mert},
  booktitle = neurips,
  optvolume = {36},
  title = {{Path regularization: A convexity and sparsity inducing regularization for parallel relu networks}},
  year = {2023}
}

@article{fan2024deep,
  author = {Fan, Simin and Pascanu, Razvan and Jaggi, Martin},
  journal = {arXiv preprint arXiv:2405.19454},
  title = {{Deep Grokking: Would Deep Neural Networks Generalize Better?}},
  year = {2024}
}

@inproceedings{feng2022rank,
  author = {Feng, Ruili and Zheng, Kecheng and Huang, Yukun and Zhao, Deli and Jordan, Michael and Zha, Zheng-Jun},
  booktitle = neurips,
  optpages = {33054--33065},
  optvolume = {35},
  title = {{Rank diminishing in deep neural networks}},
  year = {2022}
}

@article{ferbach2023proving,
  author = {Ferbach, Damien and Goujaud, Baptiste and Gidel, Gauthier and Dieuleveut, Aymeric},
  journal = {arXiv preprint arXiv:2310.19103},
  title = {{Proving linear mode connectivity of neural networks via optimal transport}},
  year = {2023}
}

@inproceedings{frankle2018lottery,
  author = {Frankle, Jonathan and Carbin, Michael},
  booktitle = iclr,
  title = {{The Lottery Ticket Hypothesis: Finding Sparse, Trainable Neural Networks}},
  year = {2018}
}

@inproceedings{frei2022implicit,
  author = {Frei, Spencer and Vardi, Gal and Bartlett, Peter and Srebro, Nathan and Hu, Wei},
  booktitle = iclr,
  title = {{Implicit Bias in Leaky {ReLU} Networks Trained on High-Dimensional Data}},
  year = {2022}
}

@article{freund1999short,
  author = {Freund, Yoav and Schapire, Robert and Abe, Naoki},
  journal = {Journal-Japanese Society For Artificial Intelligence},
  number = {771-780},
  pages = {1612},
  publisher = {JAPANESE SOC ARTIFICIAL INTELL},
  title = {{A short introduction to boosting}},
  volume = {14},
  year = {1999}
}

@article{galanti2022sgd,
  author = {Galanti, Tomer and Siegel, Zachary S and Gupte, Aparna and Poggio, Tomaso},
  journal = {arXiv preprint arXiv:2206.05794},
  title = {{{SGD} and weight decay provably induce a low-rank bias in neural networks}},
  year = {2022}
}

@inproceedings{garipov2018loss,
  author = {Garipov, Timur and Izmailov, Pavel and Podoprikhin, Dmitrii and Vetrov, Dmitry P and Wilson, Andrew G},
  booktitle = neurips,
  title = {{Loss surfaces, mode connectivity, and fast ensembling of {DNNs}}},
  volume = {31},
  year = {2018}
}

@article{ghosh2025learning,
  author = {Ghosh, Avrajit and Kwon, Soo Min and Wang, Rongrong and Ravishankar, Saiprasad and Qu, Qing},
  journal = {arXiv preprint arXiv:2502.20531},
  title = {{Learning Dynamics of Deep Linear Networks Beyond the Edge of Stability}},
  year = {2025}
}

@article{grattafiori2024llama,
  author = {Grattafiori, Aaron and Dubey, Abhimanyu and Jauhri, Abhinav and Pandey, Abhinav and Kadian, Abhishek and Al-Dahle, Ahmad and Letman, Aiesha and Mathur, Akhil and Schelten, Alan and Vaughan, Alex and others},
  journal = {arXiv preprint arXiv:2407.21783},
  title = {{The llama 3 herd of models}},
  year = {2024}
}

@article{gromov2023grokking,
  author = {Gromov, Andrey},
  journal = {arXiv preprint arXiv:2301.02679},
  title = {{Grokking modular arithmetic}},
  year = {2023}
}

@inproceedings{gunasekar2017implicit,
  author = {Gunasekar, Suriya and Woodworth, Blake E and Bhojanapalli, Srinadh and Neyshabur, Behnam and Srebro, Nati},
  booktitle = neurips,
  optvolume = {30},
  title = {{Implicit regularization in matrix factorization}},
  year = {2017}
}

@article{guo2025deepseek,
  author = {Guo, Daya and Yang, Dejian and Zhang, Haowei and Song, Junxiao and Wang, Peiyi and Zhu, Qihao and Xu, Runxin and Zhang, Ruoyu and Ma, Shirong and Bi, Xiao and others},
  journal = {arXiv preprint arXiv:2501.12948},
  title = {{Deepseek-r1: Incentivizing reasoning capability in llms via reinforcement learning}},
  year = {2025}
}

@inproceedings{hanson1988comparing,
  author = {Hanson, Stephen and Pratt, Lorien},
  booktitle = neurips,
  optvolume = {1},
  title = {{Comparing biases for minimal network construction with back-propagation}},
  year = {1988}
}

@article{harris2020array,
  author = {Charles R. Harris and K. Jarrod Millman and St{\'{e}}fan J.
van der Walt and Ralf Gommers and Pauli Virtanen and David
Cournapeau and Eric Wieser and Julian Taylor and Sebastian
Berg and Nathaniel J. Smith and Robert Kern and Matti Picus
and Stephan Hoyer and Marten H. van Kerkwijk and Matthew
Brett and Allan Haldane and Jaime Fern{\'{a}}ndez del
R{\'{i}}o and Mark Wiebe and Pearu Peterson and Pierre
G{\'{e}}rard-Marchant and Kevin Sheppard and Tyler Reddy and
Warren Weckesser and Hameer Abbasi and Christoph Gohlke and
Travis E. Oliphant},
  journal = {Nature},
  month = sep,
  number = {7825},
  optdoi = {10.1038/s41586-020-2649-2},
  optpublisher = {Springer Science and Business Media {LLC}},
  opturl = {https://doi.org/10.1038/s41586-020-2649-2},
  pages = {357--362},
  title = {{Array programming with {NumPy}}},
  volume = {585},
  year = {2020}
}

@inproceedings{hendrycks2021natural,
  author = {Hendrycks, Dan and Zhao, Kevin and Basart, Steven and Steinhardt, Jacob and Song, Dawn},
  booktitle = cvpr,
  optpages = {15262--15271},
  title = {{Natural adversarial examples}},
  year = {2021}
}

@article{hestness2017deep,
  author = {Hestness, Joel and Narang, Sharan and Ardalani, Newsha and Diamos, Gregory and Jun, Heewoo and Kianinejad, Hassan and Patwary, Md Mostofa Ali and Yang, Yang and Zhou, Yanqi},
  journal = {arXiv preprint arXiv:1712.00409},
  title = {{Deep learning scaling is predictable, empirically}},
  year = {2017}
}

@article{hinton2012deep,
  author = {Hinton, Geoffrey and Deng, Li and Yu, Dong and Dahl, George E and Mohamed, Abdel-rahman and Jaitly, Navdeep and Senior, Andrew and Vanhoucke, Vincent and Nguyen, Patrick and Sainath, Tara N and others},
  journal = {IEEE Signal processing magazine},
  number = {6},
  pages = {82--97},
  publisher = {IEEE},
  title = {{Deep neural networks for acoustic modeling in speech recognition: The shared views of four research groups}},
  volume = {29},
  year = {2012}
}

@inproceedings{ho2020denoising,
  author = {Ho, Jonathan and Jain, Ajay and Abbeel, Pieter},
  booktitle = neurips,
  optpages = {6840--6851},
  optvolume = {33},
  title = {{Denoising diffusion probabilistic models}},
  year = {2020}
}

@article{hochreiter1997long,
  author = {Hochreiter, Sepp and Schmidhuber, J{\"u}rgen},
  journal = {Neural Computation},
  number = {8},
  optpublisher = {MIT press},
  pages = {1735--1780},
  title = {{Long short-term memory}},
  volume = {9},
  year = {1997}
}

@inproceedings{hu2018does,
  author = {Hu, Weihua and Niu, Gang and Sato, Issei and Sugiyama, Masashi},
  booktitle = {International Conference on Machine Learning},
  organization = {PMLR},
  pages = {2029--2037},
  title = {{Does distributionally robust supervised learning give robust classifiers?}},
  year = {2018}
}

@article{hu2024minicpm,
  author = {Hu, Shengding and Tu, Yuge and Han, Xu and He, Chaoqun and Cui, Ganqu and Long, Xiang and Zheng, Zhi and Fang, Yewei and Huang, Yuxiang and Zhao, Weilin and others},
  journal = {arXiv preprint arXiv:2404.06395},
  title = {{Minicpm: Unveiling the potential of small language models with scalable training strategies}},
  year = {2024}
}

@article{huh2022low,
  author = {Huh, Minyoung and Mobahi, Hossein and Zhang, Richard and Cheung, Brian and Agrawal, Pulkit and Isola, Phillip},
  journal = tmlr,
  title = {{The Low-Rank Simplicity Bias in Deep Networks}},
  year = {2022}
}

@article{humayun2024deep,
  author = {Humayun, Ahmed Imtiaz and Balestriero, Randall and Baraniuk, Richard},
  journal = {arXiv preprint arXiv:2402.15555},
  title = {{Deep Networks Always Grok and Here is Why}},
  year = {2024}
}

@article{Hunter:2007,
  author = {Hunter, J. D.},
  journal = {Computing in Science \& Engineering},
  number = {3},
  optabstract = {Matplotlib is a {2D} graphics package used for {Python} for
application development, interactive scripting, and publication-quality
image generation across user interfaces and operating systems.},
  optdoi = {10.1109/MCSE.2007.55},
  optpublisher = {IEEE COMPUTER SOC},
  pages = {90--95},
  title = {{Matplotlib: {A} {2D} graphics environment}},
  volume = {9},
  year = {2007}
}

@inproceedings{ilharco2022editing,
  author = {Ilharco, Gabriel and Ribeiro, Marco Tulio and Wortsman, Mitchell and Schmidt, Ludwig and Hajishirzi, Hannaneh and Farhadi, Ali},
  booktitle = iclr,
  title = {{Editing models with task arithmetic}},
  year = {2022}
}

@inproceedings{ilyas2019adversarial,
  author = {Ilyas, Andrew and Santurkar, Shibani and Tsipras, Dimitris and Engstrom, Logan and Tran, Brandon and Madry, Aleksander},
  booktitle = neurips,
  optvolume = {32},
  title = {{Adversarial examples are not bugs, they are features}},
  year = {2019}
}

@inproceedings{ioffe2015batch,
  author = {Ioffe, Sergey and Szegedy, Christian},
  booktitle = icml,
  optorganization = {pmlr},
  optpages = {448--456},
  title = {{Batch normalization: Accelerating deep network training by reducing internal covariate shift}},
  year = {2015}
}

@article{ito2024analysis,
  author = {Ito, Akira and Yamada, Masanori and Kumagai, Atsutoshi},
  journal = {arXiv preprint arXiv:2402.04051},
  title = {{Analysis of Linear Mode Connectivity via Permutation-Based Weight Matching}},
  year = {2024}
}

@inproceedings{jacot2018neural,
  author = {Jacot, Arthur and Gabriel, Franck and Hongler, Cl{\'e}ment},
  booktitle = neurips,
  optvolume = {31},
  title = {{Neural tangent kernel: {C}onvergence and generalization in neural networks}},
  year = {2018}
}

@article{jeffares2023joint,
  author = {Jeffares, Alan and Liu, Tennison and Crabb{\'e}, Jonathan and van der Schaar, Mihaela},
  journal = {Advances in Neural Information Processing Systems},
  pages = {13559--13589},
  title = {{Joint training of deep ensembles fails due to learner collusion}},
  volume = {36},
  year = {2023}
}

@inproceedings{ji2019gradient,
  author = {Ji, Ziwei and Telgarsky, Matus},
  booktitle = iclr,
  title = {{Gradient descent aligns the layers of deep linear networks}},
  year = {2019}
}

@inproceedings{jordan2022repair,
  author = {Jordan, Keller and Sedghi, Hanie and Saukh, Olga and Entezari, Rahim and Neyshabur, Behnam},
  booktitle = iclr,
  title = {{{REPAIR}: {RE}normalizing {P}ermuted {A}ctivations for {I}nterpolation {R}epair}},
  year = {2022}
}

@article{jumper2021highly,
  author = {Jumper, John and Evans, Richard and Pritzel, Alexander and Green, Tim and Figurnov, Michael and Ronneberger, Olaf and Tunyasuvunakool, Kathryn and Bates, Russ and {\v{Z}}{\'\i}dek, Augustin and Potapenko, Anna and others},
  journal = {nature},
  number = {7873},
  pages = {583--589},
  publisher = {Nature Publishing Group UK London},
  title = {{Highly accurate protein structure prediction with AlphaFold}},
  volume = {596},
  year = {2021}
}

@inproceedings{juneja2022linear,
  author = {Juneja, Jeevesh and Bansal, Rachit and Cho, Kyunghyun and Sedoc, Jo{\~a}o and Saphra, Naomi},
  booktitle = iclr,
  title = {{Linear Connectivity Reveals Generalization Strategies}},
  year = {2022}
}

@article{kaplan2020scaling,
  author = {Kaplan, Jared and McCandlish, Sam and Henighan, Tom and Brown, Tom B and Chess, Benjamin and Child, Rewon and Gray, Scott and Radford, Alec and Wu, Jeffrey and Amodei, Dario},
  journal = {arXiv preprint arXiv:2001.08361},
  title = {{Scaling laws for neural language models}},
  year = {2020}
}

@article{karras2023analyzing,
  author = {Karras, Tero and Aittala, Miika and Lehtinen, Jaakko and Hellsten, Janne and Aila, Timo and Laine, Samuli},
  journal = {arXiv preprint arXiv:2312.02696},
  title = {{Analyzing and improving the training dynamics of diffusion models}},
  year = {2023}
}

@inproceedings{kawaguchi2016deep,
  author = {Kawaguchi, Kenji},
  booktitle = neurips,
  pages = {586--594},
  title = {{Deep learning without poor local minima}},
  volume = {29},
  year = {2016}
}

@inproceedings{kim2026pre,
  author = {Kim, Konwoo and Kotha, Suhas and Liang, Percy and Hashimoto, Tatsunori},
  booktitle = {International Conference on Learning Representations},
  pages = {74596--74636},
  title = {{Pre-training under infinite compute}},
  year = {2026}
}

@misc{kkr2025beyondbubble,
  author = {{KKR Global Macro \& Asset Allocation}},
  howpublished = {KKR Insights},
  month = {November},
  title = {{Beyond the Bubble: Why We Think AI Infrastructure Will Compound Long after the Hype}},
  url = {https://www.kkr.com/insights/ai-infrastructure},
  year = {2025}
}

@article{kondratyuk2020ensembling,
  author = {Kondratyuk, Dan and Tan, Mingxing and Brown, Matthew and Gong, Boqing},
  journal = {arXiv preprint arXiv:2005.00570},
  title = {{When ensembling smaller models is more efficient than single large models}},
  year = {2020}
}

@inproceedings{kornblith2019similarity,
  author = {Kornblith, Simon and Norouzi, Mohammad and Lee, Honglak and Hinton, Geoffrey},
  booktitle = icml,
  organization = {PMlR},
  pages = {3519--3529},
  title = {{Similarity of neural network representations revisited}},
  year = {2019}
}

@article{krizhevsky2012imagenet,
  author = {Krizhevsky, Alex and Sutskever, Ilya and Hinton, Geoffrey E},
  journal = neurips,
  title = {{Imagenet classification with deep convolutional neural networks}},
  volume = {25},
  year = {2012}
}

@inproceedings{krogh1991simple,
  author = {Krogh, Anders and Hertz, John},
  booktitle = neurips,
  optvolume = {4},
  title = {{A simple weight decay can improve generalization}},
  year = {1991}
}

@article{kumar2023grokking,
  author = {Kumar, Tanishq and Bordelon, Blake and Gershman, Samuel J and Pehlevan, Cengiz},
  journal = {arXiv preprint arXiv:2310.06110},
  title = {{Grokking as the transition from lazy to rich training dynamics}},
  year = {2023}
}

@techreport{kusano2025comparison,
  archiveprefix = {arXiv},
  author = {Kusano, Kristofer D. and Scanlon, John M. and Chen, Yin-Hsiu and McMurry, Timothy L. and Gode, Tilia and Victor, Trent},
  doi = {10.48550/arXiv.2505.01515},
  eprint = {2505.01515},
  institution = {arXiv},
  primaryclass = {cs.RO},
  title = {{Comparison of Waymo Rider-Only Crash Rates by Crash Type to Human Benchmarks at 56.7 Million Miles}},
  year = {2025}
}

@inproceedings{le2021training,
  author = {Le, Thien and Jegelka, Stefanie},
  booktitle = iclr,
  title = {{Training invariances and the low-rank phenomenon: beyond linear networks}},
  year = {2021}
}

@misc{lecun1998mnist,
  author = {LeCun, Yann},
  howpublished = {\url{http://yann. lecun. com/exdb/mnist/}},
  title = {{The {MNIST} database of handwritten digits}},
  year = {1998}
}

@article{lhoest2021datasets,
  author = {Lhoest, Quentin and del Moral, Albert Villanova and Jernite, Yacine and Thakur, Abhishek and von Platen, Patrick and Patil, Suraj and Chaumond, Julien and Drame, Mariama and Plu, Julien and Tunstall, Lewis and others},
  journal = {arXiv preprint arXiv:2109.02846},
  title = {{Datasets: A community library for natural language processing}},
  year = {2021}
}

@inproceedings{li2020towards,
  author = {Li, Zhiyuan and Luo, Yuping and Lyu, Kaifeng},
  booktitle = iclr,
  title = {{Towards Resolving the Implicit Bias of Gradient Descent for Matrix Factorization: Greedy Low-Rank Learning}},
  year = {2020}
}

@inproceedings{lin2017focal,
  author = {Lin, Tsung-Yi and Goyal, Priya and Girshick, Ross and He, Kaiming and Doll{\'a}r, Piotr},
  booktitle = {Proceedings of the IEEE international conference on computer vision},
  pages = {2980--2988},
  title = {{Focal loss for dense object detection}},
  year = {2017}
}

@article{liu1999ensemble,
  author = {Liu, Yong and Yao, Xin},
  journal = {Neural networks},
  number = {10},
  pages = {1399--1404},
  publisher = {Elsevier},
  title = {{Ensemble learning via negative correlation}},
  volume = {12},
  year = {1999}
}

@inproceedings{liu2022towards,
  author = {Liu, Ziming and Kitouni, Ouail and Nolte, Niklas S and Michaud, Eric and Tegmark, Max and Williams, Mike},
  booktitle = neurips,
  optpages = {34651--34663},
  optvolume = {35},
  title = {{Towards understanding grokking: An effective theory of representation learning}},
  year = {2022}
}

@inproceedings{liu2023omnigrok,
  author = {Liu, Ziming and Michaud, Eric J and Tegmark, Max},
  booktitle = iclr,
  title = {{Omnigrok: Grokking beyond algorithmic data}},
  year = {2023}
}

@article{lobacheva2020power,
  author = {Lobacheva, Ekaterina and Chirkova, Nadezhda and Kodryan, Maxim and Vetrov, Dmitry P},
  journal = {Advances In Neural Information Processing Systems},
  pages = {2375--2385},
  title = {{On power laws in deep ensembles}},
  volume = {33},
  year = {2020}
}

@article{loshchilov2018fixing,
  author = {Loshchilov, Ilya and Hutter, Frank},
  journal = {arXiv preprint arXiv:1711.05101},
  title = {{Fixing weight decay regularization in {A}dam}},
  year = {2017}
}

@inproceedings{lubana2023mechanistic,
  author = {Lubana, Ekdeep Singh and Bigelow, Eric J and Dick, Robert P and Krueger, David and Tanaka, Hidenori},
  booktitle = icml,
  optorganization = {PMLR},
  optpages = {22965--23004},
  title = {{Mechanistic mode connectivity}},
  year = {2023}
}

@inproceedings{lyu2023dichotomy,
  author = {Lyu, Kaifeng and Jin, Jikai and Li, Zhiyuan and Du, Simon Shaolei and Lee, Jason D and Hu, Wei},
  booktitle = iclr,
  title = {{Dichotomy of early and late phase implicit biases can provably induce grokking}},
  year = {2023}
}

@article{mandal2026q0,
  author = {Mandal, Bishwas and Berman, Shmuel and Vegesna, Akshay and Dahal, Samip},
  journal = {arXiv preprint arXiv:2606.03938},
  title = {{q0: Primitives for Hyper-Epoch Pretraining}},
  year = {2026}
}

@article{marquardt1975ridge,
  author = {Marquardt, Donald W and Snee, Ronald D},
  journal = {The American Statistician},
  number = {1},
  pages = {3--20},
  publisher = {Taylor \& Francis},
  title = {{Ridge regression in practice}},
  volume = {29},
  year = {1975}
}

@inproceedings{martin2020heavy,
  author = {Martin, Charles H and Mahoney, Michael W},
  booktitle = {Proceedings of the SIAM International Conference on Data Mining (ICDM)},
  optorganization = {SIAM},
  optpages = {505--513},
  title = {{Heavy-tailed universality predicts trends in test accuracies for very large pre-trained deep neural networks}},
  year = {2020}
}

@article{martin2021implicit,
  author = {Martin, Charles H and Mahoney, Michael W},
  journal = jmlr,
  number = {165},
  pages = {1--73},
  title = {{Implicit self-regularization in deep neural networks: Evidence from random matrix theory and implications for learning}},
  volume = {22},
  year = {2021}
}

@inproceedings{mazeika2024harmbench,
  author = {Mazeika, Mantas and Phan, Long and Yin, Xuwang and Zou, Andy and Wang, Zifan and Mu, Norman and Sakhaee, Elham and Li, Nathaniel and Basart, Steven and Li, Bo and others},
  booktitle = icml,
  title = {{{HarmBench}: {A} Standardized Evaluation Framework for Automated Red Teaming and Robust Refusal}},
  year = {2024}
}

@article{merchant2023scaling,
  author = {Merchant, Amil and Batzner, Simon and Schoenholz, Samuel S and Aykol, Muratahan and Cheon, Gowoon and Cubuk, Ekin Dogus},
  journal = {Nature},
  number = {7990},
  pages = {80--85},
  publisher = {Nature Publishing Group UK London},
  title = {{Scaling deep learning for materials discovery}},
  volume = {624},
  year = {2023}
}

@inproceedings{merity2016pointer,
  author = {Merity, Stephen and Xiong, Caiming and Bradbury, James and Socher, Richard},
  booktitle = iclr,
  title = {{Pointer Sentinel Mixture Models}},
  year = {2016}
}

@inproceedings{merrill2023tale,
  author = {Merrill, William and Tsilivis, Nikolaos and Shukla, Aman},
  booktitle = {ICLR 2023 Workshop on Mathematical and Empirical Understanding of Foundation Models},
  title = {{A Tale of Two Circuits: Grokking as Competition of Sparse and Dense Subnetworks}},
  year = {2023}
}

@inproceedings{milanesi2021implicit,
  author = {Milanesi, Paolo and Kadri, Hachem and Ayache, St{\'e}phane and Arti{\`e}res, Thierry},
  booktitle = {Proceedings of the International Joint Conference on Neural Networks (IJCNN)},
  optorganization = {IEEE},
  optpages = {1--8},
  title = {{Implicit regularization in deep tensor factorization}},
  year = {2021}
}

@article{mohan2019robust,
  author = {Mohan, Sreyas and Kadkhodaie, Zahra and Simoncelli, Eero P and Fernandez-Granda, Carlos},
  journal = {arXiv preprint arXiv:1906.05478},
  title = {{Robust and interpretable blind image denoising via bias-free convolutional neural networks}},
  year = {2019}
}

@inproceedings{mulayoff2020unique,
  author = {Mulayoff, Rotem and Michaeli, Tomer},
  booktitle = icml,
  optorganization = {PMLR},
  optpages = {7108--7118},
  title = {{Unique properties of flat minima in deep networks}},
  year = {2020}
}

@article{nanda2023progress,
  author = {Nanda, Neel and Chan, Lawrence and Lieberum, Tom and Smith, Jess and Steinhardt, Jacob},
  journal = {arXiv preprint arXiv:2301.05217},
  title = {{Progress measures for grokking via mechanistic interpretability}},
  year = {2023}
}

@article{neyshabur2014search,
  author = {Neyshabur, Behnam and Tomioka, Ryota and Srebro, Nathan},
  journal = {arXiv preprint arXiv:1412.6614},
  title = {{In search of the real inductive bias: On the role of implicit regularization in deep learning}},
  year = {2014}
}

@article{ongie2022role,
  author = {Ongie, Greg and Willett, Rebecca},
  journal = {arXiv preprint arXiv:2202.00856},
  title = {{The role of linear layers in nonlinear interpolating networks}},
  year = {2022}
}

@techreport{openai2026tenadvances,
  author = {{OpenAI}},
  day = {1},
  institution = {OpenAI},
  month = {August},
  title = {{Ten Advances in Mathematics and Theoretical Computer Science}},
  url = {https://cdn.openai.com/pdf/ten-proofs-oai.pdf},
  year = {2026}
}

@article{ouyang2022training,
  author = {Ouyang, Long and Wu, Jeffrey and Jiang, Xu and Almeida, Diogo and Wainwright, Carroll and Mishkin, Pamela and Zhang, Chong and Agarwal, Sandhini and Slama, Katarina and Ray, Alex and others},
  journal = neurips,
  pages = {27730--27744},
  title = {{Training language models to follow instructions with human feedback}},
  volume = {35},
  year = {2022}
}

@article{parhi2023deep,
  author = {Parhi, Rahul and Nowak, Robert D},
  journal = {IEEE Signal Processing Magazine},
  number = {6},
  pages = {63--74},
  publisher = {IEEE},
  title = {{Deep learning meets sparse regularization: A signal processing perspective}},
  volume = {40},
  year = {2023}
}

@inproceedings{park2024linear,
  author = {Park, Kiho and Choe, Yo Joong and Veitch, Victor},
  booktitle = icml,
  title = {{The Linear Representation Hypothesis and the Geometry of Large Language Models}},
  year = {2024}
}

@inproceedings{paszke2019pytorch,
  author = {Paszke, Adam and Gross, Sam and Massa, Francisco and Lerer, Adam and Bradbury, James and Chanan, Gregory and Killeen, Trevor and Lin, Zeming and Gimelshein, Natalia and Antiga, Luca and others},
  booktitle = neurips,
  optvolume = {32},
  title = {{{PyTorch}: {A}n imperative style, high-performance deep learning library}},
  year = {2019}
}

@inproceedings{paul2022unmasking,
  author = {Paul, Mansheej and Chen, Feng and Larsen, Brett W and Frankle, Jonathan and Ganguli, Surya and Dziugaite, Gintare Karolina},
  booktitle = iclr,
  title = {{Unmasking the Lottery Ticket Hypothesis: What's Encoded in a Winning Ticket's Mask?}},
  year = {2022}
}

@inproceedings{pennington2018emergence,
  author = {Pennington, Jeffrey and Schoenholz, Samuel and Ganguli, Surya},
  booktitle = aistats,
  optorganization = {PMLR},
  optpages = {1924--1932},
  title = {{The emergence of spectral universality in deep networks}},
  year = {2018}
}

@article{power2022grokking,
  author = {Power, Alethea and Burda, Yuri and Edwards, Harri and Babuschkin, Igor and Misra, Vedant},
  journal = {arXiv preprint arXiv:2201.02177},
  title = {{Grokking: Generalization beyond overfitting on small algorithmic datasets}},
  year = {2022}
}

@article{praggastis2022svd,
  author = {Praggastis, Brenda and Brown, Davis and Marrero, Carlos Ortiz and Purvine, Emilie and Shapiro, Madelyn and Wang, Bei},
  journal = {arXiv preprint arXiv:2208.06894},
  title = {{The {SVD} of convolutional weights: a {CNN} interpretability framework}},
  year = {2022}
}

@article{qu2024rethink,
  author = {Qu, Xingyu and Horvath, Samuel},
  journal = {arXiv preprint arXiv:2402.05966},
  title = {{Rethink Model Re-Basin and the Linear Mode Connectivity}},
  year = {2024}
}

@misc{radford2018improving,
  author = {Radford, Alec and Narasimhan, Karthik and Salimans, Tim and Sutskever, Ilya and others},
  publisher = {San Francisco, CA, USA},
  title = {{Improving language understanding by generative pre-training}},
  year = {2018}
}

@article{radford2019language,
  author = {Radford, Alec and Wu, Jeffrey and Child, Rewon and Luan, David and Amodei, Dario and Sutskever, Ilya and others},
  journal = {OpenAI blog},
  number = {8},
  pages = {9},
  title = {{Language models are unsupervised multitask learners}},
  volume = {1},
  year = {2019}
}

@article{raffel2020exploring,
  author = {Raffel, Colin and Shazeer, Noam and Roberts, Adam and Lee, Katherine and Narang, Sharan and Matena, Michael and Zhou, Yanqi and Li, Wei and Liu, Peter J},
  journal = {Journal of machine learning research},
  number = {140},
  pages = {1--67},
  title = {{Exploring the limits of transfer learning with a unified text-to-text transformer}},
  volume = {21},
  year = {2020}
}

@article{ramachandran2017searching,
  author = {Ramachandran, Prajit and Zoph, Barret and Le, Quoc V},
  journal = {arXiv preprint arXiv:1710.05941},
  title = {{Searching for activation functions}},
  year = {2017}
}

@inproceedings{rame2022diverse,
  author = {Rame, Alexandre and Kirchmeyer, Matthieu and Rahier, Thibaud and Rakotomamonjy, Alain and Gallinari, Patrick and Cord, Matthieu},
  booktitle = neurips,
  optpages = {10821--10836},
  optvolume = {35},
  title = {{Diverse weight averaging for out-of-distribution generalization}},
  year = {2022}
}

@article{ramesh2022hierarchical,
  author = {Ramesh, Aditya and Dhariwal, Prafulla and Nichol, Alex and Chu, Casey and Chen, Mark},
  journal = {arXiv preprint arXiv:2204.06125},
  title = {{Hierarchical Text-Conditional Image Generation with {CLIP} Latents}},
  year = {2022}
}

@inproceedings{razin2020implicit,
  author = {Razin, Noam and Cohen, Nadav},
  booktitle = neurips,
  optpages = {21174--21187},
  optvolume = {33},
  title = {{Implicit regularization in deep learning may not be explainable by norms}},
  year = {2020}
}

@inproceedings{ronneberger2015u,
  author = {Ronneberger, Olaf and Fischer, Philipp and Brox, Thomas},
  booktitle = {Proceeding of the International Conference on Medical Image Computing and Computer-Assisted Intervention (MICCAI)},
  optorganization = {Springer},
  optpages = {234--241},
  title = {{U-net: Convolutional networks for biomedical image segmentation}},
  year = {2015}
}

@inproceedings{roy2007effective,
  author = {Roy, Olivier and Vetterli, Martin},
  booktitle = {Proceedings of the European Signal Processing Conference (EUSIPCO)},
  optorganization = {IEEE},
  optpages = {606--610},
  title = {{The effective rank: {A} measure of effective dimensionality}},
  year = {2007}
}

@misc{rumbelow2023solidgoldmagikarp,
  author = {Rumbelow, Jessica and Watkins, Matthew},
  howpublished = {\url{https://www.lesswrong.com/posts/aPeJE8bSo6rAFoLqg/solidgoldmagikarp-plus-prompt-generation}},
  month = {February},
  note = {LessWrong},
  title = {{SolidGoldMagikarp (plus, prompt generation)}},
  year = {2023}
}

@article{russakovsky2015imagenet,
  author = {Russakovsky, Olga and Deng, Jia and Su, Hao and Krause, Jonathan and Satheesh, Sanjeev and Ma, Sean and Huang, Zhiheng and Karpathy, Andrej and Khosla, Aditya and Bernstein, Michael and others},
  journal = {International Journal of Computer Vision},
  number = {3},
  pages = {211--252},
  publisher = {Springer},
  title = {{Imagenet large scale visual recognition challenge}},
  volume = {115},
  year = {2015}
}

@inproceedings{sadrtdinov2023stay,
  author = {Sadrtdinov, Ildus and Pozdeev, Dmitrii and Vetrov, Dmitry P and Lobacheva, Ekaterina},
  booktitle = neurips,
  optvolume = {36},
  title = {{To Stay or Not to Stay in the Pre-train Basin: {I}nsights on Ensembling in Transfer Learning}},
  year = {2023}
}

@article{sagawa2019distributionally,
  author = {Sagawa, Shiori and Koh, Pang Wei and Hashimoto, Tatsunori B and Liang, Percy},
  journal = {arXiv preprint arXiv:1911.08731},
  title = {{Distributionally robust neural networks for group shifts: On the importance of regularization for worst-case generalization}},
  year = {2019}
}

@inproceedings{saha2021gradient,
  author = {Saha, Gobinda and Garg, Isha and Roy, Kaushik},
  booktitle = iclr,
  title = {{Gradient Projection Memory for Continual Learning}},
  year = {2021}
}

@inproceedings{saunshi2025reasoning,
  author = {Saunshi, Nikunj and Dikkala, Nishanth and Li, Zhiyuan and Kumar, Sanjiv and J Reddi, Sashank},
  booktitle = {International Conference on Learning Representations},
  pages = {14855--14881},
  title = {{Reasoning with latent thoughts: On the power of looped transformers}},
  volume = {2025},
  year = {2025}
}

@inproceedings{saxe2014exact,
  author = {Saxe, A and McClelland, J and Ganguli, S},
  booktitle = iclr,
  optorganization = {International Conference on Learning Represenatations 2014},
  title = {{Exact solutions to the nonlinear dynamics of learning in deep linear neural networks}},
  year = {2014}
}

@inproceedings{schotthofer2022low,
  author = {Schotth{\"o}fer, Steffen and Zangrando, Emanuele and Kusch, Jonas and Ceruti, Gianluca and Tudisco, Francesco},
  booktitle = neurips,
  optpages = {20051--20063},
  optvolume = {35},
  title = {{Low-rank lottery tickets: Finding efficient low-rank neural networks via matrix differential equations}},
  year = {2022}
}

@inproceedings{sedghi2018singular,
  author = {Sedghi, Hanie and Gupta, Vineet and Long, Philip M},
  booktitle = {International Conference on Learning Representations},
  title = {{The Singular Values of Convolutional Layers}},
  year = {2018}
}

@inproceedings{seide2011feature,
  author = {Seide, Frank and Li, Gang and Chen, Xie and Yu, Dong},
  booktitle = {2011 IEEE Workshop on Automatic Speech Recognition \& Understanding},
  organization = {IEEE},
  pages = {24--29},
  title = {{Feature engineering in context-dependent deep neural networks for conversational speech transcription}},
  year = {2011}
}

@article{sharma2023truth,
  author = {Sharma, Pratyusha and Ash, Jordan T and Misra, Dipendra},
  journal = {arXiv preprint arXiv:2312.13558},
  title = {{The truth is in there: Improving reasoning in language models with layer-selective rank reduction}},
  year = {2023}
}

@article{shazeer2017outrageously,
  author = {Shazeer, Noam and Mirhoseini, Azalia and Maziarz, Krzysztof and Davis, Andy and Le, Quoc and Hinton, Geoffrey and Dean, Jeff},
  journal = {arXiv preprint arXiv:1701.06538},
  title = {{Outrageously large neural networks: The sparsely-gated mixture-of-experts layer}},
  year = {2017}
}

@article{shenouda2023vector,
  author = {Shenouda, Joseph and Parhi, Rahul and Lee, Kangwook and Nowak, Robert D},
  journal = {arXiv preprint arXiv:2305.16534},
  title = {{Vector-valued variation spaces and width bounds for {DNNs}: Insights on weight decay regularization}},
  year = {2023}
}

@article{silver2017mastering,
  author = {Silver, David and Schrittwieser, Julian and Simonyan, Karen and Antonoglou, Ioannis and Huang, Aja and Guez, Arthur and Hubert, Thomas and Baker, Lucas and Lai, Matthew and Bolton, Adrian and others},
  journal = {nature},
  number = {7676},
  pages = {354--359},
  publisher = {Nature Publishing Group UK London},
  title = {{Mastering the game of go without human knowledge}},
  volume = {550},
  year = {2017}
}

@article{silver2018general,
  author = {Silver, David and Hubert, Thomas and Schrittwieser, Julian and Antonoglou, Ioannis and Lai, Matthew and Guez, Arthur and Lanctot, Marc and Sifre, Laurent and Kumaran, Dharshan and Graepel, Thore and others},
  journal = {Science},
  number = {6419},
  pages = {1140--1144},
  publisher = {American Association for the Advancement of Science},
  title = {{A general reinforcement learning algorithm that masters chess, shogi, and Go through self-play}},
  volume = {362},
  year = {2018}
}

@article{simonyan2014very,
  author = {Simonyan, Karen and Zisserman, Andrew},
  journal = {arXiv preprint arXiv:1409.1556},
  title = {{Very deep convolutional networks for large-scale image recognition}},
  year = {2014}
}

@inproceedings{simsek2021geometry,
  author = {Simsek, Berfin and Ged, Fran{\c{c}}ois and Jacot, Arthur and Spadaro, Francesco and Hongler, Cl{\'e}ment and Gerstner, Wulfram and Brea, Johanni},
  booktitle = icml,
  optorganization = {PMLR},
  optpages = {9722--9732},
  title = {{Geometry of the loss landscape in overparameterized neural networks: Symmetries and invariances}},
  year = {2021}
}

@inproceedings{sohl2015deep,
  author = {Sohl-Dickstein, Jascha and Weiss, Eric and Maheswaranathan, Niru and Ganguli, Surya},
  booktitle = icml,
  optorganization = {PMLR},
  optpages = {2256--2265},
  title = {{Deep unsupervised learning using nonequilibrium thermodynamics}},
  year = {2015}
}

@article{sontag1989backpropagation,
  author = {Sontag, Eduardo D and Sussmann, H{\'e}ctor J},
  journal = {Complex Systems},
  number = {1},
  pages = {91--106},
  title = {{Backpropagation can give rise to spurious local minima even for networks without hidden layers}},
  volume = {3},
  year = {1989}
}

@inproceedings{sun2017revisiting,
  author = {Sun, Chen and Shrivastava, Abhinav and Singh, Saurabh and Gupta, Abhinav},
  booktitle = {2017 IEEE International Conference on Computer Vision (ICCV)},
  organization = {IEEE},
  pages = {843--852},
  title = {{Revisiting unreasonable effectiveness of data in deep learning era}},
  year = {2017}
}

@misc{sutton2019bitter,
  author = {Sutton, Rich},
  howpublished = {\url{http://www.incompleteideas.net/IncIdeas/BitterLesson.html}},
  month = {March},
  note = {Incomplete Ideas (Blog)},
  title = {{The Bitter Lesson}},
  year = {2019}
}

@article{szegedy2013intriguing,
  author = {Szegedy, Christian and Zaremba, Wojciech and Sutskever, Ilya and Bruna, Joan and Erhan, Dumitru and Goodfellow, Ian and Fergus, Rob},
  journal = {arXiv:1312.6199},
  title = {{Intriguing properties of neural networks}},
  year = {2013}
}

@article{tan2023understanding,
  author = {Tan, Zhiquan and Huang, Weiran},
  journal = {arXiv preprint arXiv:2311.06597},
  title = {{Understanding Grokking Through A Robustness Viewpoint}},
  year = {2023}
}

@article{thilak2022slingshot,
  author = {Thilak, Vimal and Littwin, Etai and Zhai, Shuangfei and Saremi, Omid and Paiss, Roni and Susskind, Joshua},
  journal = {arXiv preprint arXiv:2206.04817},
  title = {{The slingshot mechanism: An empirical study of adaptive optimizers and the grokking phenomenon}},
  year = {2022}
}

@inproceedings{timor2023implicit,
  author = {Timor, Nadav and Vardi, Gal and Shamir, Ohad},
  booktitle = {Proceedings of the International Conference on Algorithmic Learning Theory (ALT)},
  optorganization = {PMLR},
  optpages = {1429--1459},
  title = {{Implicit regularization towards rank minimization in {ReLU} networks}},
  year = {2023}
}

@misc{tunguz2026twelvexbet,
  author = {Tunguz, Tomasz},
  day = {17},
  howpublished = {Personal Blog},
  month = {March},
  title = {{The 12x Bet on AI}},
  url = {https://tomtunguz.com/12x-bet-on-ai/},
  year = {2026}
}

@article{van2017l2,
  author = {Van Laarhoven, Twan},
  journal = {arXiv preprint arXiv:1706.05350},
  title = {{L2 regularization versus batch and weight normalization}},
  year = {2017}
}

@book{vandenberghe2004convex,
  author = {Vandenberghe, Lieven},
  title = {{Convex optimization}},
  publisher = {Cambridge University Press},
  volume = {1},
  year = {2004}
}

@inproceedings{vardi2021implicit,
  author = {Vardi, Gal and Shamir, Ohad},
  booktitle = colt,
  optorganization = {PMLR},
  optpages = {4224--4258},
  title = {{Implicit regularization in relu networks with the square loss}},
  year = {2021}
}

@inproceedings{vaswani2017attention,
  author = {Vaswani, Ashish and Shazeer, Noam and Parmar, Niki and Uszkoreit, Jakob and Jones, Llion and Gomez, Aidan N and Kaiser, {\L}ukasz and Polosukhin, Illia},
  booktitle = neurips,
  optvolume = {30},
  title = {{Attention is all you need}},
  year = {2017}
}

@inproceedings{villalobos2024position,
  author = {Villalobos, Pablo and Ho, Anson and Sevilla, Jaime and Besiroglu, Tamay and Heim, Lennart and Hobbhahn, Marius},
  booktitle = {Forty-first International Conference on Machine Learning},
  title = {{Position: Will we run out of data? Limits of LLM scaling based on human-generated data}},
  year = {2024}
}

@article{wang2018dataset,
  author = {Wang, Tongzhou and Zhu, Jun-Yan and Torralba, Antonio and Efros, Alexei A},
  journal = {arXiv preprint arXiv:1811.10959},
  title = {{Dataset distillation}},
  year = {2018}
}

@misc{wang2020ml,
  author = {Wang, Binxu and Vastola, John},
  title = {{{ML} from scratch: {S}table Diffusion, Day 2}},
  url = {https://colab.research.google.com/drive/1Y5wr91g5jmpCDiX-RLfWL1eSBWoSuLqO?usp=sharing#scrollTo=9is-DXZYwIIi},
  year = {2022}
}

@article{wang2021pufferfish,
  author = {Wang, Hongyi and Agarwal, Saurabh and Papailiopoulos, Dimitris},
  journal = {Proceedings of Machine Learning and Systems},
  pages = {365--386},
  title = {{Pufferfish: Communication-efficient models at no extra cost}},
  volume = {3},
  year = {2021}
}

@article{wang2023implicit,
  author = {Wang, Zihan and Jacot, Arthur},
  journal = {arXiv preprint arXiv:2305.16038},
  title = {{Implicit bias of {SGD} in $ L\_ $\{$2$\}$ $-regularized linear {DNNs}: {O}ne-way jumps from high to low rank}},
  year = {2023}
}

@article{wang2026beyond,
  author = {Wang, Shenzhi and Yu, Le and Gao, Chang and Zheng, Chujie and Liu, Shixuan and Lu, Rui and Dang, Kai and Chen, Xiong-Hui and Yang, Jianxin and Zhang, Zhenru and others},
  journal = {Advances in Neural Information Processing Systems},
  pages = {115452--115486},
  title = {{Beyond the 80/20 rule: High-entropy minority tokens drive effective reinforcement learning for llm reasoning}},
  volume = {38},
  year = {2026}
}

@article{Waskom2021,
  author = {Michael L. Waskom},
  journal = {Journal of Open Source Software},
  number = {60},
  optdoi = {10.21105/joss.03021},
  opturl = {https://doi.org/10.21105/joss.03021},
  pages = {3021},
  publisher = {The Open Journal},
  title = {{seaborn: statistical data visualization}},
  volume = {6},
  year = {2021}
}

@inproceedings{woodworth2020kernel,
  author = {Woodworth, Blake and Gunasekar, Suriya and Lee, Jason D and Moroshko, Edward and Savarese, Pedro and Golan, Itay and Soudry, Daniel and Srebro, Nathan},
  booktitle = {Conference on Learning Theory},
  organization = {PMLR},
  pages = {3635--3673},
  title = {{Kernel and rich regimes in overparametrized models}},
  year = {2020}
}

@article{wortsman2023stable,
  author = {Wortsman, Mitchell and Dettmers, Tim and Zettlemoyer, Luke and Morcos, Ari and Farhadi, Ali and Schmidt, Ludwig},
  journal = neurips,
  pages = {10271--10298},
  title = {{Stable and low-precision training for large-scale vision-language models}},
  volume = {36},
  year = {2023}
}

@inproceedings{xu2023benign,
  author = {Xu, Zhiwei and Wang, Yutong and Frei, Spencer and Vardi, Gal and Hu, Wei},
  booktitle = iclr,
  title = {{Benign Overfitting and Grokking in {ReLU} Networks for {XOR} Cluster Data}},
  year = {2023}
}

@inproceedings{yaras2023invariant,
  author = {Yaras, Can and Wang, Peng and Hu, Wei and Zhu, Zhihui and Balzano, Laura and Qu, Qing},
  booktitle = {NeurIPS 2023 Workshop on Mathematics of Modern Machine Learning},
  title = {{Invariant Low-Dimensional Subspaces in Gradient Descent for Learning Deep Matrix Factorizations}},
  year = {2023}
}

@article{yaras2023law,
  author = {Yaras, Can and Wang, Peng and Hu, Wei and Zhu, Zhihui and Balzano, Laura and Qu, Qing},
  journal = {arXiv preprint arXiv:2306.01154},
  title = {{The Law of Parsimony in Gradient Descent for Learning Deep Linear Networks}},
  year = {2023}
}

@inproceedings{yu2017compressing,
  author = {Yu, Xiyu and Liu, Tongliang and Wang, Xinchao and Tao, Dacheng},
  booktitle = cvpr,
  optpages = {7370--7379},
  title = {{On compressing deep models by low rank and sparse decomposition}},
  year = {2017}
}

@inproceedings{yu2023compressing,
  author = {Yu, Hao and Wu, Jianxin},
  booktitle = aaai,
  title = {{Compressing Transformers: Features Are Low-Rank, but Weights Are Not!}},
  year = {2023}
}

@inproceedings{yunis2022convexity,
  author = {Yunis, David and Patel, Kumar Kshitij and Savarese, Pedro Henrique Pamplona and Vardi, Gal and Frankle, Jonathan and Walter, Matthew and Livescu, Karen and Maire, Michael},
  booktitle = {OPT 2022: Optimization for Machine Learning (NeurIPS 2022 Workshop)},
  title = {{On Convexity and Linear Mode Connectivity in Neural Networks}},
  year = {2022}
}

@article{zangrando2024neural,
  author = {Zangrando, Emanuele and Deidda, Piero and Brugiapaglia, Simone and Guglielmi, Nicola and Tudisco, Francesco},
  journal = {arXiv preprint arXiv:2402.03991},
  title = {{Neural Rank Collapse: Weight Decay and Small Within-Class Variability Yield Low-Rank Bias}},
  year = {2024}
}

@inproceedings{zhang2018fixup,
  author = {Zhang, Hongyi and Dauphin, Yann N and Ma, Tengyu},
  booktitle = iclr,
  title = {{Fixup Initialization: Residual Learning Without Normalization}},
  year = {2018}
}

@article{zhang2018three,
  author = {Zhang, Guodong and Wang, Chaoqi and Xu, Bowen and Grosse, Roger},
  journal = {arXiv preprint arXiv:1810.12281},
  title = {{Three mechanisms of weight decay regularization}},
  year = {2018}
}

@inproceedings{zhang2025memory,
  author = {Zhang, Jianyu and Nolte, Niklas and Sadhukhan, Ranajoy and Chen, Beidi and Bottou, L{\'e}on},
  booktitle = {International Conference on Learning Representations},
  pages = {36412--36433},
  title = {{Memory mosaics}},
  volume = {2025},
  year = {2025}
}

@article{zhou2022mixture,
  author = {Zhou, Yanqi and Lei, Tao and Liu, Hanxiao and Du, Nan and Huang, Yanping and Zhao, Vincent and Dai, Andrew M and Le, Quoc V and Laudon, James and others},
  journal = {Advances in Neural Information Processing Systems},
  pages = {7103--7114},
  title = {{Mixture-of-experts with expert choice routing}},
  volume = {35},
  year = {2022}
}

@inproceedings{ziyin2022exact,
  author = {Ziyin, Liu and Li, Botao and Meng, Xiangming},
  booktitle = neurips,
  optpages = {24446--24458},
  optvolume = {35},
  title = {{Exact solutions of a deep linear network}},
  year = {2022}
}

@article{zou2023universal,
  author = {Zou, Andy and Wang, Zifan and Kolter, J Zico and Fredrikson, Matt},
  journal = {arXiv preprint arXiv:2307.15043},
  title = {{Universal and transferable adversarial attacks on aligned language models}},
  year = {2023}
}

@misc{openai2022chatgpt,
  author = {{OpenAI}},
  title = {ChatGPT},
  year = {2022},
  url = {https://chatgpt.com},
  note = {Large language model; launched November 30, 2022}
}

@misc{anthropic2025claudecode,
  author = {{Anthropic}},
  title = {Claude Code},
  year = {2025},
  url = {https://claude.ai},
  note = {Agentic command-line coding tool released in research preview February 2025 and generally available May 2025}
}

@string{aistats="Proceedings of the International Conference on Artificial Intelligence and Statistics (AISTATS)"}

@string{tmlr="Transactions on Machine Learning Research"}

@string{colt="Proceedings of the Annual Conference on Learning Theory (COLT)"}

@string{facct="{Proceedings of the ACM Conference on Fairness, Accountability, and Transparency (FAccT)},"}
\bibliographystyle{plainnat}

\clearpage
\appendix

\chapter{Convex Mode Connectivity}
\section{Task Details}\label{app:hyps}

\begin{figure}[!t]
    \centering
    \begin{subfigure}[b]{0.24\textwidth}
        \centering
        \includegraphics[width=\textwidth]{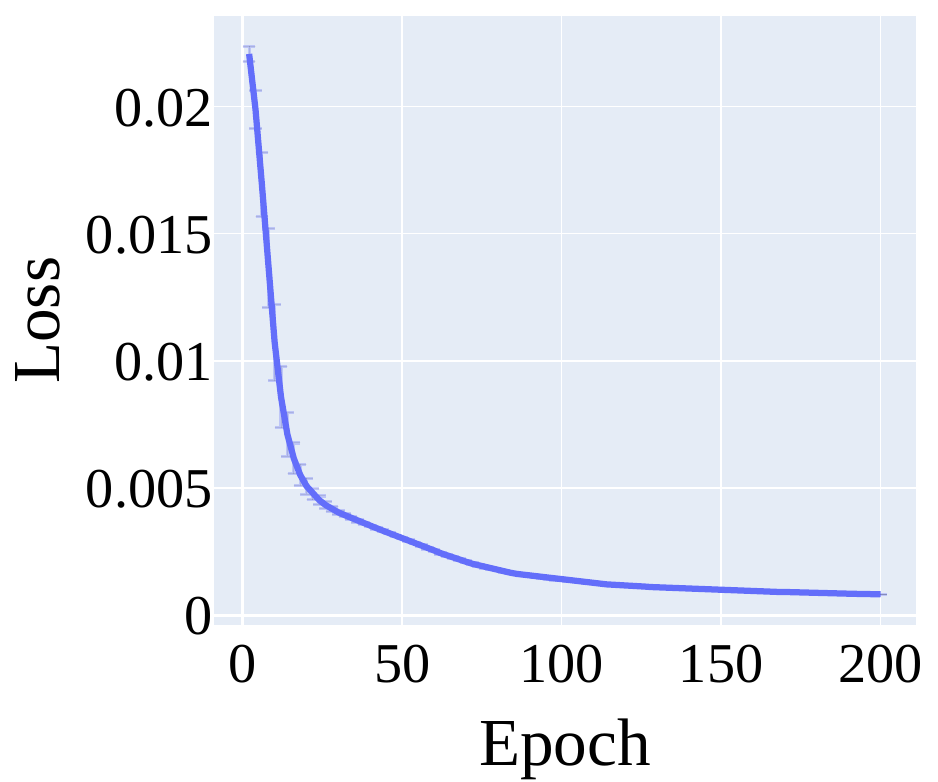}
        \caption{SIREN}
    \end{subfigure}\hfil
    \begin{subfigure}[b]{0.24\textwidth}
        \centering
        \includegraphics[width=\textwidth]{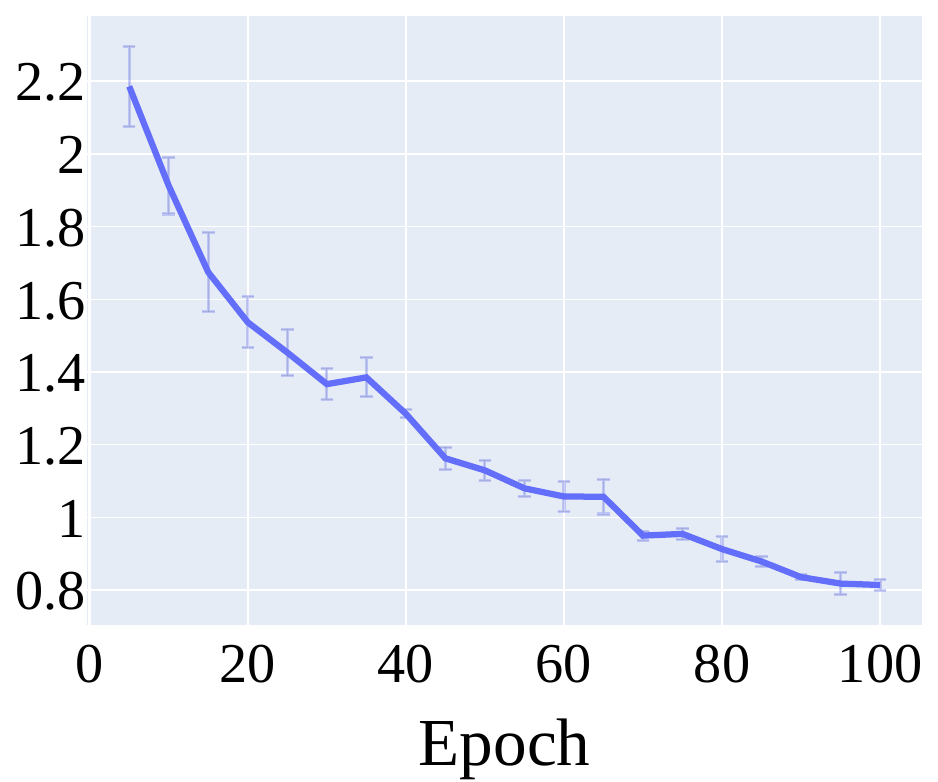}
        \caption{LeNet}
    \end{subfigure}\hfil
    \begin{subfigure}[b]{0.24\textwidth}
        \centering
        \includegraphics[width=\textwidth]{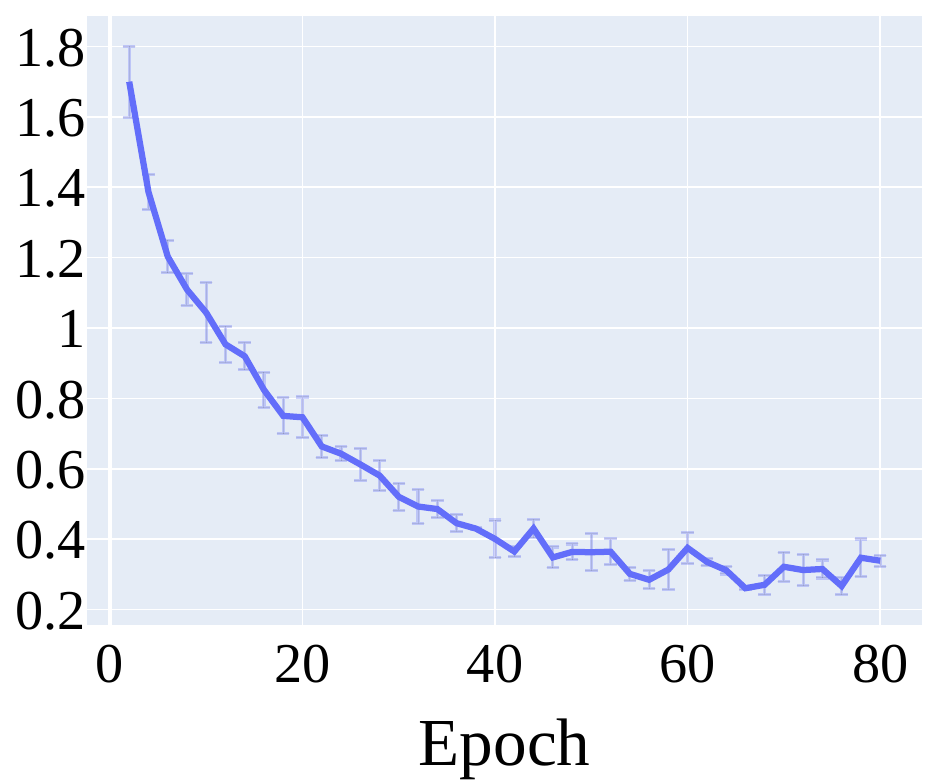}
        \caption{LeNet-m}
    \end{subfigure}\hfil
    \begin{subfigure}[b]{0.24\textwidth}
        \centering
        \includegraphics[width=\textwidth]{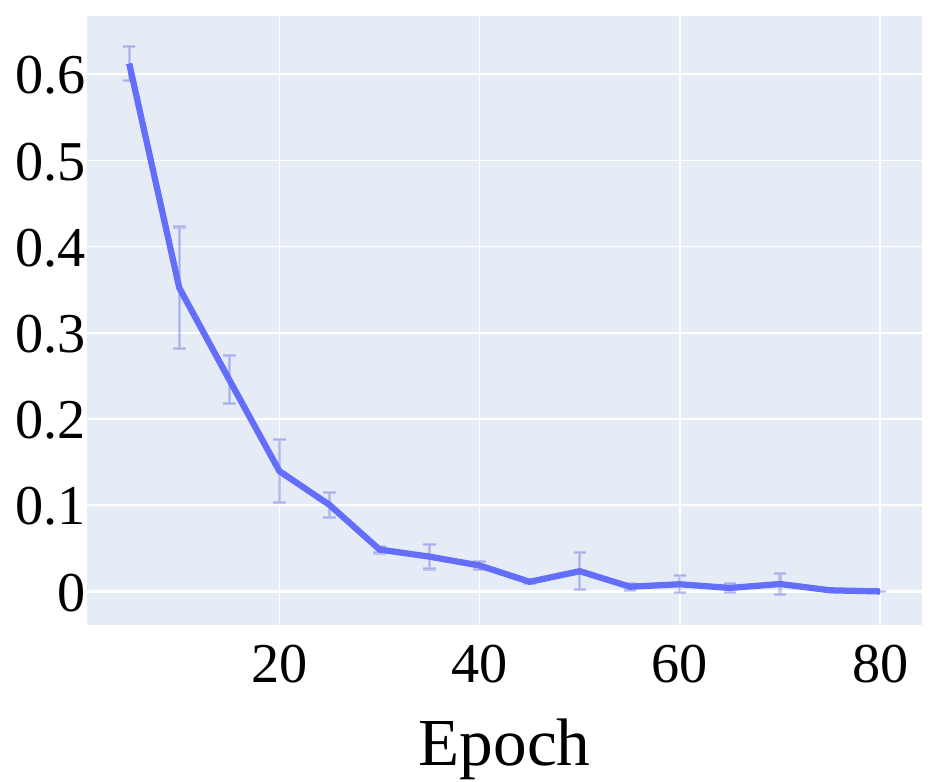}
        \caption{ResNet}
    \end{subfigure}\hfil
    \caption{Training loss curves for all tasks. When cross-referencing with Figure~\ref{fig:cmc_plot}, we see that the first split point that leads to a convex hull is quite early, long before the loss has converged, and often before it has even halved.}
    \label{fig:loss}
\end{figure}

SIREN \citep{sitzmann2020implicit} is a fully-connected network with sine activations. It was originally designed to overcome the low-frequency bias in radiance fields. Here it is tuned to reconstruct an audio waveform, where the input is the timestep, and the output is the waveform value. We choose to study it because all prior experiments on LMC were conducted with ReLU activations, and we wished to observe a system that was not piece-wise linear.  For our other models, we use LeNet-5~\citep{lecun1998gradient}, and ResNet-20~\citep{he2016deep} trained on CIFAR-10~\citep{krizhevsky2009learning}. These combinations are some of the simplest systems for which we do not see LMC at initialization \citep{nagarajan2019uniform, frankle2020linear, ainsworth2022git}. For the sake of simplicity, all learning rates are constant in our experiments. Because the SIREN task does not include generalization, we stop training when the training loss has converged. In the case of our other models, we tune hyper-parameters and stop training when the validation error has converged, which is typically the desired criterion in practice. For hyperparameters, see Table~\ref{tab:tasks}. For full training loss curves, see Figure~\ref{fig:loss}.

\begin{table}[h]
    \centering
    \caption{Summary of tasks studied in this chapter.}
    {\footnotesize
    \begin{tabular}{ccccccc}
        \toprule
        \textbf{Abbrev.} & \textbf{Task} & \textbf{Dataset} & \textbf{Model} & \textbf{Optimizer} & \textbf{LR} & \textbf{Batch size} \\
        \midrule
        SIREN & Regress.\ & Bach Waveform & 5-layer SIREN & Adam & 1e-5 & 8192 \\
        \hline
        LeNet & Classif.\ & CIFAR-10 & LeNet-5 & SGD & 1e-2 & 128 \\
        LeNet-m & Classif.\ & CIFAR-10 & LeNet-5 & SGD+mom. (0.9) \ & 1e-2 & 128 \\
        \hline
        ResNet & Classif.\ & CIFAR-10 & ResNet-20 & SGD+mom. (0.9) \ & 0.1 & 128 \\
        \hline
    \end{tabular}}
    \label{tab:tasks}
\end{table}

\section{Metric Calculation Details}\label{app:calc}

In order to calculate Hessian eigenvalues, we use a power iteration method with Hessian vector products, like in \citet{li2018visualizing}. To measure local convexity between a pair of stochastic updates from parameters $W$, we compute two stochastic updates $\Delta W_1$ and $\Delta W_2$, and measure the gap
\[ g = \mathcal{L} \left( \frac{1}{2}(W + \Delta W_1) + \frac{1}{2}(W + \Delta W_2) \right) - \left(\frac{1}{2}\mathcal{L}(W + \Delta W_1) + \frac{1}{2}\mathcal{L}(W + \Delta W_2)\right).\]
We then compute the probability that this gap is negative over many samples. This gives us a metric for the non-convexity that a single update induces, which we can measure at different points throughout training.



\chapter{Spectral Dynamics}
\section{Explanation of Balancedness}\label{sec:balancedness}

Prior work on deep linear networks~\citep{arora2019implicit, milanesi2021implicit} suggests that rank minimization may describe implicit regularization in deep matrix factorization better than simple matrix norms. See \cite{arora2018optimization} (Appendix~A) for a detailed argument. However, a critical assumption used in these works is ``balanced initialization.'' This means that for consecutive matrices $W_i$ and $W_{i+1}$ in the product matrix $\prod_j W_j$, we have $W^\top_{i+1}W_{i+1} = W_iW_i^\top$ at initialization. Decomposing these matrices with SVDs and leveraging orthogonality, this simplifies to $V_{i+1}\Sigma_{i+1}^2V_{i+1}^\top = U_i\Sigma_i^2U_i^\top$ where $U_i$ and $V_{i+1}$ are orthogonal matrices. Since these are orthogonal decompositions of the same matrix, their diagonals must be equivalent, allowing for the permutation of elements with the same value. This leads to $U_i = V_{i+1} O$ up to signs, where $O$ is a block diagonal permutation matrix that may permute the rows of equivalent diagonal elements. Notably, if all diagonal elements are distinct and $U_i$ and $V_{i+1}$ are square matrices, then $U_i = V_{i+1}$ up to signs. This gives us matching singular vectors for consecutive matrices.

\section{Experimental Details}\label{app:experimental-details}

For all experiments, we use 3 random seeds and average all plots over those 3. This is relatively small, but error bars tend to be very tight, and due to the high volume of runs required for this work we lack the resources to run much more.

In order to compute alignment we consider only pairs of consecutive layers that directly feed into each other, and ignore the influence of residual connections so as to cut down on the number of comparisons. Specifics on individual architectures are given below.

\subsection{Image Classification with VGG}\label{app:imgclass-details}

We train a VGG-16~\citep{simonyan2014very} on CIFAR-10~\citep{krizhevsky2009learning} for 164 epochs, following the hyperparameters and learning rate schedule in~\citep{frankle2020linear}, but without data augmentation as it contributes extra randomness. For the optimizer we use SGD with batch size 128, initial learning rate 0.1 and momentum of 0.9. We also decay the learning rate 3 times by a factor of 10 at epoch 82, epoch 120, and finally at epoch 160. We also use a minor amount of weight decay with coefficient 0.0001.

VGG-16 uses ReLU activations and batch normalization~\citep{ioffe2015batch}, and includes both convolutional and linear layers. For linear layers we simply compute the SVD of the weight matrix. For convolutional layers, the parameters are typically stored as a 4D tensor of shape $(c_\text{out}, c_\text{in}, h, w)$ for the output channels, input channels, height and width of the filters respectively. As the filters compute a transformation from each position and input channel to an output channel, we compute the SVD of the flattened tensor of shape $(c_\text{out}, c_\text{in} \cdot h \cdot w)$, which maps all inputs to outputs, similar to \citet{praggastis2022svd}. This is not the SVD of the entire transformation of the feature map to the next feature map~\cite{sedghi2018singular}, but rather the transformation from a set of adjacent positions to a particular position in the next layer computed by the parameters. For the individual SV evolution plot, we use the 12th convolutional layer.

In order to compute alignment of bases between consecutive convolutional layers, $V_{i+1}^\top U_i$ we need to match the dimensionality between $U_i$ and $V_{i+1}$. For convolutional layers we are presented with a question as to how to handle the spatial dimensions $h$ and $w$ as naively the input dimension of the next layer will be a factor of $h\cdot w$ larger dimension. We experimented with multiple cases, including aligning at each spatial position individually or averaging over the alignment at all spatial positions, and eventually settled at aligning the output of one layer to the center spatial input of the next layer. That is, for a 3x3 convolution mapping to a following 3x3 convolution, we compute the alignment only for position (1,1) of the next layer. This seemed reasonable to us as on average the edges of the filters showed poorer alignment overall. For the individual alignment plot, we use the alignment between the 11th and 12th convolutional layers at the center spatial position of the 12th convolutional layer.

\subsection{Image Generation with UNets}\label{app:imggen-details}

We train a UNet~\citep{ronneberger2015u} diffusion model~\citep{sohl2015deep, ho2020denoising} on MNIST~\citep{lecun1998mnist} generation. We take model design and hyperparameters from~\citep{wang2020ml}. In particular we use a 4-layer residual UNet and train with AdamW~\citep{loshchilov2018fixing} with batch size 128, and learning rate of 0.0003 for 100 epochs. This model uses swish~\citep{ramachandran2017searching} activations and a combination of linear and convolutional, as well as transposed convolutional layers.

Computing SVDs and alignment is similar to the image classification case described above, except in the case of the transposed convolutions where an extra transpose of dimensions is needed as parameters are stored with the shape $(c_\text{in}, c_\text{out}, h, w)$. For the individual SV evolution plot, we use the 3rd convolutional layer. For the alignment plot, we use the alignment between the 3rd and 4th convolutional layers at the center spatial position of the 4th convolutional layer.

\subsection{Speech Recognition with LSTMs}\label{app:speech-details}

We train a bidirectional LSTM~\citep{hochreiter1997long} for automatic speech recognition on LibriSpeech~\citep{panayotov2015librispeech}. We tune for a simple and well-performing hyperparameter setting. We use AdamW~\citep{loshchilov2018fixing} with batch size 32, learning rate 0.0003 and weight decay 0.1 for 50 epochs. We also use a cosine annealing learning rate schedule from 1 to 0 over the entire 50 epochs.

The LSTM only has matrix parameters and biases, so it is straightforward to compute SVDs of the matrices. For individual SV evolution plots, we plot the 3rd layer input parameter. In the case of alignment, we make a number of connections: first down depth for the input parameters, then connecting the previous input parameter to the current hidden parameter in both directions, then connecting the previous hidden parameter to the current input parameter. In particular the LSTM parameters are stored as a stack of 4 matrices in PyTorch, and we find alignment is highest for the "gate" submatrix, so we choose that for all plots. For the individual layer alignment, we plot alignment between the 3rd and 4th layer input parameters.

\subsection{Language Modeling with Transformers}\label{app:language-details}

We train a Transformer~\citep{vaswani2017attention} language model on Wikitext-103~\citep{merity2016pointer}. We base hyperparameter choices on the Pythia suite~\citep{biderman2023pythia}, specifically the 160 million parameter configuration with sinusoidal position embeddings, 12 layers, model dimension 768, 12 attention heads per layer, and hidden dimension 768. We use AdamW~\citep{loshchilov2018fixing} with batch size 256, learning rate 0.0006 and weight decay 0.1. We use a context length of 2048 and clip gradients to a maximum norm of 1. We also use a learning rate schedule with a linear warmup and cosine decay to 10\% of the learning rate, like \citet{biderman2023pythia}.

For SVDs, for simplicity we take the SVD of the entire $(3d_\text{model}, d_\text{model})$ parameter that computes queries, keys and values from the hidden dimension inside the attention layer, without splitting into individual heads. This is reasonable as the splitting is done after the fact internally. We also take the SVD of the output parameters, and linear layers of the MLPs, which are 2 dimensional matrices. For the individual SV evolution plot, we plot the SVs of $W_1$ of the 8th layer MLP.

For alignment, we consider the alignment of $W_Q$ and $W_K$ matrices, $W_V$ and $W_O$ matrices, computing alignment between heads individually then averaging over all heads. We also consider the alignment between $W_O$ and $W_1$ of the MLP block, between $W_1$ and $W_2$ of the MLP block, and between $W_2$ and the next attention layer. For the individual layer alignment, we plot alignment between $W_1$ and $W_2$ of the 8th layer MLP.

\subsection{Spectral Dynamics with Scale (Pythia)}\label{sec:pythia}

\begin{figure*}[h!]
  \centering
  \begin{subfigure}[b]{0.19\linewidth}
    \centering
    \includegraphics[width=\linewidth]{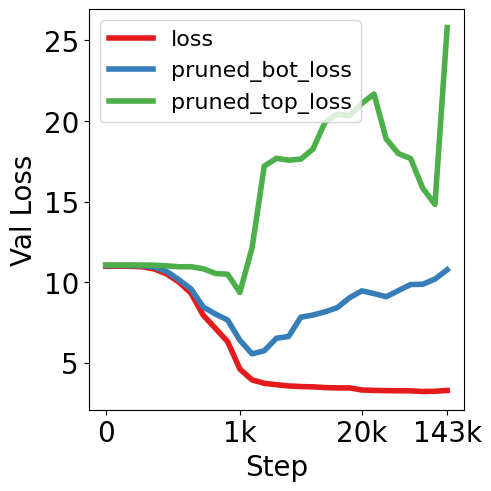}
    \\
    \includegraphics[width=\linewidth]{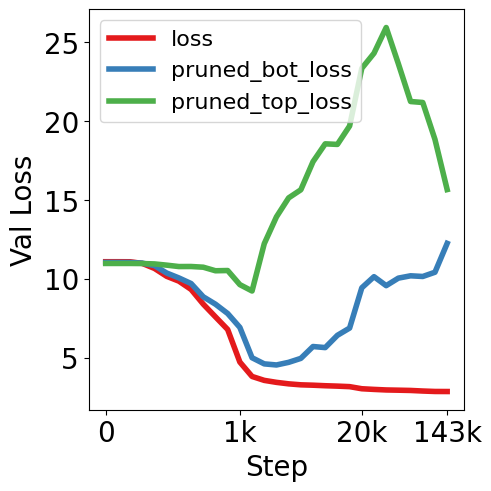}
    \\
    \includegraphics[width=\linewidth]{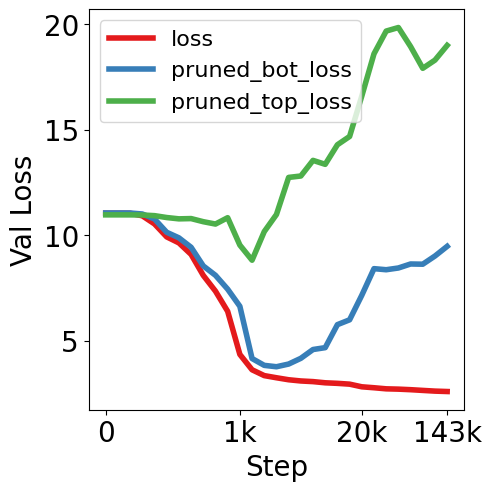}
    \\
    \includegraphics[width=\linewidth]{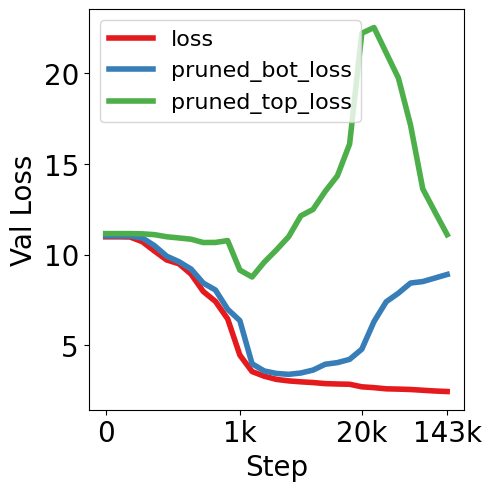}
    \caption{Val. Loss}
  \end{subfigure}
  \begin{subfigure}[b]{0.19\linewidth}
    \centering
    \includegraphics[width=\linewidth]{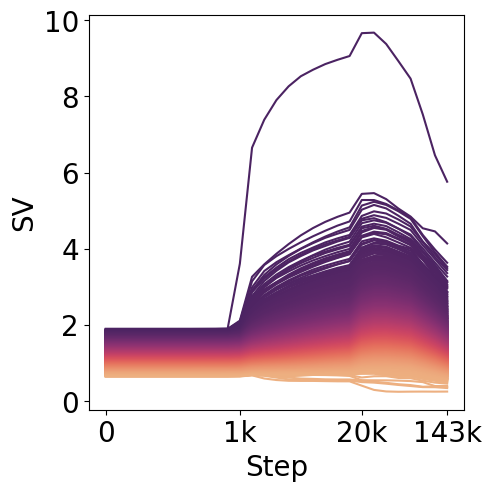}
    \\
    \includegraphics[width=\linewidth]{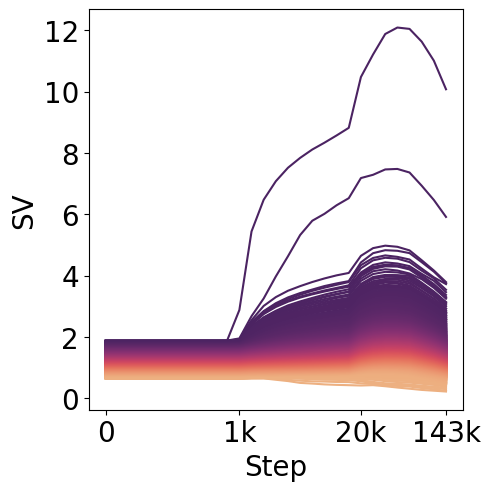}
    \\
    \includegraphics[width=\linewidth]{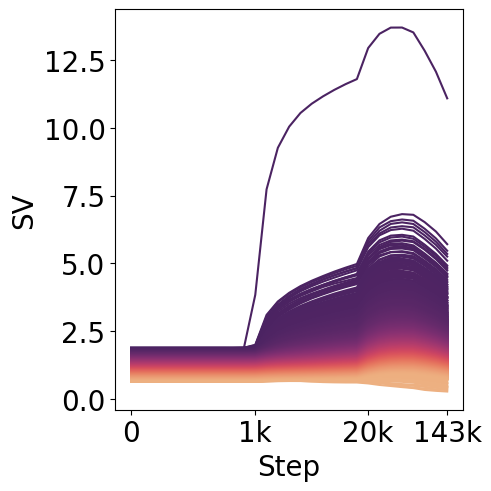}
    \\
    \includegraphics[width=\linewidth]{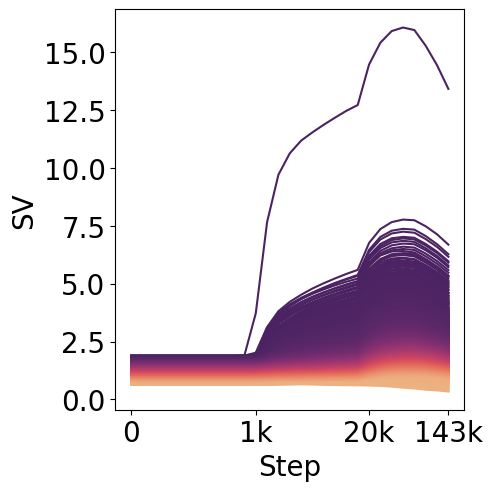}
    \caption{SVs}
  \end{subfigure}
  \begin{subfigure}[b]{0.19\linewidth}
    \centering
    \includegraphics[width=\linewidth]{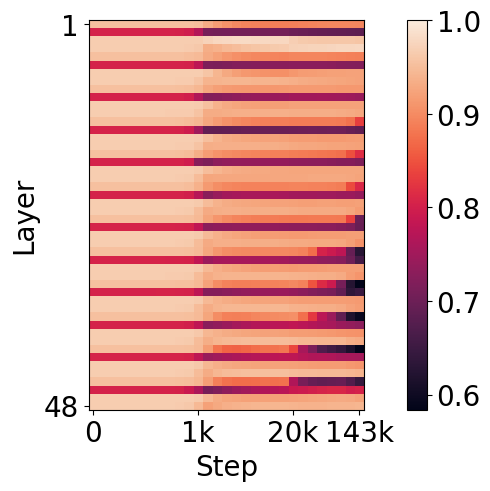}
    \\
    \includegraphics[width=\linewidth]{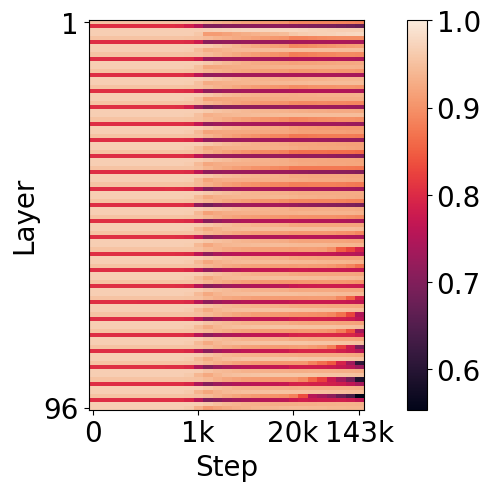}
    \\
    \includegraphics[width=\linewidth]{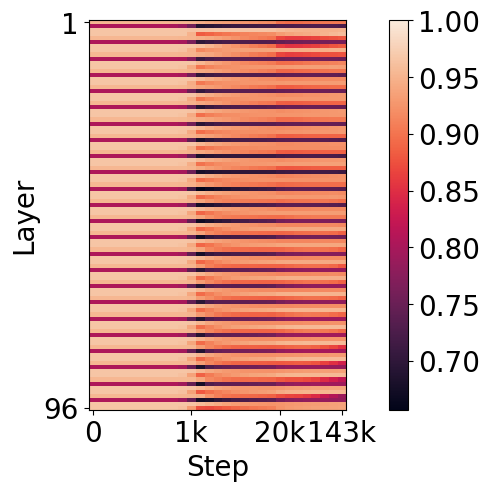}
    \\
    \includegraphics[width=\linewidth]{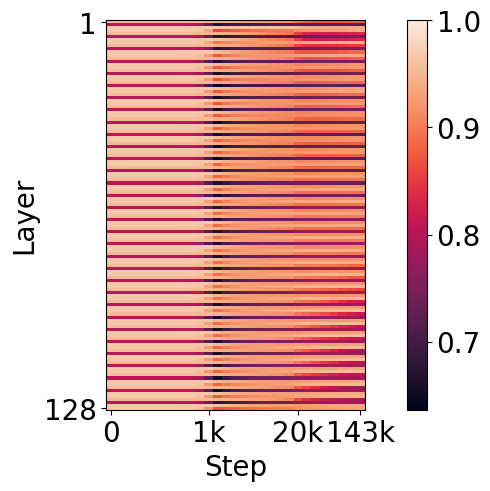}
    \caption{Eff. Rank}
  \end{subfigure}
  \begin{subfigure}[b]{0.19\linewidth}
    \centering
    \includegraphics[width=\linewidth]{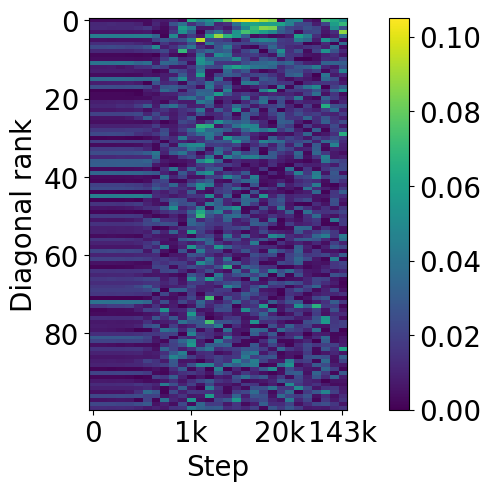}
    \\
    \includegraphics[width=\linewidth]{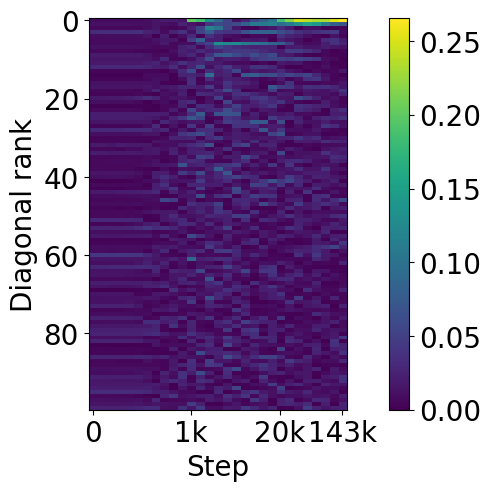}
    \\
    \includegraphics[width=\linewidth]{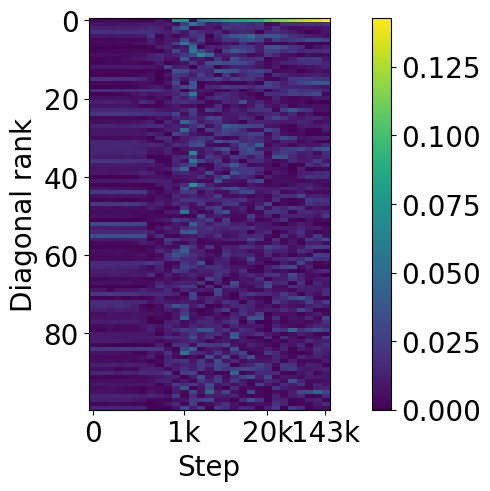}
    \\
    \includegraphics[width=\linewidth]{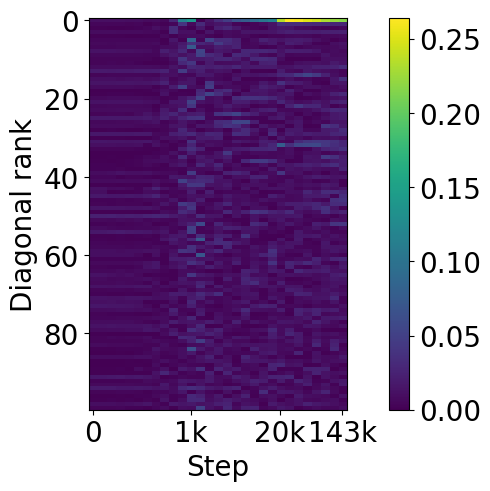}
    \caption{Alignment}
  \end{subfigure}
  \begin{subfigure}[b]{0.19\linewidth}
    \centering
    \includegraphics[width=\linewidth]{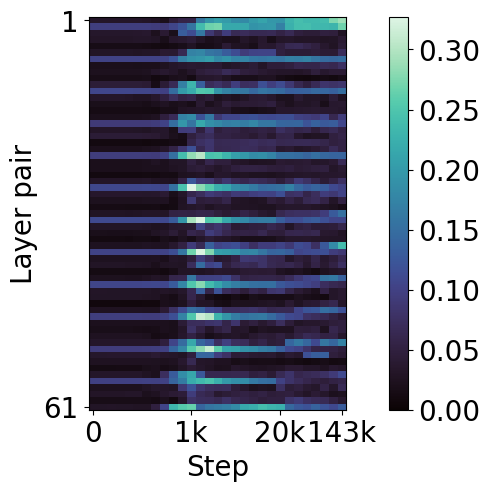}
    \\
    \includegraphics[width=\linewidth]{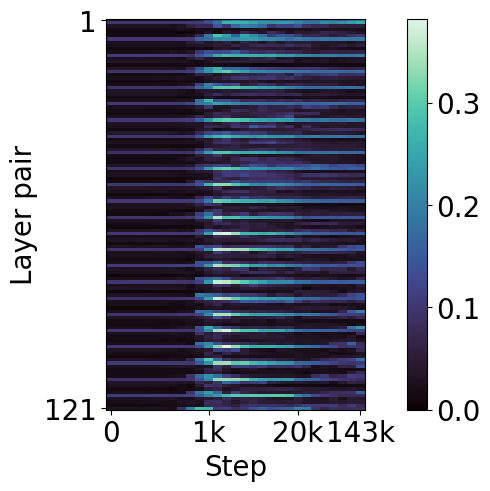}
    \\
    \includegraphics[width=\linewidth]{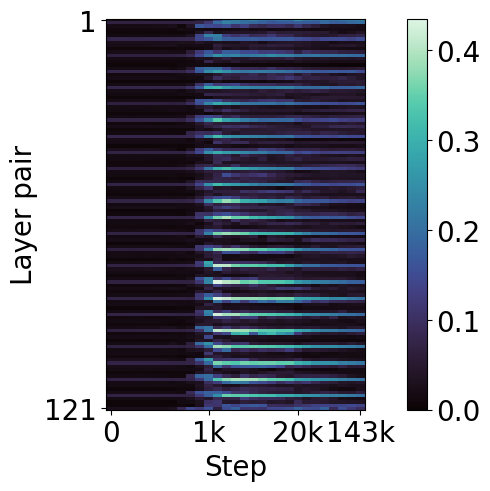}
    \\
    \includegraphics[width=\linewidth]{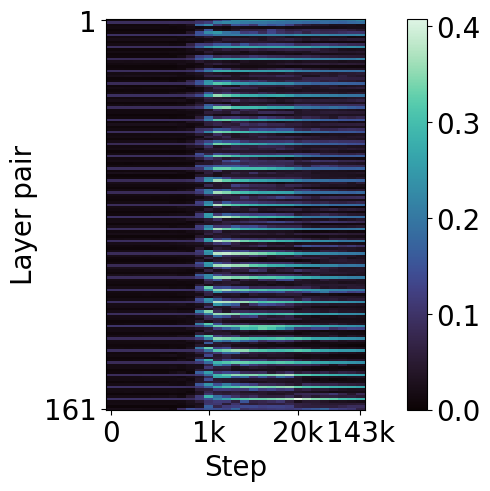}
    \caption{Align. Score}
  \end{subfigure}
  \caption{Spectral dynamics of Pythia suite. From top to bottom we examine the 160m, 410m, 1.4b and 2.8b parameter models. Notably, much less noise appears in the alignment plot with increasing scale. Presumably this could be due to the fact that larger dimensional vectors have higher probability to be orthogonal, which may play a role in making optimization easier. We see stronger alignment score (Eqn.~\ref{eqn:alignment-measure}) in all layers in the larger model, perhaps because of that cleaner signal.}
  \label{fig:pythia}
\end{figure*}

Here we apply the perspective developed in Section~\ref{sec:spectral_dynamics} to larger scale models. As we lack the resources to train these models ourselves, we leverage the Pythia~\citep{biderman2023pythia} family which provides training trajectories for language models across a range of scales (70m to 12b parameters). We are further constrained to the 2.8b parameter model at the largest due to memory requirements when computing SVDs and alignment.

In Figure~\ref{fig:pythia}, we see similar rank dynamics across a variety of scales. We select the 7th layer MLP to compare between models as it is present at all scales. We do see an unequal evolution in singular values, but also a contraction as training proceeds for longer. The difference between scales is not very obvious, but proportionally fewer of the singular values evolve to be large in the 2.8b model as opposed to the 410m model, which one can see from the thickness of the light magenta color. The lack of alignment except for the top rank is quite consistent with earlier observations, and such alignment happens much later for the largest model.

\subsection{Weight Decay Experiments}

\begin{figure*}[t!]
  \centering
  \begin{subfigure}[b]{0.24\linewidth}
    \centering
    \includegraphics[width=0.5\linewidth]{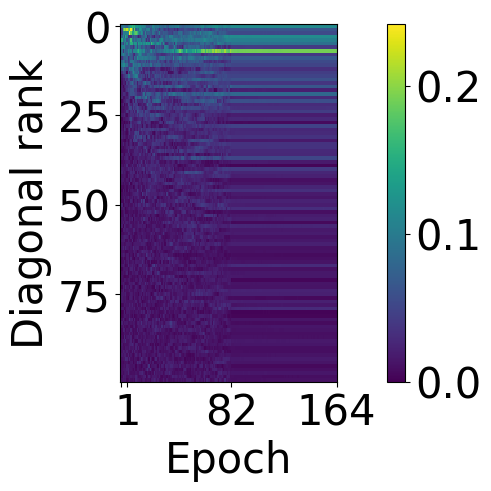}\hfil
    \includegraphics[width=0.5\linewidth]{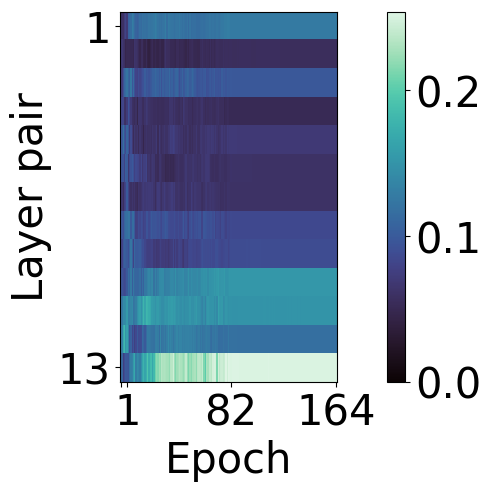}
    \\
    \includegraphics[width=0.5\linewidth]{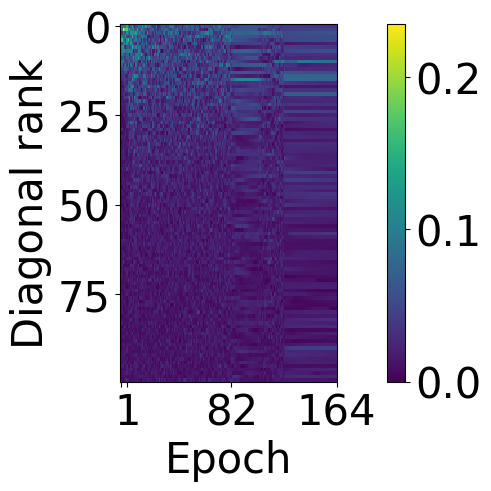}\hfil
    \includegraphics[width=0.5\linewidth]{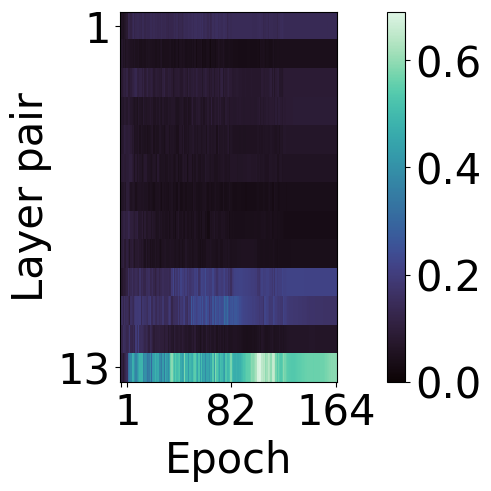}
    \\
    \includegraphics[width=0.5\linewidth]{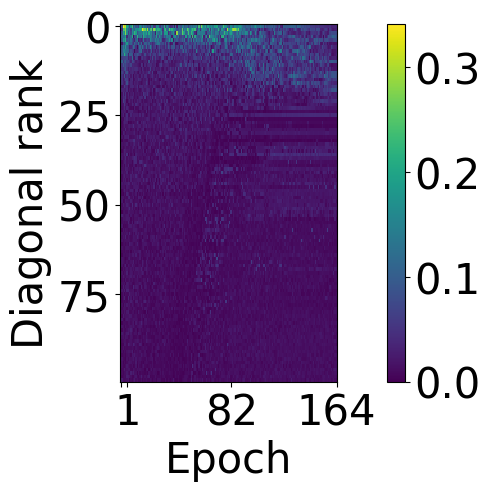}\hfil
    \includegraphics[width=0.5\linewidth]{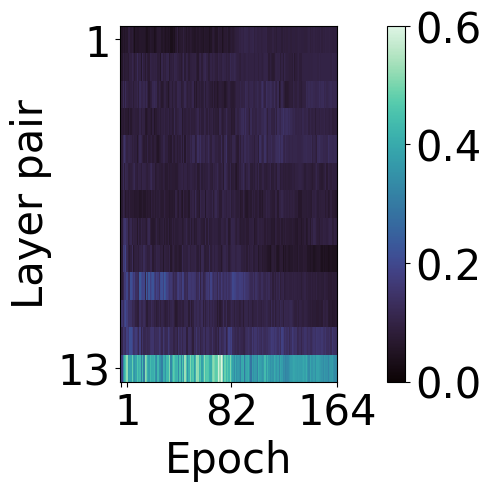}
    \\
    \includegraphics[width=0.5\linewidth]{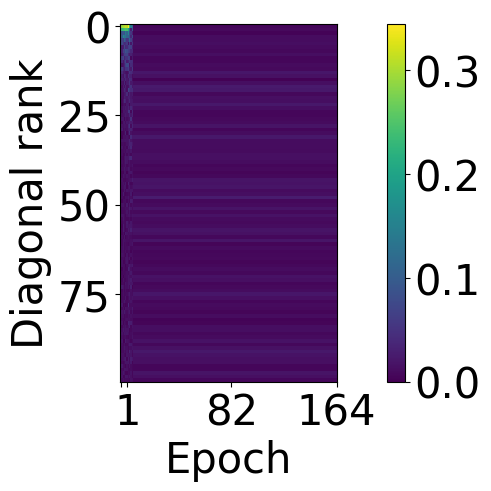}\hfil
    \includegraphics[width=0.5\linewidth]{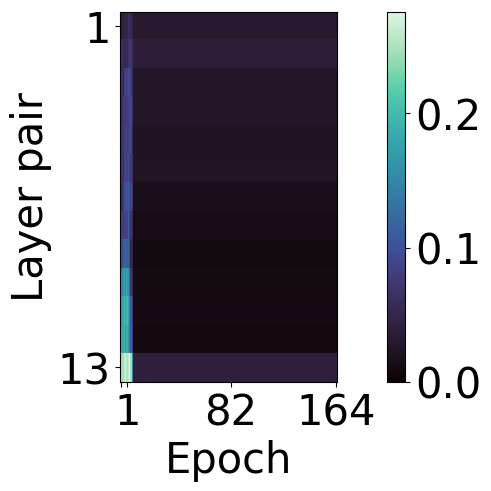}
    \caption{VGG}
  \end{subfigure}
  \begin{subfigure}[b]{0.24\linewidth}
    \centering
    \includegraphics[width=0.5\linewidth]{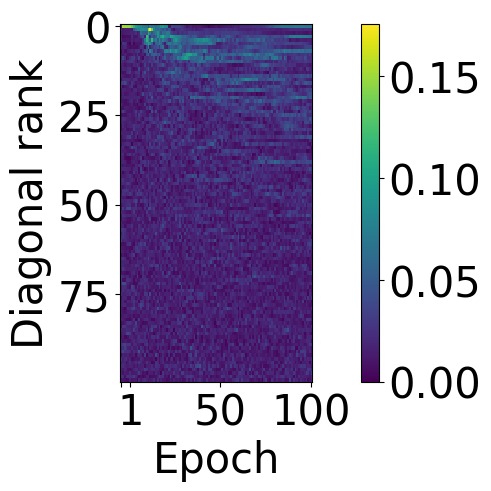}\hfil
    \includegraphics[width=0.5\linewidth]{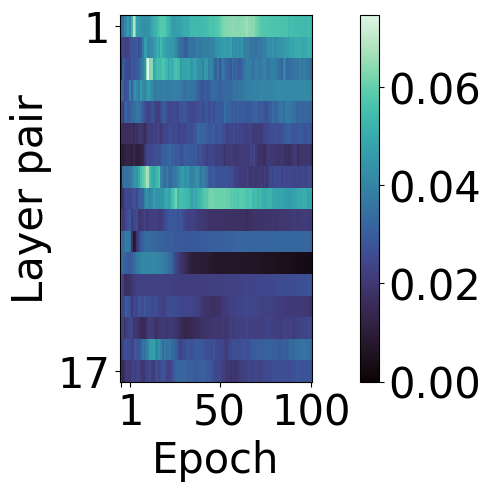}
    \\
    \includegraphics[width=0.5\linewidth]{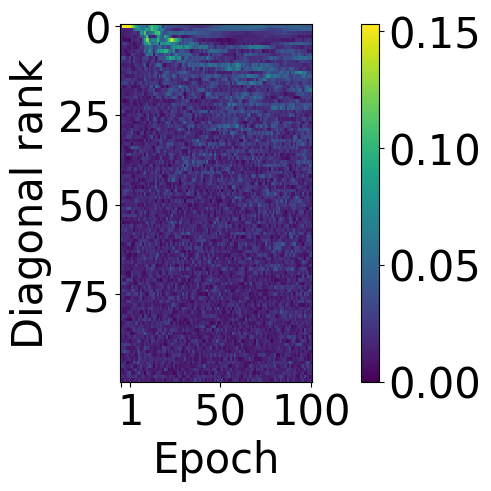}\hfil
    \includegraphics[width=0.5\linewidth]{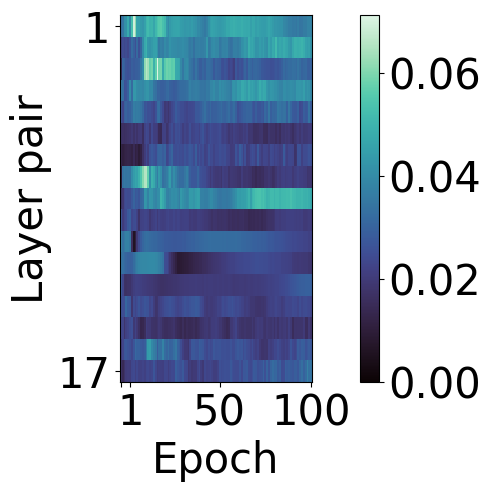}
    \\
    \includegraphics[width=0.5\linewidth]{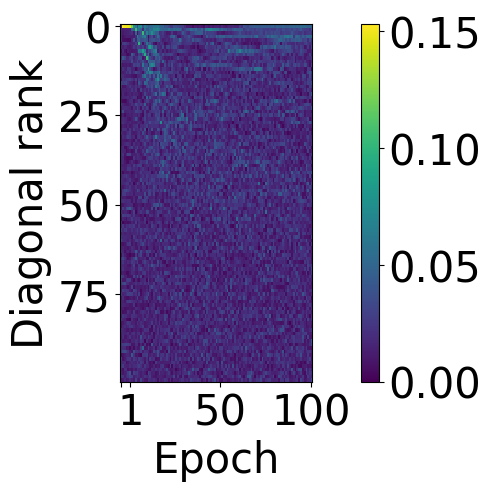}\hfil
    \includegraphics[width=0.5\linewidth]{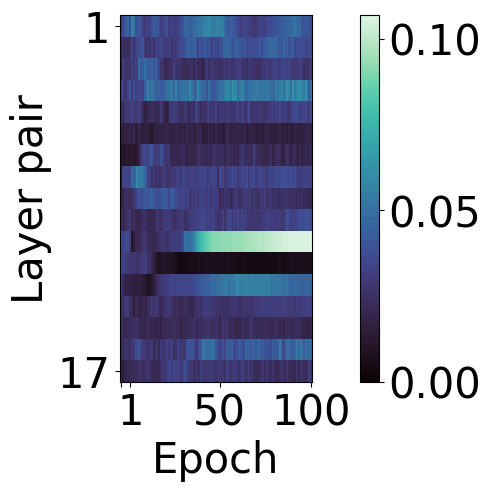}
    \\
    \includegraphics[width=0.5\linewidth]{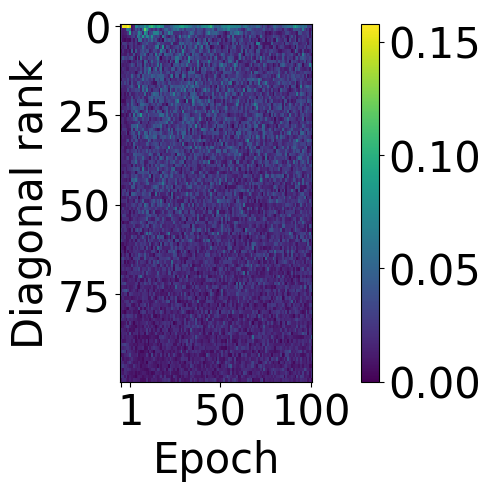}\hfil
    \includegraphics[width=0.5\linewidth]{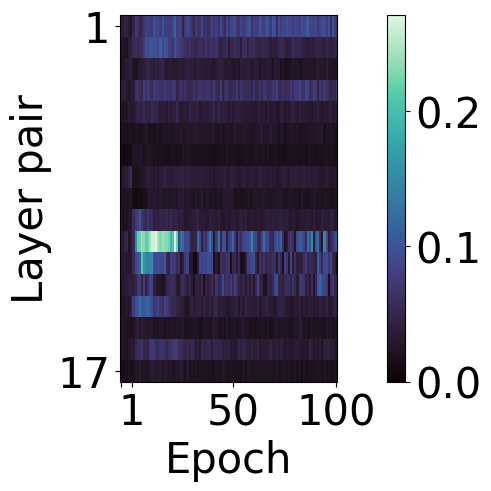}
    \caption{UNet}
  \end{subfigure}
  \begin{subfigure}[b]{0.24\linewidth}
    \centering
    \includegraphics[width=0.5\linewidth]{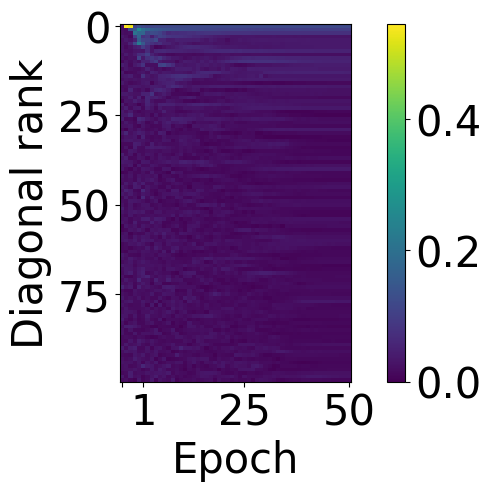}\hfil
    \includegraphics[width=0.5\linewidth]{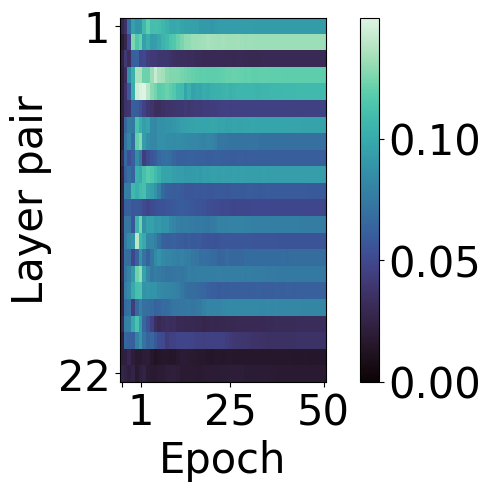}
    \\
    \includegraphics[width=0.5\linewidth]{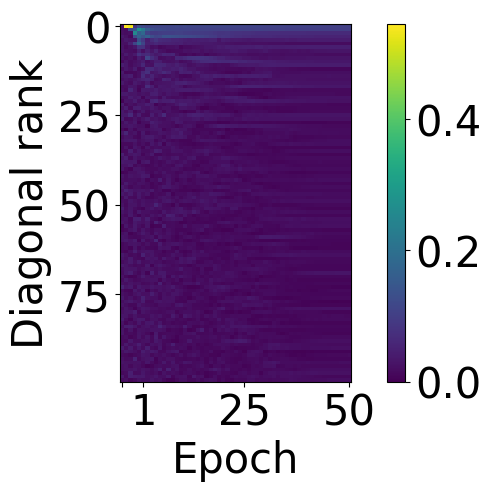}\hfil
    \includegraphics[width=0.5\linewidth]{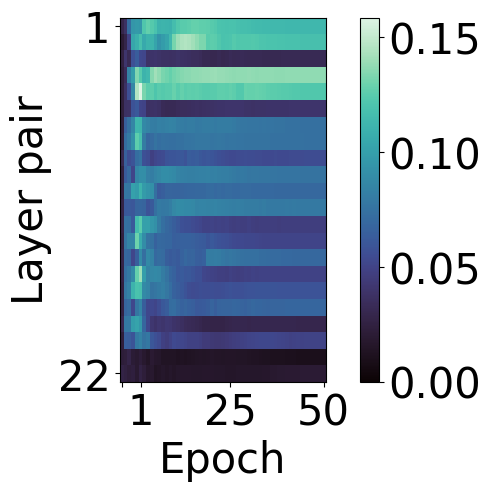}
    \\
    \includegraphics[width=0.5\linewidth]{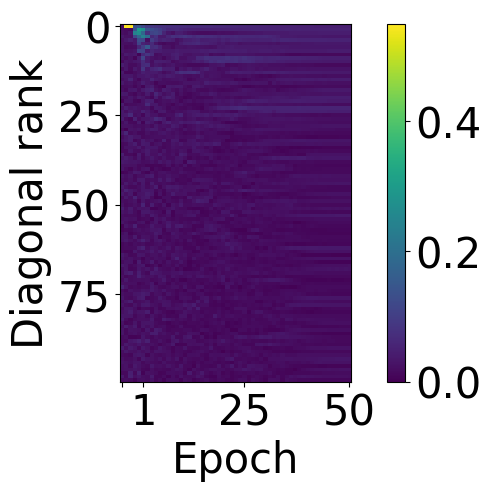}\hfil
    \includegraphics[width=0.5\linewidth]{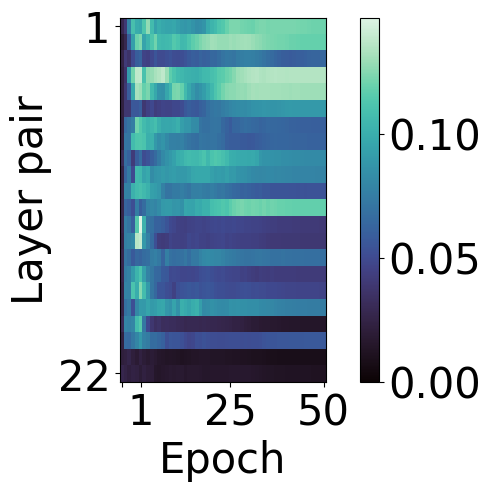}
    \\
    \includegraphics[width=0.5\linewidth]{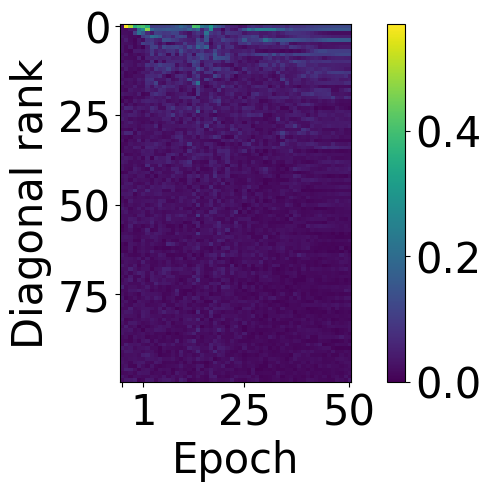}\hfil
    \includegraphics[width=0.5\linewidth]{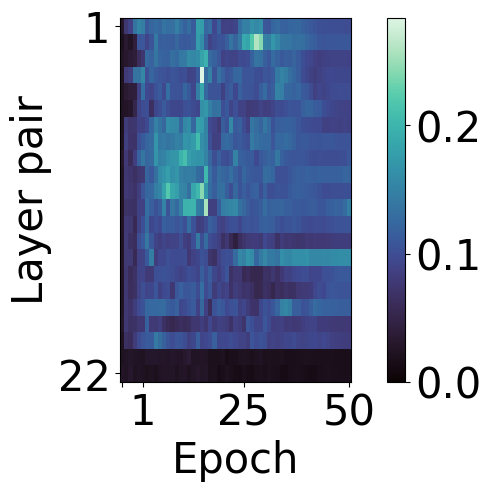}
    \caption{LSTM}
  \end{subfigure}
  \begin{subfigure}[b]{0.24\linewidth}
    \centering
    \includegraphics[width=0.5\linewidth]{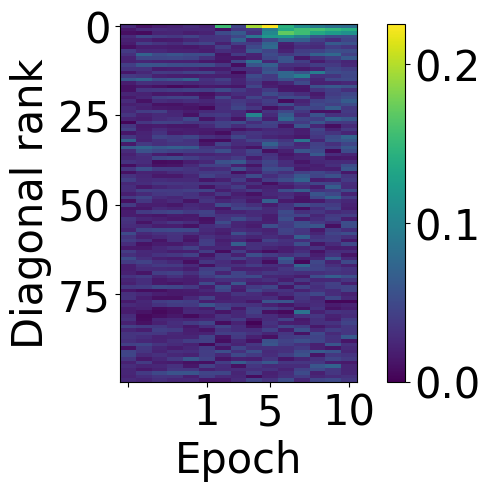}\hfil
    \includegraphics[width=0.5\linewidth]{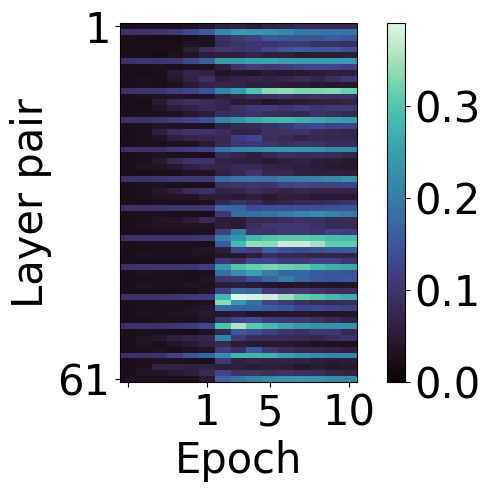}
    \\
    \includegraphics[width=0.5\linewidth]{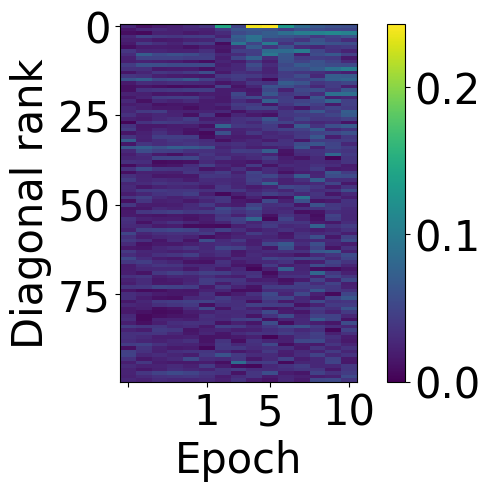}\hfil
    \includegraphics[width=0.5\linewidth]{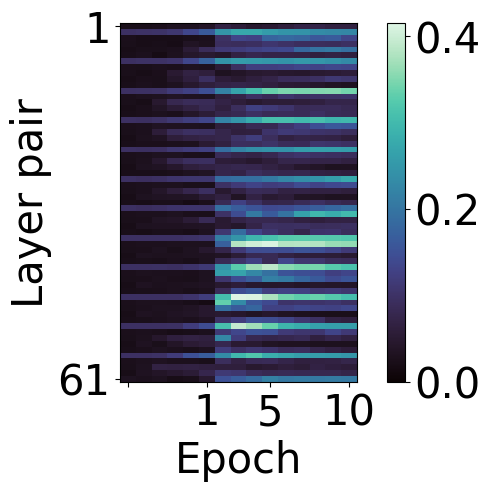}
    \\
    \includegraphics[width=0.5\linewidth]{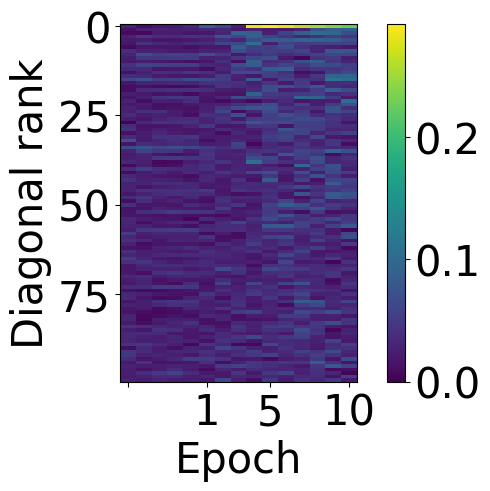}\hfil
    \includegraphics[width=0.5\linewidth]{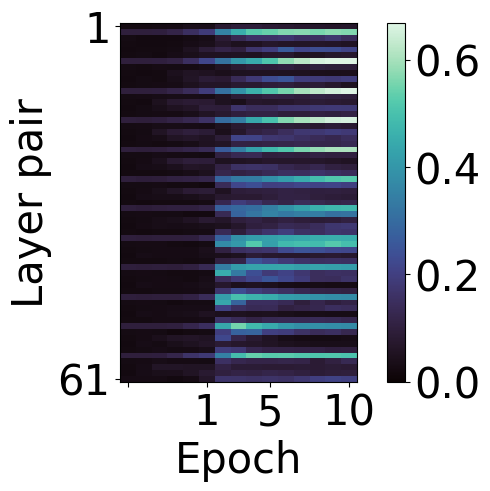}
    \\
    \includegraphics[width=0.5\linewidth]{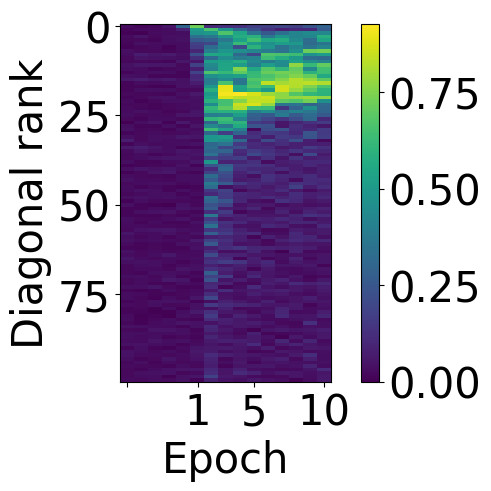}\hfil
    \includegraphics[width=0.5\linewidth]{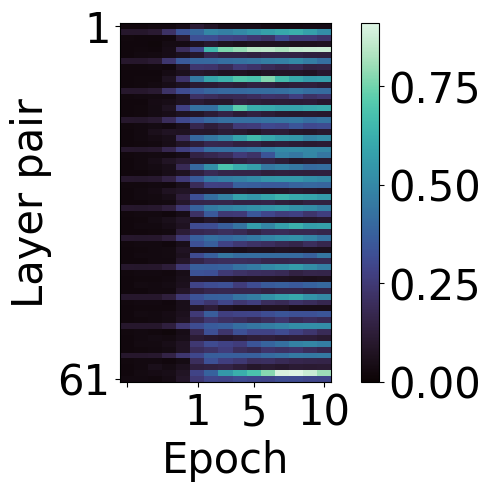}
    \caption{Transformer}
  \end{subfigure}
  \caption{Diagonal of alignment for a single pair over time (Eqn.~\ref{eqn:alignment-matrix}) and alignment metric across pairs of matrices over time (Eqn.~\ref{eqn:alignment-measure}) where the y-axis represents depth.  From top to bottom, for VGG we use coefficients $\{ 0, 0.001, 0.01, 0.1\}$, while for other networks we use coefficients $\{0, 0.1, 1, 10\}$. We see that the maximum alignment magnitude is higher with large weight decay, and in particular, the Transformer has the strongest alignment even when nonlinearities separate the MLP layers.}
  \label{fig:wd-alignment-score}
\end{figure*}

\begin{figure}[!t]
  \centering
  \begin{subfigure}[b]{0.24\linewidth}
    \centering
    \includegraphics[width=\linewidth]{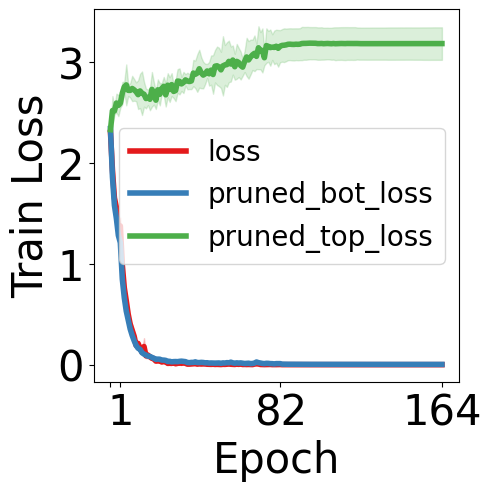}
    \\
    \includegraphics[width=\linewidth]{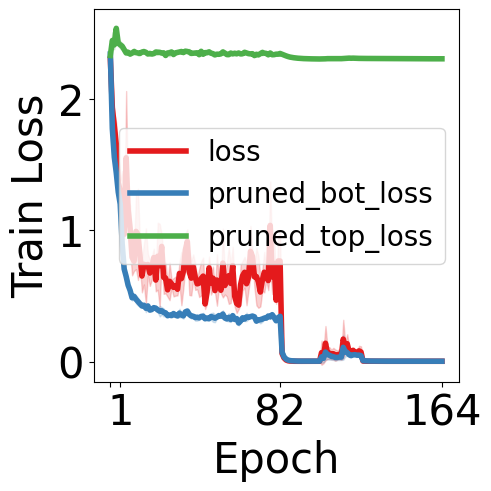}
    \\
    \includegraphics[width=\linewidth]{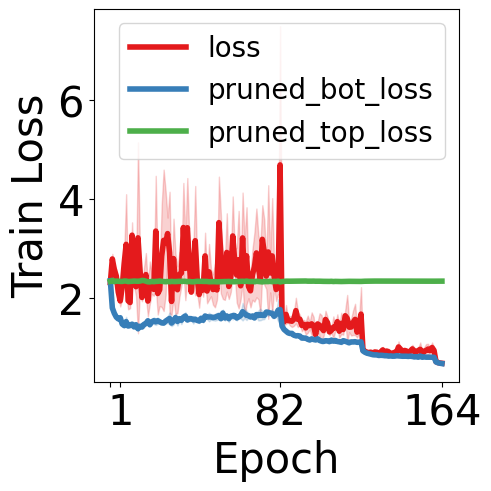}
    \\
    \includegraphics[width=\linewidth]{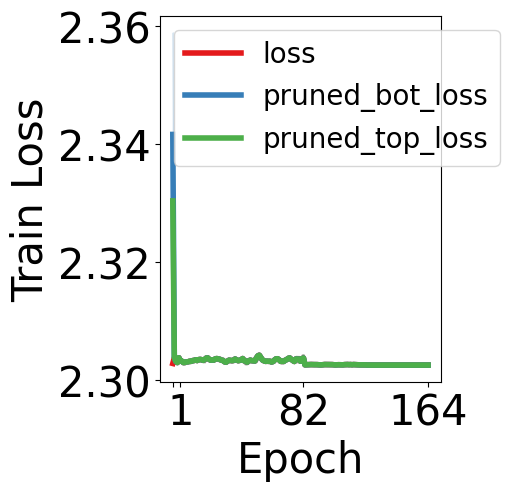}
    \caption{VGG}
  \end{subfigure}
  \begin{subfigure}[b]{0.24\linewidth}
    \centering
    \includegraphics[width=\linewidth]{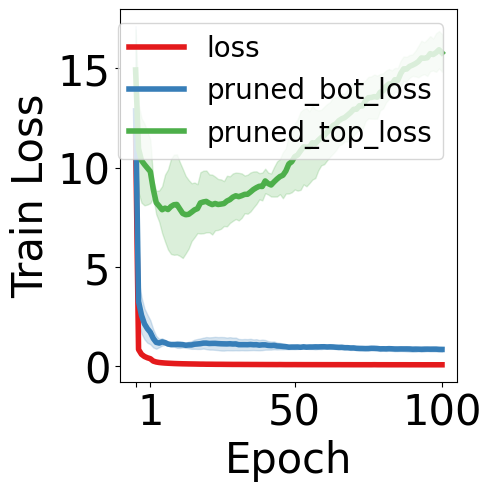}
    \\
    \includegraphics[width=\linewidth]{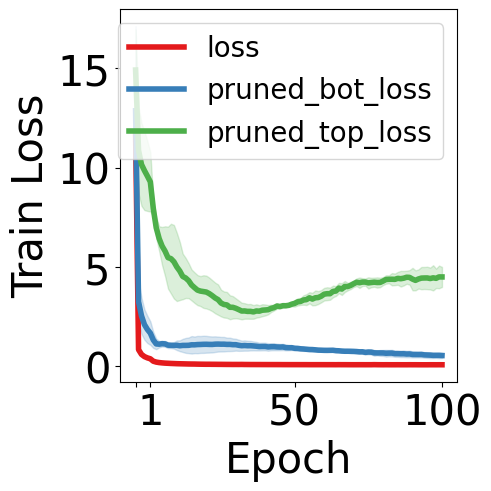}
    \\
    \includegraphics[width=\linewidth]{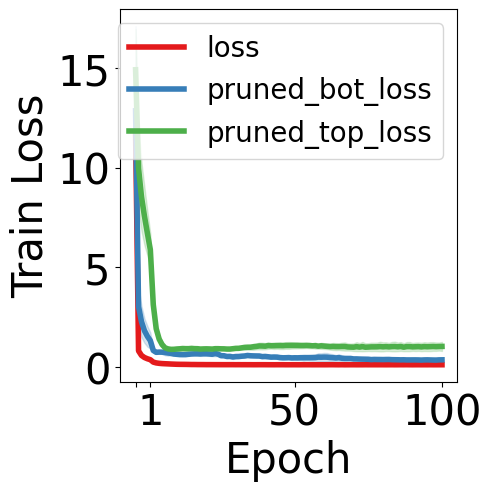}
    \\
    \includegraphics[width=\linewidth]{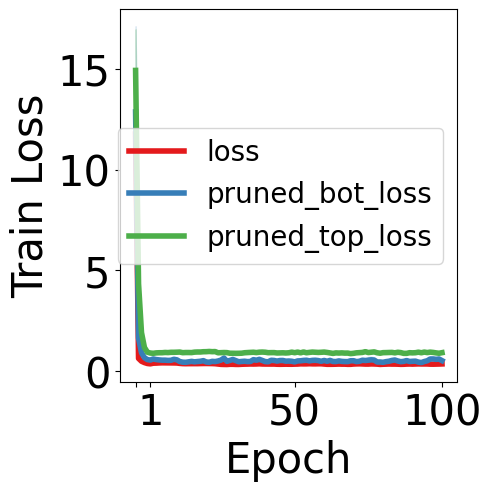}
    \caption{UNet}
  \end{subfigure}
  \begin{subfigure}[b]{0.24\linewidth}
    \centering
    \includegraphics[width=\linewidth]{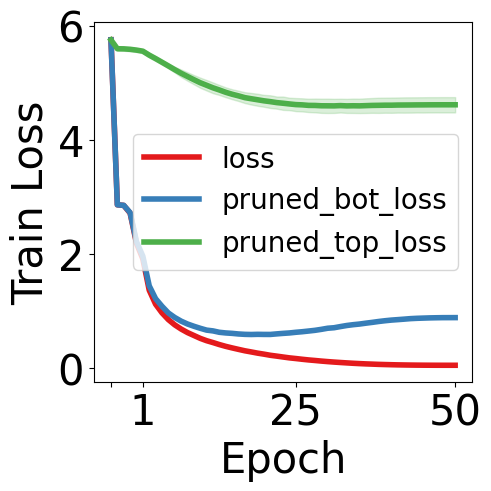}
    \\
    \includegraphics[width=\linewidth]{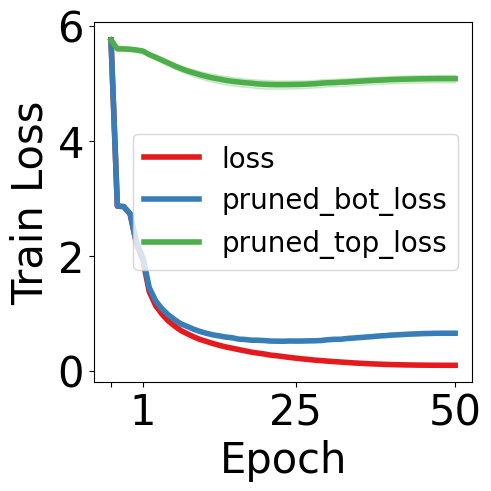}
    \\
    \includegraphics[width=\linewidth]{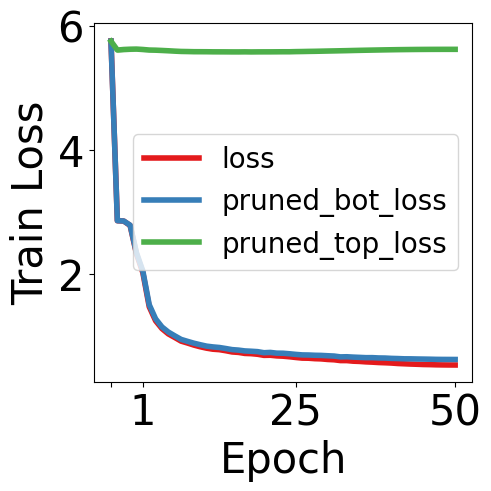}
    \\
    \includegraphics[width=\linewidth]{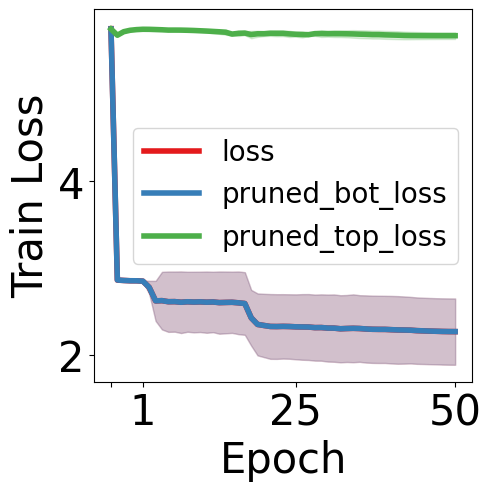}
    \caption{LSTM}
  \end{subfigure}
  \begin{subfigure}[b]{0.24\linewidth}
    \centering
    \includegraphics[width=\linewidth]{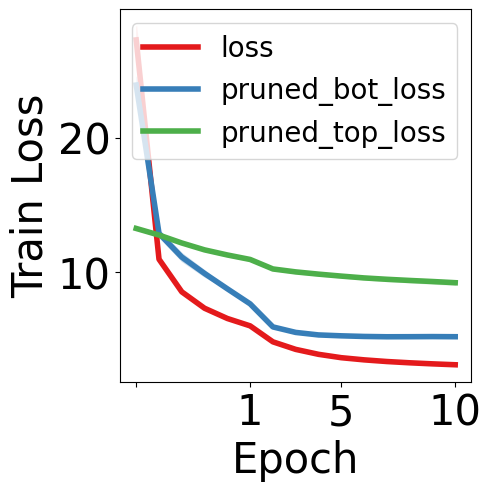}
    \\
    \includegraphics[width=\linewidth]{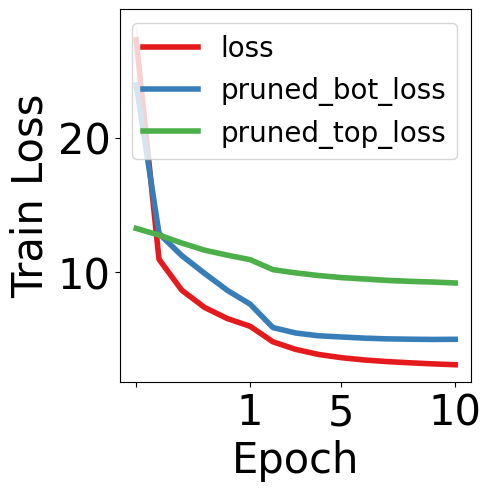}
    \\
    \includegraphics[width=\linewidth]{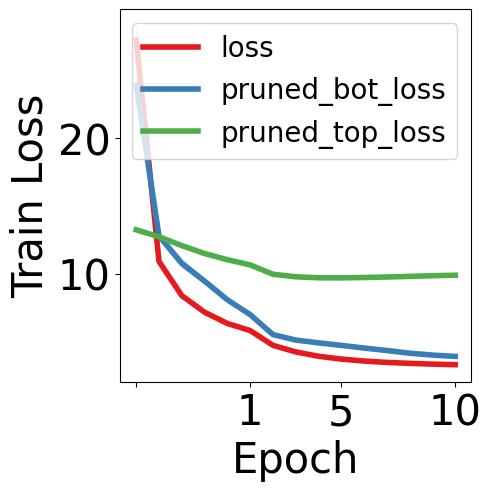}
    \\
    \includegraphics[width=\linewidth]{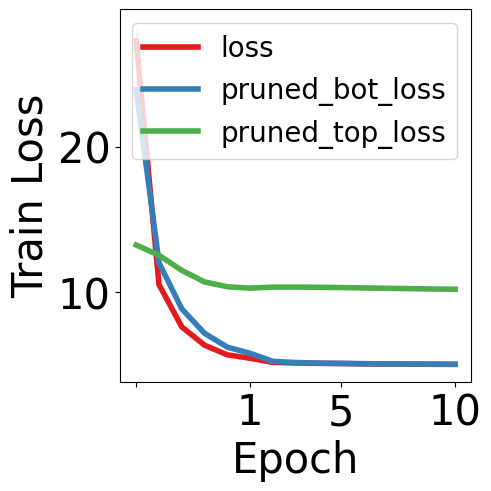}
    \caption{Transformer}
  \end{subfigure}
  \caption{Training loss over time, where the rows use differing amounts of weight decay. From top to bottom, for VGG we use coefficients $\{ 0, 0.001, 0.01, 0.1 \}$, while for other networks we use coefficients $\{0, 0.1, 1, 10\}$. We see that it is still possible to achieve low training loss under high weight decay, and as we increase the amount of weight decay, the gap between pruned and unpruned parameters closes, lending support to the idea that the parameters become lower rank.}
  \label{fig:weight-decay-performance}
\end{figure}

All tasks are trained in exactly the same fashion as mentioned previously, with increasing weight decay in the set $\{0, 0.0001, 0.001, 0.01, 0.1, 1.0, 10.0\}$. For ease of presentation we consider a subset of settings across tasks. In Figure~\ref{fig:weight-decay-performance} we include trained model performance and pruned model performance to show that, even with high levels of weight decay, models do not entirely break down. More so, the approximation of the pruned model to the full model gets better with higher weight decay.

\subsection{Grokking Experiments}

\begin{figure*}[!ht]
  \centering
  \begin{subfigure}[b]{0.24\linewidth}
    \centering
    \includegraphics[width=\linewidth]{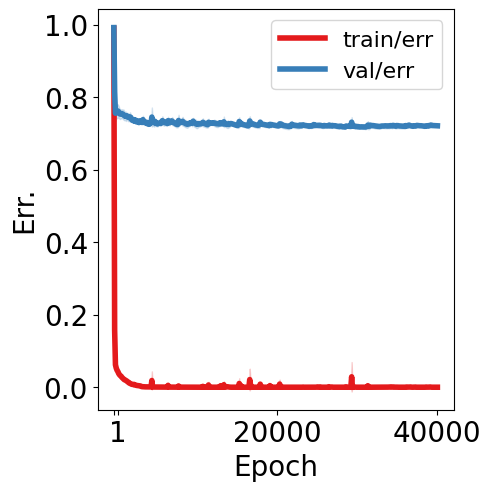}
    \\
    \includegraphics[width=\linewidth]{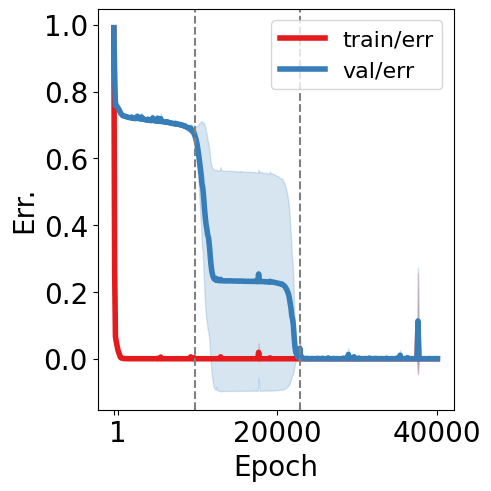}
    \\
    \includegraphics[width=\linewidth]{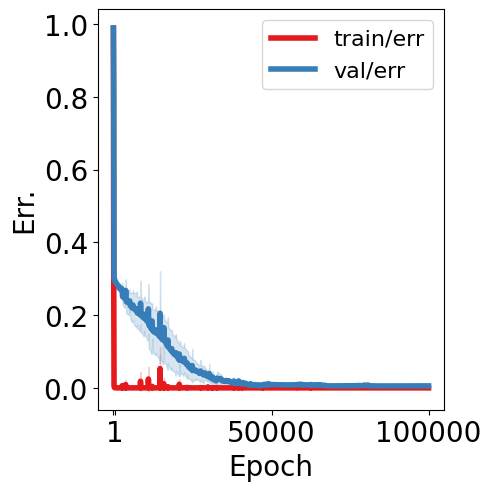}
    \\
    \includegraphics[width=\linewidth]{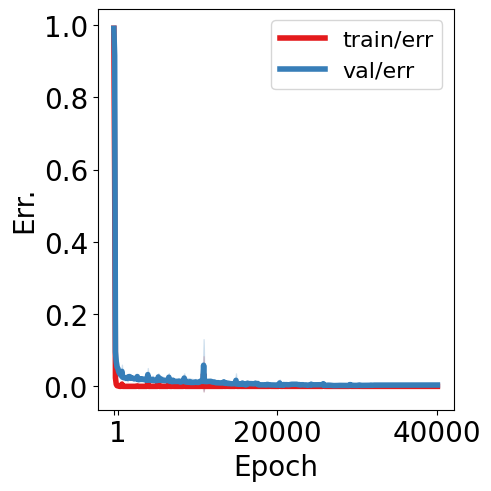}
    \caption{Error}
  \end{subfigure}\hfil
  \begin{subfigure}[b]{0.24\linewidth}
    \centering
    \includegraphics[width=\linewidth]{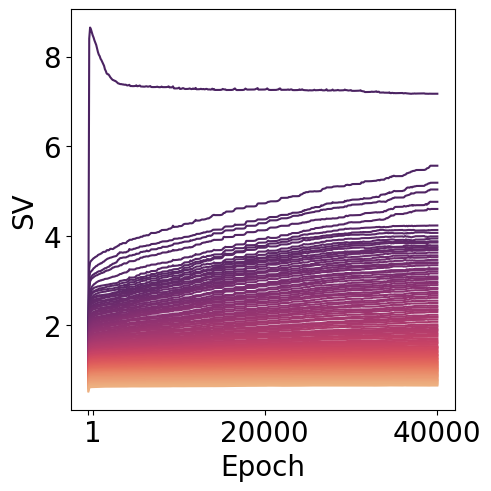}
    \\
    \includegraphics[width=\linewidth]{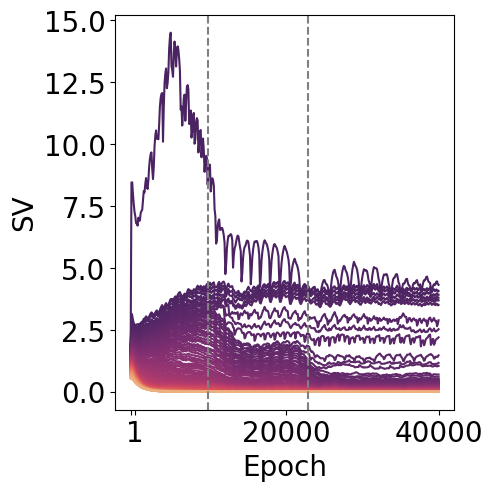}
    \\
    \includegraphics[width=\linewidth]{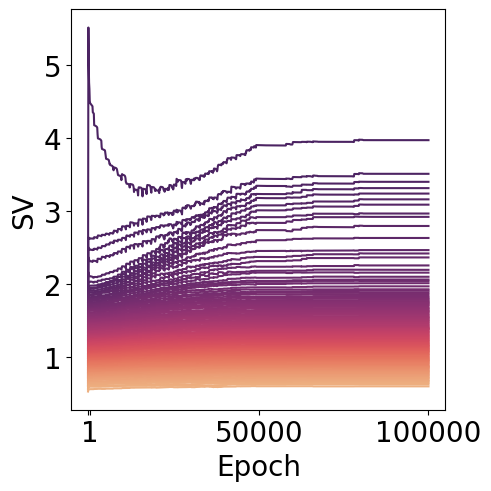}
    \\
    \includegraphics[width=\linewidth]{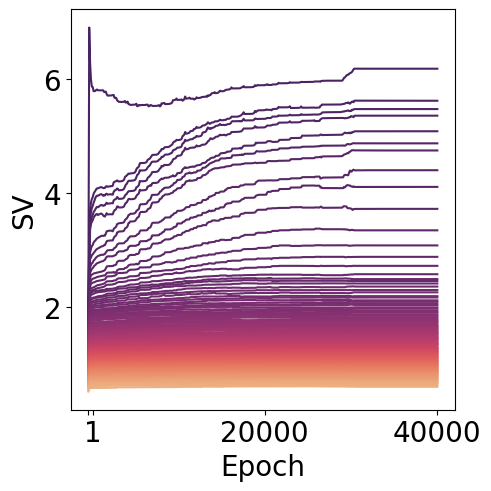}
    \caption{SV Evolution}
  \end{subfigure}\hfil
  \begin{subfigure}[b]{0.24\linewidth}
    \centering
    \includegraphics[width=\linewidth]{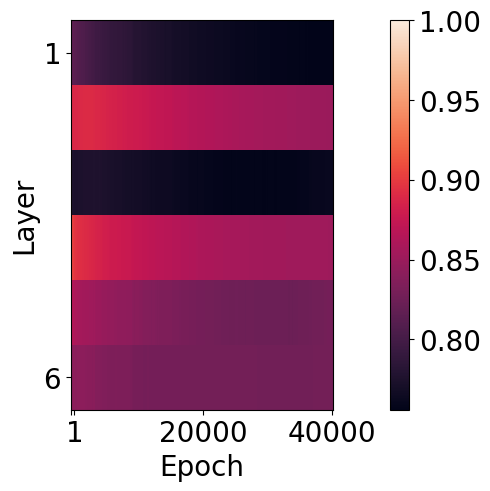}
    \\
    \includegraphics[width=\linewidth]{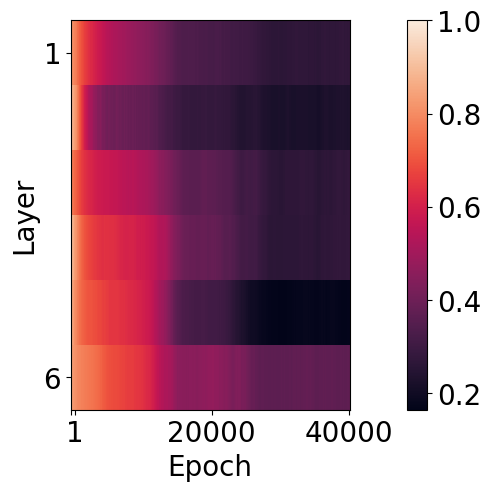}
    \\
    \includegraphics[width=\linewidth]{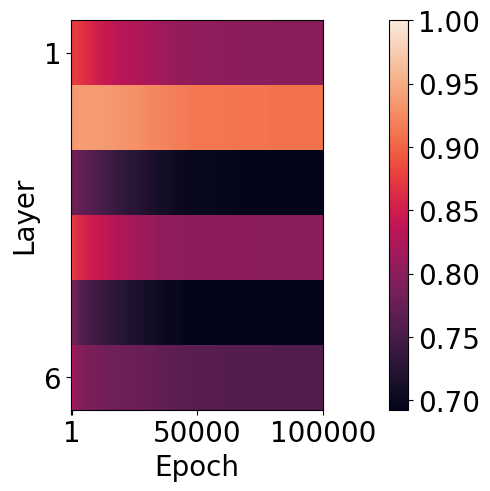}
    \\
    \includegraphics[width=\linewidth]{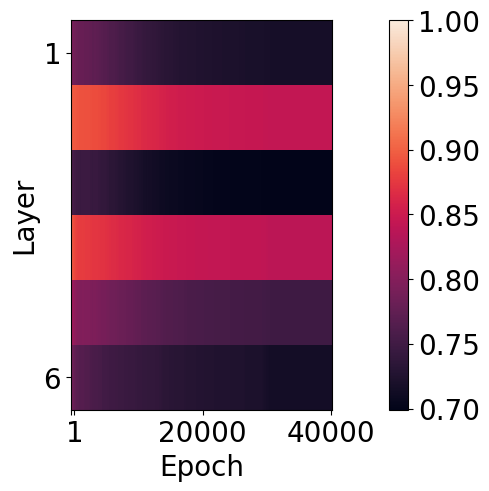}
    \caption{Effective Rank}
  \end{subfigure}\hfil
  \begin{subfigure}[b]{0.24\linewidth}
    \centering
    \includegraphics[width=\linewidth]{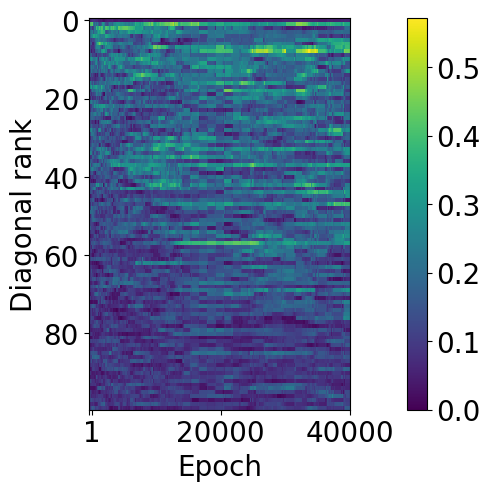}
    \\
    \includegraphics[width=\linewidth]{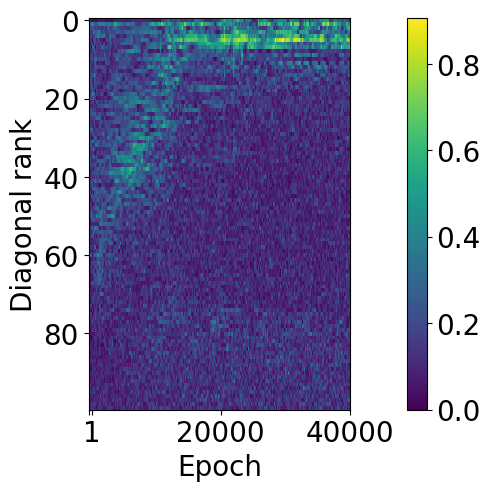}
    \\
    \includegraphics[width=\linewidth]{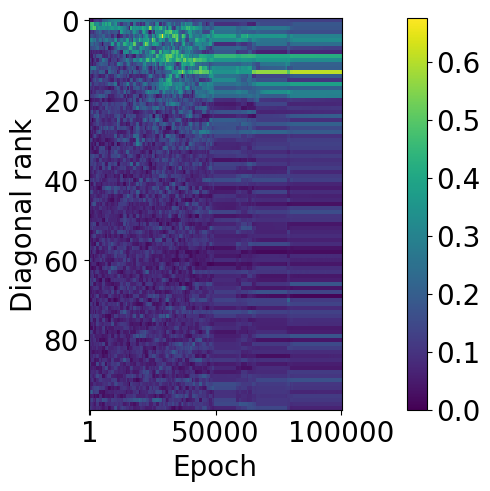}
    \\
    \includegraphics[width=\linewidth]{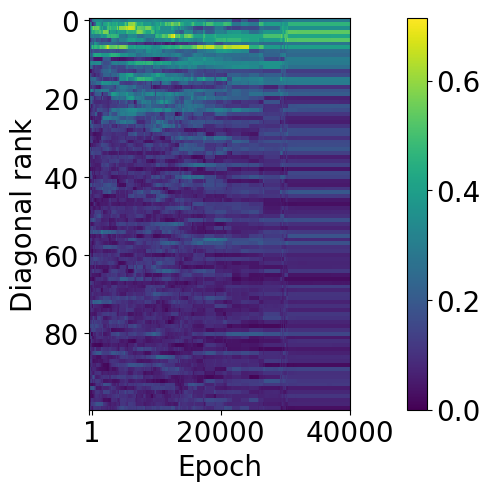}
    \caption{Alignment}
  \end{subfigure}
  \caption{\textbf{Grokking and Spectral Dynamics in Modular addition.} \textbf{Top row:} 30\% data and no weight decay. \textbf{2nd row:} 30\% data and weight decay 1.0 (grokking), using hyperparameters from \citet{nanda2023progress}. \textbf{3rd row:} 70\% data with no weight decay (slingshot), using hyperparameters from \citet{thilak2022slingshot}. \textbf{Bottom row:} 90\% data and no weight decay. \textbf{(a):} Training and validation error. \textbf{(b):} Singular value evolution is visualized for the first attention parameter, where each line represents a single singular value and the color represents the rank. \textbf{(c):} Effective rank of all layers (Eqn.~\ref{eqn:normalized-effective-rank}). \textbf{(d):} Alignment (Eqn.~\ref{eqn:alignment-matrix}) between the embedding and the first attention parameter is also visualized, where the y-axis corresponds to index $i$ of the diagonal. One can see that grokking co-occurs with low-rank weights. In addition, there is an alignment that begins early in training that evolves up the diagonal. Without weight decay and with less data, neither grokking nor the other phenomena occur during the entire training budget, but using more data, even without weight decay, leads to low-rank solutions from the beginning of training. The slingshot case follows a similar trend, though the validation loss is gradually fit. Across cases with good generalization, parameters are lower rank, and alignment is also more prevalent in the top ranks.}
  \label{fig:grokking-modadd}
\end{figure*}

For the Transformer, we mostly follow the settings and architecture of \citet{nanda2023progress}, except we use sinusoidal positional encodings instead of learned.

For the slingshot case we follow hyperparameter settings in \citet{thilak2022slingshot}, Appendix B except with the 1-layer architecture from \citet{nanda2023progress} instead of the 2-layer architecture specified. We perform addition modulo 97. The original grokking plot in \citet{thilak2022slingshot} appears much more dramatic as it log-scales the x-axis, which we do not do here for clarity.

In the case of the deep MLP, we follow \citet{fan2024deep}, where we use a 12-layer MLP with ReLU activations and width 400, trained on MSE loss on MNIST~\citep{lecun1998mnist}. We use 2000 examples, a batch size of 100, weight decay 0.01, and initialization scale 8~\citep{liu2023omnigrok}.

\subsection{Random Label Experiments}

We train a 4-layer MLP on CIFAR10~\citep{krizhevsky2009learning} with either completely random labels, or the true labels. We use SGD with momentum of 0.9 and constant learning rate of 0.001, and train for 300 epochs to see the entire trend of training. The major difference to the setting of \citet{zhang2021understanding} is the use of a constant learning rate, as their use of a learning rate schedule might conflate the results.

For the VGG case, we follow our previous hyperparameters, except we leave out weight decay and learning rate scheduling, instead using a constant learning rate of 0.01.

For the LSTM case, we follow our previous hyperparameters, and extend the training budget to 200 epochs allow for the random label setting to train longer. In this case, our network does not have sufficient capacity to memorize the data completely.

\subsection{Magnitude Pruning Experiments}

We use the same VGG setup as described previously. In this case we train until the end, then compute a global magnitude mask. To do this we flatten all linear and convolutional weights into a single vector, except for the last linear layer, and sort by magnitude. Then we keep the top 5\% of weights globally, and reshape back to the layerwise masks. This results in different sparsity levels for different layers, so when generating the random masks, we use the per-layer sparsities that resulted from the global magnitude mask.

To retrain the network, we rewind to epoch 4 (after the point of LMC~\citep{frankle2020linear}), then continue training with the mask, always setting other weights and their gradients to 0. We average all results over 3 random seeds.

For the LSTM we follow exactly the same procedure, except our mask only reaches a level of 25\% sparsity, due to larger performance degradations with higher sparsity.

\subsection{LMC Experiments}\label{app:lmc-details}

We save 5 evenly-spaced checkpoints in the first epoch, as well as at the end of the next 4 epochs for 10 initializations in total. We train 3 trunks, and split 3 branches from each trunk for a total of 9 branches which we average all plots over.

Following \citet{neyshabur2020being}, we compute the barrier between checkpoints as follows: given $W^{(1)}(T)$ and $W^{(2)}(T)$ that were branched from $W(t)$ we compute
\begin{equation}\label{eqn:lmc-barrier}
    b(t) = \max_{\alpha \in [0, 1]} \left [ \mathcal{L}((1-\alpha)W^{(1)}(T) + \alpha W^{(2)}(T)) - ((1-\alpha)\mathcal{L}(W^{(1)}(T)) + \alpha \mathcal{L}(W^{(2)}(T))) \right ]
\end{equation}
when this quantity is 0, we consider the checkpoints to exhibit LMC.

We recompute batch normalization parameters after interpolating for VGG-16, and group normalization parameters for the UNet, as these do not necessarily interpolate well~\citep{frankle2020linear}. We also compute singular vector agreement for the same parameter between either branch endpoint.

To plot the singular vector (dis)agreement and LMC between different modes, we make 11 evenly spaced measurements interpolating between branch endpoints that had the same split epoch, and the same branch seed, but different trunk initializations.

\subsection{Perturbed LMC Experiments}

We perturb all weights $W$ after the point of dynamics stability where we expect to see LMC at the end of training (epoch 4 is sufficiently late in all cases) using randomly sampled normal perturbations $\epsilon \sim \mathcal{N}(0, I)$ with $\lVert \epsilon \rVert = \eta \lVert W \rVert$ where $\eta \in \{0.0, 0.1, 0.25, 0.5, 1.0, 2.5\}$. We do not perturb the output layer, as this has a very substantial effect on the optimization. We also do not perturb the input layer for the Transformer as it is too computationally expensive for our resources.

\section{Limitations}

There are a few key limitations to our study. As mentioned, we lack the computational resources to run more than 3 random seeds per experiment, though we do find error bars to be quite tight in general (except for the generalization epoch in the grokking experiments). In addition, as discussed we ignore 1D parameters like biases and normalization in the neural networks, which may play additional roles. Due to computational constraints we do not consider alignment of layers across residual connections as this quickly becomes combinatorial in depth, thus there may be other interesting interactions that we do not observe. Finally, due to computational constraints we are unable to investigate full results on larger models than the 12 layer Transformer, which may have different behavior, but the results on the Pythia suite in Appendix~\ref{sec:pythia} are encouraging.

\section{Additional Experiments}

\subsection{Effect of initialization scale}\label{app:initscale}

\begin{figure*}[h!]
  \centering
  \begin{subfigure}[b]{0.19\linewidth}
    \centering
    \includegraphics[width=\linewidth]{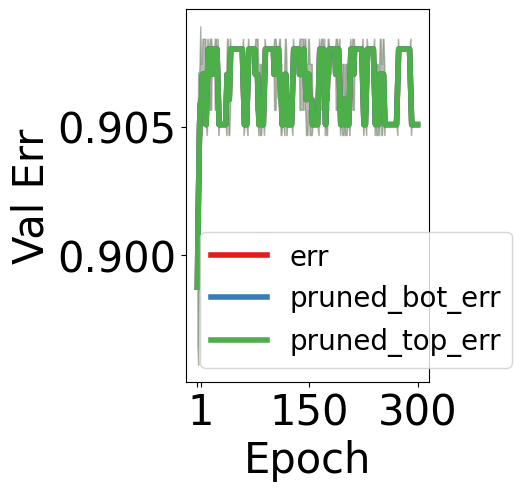}
    \\
    \includegraphics[width=\linewidth]{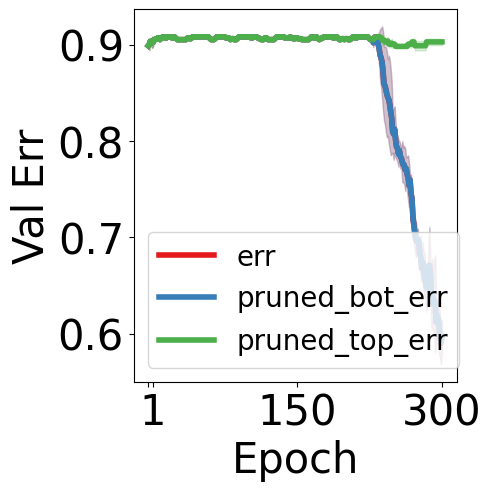}
    \\
    \includegraphics[width=\linewidth]{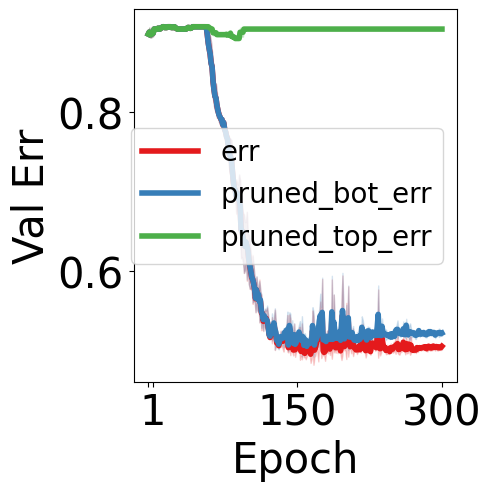}
    \\
    \includegraphics[width=\linewidth]{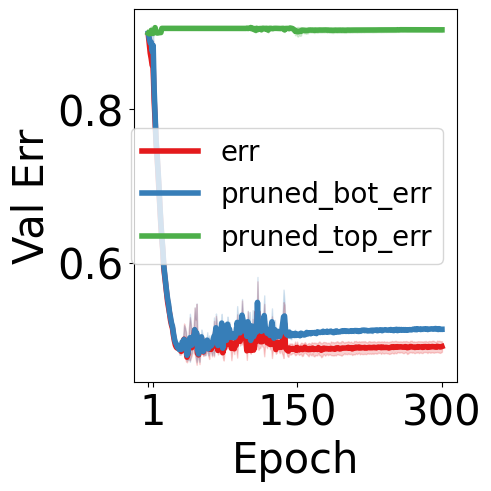}
    \\
    \includegraphics[width=\linewidth]{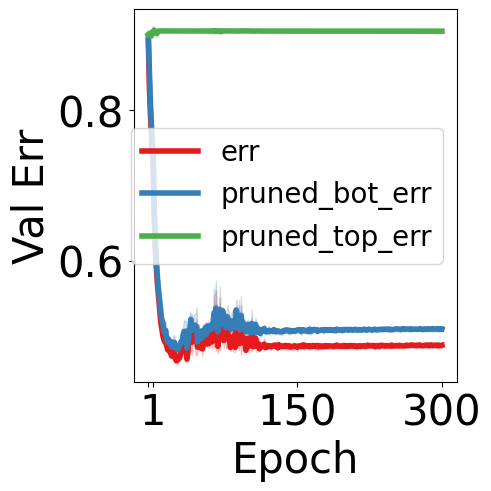}
    \caption{Val. Err}
  \end{subfigure}
  \begin{subfigure}[b]{0.19\linewidth}
    \centering
    \includegraphics[width=\linewidth]{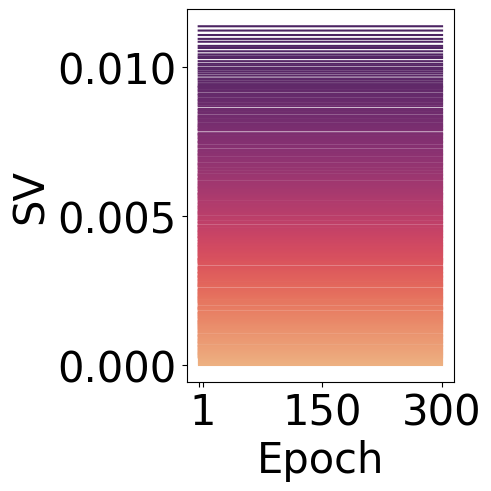}
    \\
    \includegraphics[width=\linewidth]{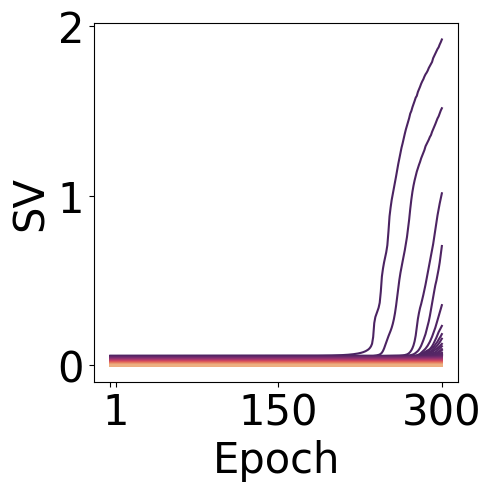}
    \\
    \includegraphics[width=\linewidth]{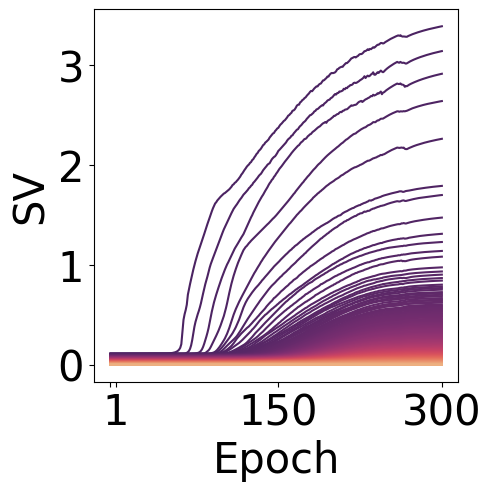}
    \\
    \includegraphics[width=\linewidth]{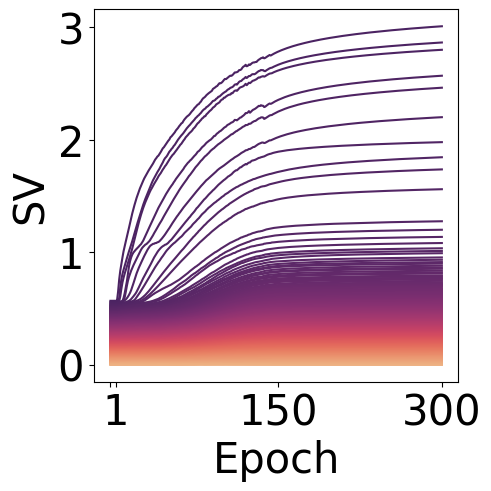}
    \\
    \includegraphics[width=\linewidth]{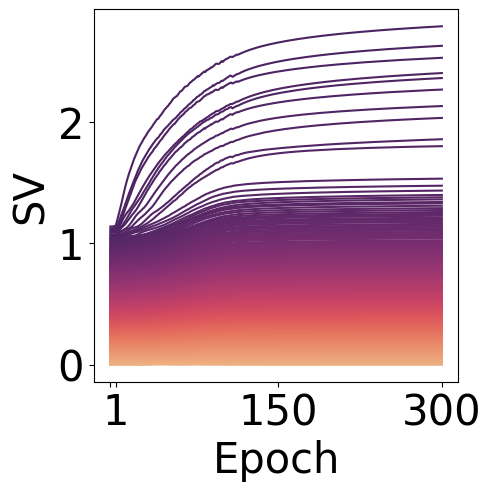}
    \caption{SVs}
  \end{subfigure}
  \begin{subfigure}[b]{0.19\linewidth}
    \centering
    \includegraphics[width=\linewidth]{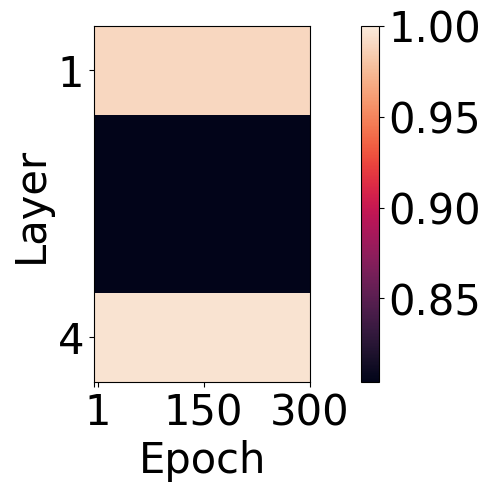}
    \\
    \includegraphics[width=\linewidth]{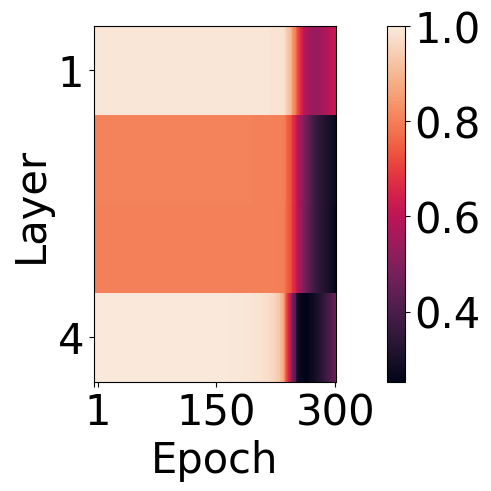}
    \\
    \includegraphics[width=\linewidth]{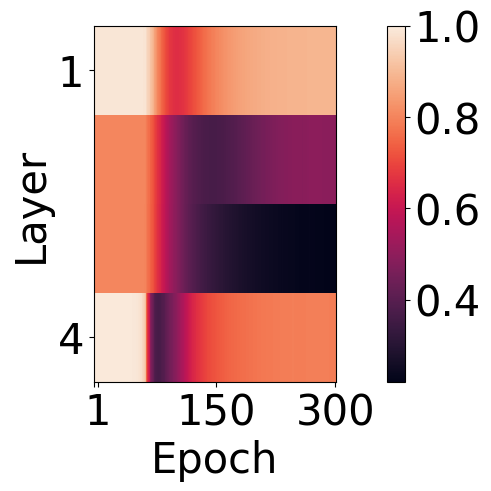}
    \\
    \includegraphics[width=\linewidth]{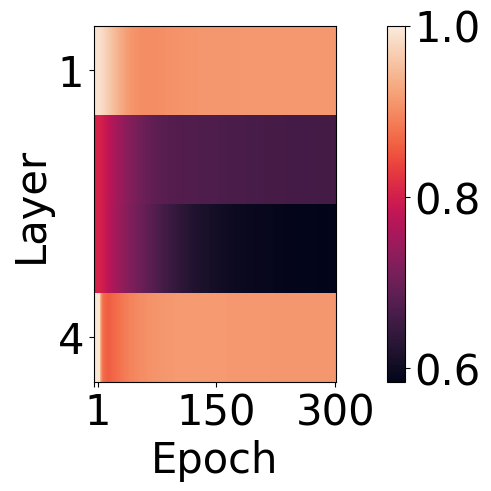}
    \\
    \includegraphics[width=\linewidth]{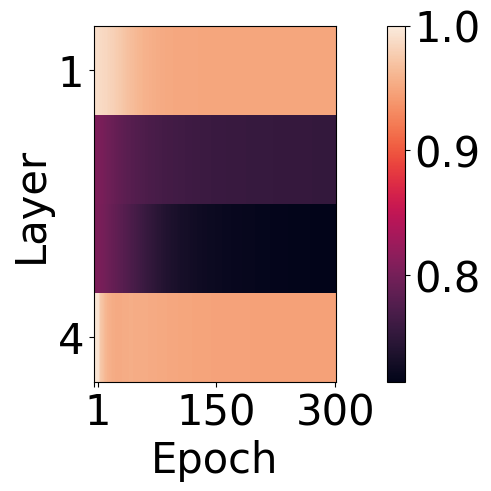}
    \caption{Eff. Rank}
  \end{subfigure}
  \begin{subfigure}[b]{0.19\linewidth}
    \centering
    \includegraphics[width=\linewidth]{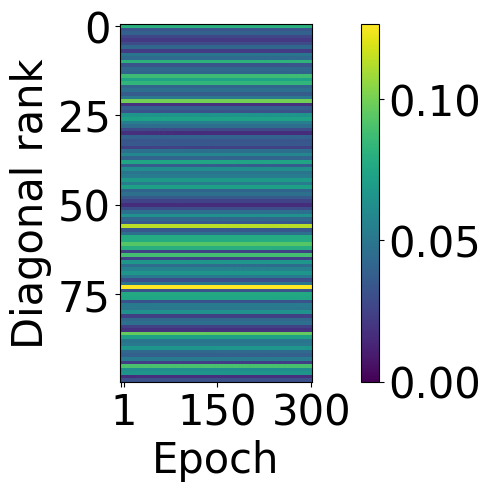}
    \\
    \includegraphics[width=\linewidth]{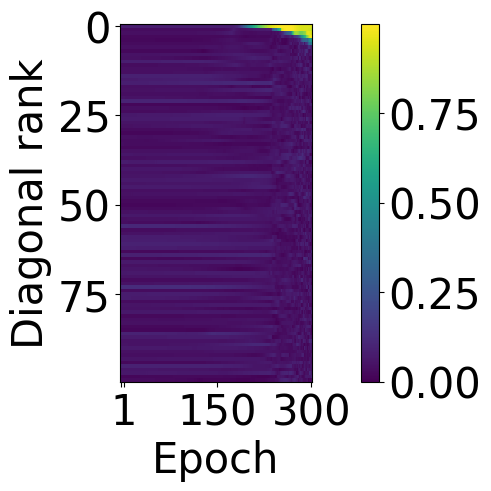}
    \\
    \includegraphics[width=\linewidth]{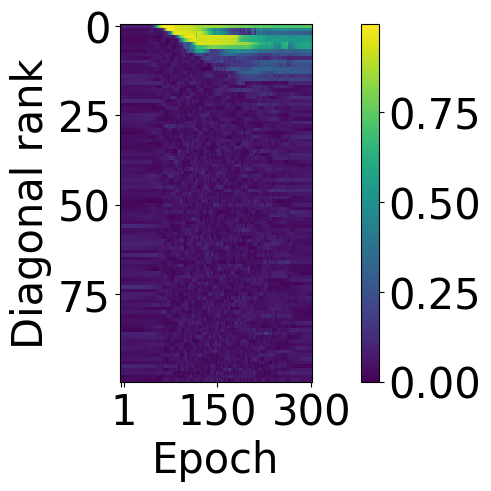}
    \\
    \includegraphics[width=\linewidth]{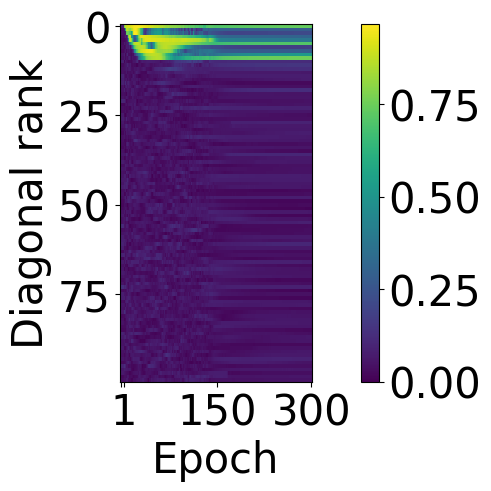}
    \\
    \includegraphics[width=\linewidth]{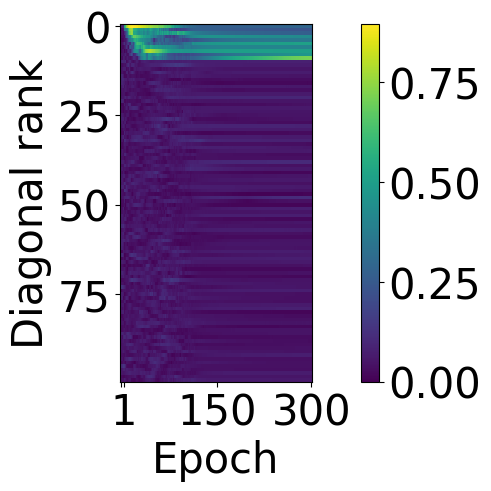}
    \caption{Alignment}
  \end{subfigure}
  \begin{subfigure}[b]{0.19\linewidth}
    \centering
    \includegraphics[width=\linewidth]{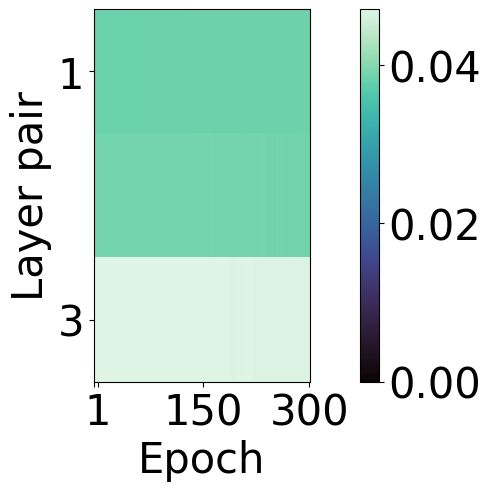}
    \\
    \includegraphics[width=\linewidth]{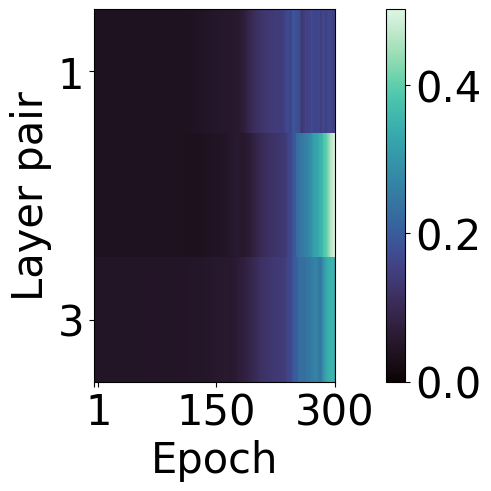}
    \\
    \includegraphics[width=\linewidth]{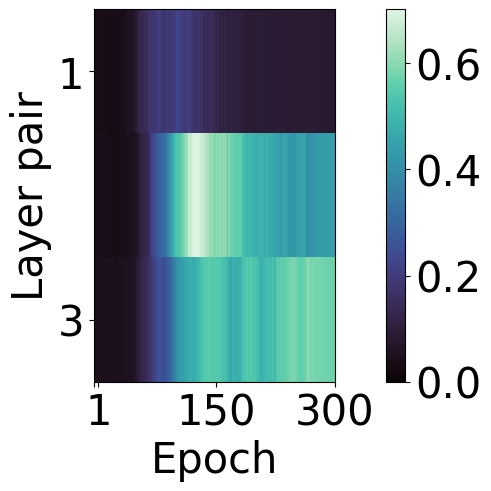}
    \\
    \includegraphics[width=\linewidth]{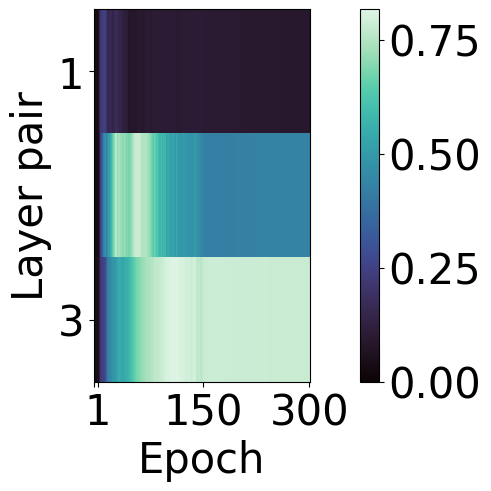}
    \\
    \includegraphics[width=\linewidth]{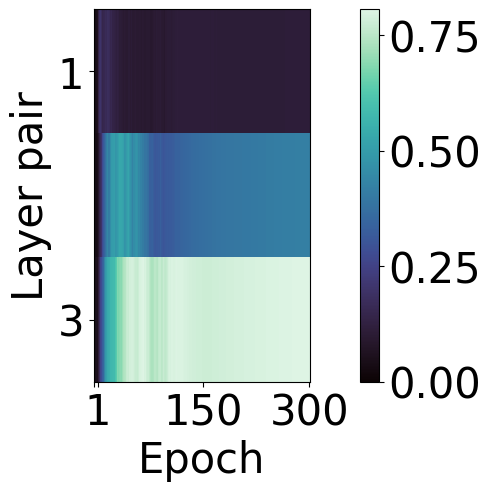}
    \caption{Align. Score}
  \end{subfigure}
  \caption{Varying initialization scale affects the dynamics of an MLP trained on CIFAR10. From top to bottom we use default PyTorch initialization multiplied by the constants $\{0.001, 0.01, 0.05, 0.1, 0.5, 1.0\}$. Notably, alignment occurs slightly more strongly with smaller initialization, as predicted by \citet{woodworth2020kernel} for deep linear models.}
  \label{fig:initscale}
\end{figure*}

In Figure~\ref{fig:initscale} we explore the effect of initialization scale. Prior work~\citep{woodworth2020kernel} showed that in deep linear systems the choice of initialization modulated the effect of incremental learning of ranks, with smaller initialization increasing this effect. We observe that alignment is slightly stronger with smaller initialization, mirroring the theory in deep linear networks.

\subsection{Effect of learning rate}\label{app:learning-rate}

\begin{figure*}[h!]
  \centering
  \begin{subfigure}[b]{0.19\linewidth}
    \centering
    \includegraphics[width=\linewidth]{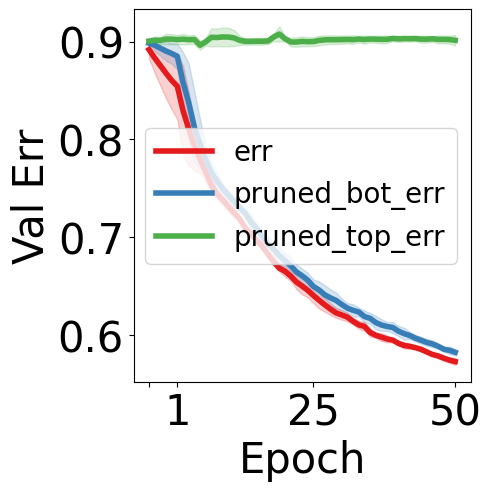}
    \\
    \includegraphics[width=\linewidth]{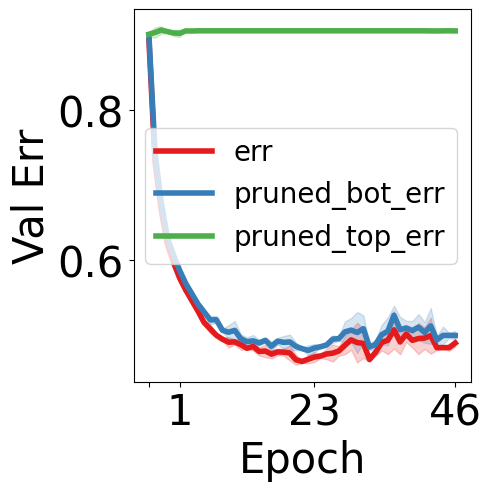}
    \\
    \includegraphics[width=\linewidth]{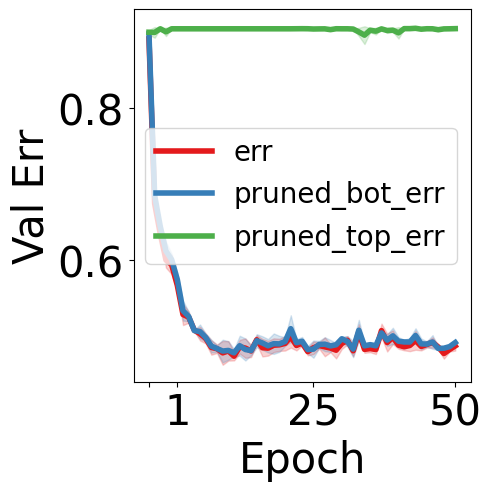}
    \caption{Val. Err}
  \end{subfigure}
  \begin{subfigure}[b]{0.19\linewidth}
    \centering
    \includegraphics[width=\linewidth]{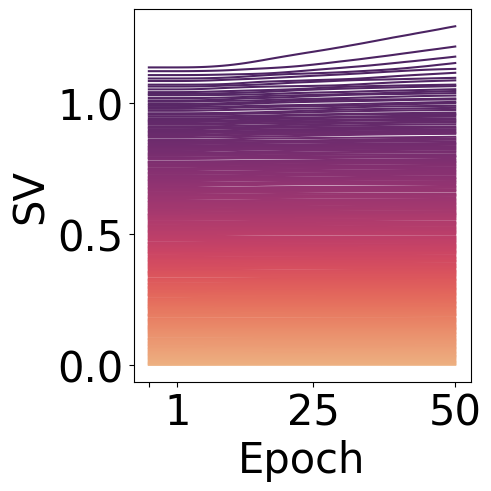}
    \\
    \includegraphics[width=\linewidth]{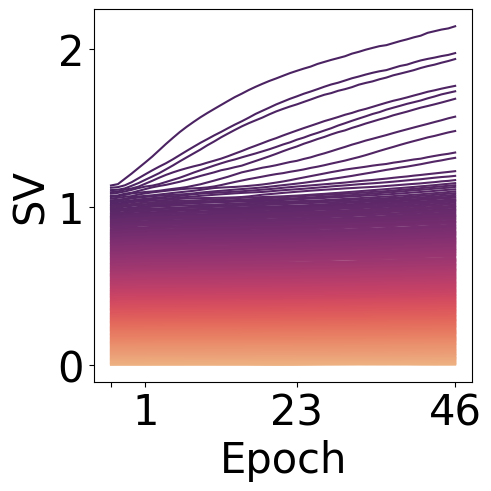}
    \\
    \includegraphics[width=\linewidth]{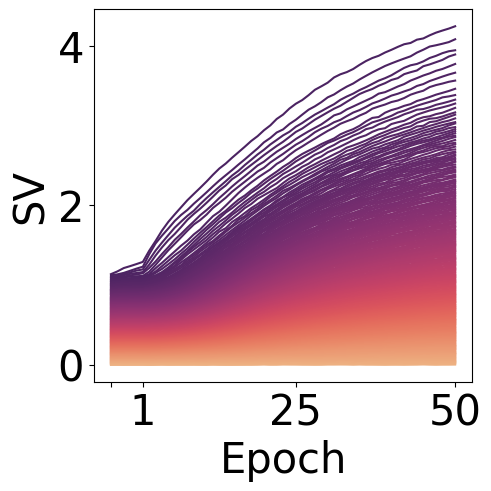}
    \caption{SVs}
  \end{subfigure}
  \begin{subfigure}[b]{0.19\linewidth}
    \centering
    \includegraphics[width=\linewidth]{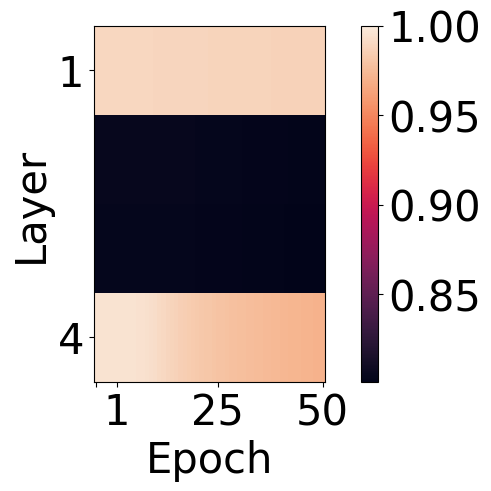}
    \\
    \includegraphics[width=\linewidth]{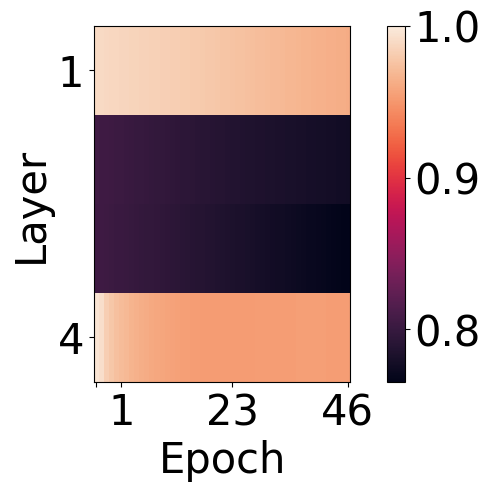}
    \\
    \includegraphics[width=\linewidth]{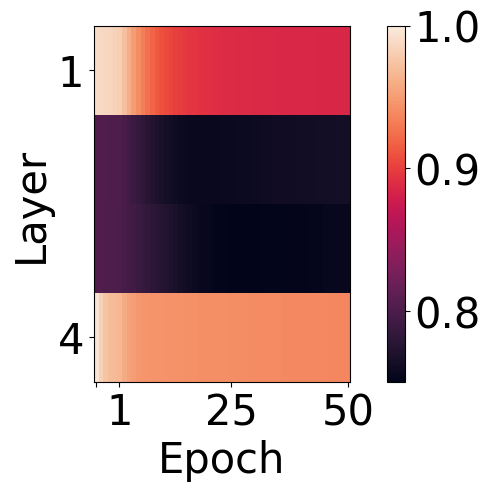}
    \caption{Eff. Rank}
  \end{subfigure}
  \begin{subfigure}[b]{0.19\linewidth}
    \centering
    \includegraphics[width=\linewidth]{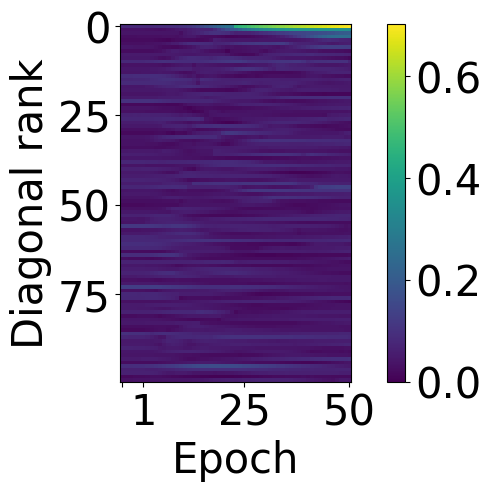}
    \\
    \includegraphics[width=\linewidth]{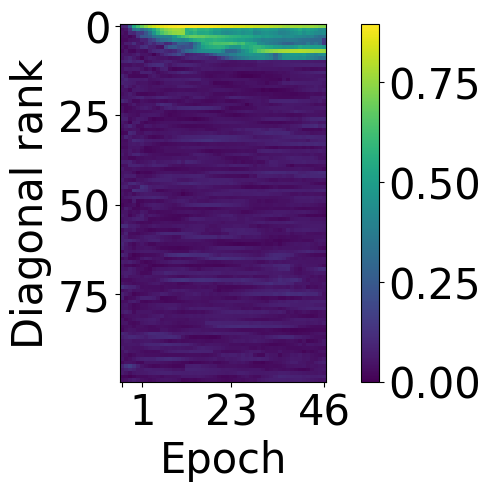}
    \\
    \includegraphics[width=\linewidth]{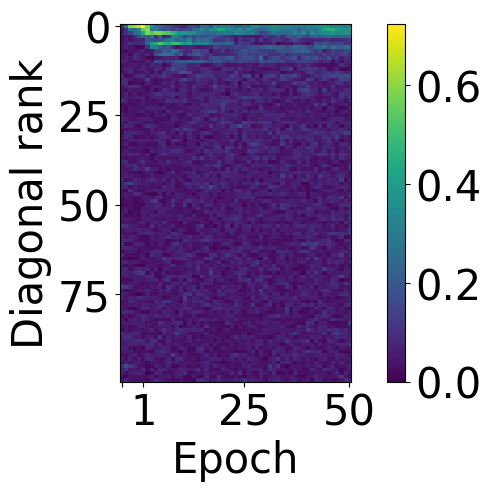}
    \caption{Alignment}
  \end{subfigure}
  \begin{subfigure}[b]{0.19\linewidth}
    \centering
    \includegraphics[width=\linewidth]{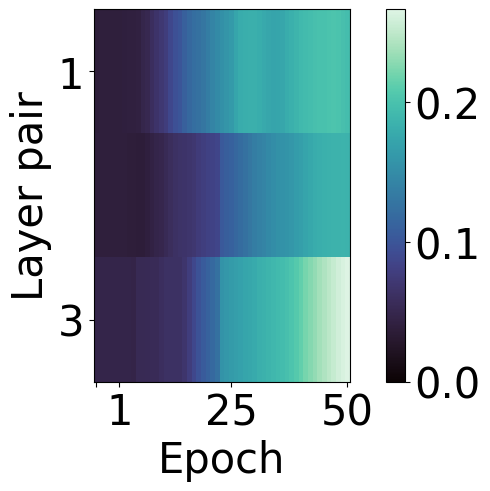}
    \\
    \includegraphics[width=\linewidth]{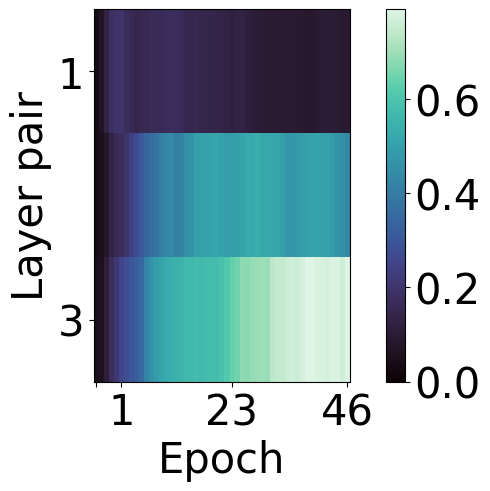}
    \\
    \includegraphics[width=\linewidth]{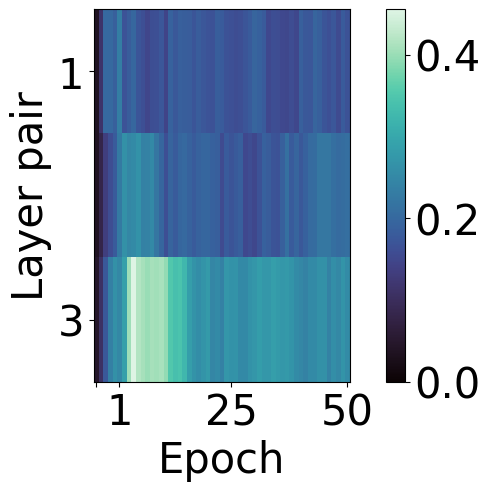}
    \caption{Align. Score}
  \end{subfigure}
  \caption{The effect of varying learning rate on an MLP trained on CIFAR10. From top to bottom we use learning rates $\{0.001, 0.01, 0.1\}$. Notably, alignment increases with larger learning rate, agreeing with the deep linear setting in \citet{ghosh2025learning}.}
  \label{fig:learning-rate}
\end{figure*}

In Figure~\ref{fig:learning-rate} we explore the effect of learning rate on dynamics for an MLP trained on CIFAR10. \citet{ghosh2025learning} proved that balancedness may develop in deep linear systems when the learning rate is sufficiently large. We see here that the trend is mixed: learning rate 0.01 shows the strongest balancedness, and going above or below has a weaker effect.

\subsection{Individual seed plots for grokking}\label{app:grokking-multiseed}

\begin{figure*}[!t]
  \centering
  \begin{subfigure}[b]{0.24\linewidth}
    \centering
    \includegraphics[width=\linewidth]{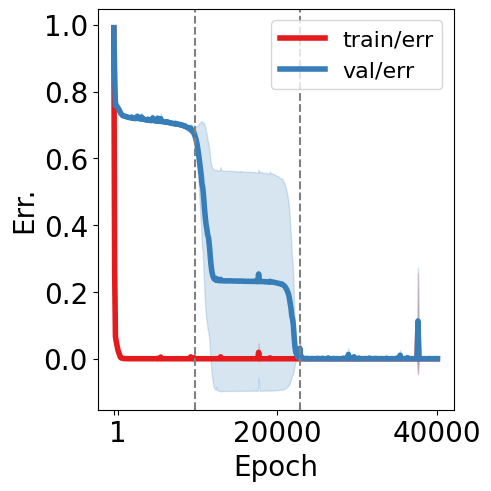}
    \caption{Error}
  \end{subfigure}\hfil
  \begin{subfigure}[b]{0.24\linewidth}
    \centering
    \includegraphics[width=\linewidth]{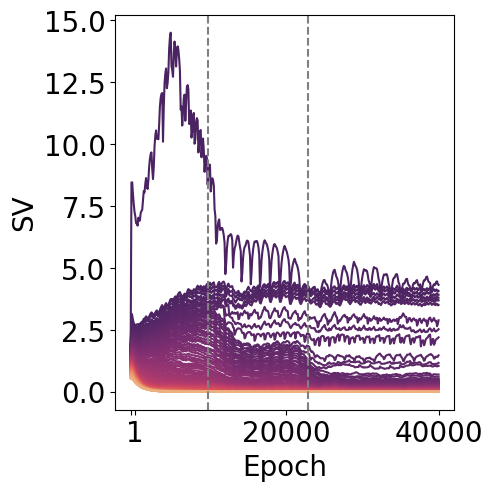}
    \caption{SV Evolution}
  \end{subfigure}\hfil
  \begin{subfigure}[b]{0.24\linewidth}
    \centering
    \includegraphics[width=\linewidth]{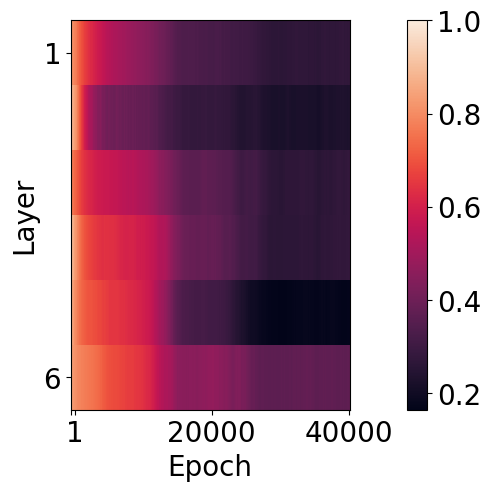}
    \caption{Effective Rank}
  \end{subfigure}\hfil
  \begin{subfigure}[b]{0.24\linewidth}
    \centering
    \includegraphics[width=\linewidth]{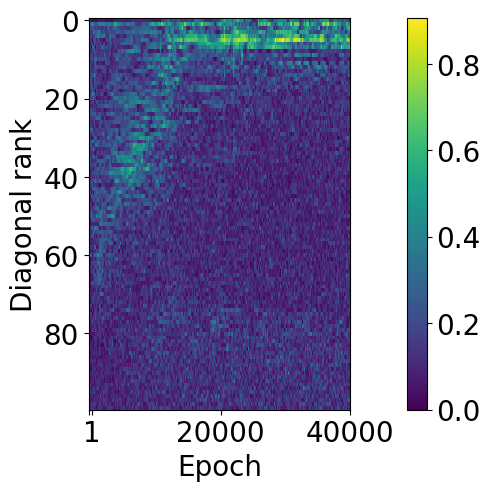}
    \caption{Alignment}
  \end{subfigure}
  \caption{Plots for averages of 3 seeds of grokking modular addition with Transformers (Figure~\ref{fig:grokking}). We see a stark transition to low-rank when validation error decreases.}
  \label{fig:grokking-multiseed}
\end{figure*}

For clarity, in Figure~\ref{fig:grokking-multiseed} we show an average of 3 seeds of grokking modular addition with Transformers (see Figure~\ref{fig:grokking}). The transition to low error and low rank coincide.

\subsection{Full alignment matrix evolution}\label{app:alignment-evolution}

\begin{figure*}[h!]
  \centering
  \begin{subfigure}[b]{0.14\linewidth}
    \centering
    \includegraphics[width=\linewidth]{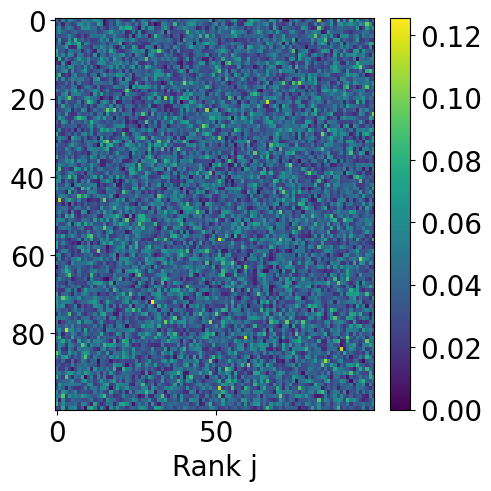}
    \caption{Init}
  \end{subfigure}\hfil
  \begin{subfigure}[b]{0.14\linewidth}
    \centering
    \includegraphics[width=\linewidth]{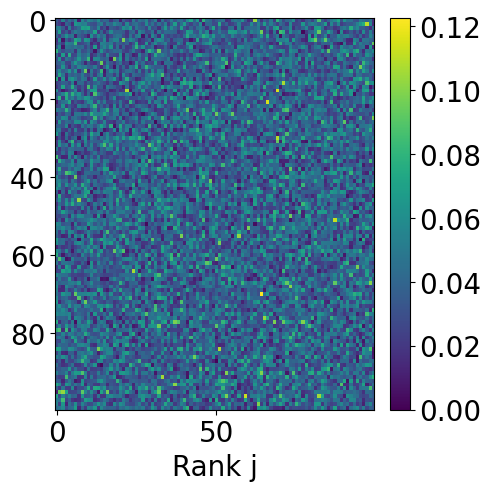}
    \caption{Ep. 2500}
  \end{subfigure}\hfil
  \begin{subfigure}[b]{0.14\linewidth}
    \centering
    \includegraphics[width=\linewidth]{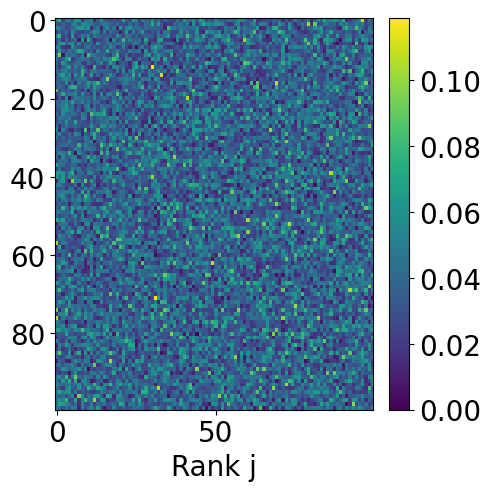}
    \caption{Ep. 5000}
  \end{subfigure}\hfil
  \begin{subfigure}[b]{0.14\linewidth}
    \centering
    \includegraphics[width=\linewidth]{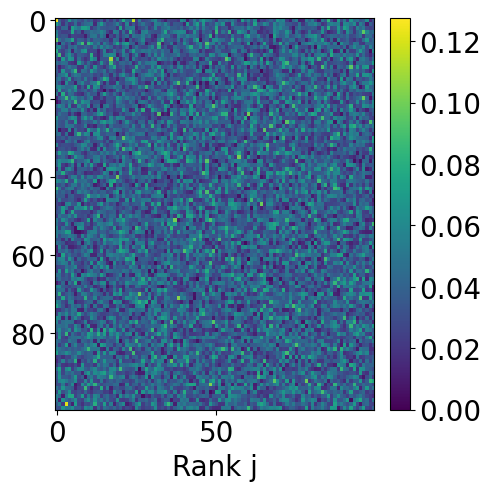}
    \caption{Ep. 5500}
  \end{subfigure}\hfil
  \begin{subfigure}[b]{0.14\linewidth}
    \centering
    \includegraphics[width=\linewidth]{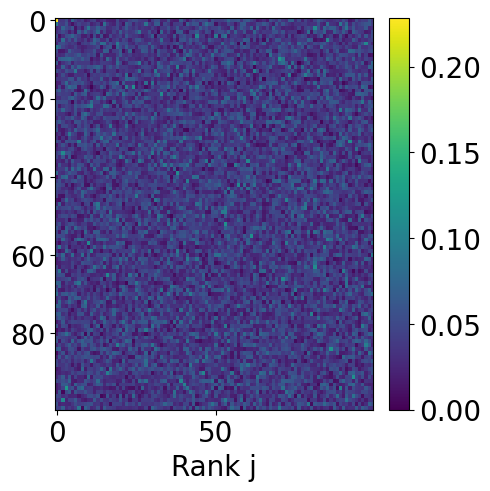}
    \caption{Ep. 6000}
  \end{subfigure}\hfil
  \begin{subfigure}[b]{0.14\linewidth}
    \centering
    \includegraphics[width=\linewidth]{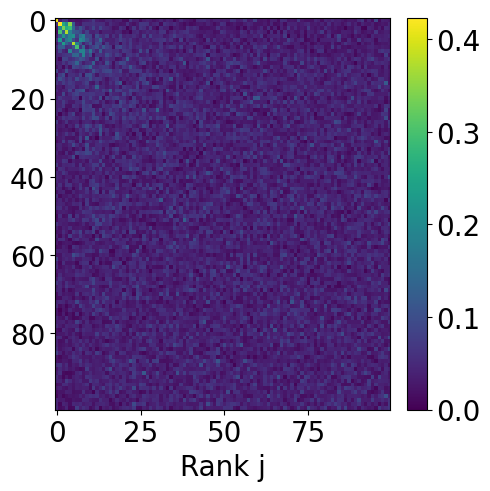}
    \caption{Ep. 7000}
  \end{subfigure}\hfil
  \begin{subfigure}[b]{0.14\linewidth}
    \centering
    \includegraphics[width=\linewidth]{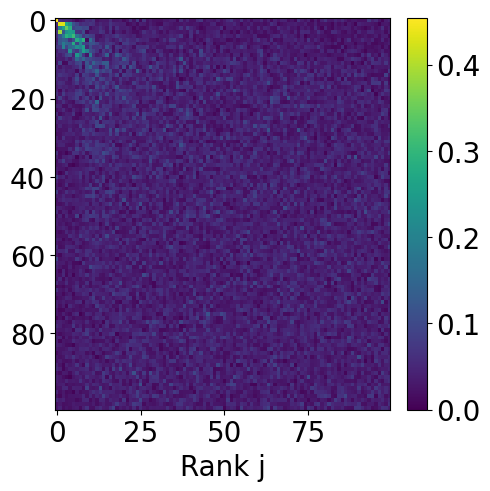}
    \caption{Ep. 8000}
  \end{subfigure}\hfil
    \caption{Alignment matrix (Eqn.~\ref{eqn:alignment-matrix}) over time. We see that the majority of signal is concentrated in the diagonal, an example of why we consider the diagonal in the rest of the work.}
    \label{fig:alignment-matrix-evolution}
\end{figure*}

We previously plotted only the diagonal of the alignment matrix (Eqn.~\ref{eqn:alignment-matrix}) instead of considering off-diagonal elements. In Figure~\ref{fig:alignment-matrix-evolution} we provide the evolution of the entire alignment matrix between two layers throughout training for an MLP on CIFAR10 so as to explain why: anecdotally we did not observe much signal off-diagonal, hence the focus on the upper diagonal in Eqns~\ref{eqn:alignment-matrix} and~\ref{eqn:alignment-measure}.

\subsection{Image classification}\label{app:image-classification}

\begin{figure*}[h!]
  \centering
  \begin{subfigure}[b]{0.19\linewidth}
    \centering
    \includegraphics[width=\linewidth]{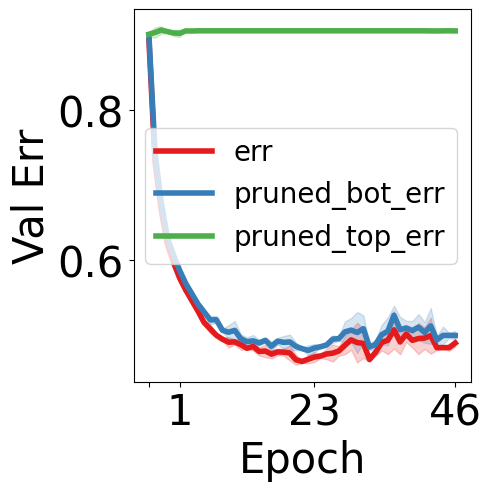}
    \\
    \includegraphics[width=\linewidth]{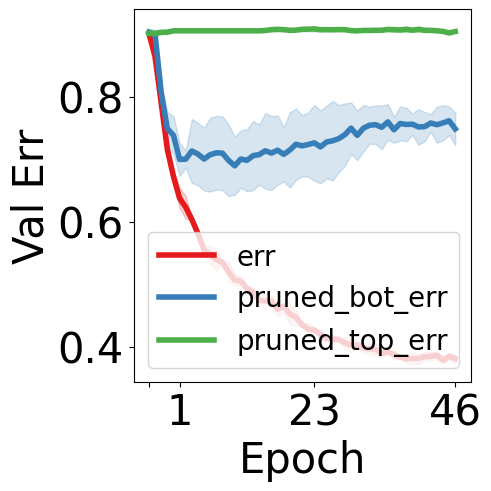}
    \\
    \includegraphics[width=\linewidth]{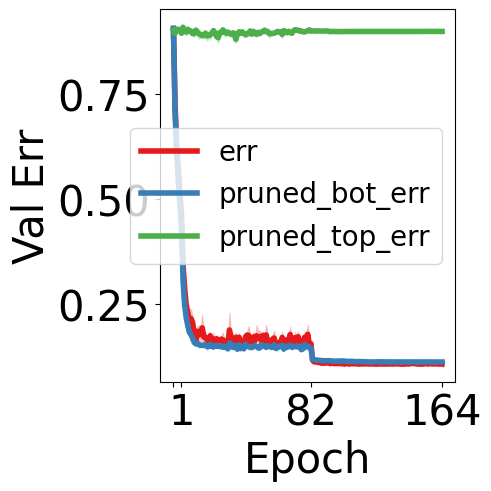}
    \caption{Val. Err}
  \end{subfigure}
  \begin{subfigure}[b]{0.19\linewidth}
    \centering
    \includegraphics[width=\linewidth]{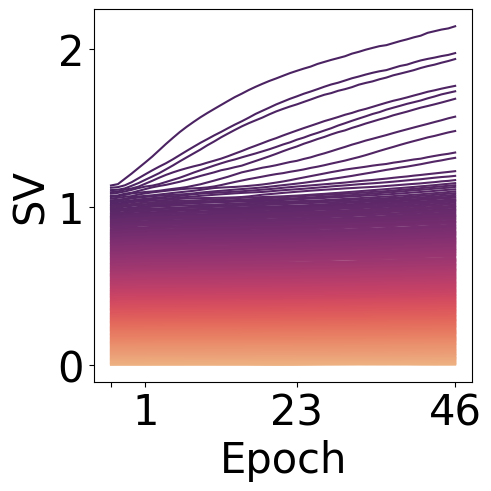}
    \\
    \includegraphics[width=\linewidth]{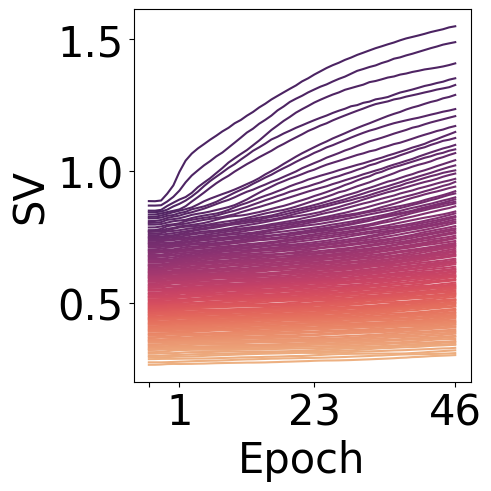}
    \\
    \includegraphics[width=\linewidth]{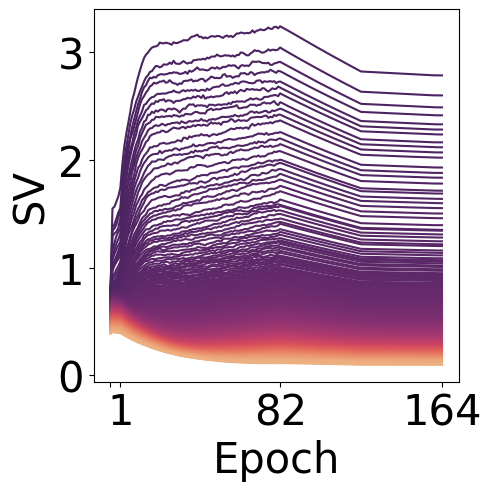}
    \caption{SVs}
  \end{subfigure}
  \begin{subfigure}[b]{0.19\linewidth}
    \centering
    \includegraphics[width=\linewidth]{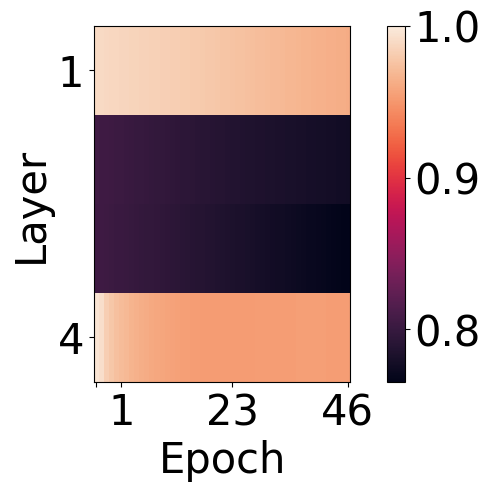}
    \\
    \includegraphics[width=\linewidth]{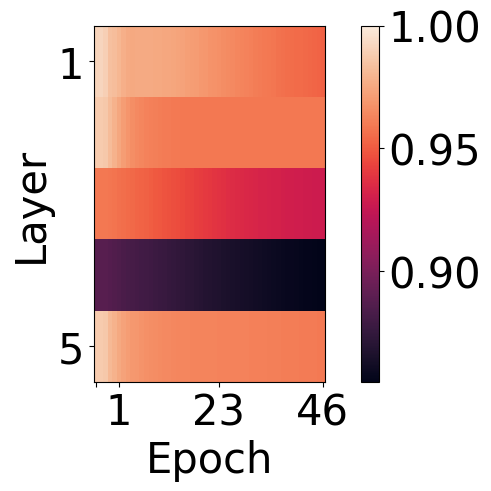}
    \\
    \includegraphics[width=\linewidth]{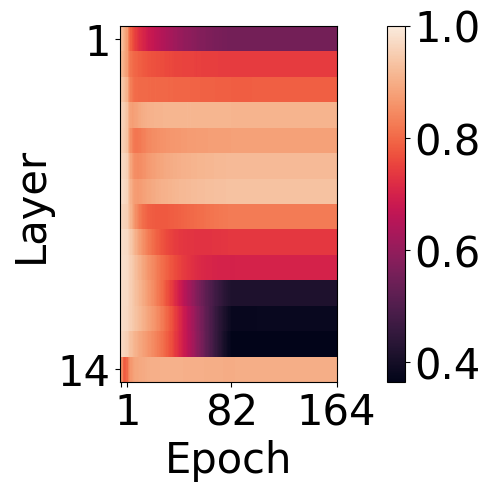}
    \caption{Eff. Rank}
  \end{subfigure}
  \begin{subfigure}[b]{0.19\linewidth}
    \centering
    \includegraphics[width=\linewidth]{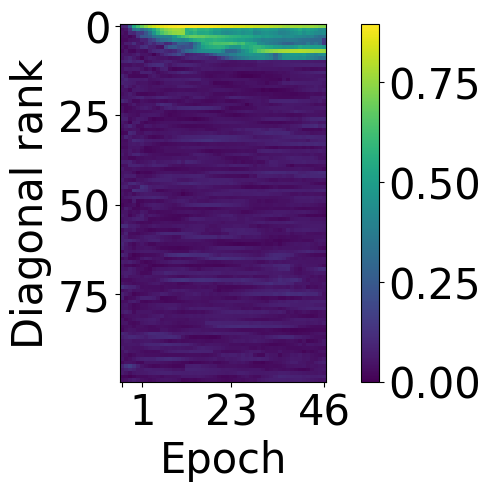}
    \\
    \includegraphics[width=\linewidth]{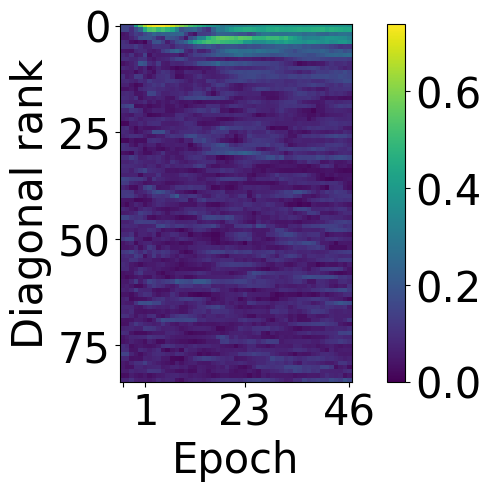}
    \\
    \includegraphics[width=\linewidth]{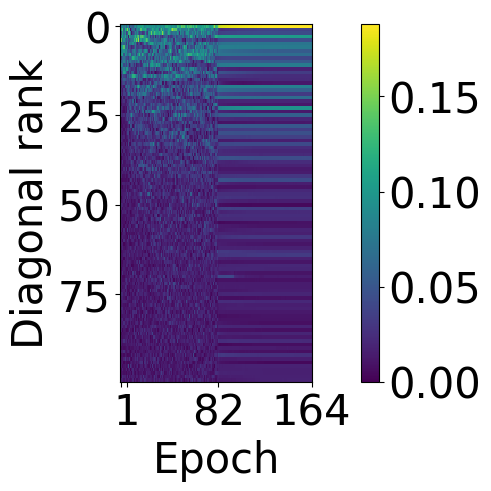}
    \caption{Alignment}
  \end{subfigure}
  \begin{subfigure}[b]{0.19\linewidth}
    \centering
    \includegraphics[width=\linewidth]{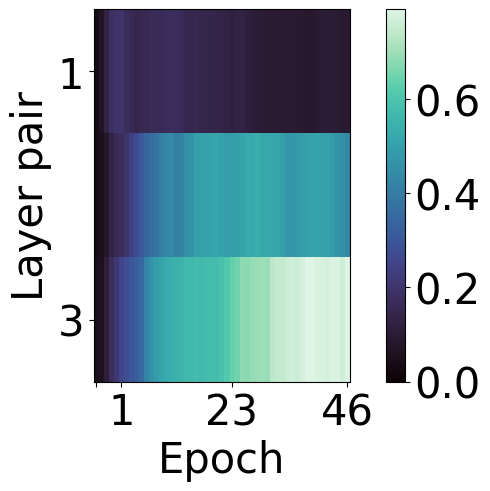}
    \\
    \includegraphics[width=\linewidth]{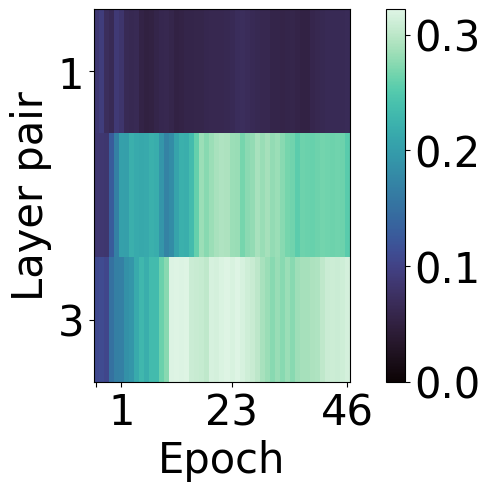}
    \\
    \includegraphics[width=\linewidth]{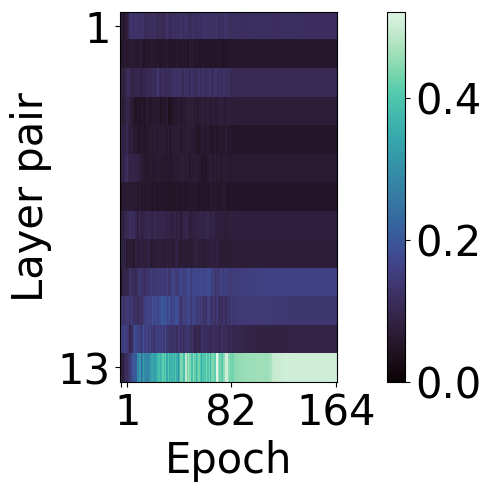}
    \caption{Align. Score}
  \end{subfigure}
  \caption{Different architectures all trained on CIFAR10. From top to bottom, MLP, LeNet-5 and VGG-16. We see roughly consistent spectral dynamics across architectures within the same task.}
  \label{fig:image-class}
\end{figure*}

In Figure~\ref{fig:image-class} we provide the behavior for all networks tested in CIFAR10 side-by-side to facilitate comparison. We see across architectures that there is qualitative agreement in the trend toward rank minimization.

\subsection{}

\begin{figure*}[!t]
  \centering
  \begin{subfigure}[b]{0.24\linewidth}
    \centering
    \includegraphics[width=0.5\linewidth]{./spectral_dynamics/figs/vgg/weight_decay_0.0-sv-model.layers.11.conv.weight_sv.png}\hfil
    \includegraphics[width=0.5\linewidth]{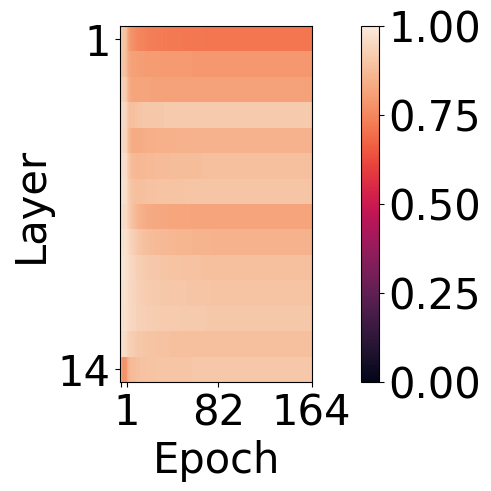}
    \\
    \includegraphics[width=0.5\linewidth]{./spectral_dynamics/figs/vgg/weight_decay_0.001-sv-model.layers.11.conv.weight_sv.png}\hfil
    \includegraphics[width=0.5\linewidth]{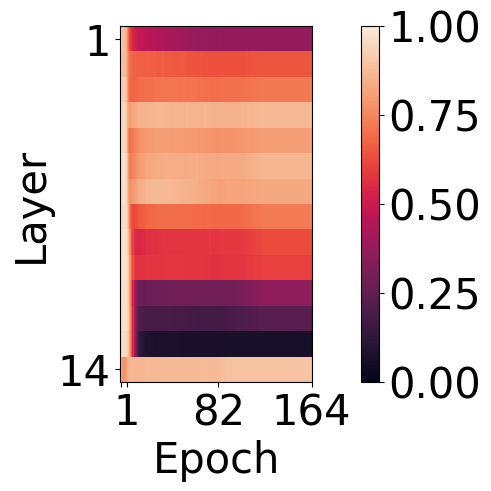}
    \\
    \includegraphics[width=0.5\linewidth]{./spectral_dynamics/figs/vgg/weight_decay_0.01-sv-model.layers.11.conv.weight_sv.png}\hfil
    \includegraphics[width=0.5\linewidth]{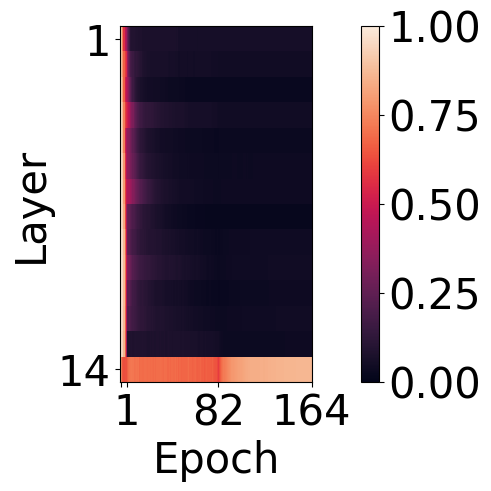}
    \\
    \includegraphics[width=0.5\linewidth]{./spectral_dynamics/figs/vgg/weight_decay_0.1-sv-model.layers.11.conv.weight_sv.png}\hfil
    \includegraphics[width=0.5\linewidth]{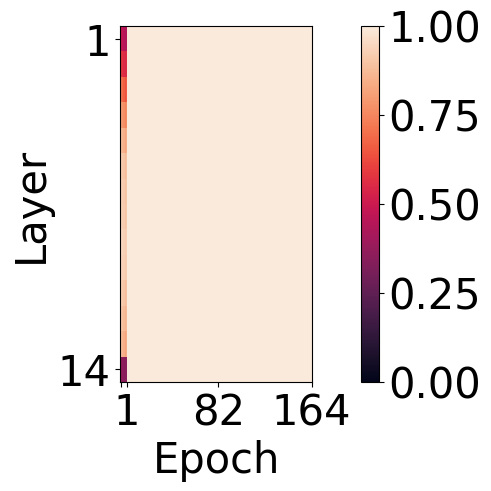}
    \caption{VGG}
  \end{subfigure}
  \begin{subfigure}[b]{0.24\linewidth}
    \centering
    \includegraphics[width=0.5\linewidth]{./spectral_dynamics/figs/unet/weight_decay_0.0-sv-model.conv3.weight_sv.png}\hfil
    \includegraphics[width=0.5\linewidth]{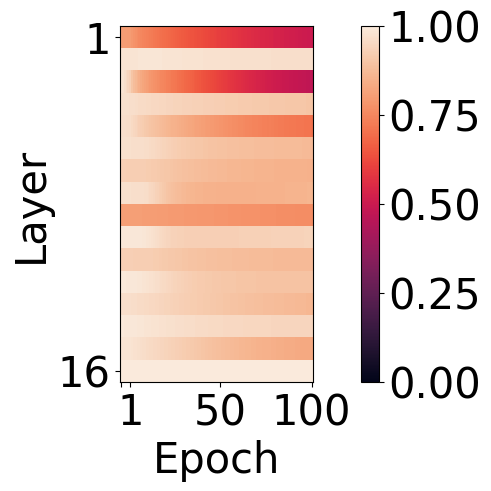}
    \\
    \includegraphics[width=0.5\linewidth]{./spectral_dynamics/figs/unet/weight_decay_0.1-sv-model.conv3.weight_sv.png}\hfil
    \includegraphics[width=0.5\linewidth]{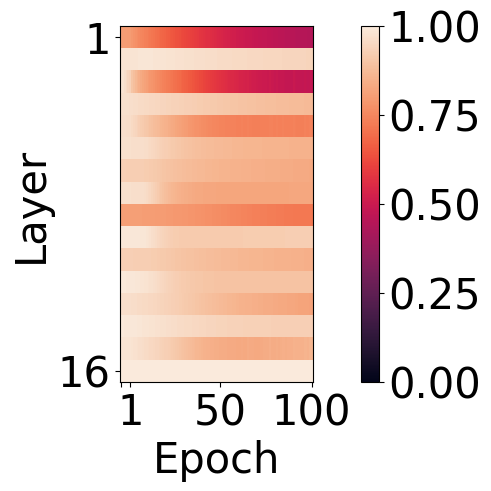}
    \\
    \includegraphics[width=0.5\linewidth]{./spectral_dynamics/figs/unet/weight_decay_1.0-sv-model.conv3.weight_sv.png}\hfil
    \includegraphics[width=0.5\linewidth]{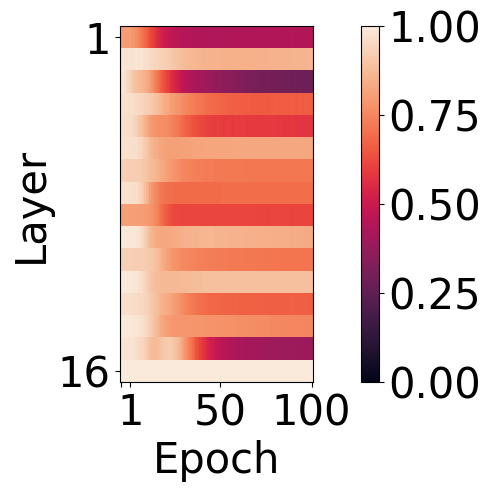}
    \\
    \includegraphics[width=0.5\linewidth]{./spectral_dynamics/figs/unet/weight_decay_10.0-sv-model.conv3.weight_sv.png}\hfil
    \includegraphics[width=0.5\linewidth]{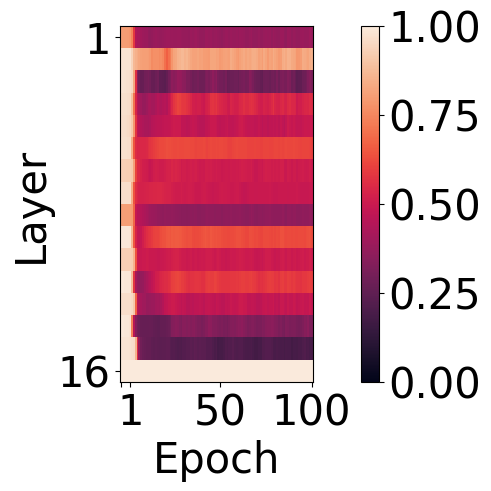}
    \caption{UNet}
  \end{subfigure}
  \begin{subfigure}[b]{0.24\linewidth}
    \centering
    \includegraphics[width=0.5\linewidth]{./spectral_dynamics/figs/lstm/weight_decay_0.0-sv-model.lstm.weight_ih_l2_sv.png}\hfil
    \includegraphics[width=0.5\linewidth]{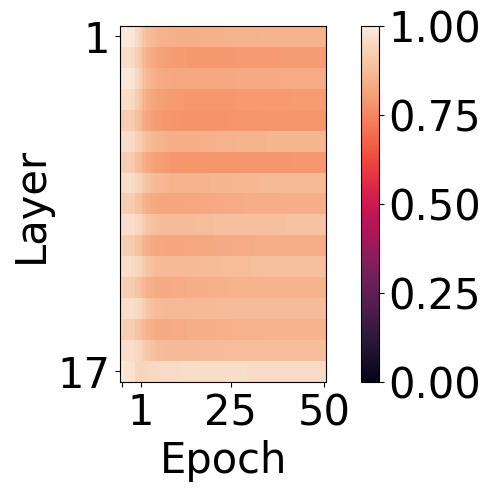}
    \\
    \includegraphics[width=0.5\linewidth]{./spectral_dynamics/figs/lstm/weight_decay_0.1-sv-model.lstm.weight_ih_l2_sv.png}\hfil
    \includegraphics[width=0.5\linewidth]{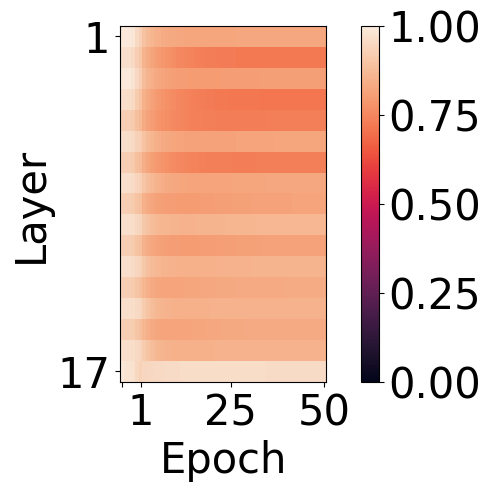}
    \\
        \includegraphics[width=0.5\linewidth]{./spectral_dynamics/figs/lstm/weight_decay_1.0-sv-model.lstm.weight_ih_l2_sv.png}\hfil
    \includegraphics[width=0.5\linewidth]{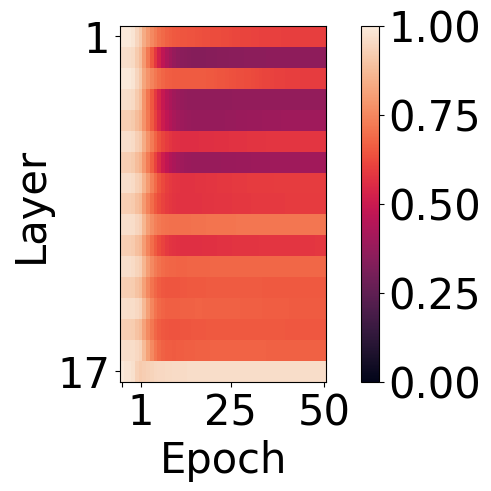}
    \\
    \includegraphics[width=0.5\linewidth]{./spectral_dynamics/figs/lstm/weight_decay_10.0-sv-model.lstm.weight_ih_l2_sv.png}\hfil
    \includegraphics[width=0.5\linewidth]{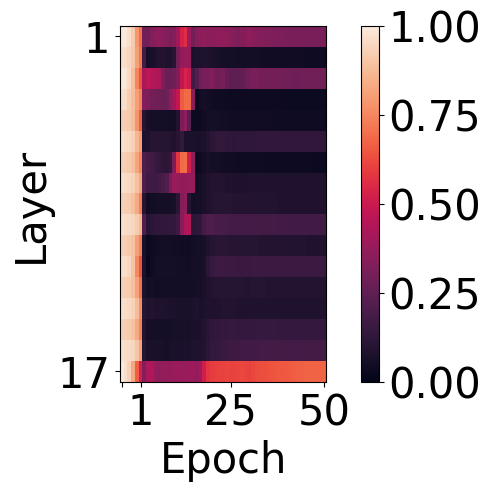}
    \caption{LSTM}
  \end{subfigure}
  \begin{subfigure}[b]{0.24\linewidth}
    \centering
    \includegraphics[width=0.5\linewidth]{./spectral_dynamics/figs/tfmr/weight_decay_0.0-sv-model.transformer_encoder.layers.7.linear1.weight_sv.png}\hfil
    \includegraphics[width=0.5\linewidth]{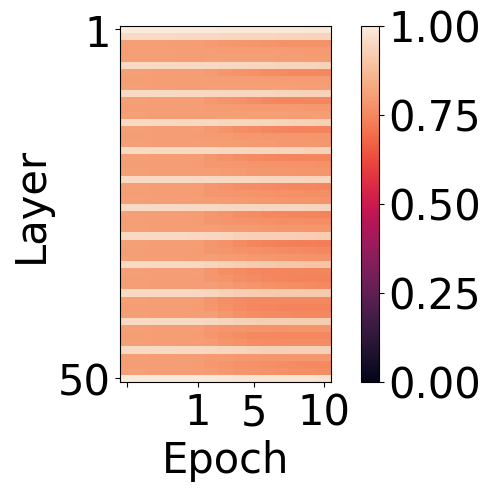}
    \\
    \includegraphics[width=0.5\linewidth]{./spectral_dynamics/figs/tfmr/weight_decay_0.1-sv-model.transformer_encoder.layers.7.linear1.weight_sv.png}\hfil
    \includegraphics[width=0.5\linewidth]{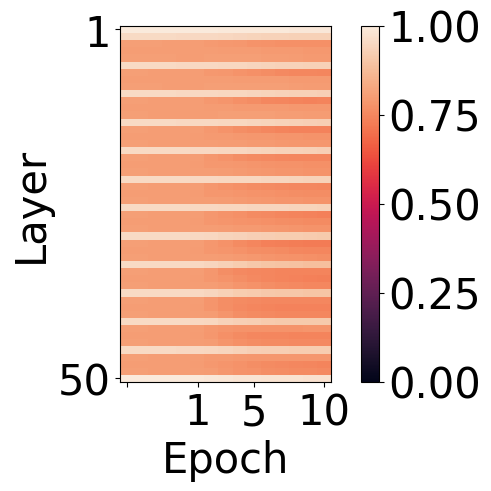}
    \\
    \includegraphics[width=0.5\linewidth]{./spectral_dynamics/figs/tfmr/weight_decay_1.0-sv-model.transformer_encoder.layers.7.linear1.weight_sv.png}\hfil
    \includegraphics[width=0.5\linewidth]{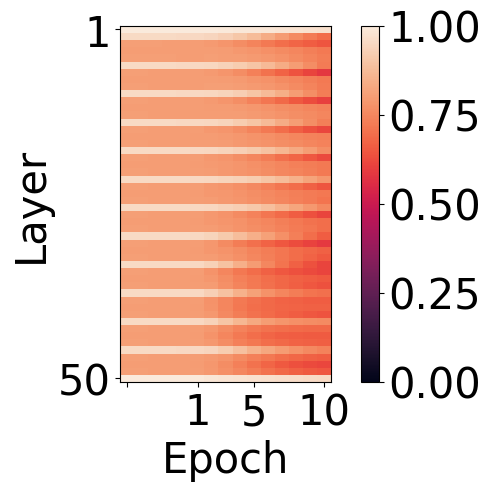}
    \\
    \includegraphics[width=0.5\linewidth]{./spectral_dynamics/figs/tfmr/weight_decay_10.0-sv-model.transformer_encoder.layers.7.linear1.weight_sv.png}\hfil
    \includegraphics[width=0.5\linewidth]{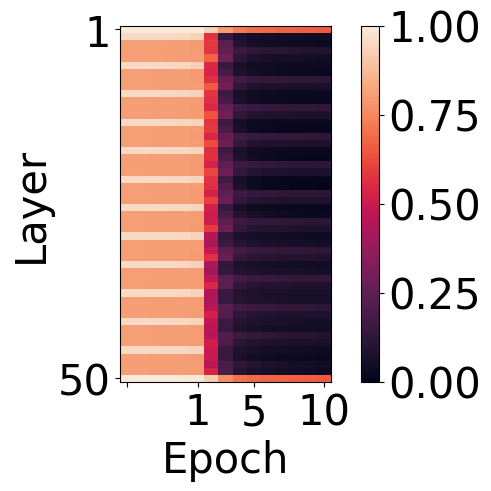}
    \caption{Transformer}
  \end{subfigure}
  \caption{Duplicate of Figure~\ref{fig:wd-effective-rank} with shared color scaling between settings.}
  \label{fig:wd-effective-rank-shared}
\end{figure*}

\section{Compute Resources}

All experiments are performed on an internal cluster with on the order of 100 NVIDIA RTX 2080 Ti GPUs or newer. All experiments run on a single GPU in less than 8 hours, though it is extremely helpful to parallelize across machines. We estimate that end-to-end it might take a few days on these resources to rerun all of the experiments in this chapter. Additionally, the storage requirements for all of the checkpoints will take on the order of 10 terabytes.

\section{Code Sources}

We use PyTorch~\citep{paszke2019pytorch} and NumPy~\citep{harris2020array} for all experiments and Weights \& Biases~\citep{biewald2020experiment} for experiment tracking. We make plots with Matplotlib~\citep{Hunter:2007} and Seaborn~\citep{Waskom2021}. We also use HuggingFace Datasets~\citep{lhoest2021datasets} for Wikitext-103~\citep{merity2016pointer}.

\chapter{Diverse Feature Learning}
For all results we train 5 seeds to get an estimate of the error.

\section{PGD Experiments with Image Classification}\label{app:mnist_cifar_details}

We train simple 4-layer MLPs with hidden dimension of 256 on the MNIST-CIFAR binary classification task. We use Adam with a tuned learning rate of 0.005 and a batch size of 128 without regularization. We train for 50 rounds of PGD or until training stops (as no ranks are left). We apply PGD to all layers except the last due to the low rank of the last layer. Each round of PGD projects off of 3 additional ranks. We do not apply PGD to biases as when ablating this we found no difference in the results.

\section{CKA Experiments with Image Classification}\label{app:imgclass_ensembles}

For all of our PGD and CKA experiments with image classification from ImageNet to CIFAR100 we use ResNet-18 with the Adam optimizer, a learning rate of 3e-4, batch size of 512 and weight decay of 1e-4. For speed of experimentation we restrict to 10 of the 1000 classes in ImageNet. For CKA and PGD jobs we initially start with PGD and CKA at
all layers, then sweep this choice over different subsets of blocks and layers. In the case of PGD instead of an absolute number of ranks, we project off 5\% of the probability mass of the singular values at each round. We do this as ResNet-18 has very different layer dimensions in different places. For CKA, the final best configuration corresponds to applying CKA after projection layers back into the residual stream (the second convolutional models, \texttt{.conv2}, in each block). We also sweep the loss coefficient $\gamma$ on a log grid from 0.1 to 1000, landing on an optimal choice of 10.

In each round, we train 50 epochs, and we train for 10 rounds. In the case of the larger model, we increase the width by 4x in every dimension, following~\citet{zhang2023learning}, which is more than 10 times the parameters and compute.

\section{Ensembling in Language Modeling}\label{app:language_ensembles}

For all experiments with language models, we train a simple Transformer~\citet{vaswani2017attention} language model with 8 layers, 512 hidden dimension, 8 heads, and the GPTNeoXTokenizer~\citep{black2022gpt} on BabiStories~\citep{zhang2025memory}. We train 6 rounds for 10 epochs each, with a batch size of 128, a sequence length of 128 and a warmup stable decay~\citep{hu2024minicpm} learning rate schedule. We compare ensembles to a model 3x as wide, which due to embeddings is a similar number of parameters but more compute.

For ensembling evaluations, except where otherwise noted, we average the probability distributions of the models before measuring loss.

We provide specific details for tuning the different diversity methods below.

\subsection{CKA}

For CKA, we experiment with which layers to place the loss on, from attention only, mlp only, early blocks, late blocks, middle blocks, and all layers. We also tune the coefficient on a log-spaced grid from 0.1 to 30.0 where it diverges. The best setting corresponds to only the mlp layers, which may make sense as this is the final output of each Transformer block, similar to the final convolution layers in residual blocks in the image classification experiments.

\subsection{NCL}

For NCL experiments, we tune the additional $\gamma$ coefficient on a log spaced grid from 0.01 to 10, experiment with whether to isolate NCL only to the correct class, or to the top-k values tuned from 1 to 5 to 10. For the top-p version of the experiments, we similarly tune $\gamma$ from 0.01 to 10, while considering a top-p threshold in $\{0.05, 0.1, 0.25, 0.5, 0.75, 0.9\}$.

\subsection{Focal weighting}

For focal weighting experiments, we tune the weighting coefficient from 0.01 to 10 on a log-spaced grid. We also tune the top-p version similarly to the case of NCL with a value in $\{0.05, 0.1, 0.25, 0.5, 0.75, 0.9\}$.

\subsection{Entropy weighting}

For entropy weighting experiments, we proceed with the same tuning process as for focal weighting.

\section{Ensemble Weighting}\label{app:ensemble_weighting}

For experiments on ensemble weighting, we run evaluations with a temperature of 0.01 to 100 on a log-spaced grid, and consider interpolating with the uniform distribution with a coefficient $\alpha \in \{0.2, 0.5, 1.0\}$.

\end{document}